\documentclass{article}

\usepackage{iclr2024_conference}
\usepackage{times}
\iclrfinalcopy

\usepackage{hyperref}
\usepackage{url}

\usepackage{amsmath}
\usepackage{tikz}
\usetikzlibrary{fit}
\usetikzlibrary{shapes,arrows}
\usetikzlibrary{arrows,calc,positioning}
\usepackage{algorithmic}
\usepackage{nicefrac} 
\usepackage{comment}
\usepackage{mathtools}

\usepackage{scalefnt,letltxmacro}
\LetLtxMacro{\oldtextsc}{\textsc}
\renewcommand{\textsc}[1]{\oldtextsc{\scalefont{1.10}#1}}
\usepackage[acronym,nowarn]{glossaries}
\setacronymstyle{long-sc-short}
\glsdisablehyper
\makeglossaries
\usepackage{xspace}
\usepackage{float}
\usepackage{physics}
\usepackage[normalem]{ulem}

\usepackage{enumitem}

\usepackage{amssymb}
\usepackage{mathtools}
\usepackage{amsfonts}
\usepackage{amsmath}
\usepackage{amsthm,thmtools}
\usepackage{booktabs}
\usepackage[]{microtype}
\usepackage{array}
\usepackage{multirow}
\usepackage{dsfont}
\usepackage{subcaption}
\usepackage{algorithm2e}
\usepackage{cleveref}

\usepackage{graphicx} 
\usepackage{booktabs}
\usepackage{colortbl}
\usepackage{multirow}
\usepackage{longtable}
\usepackage{xcolor}
\usepackage{lscape}
\usepackage{rotating}
\newacronym{MAP}{map}{maximum-a-posteriori}
\newacronym{MLE}{mle}{maximum likelihood estimation}
\newacronym{MNLL}{mnll}{mean negative loglikelihood}
\newacronym{NLL}{nll}{negative loglikelihood}
\newacronym{LL}{ll}{log-likelihood}
\newacronym{RMSE}{rmse}{root mean square error}
\newacronym{ECE}{ece}{expected calibration error}
\newacronym{SNR}{snr}{signal-to-noise ratio}
\newacronym{FID}{fid}{Fr\'echet Inception Distance}
\newacronym{FAD}{fad}{Fr\'echet Audio Distance}
\newacronym{FMD}{fmd}{Fr\'echet Modality Distance}
\newacronym{BPD}{bpd}{bit per dimension}
\newacronym{NFE}{nfe}{neural function evaluations}

\newacronym{AE}{ae}{autoencoder}
\newacronym{WAE}{wae}{Wasserstein Autoencoder}
\newacronym{VAE}{vae}{Variational Autoencoder}
\newacronym{BAE}{bae}{Bayesian autoencoder}
\newacronym{CDF}{cdf}{cumulative density function}
\newacronym{GAN}{gan}{Generative Adversarial Network}
\newacronym{DPGMM}{dpgmm}{Dirichlet process Gaussian mixture model}

\newacronym{MC}{mc}{Monte Carlo}
\newacronym{SDE}{sde}{Stochastic Differential Equation}
\newacronym{CNF}{cnf}{Continuous Normaxlizing Flow}
\newacronym{ODE}{ode}{Ordinary Differential Equation}
\newacronym{MCMC}{mcmc}{Markov chain Monte Carlo}
\newacronym{HMC}{hmc}{Hamiltonian Monte Carlo}
\newacronym{MH}{mh}{Metropolis-Hastings}
\newacronym{NUTS}{nuts}{no-u-turn sampler}
\newacronym{SGHMC}{sghmc}{stochastic gradient Hamiltonian Monte Carlo}

\newacronym{DGP}{dgp}{deep Gaussian process} 
\newacronym{GPLVM}{gplvm}{Gaussian process latent variable model}
\newacronym{DPMM}{dpmm}{Dirichlet Process Mixture Model}

\newacronym{VFE}{vfe}{variational free energy}

\newacronym[firstplural=Gaussian Processes]{GP}{gp}{Gaussian Process}

\newacronym{VI}{vi}{variational inference}

\newacronym{PDE}{pde}{Partial Differential Equation}
\newacronym{ELBO}{elbo}{evidence lower bound}
\newacronym{NELBO}{nelbo}{negative evidence lower bound}
\newacronym{ELL}{ell}{expected log likelihood}
\newacronym{KL}{kl}{Kullback-Leibler}
\newacronym{AUC}{auc}{area under the curve}

\newacronym[firstplural=Bayesian neural networks]{BNN}{bnn}{Bayesian neural network}
\newacronym[firstplural=deep neural networks]{DNN}{dnn}{deep neural network}
\newacronym[]{CNN}{cnn}{convolutional neural network}
\newacronym{MLP}{mlp}{multilayer perceptron}
\newacronym{NN}{nn}{neural network}
\newacronym{RELU}{ReLU}{rectified linear unit}

\newacronym{NF}{nf}{normalizing flow}

\newacronym{RBF}{rbf}{radial basis function}
\newacronym{ARD}{ard}{automatic relevance determination}

\newacronym{RKHS}{rkhs}{reproducing kernel Hilbert space}

\newacronym{OT}{ot}{optimal transport}
\newacronym{WD}{wd}{Wasserstein distance}
\newacronym{SWD}{swd}{sliced-Wasserstein distance}
\newacronym{DSWD}{dswd}{distributional sliced-Wasserstein distance}

\newcommand{\polymnist}{\textsc{polymnist}\xspace}
\newcommand{\cub}{\textsc{cub}\xspace}
\newcommand{\mnist}{\textsc{mnist}\xspace}
\newcommand{\mhd}{\textsc{mhd}\xspace}
\newcommand{\svhn}{\textsc{svhn}\xspace}

\newacronym{MLD}{mld}{Multi-modal Latent Diffusion}
\newacronym{MLD Inpaint}{mld in-paint}{Multi-modal Latent Diffusion with In-painting}
\newacronym{MLD Uni}{mld uni}{Multi-modal Latent Diffusion UniDiffuser }
\newacronym{MOPOE}{mopoe}{Mixture of Product of Experts}
\newacronym{MVAE}{mvae}{Product of Experts}
\newacronym{MMVAE}{mmvae}{Mixture of Expert}
\newacronym{NEXUS}{nexus}{Hierarchical Genertive Model}

\newacronym{MMVAEplus}{MMVAE+}{}

\newacronym{MVTCAE}{mvtcae}{Multi-view Total Correlation Autoencoder}
\newacronym{CLIP-S}{clip-s}{CLIP-Score}
\newacronym{MHD}{mhd}{The Multimodal Handwritten Digits data-set}
\newacronym{VPSDE}{vpsde}{Variance preserving SDE}
\newacronym{EMA}{ema}{Exponential moving average}
\newacronym{MSE}{mse}{Mean square error}

\newcommand{\g}{\,|\,}
\renewcommand{\d}[1]{\ensuremath{\operatorname{d}\!{#1}}}

\DeclarePairedDelimiterX{\infdivx}[2]{[}{]}{%
#1\;\delimsize\|\;#2%
}

\newcommand{\defeq}{\stackrel{\text{\tiny def}}{=}}

\usepackage{algorithm2e}
\RestyleAlgo{ruled} 
\usepackage{dsfont}

\title{Multi-modal Latent Diffusion}

\author{Mustapha Bounoua  \\\small{Renault Software Factory}:
 \\ \small{Department of Data Science}
  \\ \small{EURECOM, France}  \\ \small{\texttt{bounoua@eurecom.fr}}  \And
	 Giulio Franzese \\ \small{Department of Data Science} \\
  \small{EURECOM, France} \And
	 Pietro Michiardi \\ \small{Department of Data Science }
  \\ \small{EURECOM, France}
}

\begin{document}

\maketitle

\begin{abstract}
Multi-modal data-sets are ubiquitous in modern applications, and multi-modal Variational Autoencoders are a popular family of models that aim to learn a joint representation of the different modalities. However, existing approaches suffer from a coherence--quality tradeoff, where models with good generation quality lack generative coherence across modalities, and vice versa. We discuss the limitations underlying the unsatisfactory performance of existing methods, to motivate the need for a different approach. We propose a novel method that uses a set of independently trained, uni-modal, deterministic autoencoders. Individual latent variables are concatenated into a common latent space, which is fed to a masked diffusion model to enable generative modeling. We also introduce a new multi-time training method to learn the conditional score network for multi-modal diffusion. Our methodology substantially outperforms competitors in both generation quality and coherence, as shown through an extensive experimental campaign. 

\end{abstract}

\section{Introduction}\label{sec:introduction}

Multi-modal generative modelling is a crucial area of research in machine learning that aims to develop models capable of generating data according to multiple modalities, such as images, text, audio, and more. This is important because real-world observations are often captured in various forms, and combining multiple modalities describing the same information can be an invaluable asset. For instance, images and text can provide complementary information in describing an object, audio and video can capture different aspects of a scene. Multi-modal generative models can also help in tasks such as data augmentation \citep{he2023is, azizi2023synthetic, sariyildiz2023fake}, missing modality imputation \citep{pmlr-v97-antelmi19a, 9680207, zhang2023unified, Tran_2017_CVPR}, and conditional generation \citep{huang2022eccv, lee2018harmonizing}.

Multi-modal models have flourished over the past years and have seen a tremendous interest from academia and industry, especially in the content creation sector. Whereas most recent approaches focus on specialization, by considering text as primary input to be associated mainly to images \citep{Rombach_2022_CVPR, saharia2022photorealistic, ramesh2022hierarchical, tao2022dfgan, Wu_2022_CVPR, nichol2022glide, chang2023muse} and videos \citep{blattmann2023align, hong2023cogvideo, singer2022makeavideo}, in this work we target an established literature whose scope is more general, and in which all modalities are considered equally important. A large body of work rely on extensions of the \gls{VAE} \citep{kingmaW13vae} to the multi-modal domain: initially interested in learning joint latent representation of multi-modal data, such works have mostly focused on generative modeling.
Multi-modal generative models aim at \textit{high-quality} data generation, as well as generative \textit{coherence} across all modalities. These objectives apply to both joint generation of new data, and to conditional generation of missing modalities, given a disjoint set of available modalities.

In short, multi-modal \glspl{VAE} rely on combinations of uni-modal \glspl{VAE}, and the design space consists mainly in the way the uni-modal latent variables are combined, to construct the joint posterior distribution. Early work such as \cite{mvaeWu2018} adopt a product of experts approach, whereas others \cite{mmvaeShi2019} consider a mixture of expert approach. Product-based models achieve high generative quality, but suffer in terms of both joint and conditional coherence. This was found to be due to experts mis-calibration issues \citep{mmvaeShi2019, sutter2021generalized}. On the other hand, mixture-based models produce coherent but qualitatively poor samples.
A first attempt to address the so called \textbf{coherence--quality tradeoff} \citep{Daunhawer2021} is represented by the mixture of product of experts approach \citep{sutter2021generalized}. However recent comparative studies \citep{Daunhawer2021} show that none of the existing approaches fulfill both the generative quality and  coherence criteria. A variety of techniques aim at finding a better operating point, such as contrastive learning techniques \citep{contrastShi2020}, hierarchical schemes \citep{Vasco2022}, total correlation based calibration of single modality encoders \citep{tcmvaehwang21}, or different training objectives \cite{jsdSutter2020}. More recently, the work in \citep{palumbo2023mmvae} considers explicitly separated shared and private latent spaces to overcome the aforementioned limitations.

By expanding on results presented in \citep{Daunhawer2021}, in \Cref{sec:motivation} we further investigate the tradeoff between generative coherence and quality, and argue that it is intrinsic to all variants of multi-modal \glspl{VAE}. We indicate two root causes of such problem: latent variable collapse \citep{alemi2018fixing, dieng2019avoiding} and information loss due to mixture sub-sampling. To tackle these issues, in this work, we propose in \Cref{sec:method} a new approach which uses a set of independent, uni-modal \textit{deterministic} auto-encoders whose latent variables are simply concatenated in a joint latent variable. Joint and conditional generative capabilities are provided by an additional model that learns a probability density associated to the joint latent variable.
We propose an extension of score-based diffusion models \citep{Song2020} to operate on the multi-modal latent space. We thus derive both forward and backward dynamics that are compatible with the multi-modal nature of the latent data. In \cref{sec:mld_var} we propose a novel method to train the multi-modal score network, such that it can both be used for joint and conditional generation. Our approach is based on a guidance mechanism, which we compare to alternatives. We label our approach \gls{MLD}.

Our experimental evaluation of \gls{MLD} in \Cref{sec:experiments} provides compelling evidence of the superiority of our approach for multi-modal generative modeling. We compare \gls{MLD} to a large variety of \gls{VAE}-based alternatives, on several real-life multi-modal data-sets, in terms of generative quality and both joint and conditional coherence. Our model outperforms alternatives in all possible scenarios, even those that are notoriously difficult because modalities might be only loosely correlated. 
Note that recent work also explore the joint generation of multiple modalities \cite{Ruan2022, Hu2022}, but such approaches are application specific, e.g. text-to-image, and essentially only target two modalities. When relevant, we compare our method to additional recent alternatives to multi-modal diffusion \citep{bao2023transformer, wesego2023scorebased}, and show superior performance of \gls{MLD}.

\section{Limitations of Multi-modal VAEs}\label{sec:motivation}

In this work, we consider multi-modal \glspl{VAE} \citep{mvaeWu2018, mmvaeShi2019, sutter2021generalized, palumbo2023mmvae} as the standard modeling approach to tackle both joint and conditional generation of multiple modalities. Our goal here is to motivate the need to go beyond such a standard approach, to overcome limitations that affect multi-modal \glspl{VAE}, which result in a trade-off between generation quality and generative coherence \citep{Daunhawer2021, palumbo2023mmvae}.

Consider the random variable $X=\{X^{1},\dots,X^{M}\} \sim p_{D}(x^1,\dots,x^M)$, consisting in the set of $M$ of modalities sampled from the (unknown) multi-modal data distribution $p_{D}$. We indicate the marginal distribution of a single modality by $X^i \sim p_D^i(x^i)$ and the collection of a generic subset of modalities by $X^{A}\sim p_D^A(x^A)$, with $X^{A} \defeq \{X^i \}_{i\in A}$, where $A\subset \{1,\dots,M\}$ is a set of indexes. For example: given $A=\{1,3,5\}$, then $X^{A} = \{X^1, X^3, X^5\}$.

We begin by considering uni-modal \glspl{VAE} as particular instances of the Markov chain $X\rightarrow Z\rightarrow \hat{X}$, where $Z$ is a latent variable and $\hat{X}$ is the generated variable. Models are specified by the two conditional distributions, called the encoder $Z\g_{X=x}\sim q_\psi(z\g x)$, and the decoder $\hat{X}\g_{Z=z}\sim p_\theta(\hat{x}\g z)$. 
Given a prior distribution $p_n(z)$, the objective is to define a generative model whose samples are distributed as closely as possible to the original data. 

In the case of multi-modal \glspl{VAE}, we consider the general family of \gls{MOPOE} \citep{sutter2021generalized}, which includes as particular cases many existing variants, such as \gls{MVAE} \citep{mvaeWu2018} and \gls{MMVAE} \citep{mmvaeShi2019}. Formally,  a collection of $K$ arbitrary subsets of modalities $S=\{A_1,\dots A_K\}$, along with weighting coefficients $\omega_i\geq 0,\sum_{i=1}^{K}\omega_i=1$, define the posterior $q_\psi(z\g x)=\sum_i \omega_i q^i_{\psi^{A_i}}(z\g x^{A_i})$, with $\psi=\{\psi^1,\dots,\psi^K\}$. 
To lighten the notation, we use $q_{\psi^{A_i}}$ in place of $q^i_{\psi^{A_i}}$ noting that the various $q^i_{\psi^{A_i}}$ can have both different parameters $\psi^{A_i}$ and functional form. For example, in the \gls{MOPOE} \citep{sutter2021generalized} parametrization, we have: $q_{\psi^{A_i}}(z\g x^{A_i})=\prod_{j\in A_i}q_{\psi^j}(z\g x^{j})$. Our exposition is more general and not limited to this assumption.
The selection of the posterior can be understood as the result induced by the two step procedure where i) each subset of modalities $A_i$ is encoded into specific latent variables $Y_i\sim q_{\psi^{A_i}}(\cdot \g x^{A_i})$ and ii) the latent variable $Z$ is obtained as $Z=Y_i$ with probability $\omega_i$. 
Optimization is performed w.r.t. the following \gls{ELBO} \citep{Daunhawer2021, sutter2021generalized}:
\begin{equation}\label{eq:elbo_M}
    \mathcal{L}=\sum\limits_i \omega_i \int p_D(x)q_{\psi^{A_i}}(z\g x^{A_i})\log p_{\theta}(x|z)-\log\frac{q_{\psi^{A_i}}(z\g x^{A_i})}{p_n(z)}\dd z\dd x.
\end{equation}



A well-known limitation called the latent collapse problem \citep{alemi2018fixing,dieng2019avoiding} affects the quality of latent variables $Z$. Consider the hypothetical case of arbitrary flexible encoders and decoders: then, posteriors with zero mutual information with respect to model inputs are valid maximizers of \Cref{eq:elbo_M}. To prove this, it is sufficient to substitute the posteriors $q_{\psi^{A_i}}(z\g x^{A_i})=p_n(z)$ and $p_\theta(x|z)=p_D(x)$ into the \Cref{eq:elbo_M} to observe that the optimal value $\mathcal{L}=\int p_D(x)\log p_D(x)\dd x$ is achieved \citep{alemi2018fixing,dieng2019avoiding}.
The problem of information loss is exacerbated in the case of multi-modal \glspl{VAE} \citep{Daunhawer2021}. 
Intuitively, even if the encoders $q_{\psi^{A_i}}(z\g x^{A_i})$ carry relevant information about their inputs $X^{A_i}$, step ii) of the multi-modal encoding procedure described above induces a further information bottleneck. A fraction $\omega_i$ of the time, the latent variable $Z$ will be a copy of $Y_i$, that only provides information about the subset $X^{A_i}$. No matter how good the encoding step is, the information about $X^{\{1,\dots,M\}\setminus A}$ that is not contained in $X^{A_i}$ cannot be retrieved.

Furthermore, if the latent variable carries zero mutual information w.r.t. the multi-modal input, a coherent \textit{conditional} generation of a set of modalities given others is impossible, since $\hat{X}^{A_1}\perp X^{A_2}$ for any generic sets $A_1,A_2$. While the factorization $p_\theta(x\g z) = \prod_{i=1}^M p_{\theta^i}(x^i\g z)$, $\theta=\{\theta_1,\dots,\theta_M\}$ --- where we use $p_{\theta^i}$ instead of $p^i_{\theta^i}$ to unclutter the notation --- could enforce preservation of information and guarantee a better quality of the \textit{jointly} generated data, in practice, the latent collapse phenomenon induces multi-modal \glspl{VAE} to converge toward sub-optimal operating regime. When the posterior $q_\psi(z\g x)$ collapses onto the uninformative prior $p_n(z)$, the \gls{ELBO} in \Cref{eq:elbo_M} reduces to the sum of modality independent reconstruction terms 
$\sum\limits_i \omega_i \sum_{j\in A_i}\int p_D^{j}(x^{j})p_n(z)\left(\log p_{\theta^j}(x^j|z)\right)\dd z\dd x^{j}$. In this case, flexible decoders can similarly ignore the latent variable and converge to the solution $p_{\theta^j}(x^j|z)=p_D^{j}(x^{j})$ where, paradoxically, the quality of the approximation of the various marginal distributions is extremely high, while there is a complete lack of joint coherence. 

General principles to avoid latent collapse consist in explicitly forcing the learning of informative encoders $q_\theta(z\g x)$ via $\beta-$annealing of the \gls{KL} term in the \gls{ELBO} and the reduction of the representational power of encoders and decoders. While $\beta-$annealing has been explored in the literature \citep{mvaeWu2018} with limited improvements, reducing the flexibility of encoders/decoders clearly impacts the generation quality.
Hence the presence of a trade-off: to improve coherence, the flexibility of encoders/decoders should be constrained, which in turns hurt generative quality. This trade-off has been recently addressed in the literature of multi-modal \glspl{VAE} \citep{Daunhawer2021, palumbo2023mmvae}, but our experimental results in \Cref{sec:experiments} indicate that there is ample room for improvement, and that a new approach is truly needed.

\section{Our Approach: Multi-modal Latent Diffusion}\label{sec:method}
We propose a new method for multi-modal generative modeling that, by design, does not suffer from the limitations discussed in \Cref{sec:motivation}. Our objective is to enable both high-quality and coherent joint/conditional data generation, using a simple design (see \Cref{apdx:mld_details} for a schematic representation).
As an overview, we use deterministic uni-modal autoencoders, whereby each modality $X^i$ is encoded through its encoder $e_{\psi^i}$, which is a short form for $e^i_{\psi^i}$, into the modality specific latent variable $Z^i$ and decoded into the corresponding $\hat{X}^i=d_{\theta^i}(Z^i)$. Our approach can be interpreted as a latent variable model where the different latent variables $Z^i$ are concatenated as $Z=[Z^1,\dots,Z^M]$. This corresponds to the parametrization of the two conditional distributions as $q_\psi(z\g x)=\prod\limits_{i=1}^{M} \delta(z^i-e_{\psi^i}(x^i))$ and $p_\theta(\hat{x}\g z)=\prod\limits_{i=1}^{M}\delta(\hat{x}^i-d_{\theta^i}(z^i))$,  respectively.
Then, in place of an \gls{ELBO}, we optimize the parameters of our autoencoders by minimizing the following sum of modality specific losses: 
\begin{equation}\label{eq:AE}
    \mathcal{L}=\sum\limits_{i=1}^{M}\mathcal{L}_i,\quad \mathcal{L}_i=\int p_D^i(x^i)  l^i(x^i-d_{\theta^i}(e_{\psi^i}(x^i)))\dd x^i,
\end{equation}
where $l^i$ can be any valid distance function, e.g, the square norm $\norm{\cdot}^2$.
Parameters $\psi^i,\theta^i$ are modality specific: then, minimization of \Cref{eq:AE} corresponds to individual training of the different autoencoders.
Since the mapping from input to latent is deterministic, there is no loss of information between $X$ and $Z$.\footnote{Since the measures are not absolutely continuous w.r.t the Lebesgue measure, mutual information is $+\infty$.} Moreover, this choice avoids any form of interference in the back-propagated gradients corresponding to the uni-modal reconstruction losses. Consequently gradient conflicts issues \citep{Javaloy2022}, where stronger modalities pollute weaker ones, are avoided.

To enable such a simple design to become a generative model, it is sufficient to generate samples from the induced latent distribution $Z\sim q_\psi(z)= \int p_D(x) q_\psi(z\g x)\dd x$ and decode them as $\hat{X}=d_\theta(Z)=[d_{\theta^1}(Z^1),\dots,d_{\theta^M}(Z^M)]$. To obtain such samples, we follow the two-stage procedure described in \cite{loaizadiagnosing,tran2021model}, where samples from the lower dimensional $q_\psi(z)$ are obtained through an appropriate generative model. We consider score-based diffusion models in latent space \citep{Rombach_2022_CVPR,Vahdat2021} to solve this task, and call our approach Multi-modal Latent Diffusion (\gls{MLD}). 
It may be helpful to clarify, at this point, that the two-stage training of \gls{MLD} is carried out separately. Uni-modal deterministic autoencoders are pre-trained first, followed by the training of the score-based diffusion model, which is explained in more detail later.


To conclude the overview of our method, for joint data generation, one can sample from noise, perform backward diffusion, and then decode the generated multi-modal latent variable to obtain the corresponding data samples. For conditional data generation, given one modality, the reverse diffusion is guided by this modality, while the other modalities are generated by sampling from noise. The generated latent variable is then decoded to obtain data samples of the missing modality.

\subsection{Joint and Conditional Multi-modal Latent Diffusion Processes}\label{sec:mld}
In the first stage of our method, the deterministic encoders project the input modalities $X^i$ into the corresponding latent spaces $Z^i$. This transformation induces a distribution $q_\psi(z)$ for the latent variable $Z=[Z^1,\dots,Z^M]$, resulting from the concatenation of uni-modal latent variables.

\noindent \textbf{Joint generation.} To generate a new sample for all modalities we use a simple score-based diffusion model in latent space \citep{diff,Song2020,Vahdat2021,loaizadiagnosing,tran2021model}. This requires reversing a stochastic noising process, starting from a simple, Gaussian distribution. 
Formally, the noising process is defined by a \gls{SDE} of the form:
\begin{equation}\label{eq:fw_sde_j}
    \dd R_t=\alpha(t)R_t\dd t+g(t)\dd W_t,\,\,\, R_0\sim q(r,0),
\end{equation}
where $\alpha(t)R_t$ and $g(t)$ are the drift and diffusion terms, respectively, and $W_t$ is a Wiener process. 
The time-varying probability density $q(r,t)$ of the stochastic process at time $t\in[0,T]$, where $T$ is finite, satisfies the Fokker-Planck equation \citep{oksendal2013stochastic}, with initial conditions $q(r,0)$. We assume uniqueness and existence of a stationary distribution $\rho(r)$ for the process \Cref{eq:fw_sde_j}.\footnote{This is not necessary for the validity of the method \cite{song2021maximum}} 
The forward diffusion dynamics depend on the initial conditions $R_0\sim q(r,0)$. We consider $R_0=Z$ to be the initial condition for the diffusion process, which is equivalent to $q(r,0)=q_\psi(r)$. Under loose conditions \citep{anderson1982reverse}, a time-reversed stochastic process exists, with a new \gls{SDE} of the form:
\begin{equation}\label{eq:bw_sde_j}
  \dd R_t=\left(-\alpha(T-t)R_t+g^2(T-t)\nabla\log(q(R_t,T-t))\right)\dd t+g(T-t)\dd W_t, \,\,\, R_0\sim q(r,T),  
\end{equation}
indicating that, in principle, simulation of \Cref{eq:bw_sde_j} allows to generate samples from the desired distribution $q(r,0)$. In practice, we use a \textbf{parametric score network} $s_\chi(r,t)$ to approximate the true score function, and we approximate $q(r,T)$ with the stationary distribution $\rho(r)$. Indeed, the generated data distribution $q(r,0)$ is close (in KL sense) to the true density as described by \cite{song2021maximum, howmuch}: 
\begin{equation}\label{elbo_diff_j}
\mathrm{KL}[q_\psi(r)\g\g q(r,0)]\leq \frac{1}{2}\int\limits_0^T g^2(t)\mathbb{E}[\norm{s_\chi(R_t,t)-\nabla\log q(R_t,t)}^2]\dd t+KL[q(r,T)||\rho(r)],     
\end{equation}

where the first term on the r.h.s is referred to as score-matching objective, and is the loss over which the score network is optimized, and the second is a vanishing term for $T\rightarrow\infty$.

To conclude, joint generation of all modalities is achieved through the simulation of the reverse-time \gls{SDE} in \Cref{eq:bw_sde_j}, followed by a simple decoding procedure. Indeed, optimally trained decoders (achieving zero in \Cref{eq:AE}) can be used to transform $Z\sim q_\psi(z)$ into samples from $\int p_\theta(x\g z)q_\psi(z)\dd z=p_D(x)$.

\noindent \textbf{Conditional generation.} Given a generic partition of all modalities into non overlapping sets $A_1 \cup A_2$, where $A_2=\left(  {\{1,\dots,M\}\setminus A_1}  \right)$, conditional generation requires samples from the conditional distribution $q_\psi(z^{A_1}\g z^{A_2})$, which are based on \textit{masked} forward and backward diffusion processes. 

Given conditioning latent modalities $z^{A_2}$, we consider a modified forward diffusion process with initial conditions $R_0=\mathcal{C}(R^{A_1}_0,R^{A_2}_0)$, with $R^{A_1}_0\sim q_\psi(r^{A_1}\g z^{A_2}),R^{A_2}_0=z^{A_2}$. The composition operation $\mathcal{C}(\cdot)$ concatenates generated ($R^{A_1}$) and conditioning latents ($z^{A_2}$). As an illustration, consider $A_1 = \{1,3,5\}$, such that $X^{A_1} = \{X^1, X^3, X^5 \}$, and $A_2 = \{2,4,6\}$ such that $X^{A_2} = \{X^2, X^4, X^6 \}$. Then, $R_0=\mathcal{C}(R^{A_1}_0,R^{A_2})=\mathcal{C}(R^{A_1}_0,z^{A_2})=[ R^1_0,z^2,R^3_0,z^4,R^5_0,z^6]$.

More formally, we define the masked forward diffusion \gls{SDE}:
\begin{equation}\label{eq:cond_fw_sde_c}
    \dd R_t=m(A_1)\odot\left[ \alpha(t)R_t\dd t+g(t)\dd W_t\right],\,\,\, q(r,0)= q_\psi(r^{A_1}\g z^{A_2})\delta(r^{A_2}-z^{A_2}).
\end{equation}

The mask $m(A_1)$ contains $M$ vectors $u^i$, one per modality, and with the corresponding cardinality. If modality $j \in A_1$, then $u^j = \mathbf{1}$, otherwise $u^j = \mathbf{0}$. Then, the effect of masking is to ``freeze'' throughout the diffusion process the part of the random variable $R_t$ corresponding to the conditioning latent modalities $z^{A_2}$. We naturally associate to this modified forward process the conditional time varying density $q(r,t\g z^{A_2})=q(r^{A_1},t\g z^{A_2})\delta(r^{A_2}-z^{A_2})$.

To sample from $q_\psi(z^{A_1}\g z^{A_2})$, we derive the reverse-time dynamics of \Cref{eq:cond_fw_sde_c} as follows:
\begin{equation}\label{eq:cond_bw_sde_c}
  \dd R_t=m(A_1)\odot\left[\left(-\alpha(T-t)R_t+g^2(T-t)\nabla\log(q(R_t,T-t\g z^{A_2}))\right)\dd t+g(T-t)\dd W_t\right], 
\end{equation}
with initial conditions $R_0=\mathcal{C}(R^{A_1}_0,z^{A_2})$ and $R^{A_1}_0\sim q(r^{A_1},T\g z^{A_2})$. Then, we approximate $q(r^{A_1},T\g z^{A_2})$ by its corresponding steady state distribution $\rho(r^{A_1})$, and the true (conditional) score function $\nabla\log(q(r,t\g z^{A_2}))$ by a conditional score network $s_\chi(r^{A_1},t\g z^{A_2})$.

\section{Guidance Mechanisms to Learn the Conditional Score Network}\label{sec:mld_var}

A correctly optimized score network $s_\chi(r,t)$ allows, through simulation of \Cref{eq:bw_sde_j}, to obtain samples from the joint distribution $q_\psi(z)$. Similarly, a \textit{conditional} score network $s_\chi(r^{A_1},t\g z^{A_2})$ allows, through the simulation of \Cref{eq:cond_bw_sde_c}, to sample from $q_\psi(z^{A_1}\g z^{A_2})$. In \Cref{sec:multitime} we extend guidance mechanisms used in classical diffusion models to allow multi-modal conditional generation. A na\"ive alternative is to rely on the unconditional score network $s_\chi(r,t)$ for the conditional generation task, by casting it as an \textit{in-painting} objective. Intuitively, any missing modality could be recovered in the same way as a uni-modal diffusion model can recover masked information.
In \Cref{sec:inpainting} we discuss the implicit assumptions underlying in-painting from an information theoretic perspective, and argue that, in the context of multi-modal data, such assumptions are difficult to satisfy. Our intuition is corroborated by ample empirical evidence, where our method consistently outperform alternatives.

\subsection{Multi-time Diffusion}
\label{sec:multitime}

We propose a modification to the classifier-free guidance technique \citep{ho2022classifier} to learn a score network that can generate conditional and unconditional samples from any subset of modalities. Instead of training a separate score network for each possible combination of conditional modalities, which is computationally infeasible, we use a single architecture that accepts all modalities as inputs and a \textit{multi-time vector} $\tau=[t_1,\dots,t_M]$. The multi-time vector serves two purposes: it is both a conditioning signal and the time at which we observe the diffusion process. 

\noindent \textbf{Training:} learning the conditional score network relies on randomization. As discussed in \Cref{sec:mld}, we consider an arbitrary partitioning of all modalities in two disjoint sets, $A_1$ and $A_2$. The set $A_2$ contains randomly selected conditioning modalities, while the remaining modalities belong to set $A_1$. Then, during training, the parametric score network estimates $\nabla\log(q(r,t\g z^{A_2}))$, whereby the set $A_2$ is randomly chosen at every step. This is achieved by the \textit{masked diffusion process} from \Cref{eq:cond_fw_sde_c}, which only diffuses modalities in $A_1$. More formally, the score network input is $R_t=\mathcal{C}(R^{A_1}_t,Z^{A_2})$, along with a multi-time vector ${\tau}(A_1,t) = t\begin{bmatrix} \mathds{1}(1\in A_1),\dots,\mathds{1}(M\in A_1) \end{bmatrix}$.
As a follow-up of the example in \Cref{sec:mld}, given $A_1 = \{1,3,5\}$, such that $X^{A_1} = \{X^1, X^3, X^5 \}$, and $A_2 = \{2,4,6\}$ such that $X^{A_2} = \{X^2, X^4, X^6 \}$, then, ${\tau}(A_1,t) = [t,0,t,0,t,0]$.

More precisely, the algorithm for the multi-time diffusion training (see \ref{apdx:mld_details} for the pseudo-code) proceeds as follows. At each step, a set of conditioning modalities $A_2$ is sampled from a predefined distribution $\nu$, where $\nu(\emptyset)\defeq\Pr(A_2=\emptyset)=d$, and $\nu(U)\defeq\Pr(A_2=U)=\nicefrac{(1-d)}{(2^M-1)}$ with $U\in \mathcal{P}(\{1,\dots,M\})\setminus\emptyset$, where $\mathcal{P}(\{1,\dots,M\})$ is the powerset of all modalities. The corresponding set $A_1$ and mask $m(A_1)$ are constructed, and a sample $X$ is drawn from the training data-set. The corresponding latent variables $Z^{A_1}=\{e^i_{\psi}(X^i)\}_{i\in A_1}$ and $Z^{A_2}=\{e^i_{\psi}(X^i)\}_{i\in A_2}$ are computed using the pre-trained encoders, and a diffusion process starting from $R_0 = \mathcal{C}(Z^{A_1}, Z^{A_2})$ is simulated for a randomly chosen diffusion time $t$, using the conditional forward SDE with the mask $m(A_1)$. The score network is then fed the current state $R_t$ and multi-time vector $\tau(A_1,t)$, and the difference between the score network's prediction and the true score is computed, applying the mask $m(A_1)$. The score network parameters are updated using stochastic gradient descent, and this process is repeated for a total of $L$ training steps. Clearly, when $A_2 = \emptyset$, training proceeds as for an un-masked diffusion process, since the mask $m(A_1)$ allows all latent variables to be diffused.

\noindent \textbf{Conditional generation:} any valid numerical integration scheme for \Cref{eq:cond_bw_sde_c} can be used for conditional sampling (see \ref{apdx:mld_details} for an implementation using the Euler-Maruyama integrator). First, conditioning modalities in the set $A_2$ are encoded into the corresponding latent variables $z^{A_2}=\{e^j(x^{j})\}_{j\in A_2}$. Then, numerical integration is performed with step-size $\Delta t=\nicefrac{T}{N}$, starting from the initial conditions $R_0=\mathcal{C}(R^{A_1}_0,z^{A_2})$, with $R^{A_1}_0\sim \rho(r^{A_1})$. At each integration step, the score network $s_\chi$ is fed the current state of the process and the multi-time vector $\tau(A_1,\cdot)$. Before updating the state, the masking is applied. Finally, the generated modalities are obtained thanks to the decoders as $\hat{X}^{A_1}=\{d^j_\theta(R^{j}_T)\}_{j\in A_1}$.
Inference time conditional generation is not randomized: conditioning modalities are the ones that are available, whereas the remaining are the ones we wish to generate.



 Any-to-any multi-modality has been recently studied through the composition of modality-specific diffusion models \citep{tang2023anytoany}, by designing cross-attention and training procedures that allow arbitrary conditional generation. The work by \citet{tang2023anytoany} relies on latent interpolation of input modalities, which is akin to mixture models, and uses it as conditioning signal for individual diffusion models. This is substantially different from the joint nature of the multi-modal latent diffusion we present in our work: instead of forcing entanglement through cross-attention between score networks, our model relies on joint diffusion process, whereby modalities naturally co-evolve according to the diffusion process. Another recent work \citep{wu2023nextgpt} targets multi-modal conversational agents, whereby the strong, underlying assumption is to consider one modality, i.e., text, as a guide for the alignment and generation of other modalities. Even if conversational objectives are orthogonal to our work, techniques akin to instruction following for cross-generation, are an interesting illustration of the powerful capabilities of in-context learning of LLMs \citep{xie2022an, min2022rethinking}.

\subsection{In-painting and its implicit assumptions}\label{sec:inpainting}
Under certain assumptions, given an unconditional score network $s_\chi(r,t)$ that approximates the true score $\nabla\log q(r,t)$, it is possible to obtain a conditional score network $s_\chi(r^{A_1},t\g z^{A_2})$, to approximate $\nabla\log q(r^{A_1},t\g z^{A_2})$. We start by observing the equality:

\begin{footnotesize}
\begin{equation}\label{eq:inp}
    q(r^{A_1},t\g z^{A_2})=\int q(\mathcal{C}(r^{A_1},r^{A_2}),t\g z^{A_2}) \d r^{A_2}=\int \frac{q(z^{A_2}\g \mathcal{C}(r^{A_1},r^{A_2}),t)}{q_\psi(z^{A_2})}q(\mathcal{C}(r^{A_1},r^{A_2}),t)\d r^{A_2},
\end{equation}
\end{footnotesize}

where, with a slight abuse of notation, we indicate with $q(z^{A_2}\g \mathcal{C}(r^{A_1},r^{A_2}),t)$ the density associated to the event: the portion corresponding to $A_2$ of the latent variable $Z$ is equal to $z^{A_2}$ given that the whole diffused latent $R_t$ at time $t$, is equal to $\mathcal{C}(r^{A_1},r^{A_2})$. In the literature, the quantity $q(z^{A_2}\g \mathcal{C}(r^{A_1},r^{A_2}),t)$ is typically approximated by dropping its dependency on $r^{A_1}$. This approximation can be used to manipulate \Cref{eq:inp} as $q(r^{A_1},t\g z^{A_2})\simeq \int q(r^{A_2},t\g z^{A_2})q(r^{A_1},t|r^{A_2},t)\d r$. Further Monte-Carlo approximations \citep{Song2020,lugmayr2022repaint} of the integral allow implementation of a practical scheme, where an approximate conditional score network is used to generate conditional samples. This approach, known in the literature as \textit{in-painting}, provides high quality results in several \textit{uni-modal} application domains \citep{Song2020,lugmayr2022repaint}. 

The \gls{KL} divergence between $q(z^{A_2}\g \mathcal{C}(r^{A_1},r^{A_2}),t)$ and $q(z^{A_2}\g r^{A_2},t)$ quantifies, fixing $r^{A_1},r^{A_2}$, the discrepancy between the true and approximated conditional probabilities. Similarly, the expected \gls{KL} divergence $\Delta=\int q(r,t)\mathrm{KL}[q(z^{A_2}\g \mathcal{C}(r^{A_1},r^{A_2}),t)\g\g q(z^{A_2}\g r^{A_2},t)]\dd r$, provides information about the average discrepancy. Simple manipulations allow to recast this as a discrepancy in terms of mutual information $\Delta=I(Z^{A_2};R^{A_1}_t,R^{A_2}_t)-I(Z^{A_2};R^{A_2}_t)$. Information about $Z^{A_2}$ is contained in $R^{A_2}_t$, as the latter is the result of a diffusion with the former as initial conditions, corresponding to the Markov chain $R^{A_2}_t\rightarrow Z^{A_2}$, and in $R^{A_1}_t$ through the Markov chain $Z^{A_2} \rightarrow Z^{A_1} \rightarrow R^{A_1}_t$. The positive quantity $\Delta$ is close to zero whenever the rate of loss of information w.r.t initial conditions is similar for the two subsets $A_1,A_2$. In other terms, $\Delta\simeq 0$ whenever out of the whole $R_t$, the portion $R^{A_2}_t$ is a sufficient statistic for $Z^{A_2}$.  

The assumptions underlying the approximation are in general not valid in the case of multi-modal learning, where the robustness to stochastic perturbations of latent variables corresponding to the various modalities can vary greatly. Our claim are supported empirically by an ample analysis on real data in \ref{apdx:mld_variants}, where we show that multi-time diffusion approach consistently outperforms in-painting.

\section{Experiments}\label{sec:experiments}
\label{sec:exp}

We compare our method \gls{MLD} to \gls{MVAE} \cite{mvaeWu2018}, \gls{MMVAE} \cite{mmvaeShi2019}, \gls{MOPOE} \cite{sutter2021generalized}, \gls{NEXUS} \cite{Vasco2022} and \gls{MVTCAE} \cite{tcmvaehwang21}, \gls{MMVAEplus} \cite{palumbo2023mmvae} re-implementing competitors in the same code base as our method, and selecting their best hyper-parameters (as indicated by the authors). For fair comparison, we use the same encoder/decoder architecture for all the models. For \gls{MLD}, the score network is implemented using a simple stacked \gls{MLP} with skip connections (see \ref{apdx:mld_details} for more details).

\noindent \textbf{Evaluation metrics.} \textit{Coherence} is measured as in \cite{mmvaeShi2019, sutter2021generalized, palumbo2023mmvae}, using pre-trained classifiers on the generated data and checking the consistency of their outputs. \textit{Generative quality} is computed using \gls{FID} \cite{fid} and \gls{FAD} \cite{fad} scores for images and audio respectively. Full details on the metrics are included in \ref{apdx:dataset_eval}. All results are averaged over 5 seeds (We report standard deviation in \ref{apdx:additionnal_res}).

\noindent \textbf{Results.} Overall, \gls{MLD} largely outperforms alternatives from the literature, \textbf{both} in terms of coherence and generative quality. \gls{VAE}-based models suffer from a coherence--quality trad-off and modality collapse for highly heterogeneous data-sets. We proceed to show this on several standard benchmarks from the multi-modal \gls{VAE}-based literature (see \ref{apdx:dataset_eval} for details on the data-sets).

\begin{figure}[h]
     \centering

\begin{subfigure}{0.15\textwidth}
         \centering
         \includegraphics[width=\linewidth]{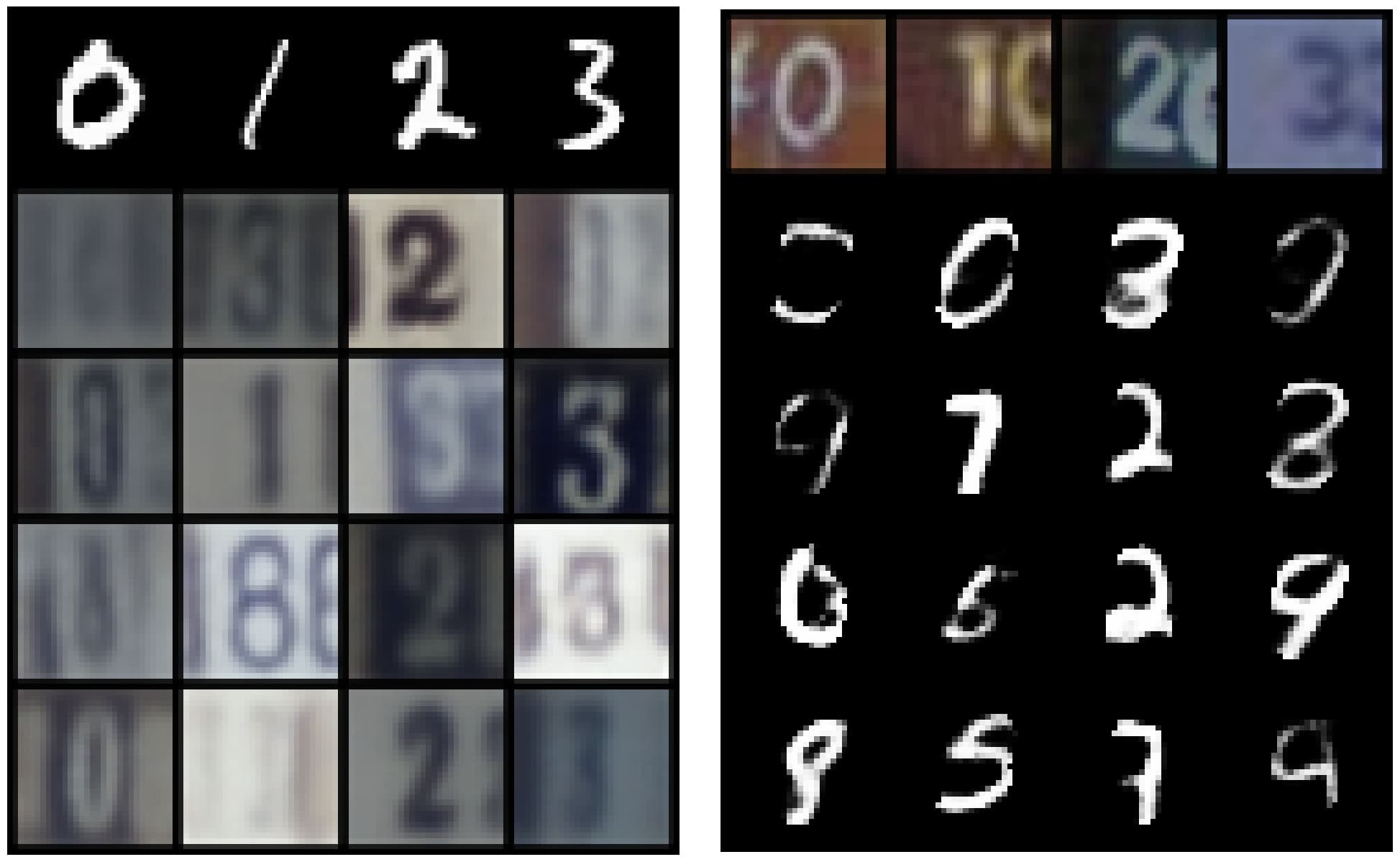}
    
\caption*{\gls{MVAE}}      
\end{subfigure}
\begin{subfigure}{0.15\textwidth}
         \centering
         \includegraphics[width=\linewidth]{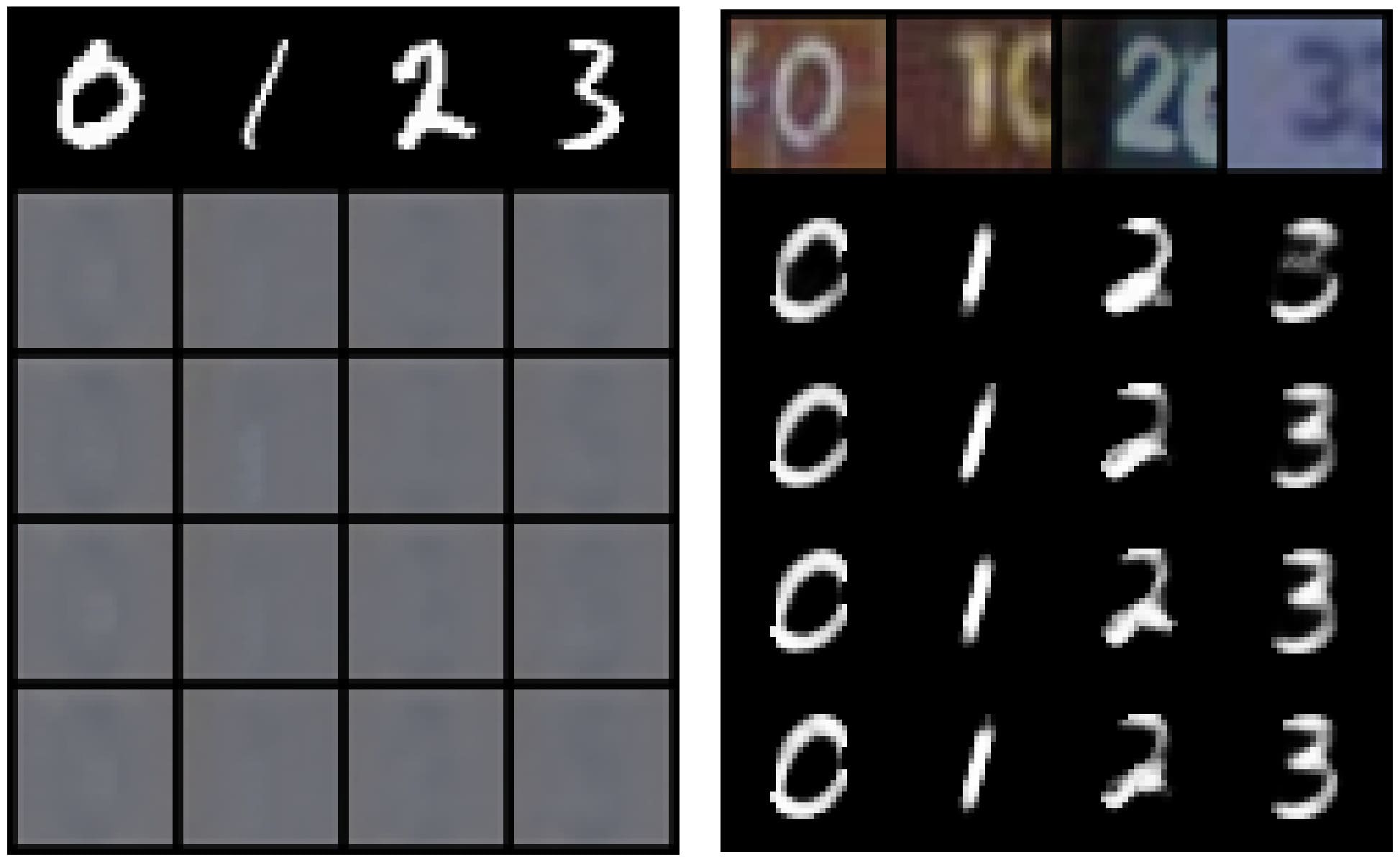}
         \caption*{\gls{MMVAE}}
\end{subfigure}
\begin{subfigure}{0.15\textwidth}
         \centering
         \includegraphics[width=\linewidth]{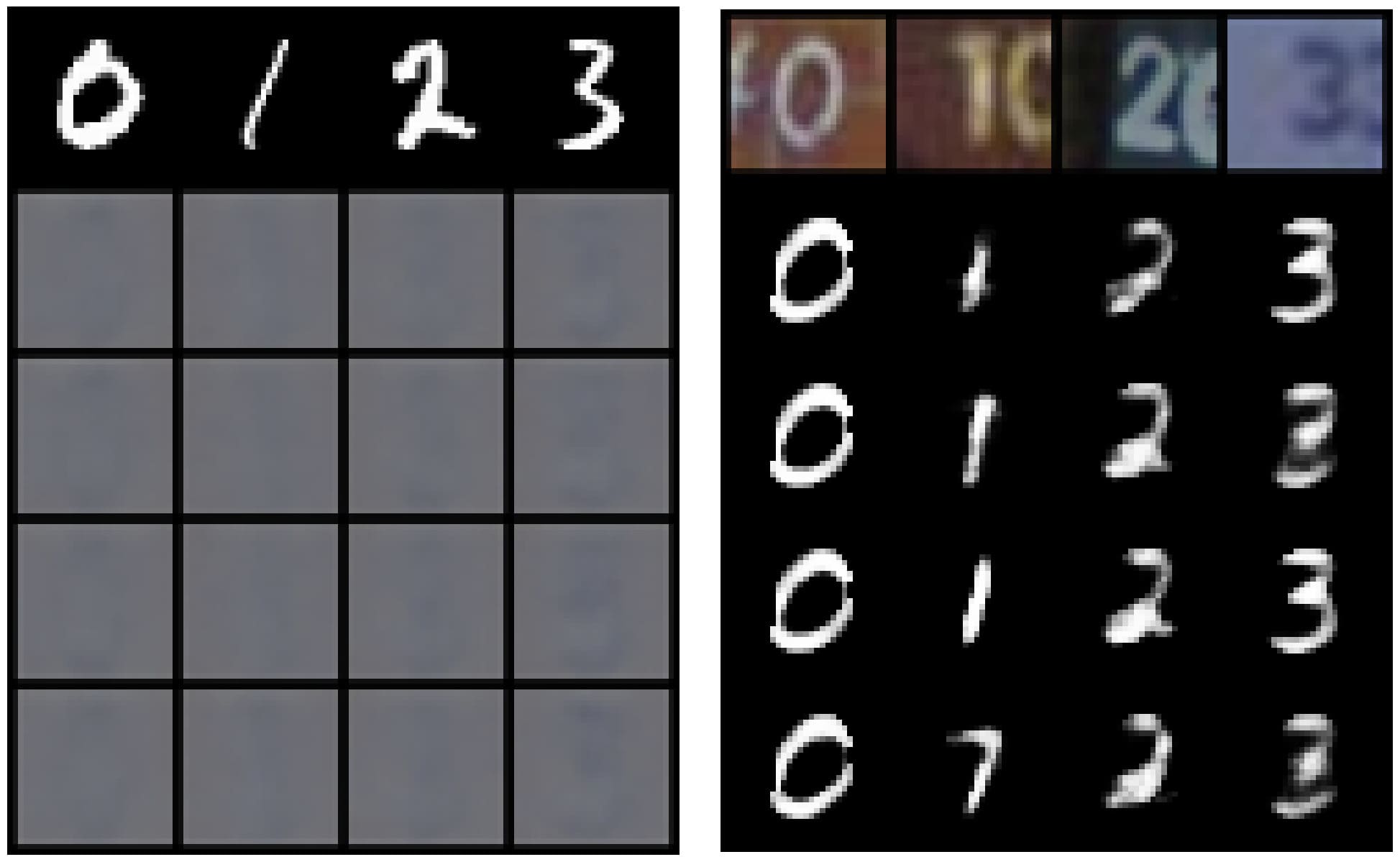}
         \caption*{\gls{MOPOE}}
     \end{subfigure}
\begin{subfigure}{0.15\textwidth}
         \centering
         \includegraphics[width=\linewidth]{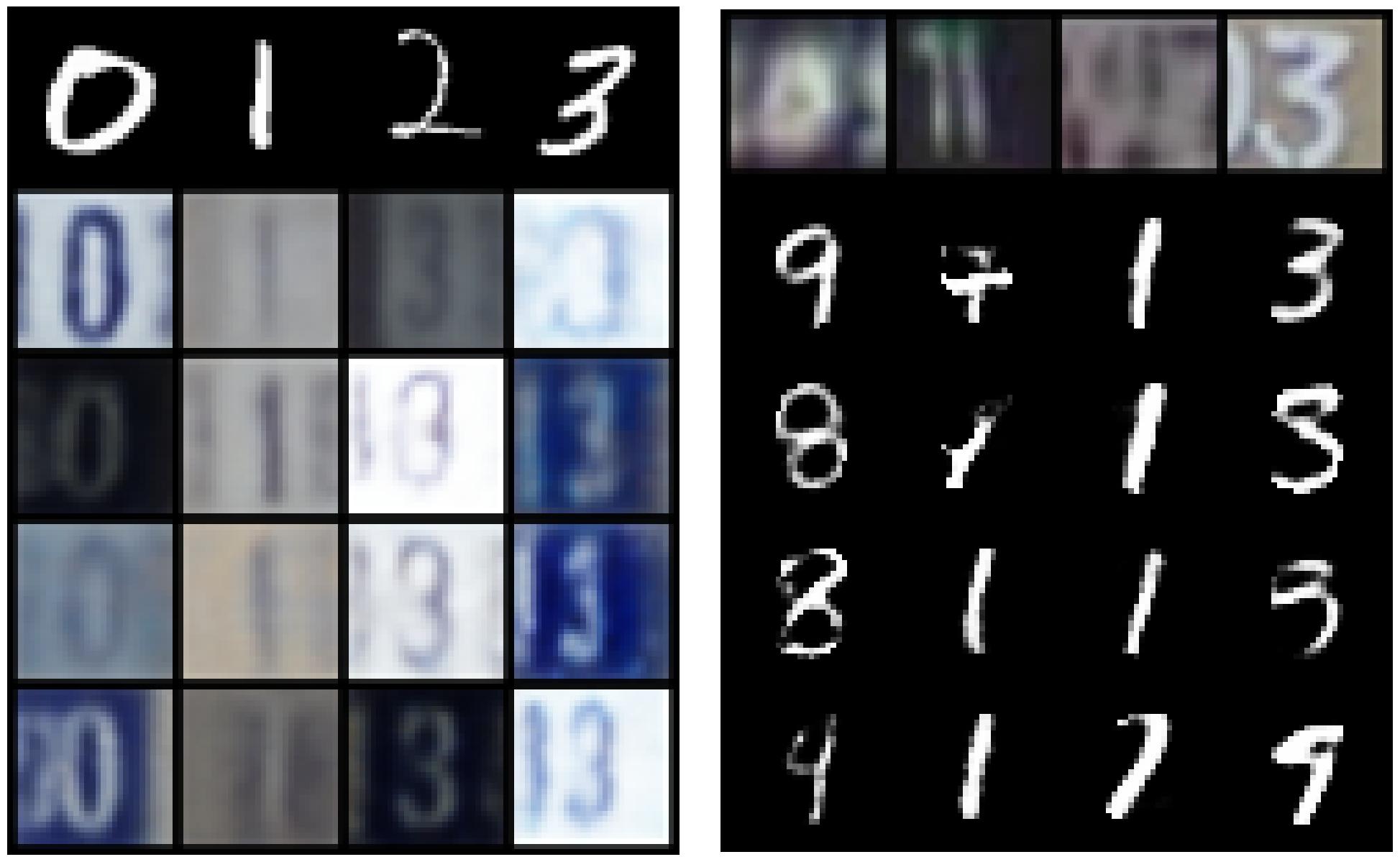}
    \caption*{\gls{MMVAEplus}(10)}
    \end{subfigure}
\begin{subfigure}{0.15\textwidth}
         \centering
         \includegraphics[width=\linewidth]{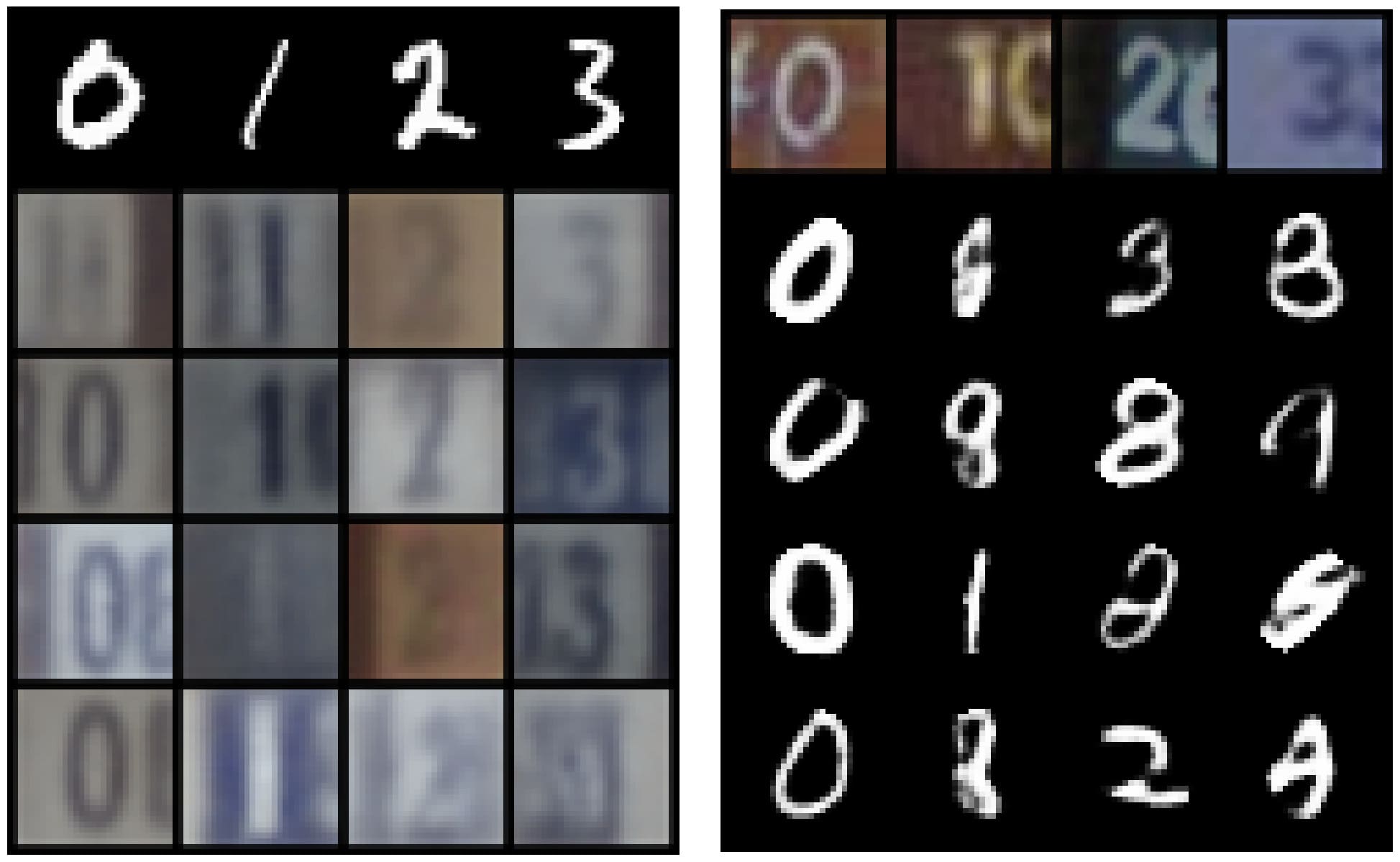}
         \caption*{\gls{MVTCAE}}
\end{subfigure}
  \begin{subfigure}{0.15\textwidth}
         \centering
         \includegraphics[page=1,width=\linewidth]{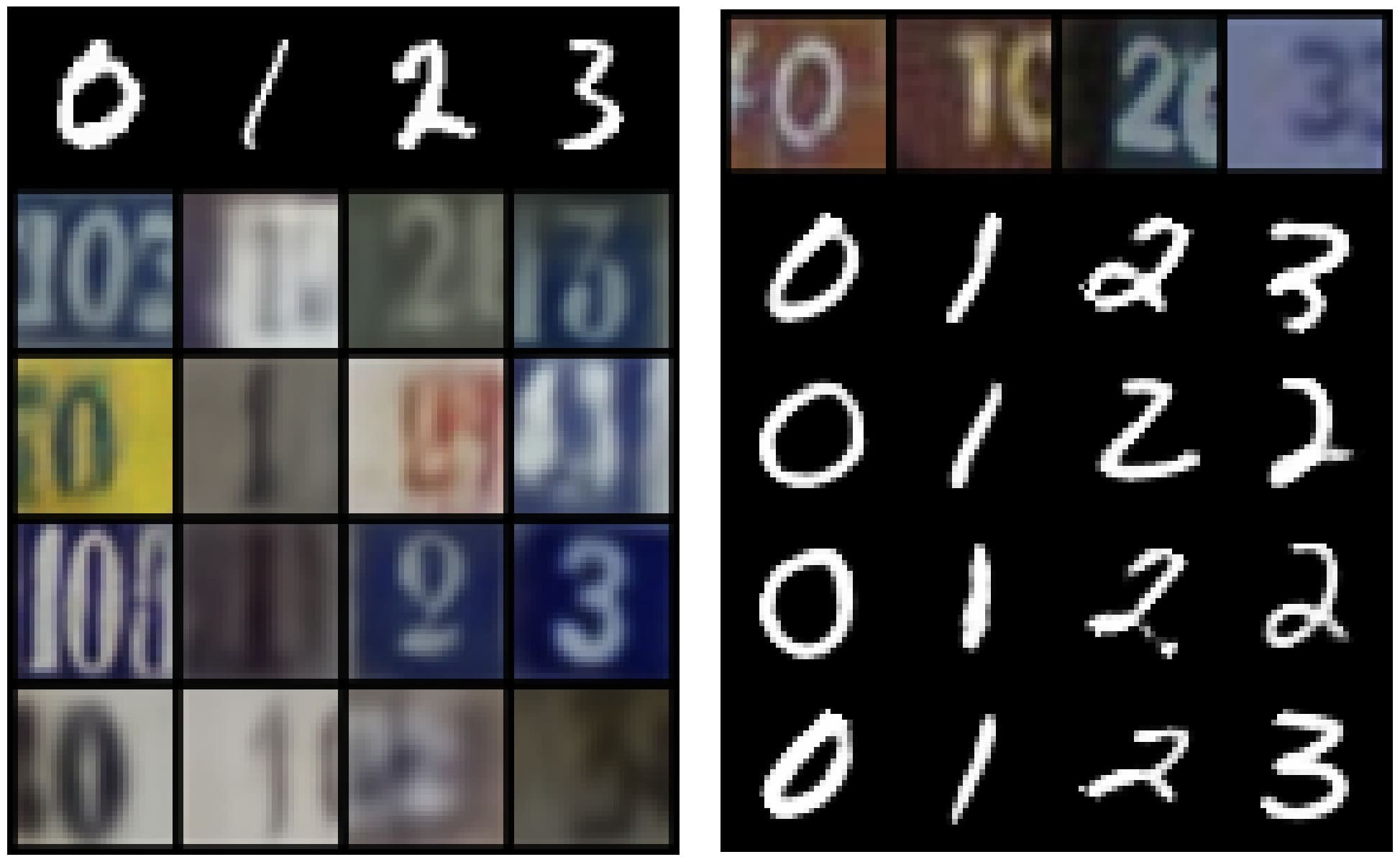}
         \caption*{\textbf{\gls{MLD} (ours)}}
     \end{subfigure}

\caption{Qualitative results for \textbf{\mnist-\svhn}. For each model we report: \mnist to \svhn conditional generation in the left,   \svhn to \mnist conditional generation in the right.}
        \label{fig:cond_joint_ms}
\end{figure}

The first data-set we consider is \textbf{\mnist-\svhn} (\citep{mmvaeShi2019}), where the two modalities differ in complexity.  
High variability, noise and ambiguity makes attaining good coherence for the \svhn modality a challenging task. Overall, \gls{MLD} outperforms all \gls{VAE}-based alternatives in terms of coherency, especially in terms of joint generation and conditional generation of \mnist given \svhn, see \Cref{coh_quality:ms}. Mixture models (\gls{MMVAE}, \gls{MOPOE}) suffer from modality collapse (poor \svhn generation), whereas product of experts (\gls{MVAE}, \gls{MVTCAE}) generate better quality samples at the expense of \svhn to \mnist conditional coherence. Joint generation is poor for all \gls{VAE} models. Interestingly, these models also fail at \svhn self-reconstruction which we discuss in \ref{apdx:additionnal_res}. \gls{MLD} achieves the best performance also in terms of generation quality, as confirmed also by qualitative results (\Cref{fig:cond_joint_ms}) showing for example how \gls{MLD} conditionally generates multiple \svhn digits within one sample, given the input \mnist image, whereas other methods fail to do so.

\begin{table}[h]
\caption{Generation coherence and quality for \textbf{\mnist-\svhn} ( M :\mnist, S: \svhn ). The generation quality is measured in terms of \gls{FMD}  for \mnist and \gls{FID} for \svhn.}
\centering
\tiny
\begin{tabular}{c|c|cc|cc|cc}
\toprule
\multirow{2}{*}{ Models }  & \multicolumn{3}{c}{Coherence (\%$\uparrow$) } &   \multicolumn{4}{c}{Quality ($\downarrow$)} \\
    \cmidrule{2-8}
    & Joint &  M $\rightarrow$ S & S $\rightarrow$ M &
     Joint(M)  &  Joint(S) &M $\rightarrow$ S &  S $\rightarrow$ M  \\
    \midrule
     \gls{MVAE} &  
     $38.19$ &
     $48.21$  &
     $ 28.57$ & 
     $13.34$ &
     $68.9$   & 
     $ \underline{68.0}  $ &
     $ 13.66 $ 
    
     \\
    \gls{MMVAE} &  $37.82$ & $11.72$ & $ 67.55$ & $25.89$ &
    $ 146.82  $& $393.33$  &
    $ 53.37 $ 
 
    \\
    \gls{MOPOE} & $ 39.93$ & $12.27$ & $68.82$ &$20.11$ &
    $ 129.2  $&$373.73$&
    $ 43.34 $
        \\

    \gls{NEXUS} &  $ 40.0$ & $16.68$ & $\underline{70.67}$ &  $13.84$ &
        $ 98.13  $ &$281.28$&
 $ 53.41 $
   \\

    \gls{MVTCAE} & $\underline{48.78} $& $\underline{81.97}$ & $49.78 $&$\underline{12.98}$ &
    $ \textbf{52.92} $&$69.48$ &
    $ \underline{13.55} $ 
\\ 
\gls{MMVAEplus} &17.64 &13.23 & 29.69 &26.60 & 121.77 & 240.90&35.11
\\
  \gls{MMVAEplus}(K=10) & 41.59 & 55.3
& 56.41 & 19.05 & 67.13 & 75.9 & 18.16

      \\
     
    \midrule
    \textbf{\gls{MLD} (ours)}  & $\textbf{85.22}$ & 
    $\textbf{83.79} $&
    $ \textbf{79.13}$ &
    $\textbf{3.93}$ &
    $\underline{56.36}$   &
    $\textbf{57.2}$  &
    $\textbf{3.67} $ 
    \\
    \bottomrule
   
\end{tabular}
\label{coh_quality:ms}
\end{table}

The Multi-modal Handwritten Digits data-set (\textbf{\mhd}) \citep{Vasco2022} contains gray-scale digit images, motion trajectory of the hand writing and sounds of the spoken digits. In our experiments, we do not use the label as a forth modality. While digit image and trajectory share a good amount of information, the sound modality contains a lot more of modality specific variation. Consequently, conditional generation involving the sound modality, along with joint generation, are challenging tasks.
Coherency-wise (\Cref{coh:mhd}) \gls{MLD} outperforms all the competitors where the biggest difference is seen in joint and sound to other modalities generation (in the latter task \gls{MVTCAE} performs better than other competitors but is still worse than \gls{MLD}). \gls{MLD} dominates alternatives also in terms of generation quality (\Cref{qua:mhd}). 
This is true both for image, sound modalities, for which some \gls{VAE}-based models suffer in producing high quality results, demonstrating the limitation of these methods in handling highly heterogeneous modalities. \gls{MLD}, in the other hand, achieves high generation quality for all modalities, possibly due to the independent training of the autoencoders avoiding interference.

\begin{table}[ht]
\centering
\tiny
\caption{Generation Coherence (\%) for \textbf{\mhd} (Higher is better). Line above refer to the generated modality while the observed modalities subset are presented below.
}  

\begin{tabular}{c|c|ccc|ccc|ccc}
\toprule
\multirow{2}{*}{ Models }  & \multirow{2}{*}{ Joint }  & \multicolumn{3}{c}{I (Image)}  & \multicolumn{3}{c}{T (Trajectory)}  & \multicolumn{3}{c}{S (Sound) } 
\\
\cmidrule{3-11}
    &&  T & S  & T,S  & I & S  & I,S & I & T & I,T \\
    \midrule
    \gls{MVAE} &$37.77$ &
    
    $11.68$ &
    $26.46$ & 
    $ 28.4$ & 
    
    $95.55$  &
    $26.66$ &
     $96.58$ &  
    
    $58.87$ & 
    $10.76$  & 
    $58.16$    \\

     \gls{MMVAE} &    
    $34.78$ &
    
    $\textbf{99.7}$ &
    $69.69 $ & 
    $ 84.74$ & 
    $\underline{99.3}$  &
    $85.46 $  & 
    $92.39$ & 
    $49.95$ & 
    $50.14$  & 
    $50.17$   \\
   
     \gls{MOPOE}   &   $48.84$ &
    $\underline{99.64}$ & $68.67$ & $ \underline{99.69}$ & 
    $99.28$  & $\underline{87.42}$  & $99.35$ &  
    $50.73$ & $51.5$  & $56.97$   \\

     \gls{NEXUS} &  $26.56$ &
    $94.58$ & $\underline{83.1}$ & $ 95.27$ & 
    $88.51$  & $76.82$  & $93.27$ &  
    $70.06$ & $75.84$  & $89.48$ 
    \\
      \gls{MVTCAE} &  $42.28$ 
     &
    $99.54$ & $72.05$ & $ 99.63$ & 
    $99.22$  & $72.03$  & $\underline{99.39}$ &  
    $\underline{92.58}$ & $\underline{93.07}$  & $\underline{94.78}$  \\

      \gls{MMVAEplus} &  $41.67$ 
     &
    $98.05$ & $84.16$ & $ 91.88_{\pm }$ & 
    $97.47$  & $81.16$  & $89.31$ &  
    $64.34$ & 
    $65.42$  &
    $64.88$  \\

   \gls{MMVAEplus}(k=10) &  $42.60$ 
     &
    $99.44$ & $\textbf{89.75}$ & $ 94.7$ & 
    $99.44$  
    & $\textbf{89.58}$  & $95.01$ &  
    $87.15$ & 
    $87.99$  &
    $87.57$  
    
\\
    \midrule
\textbf{\gls{MLD} (ours)} &  
    $\textbf{98.34}$ &
    $99.45$ &
    $\underline{88.91}$ & 
    $ \textbf{99.88}$ & 
    
    $\textbf{99.58}$  & 
    $\underline{88.92}$  & 
    $\textbf{99.91}$ &  
    
    $\textbf{97.63}$ & 
    $\textbf{97.7}$ 
    & $\textbf{98.01}$  \\
    \bottomrule

\end{tabular}

 \label{coh:mhd}

\end{table}

\begin{table}[h]
\centering
\tiny
\caption{Generation quality for \textbf{\mhd} in terms of \gls{FMD} for image and trajectory modalities and \gls{FAD} for the sound modality (Lower is better).}
\resizebox{0.9\textwidth}{!}{\begin{tabular}{c|cccc|cccc|cccc}
\toprule
\multirow{2}{*}{ Models }  & \multicolumn{4}{c}{I (Image)}  & \multicolumn{4}{c}{T (Trajectory)}  & \multicolumn{4}{c}{S (Sound) } 
\\
\cmidrule{2-13}
   & Joint & T & S  & T,S    &  Joint  & I & S  & I,S  & Joint &  I & T & I,T   \\
    \midrule
    \gls{MVAE} &
    $\underline{94.9}$  
    & $93.73$ &
    $92.55$  &
    $91.08$  &

    $39.51$   &
    $20.42$ &
    $38.77$  &
    $19.25$  &

    $14.14$ &
  $\underline{14.13}$ &
    $14.08$  &
    $14.17$   

    \\
  \gls{MMVAE} 
  &$224.01$   
  &  $22.6$ &
    $789.12$  &
    $170.41$  &

    $16.52$   &
    $\textbf{0.5}$ &
    $30.39$  &
    $6.07$  &

     $22.8$  &
    $22.61$ &
    $23.72$  &
    $23.01$

\\
    \gls{MOPOE} & 
    $147.81$   & 
    $16.29$ &
    $838.38$  &
    $15.89$  &

$\underline{13.92}$   & 
$\underline{0.52}$ &
    $33.38$  &
    $\textbf{0.53}$  &

    $18.53$ &
    $24.11$ &
    $24.1$  &
    $23.93$ 
    
    \\
    \gls{NEXUS} &   
    $281.76$   &     
    $116.65$ &
    $282.34$  &
    $117.24$  &

    $18.59$   &
    $6.67$ &
    $33.01$  &
    $7.54$  &

    $\underline{13.99}$ &
    $19.52$ &
    $18.71$  &
    $16.3$    

    \\
    \gls{MVTCAE} &
    $121.85$  &   $\underline{5.34}$ &
    $\underline{54.57}$  &
    $\underline{3.16}$  &
     
    $19.49$   &
    $0.62$ &
    $\underline{13.65}$  &
    $0.75$  &

    $15.88$   &
    $14.22$ &
    $\underline{14.02}$  &
    $\underline{13.96}$  
    \\

   \gls{MMVAEplus} &    
   
   $97.19$   &  
   $2.80$ &
    $128.56$  &
    $ 114.3$  &

        $22.37$   & 
    $1.21$ &
    $21.74$  &
    $15.2$  &

        $16.12$   &
        $17.31$ &
    $17.92$  &
    $17.56$
    
    \\

     \gls{MMVAEplus}(K=10) &    
   
   $85.98$   &  
   $1.83$ &
    $70.72$  &
    $62.43$  &

        $21.10$   & 
    $1.38$ &
    $8.52$  &
    $7.22$  &

   $14.58$   &  
   $14.33$ &
    $14.34$  &
    $14.32$  
    \\
    \midrule
    \gls{MLD} 
    &   
    $\textbf{7.98}$ &
    $\textbf{1.7}$ &
    $\textbf{4.54}$ & 
    $\textbf{1.84}$  &

    $\textbf{3.18}$ 
      & $0.83$ 
     & $\textbf{2.07}$ 
       & $\underline{0.6}$

       & $ \textbf{2.39} $
        & $\textbf{2.31}$ 
     & $\textbf{2.33}$ 
       & $\textbf{	2.29}$ 
       
       \\

    \bottomrule
   
\end{tabular}
}

 \label{qua:mhd}

\end{table}

\begin{figure} [h]
\centering
\begin{subfigure}{0.24\textwidth}
\centering
     \begin{subfigure}{1\textwidth}
         \centering
         \includegraphics[page=1,width=\linewidth]{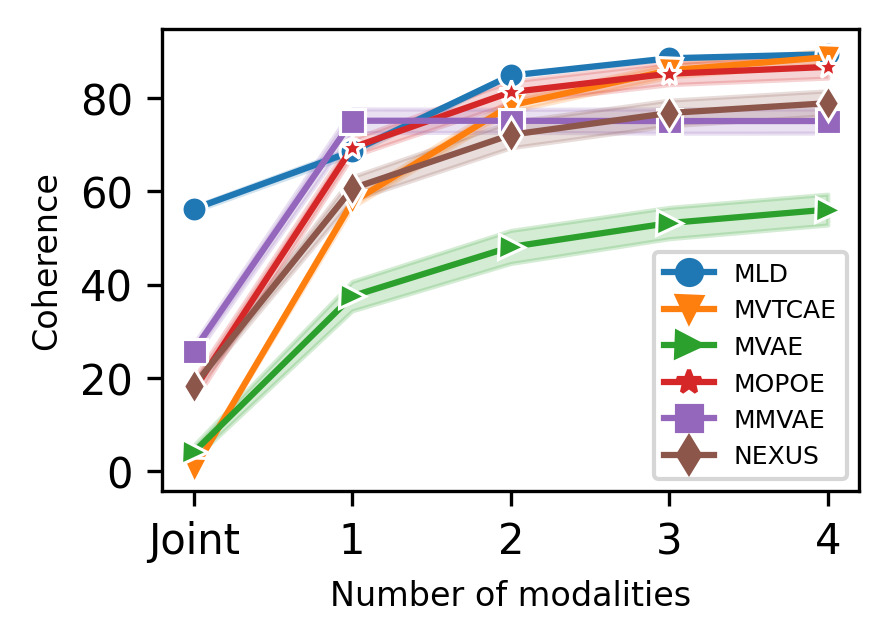}
   
     \end{subfigure}
     
     \begin{subfigure}{1\textwidth}
         \centering
         \includegraphics[width=\linewidth]{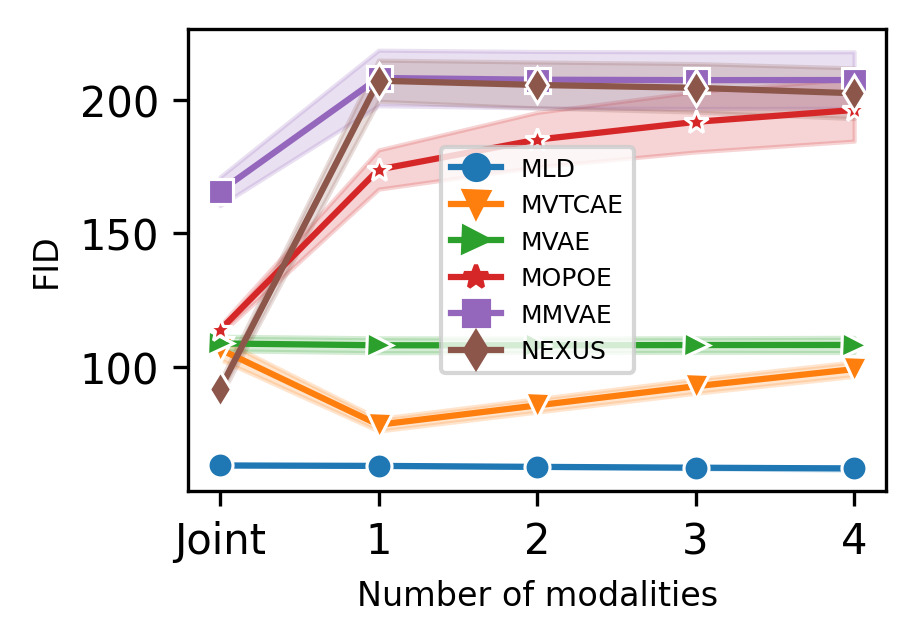}
   
     \end{subfigure}

\label{mmnist_quantitavie}
     \end{subfigure}
\begin{subfigure}{0.73\textwidth}
     \centering
     \begin{subfigure}{0.25\textwidth}
         \centering
         \includegraphics[width=\linewidth]{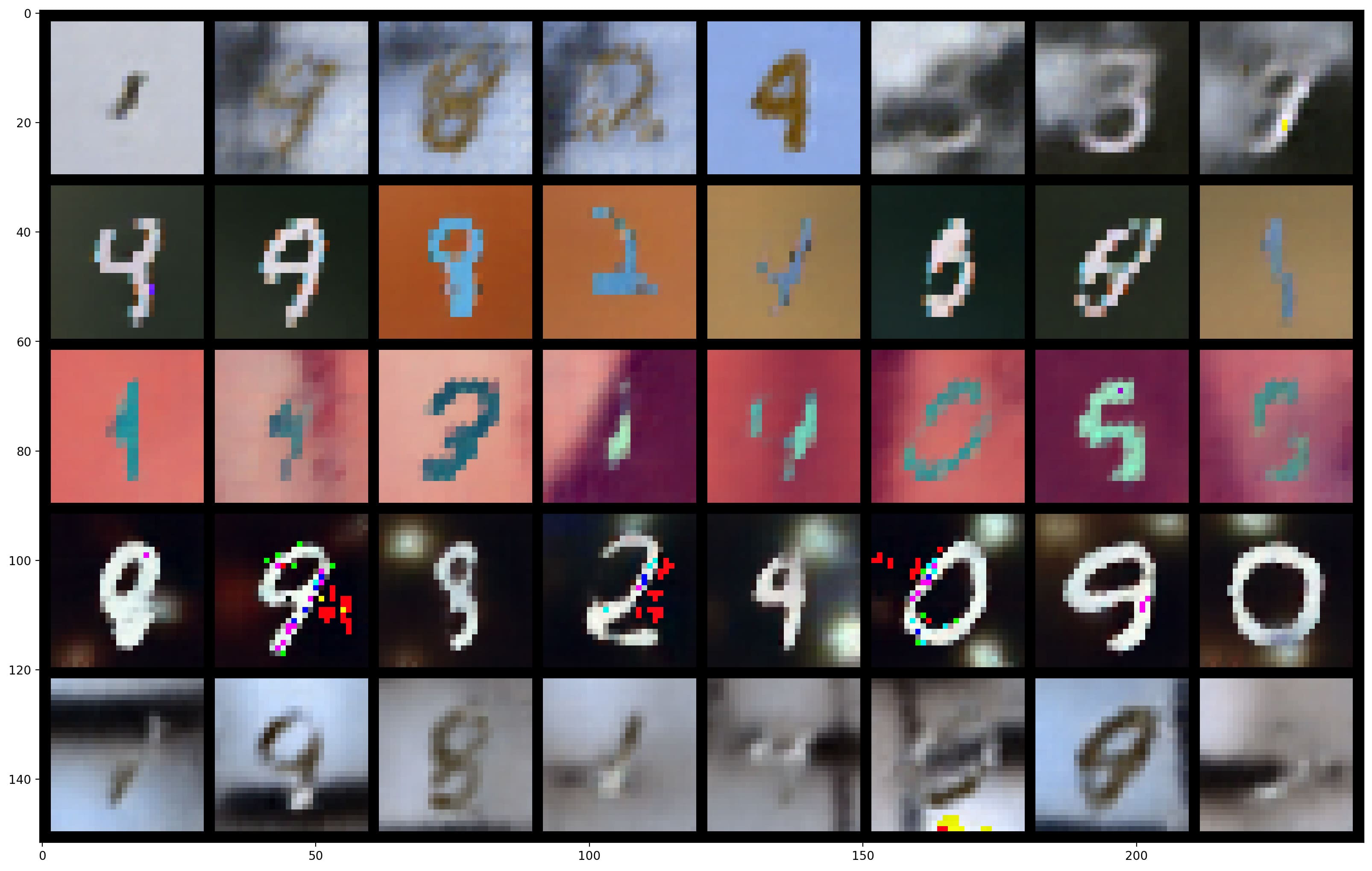}
         \caption*{\gls{MVAE}}
         
     \end{subfigure}
     \begin{subfigure}{0.25\textwidth}
         \centering
         \includegraphics[width=\linewidth]{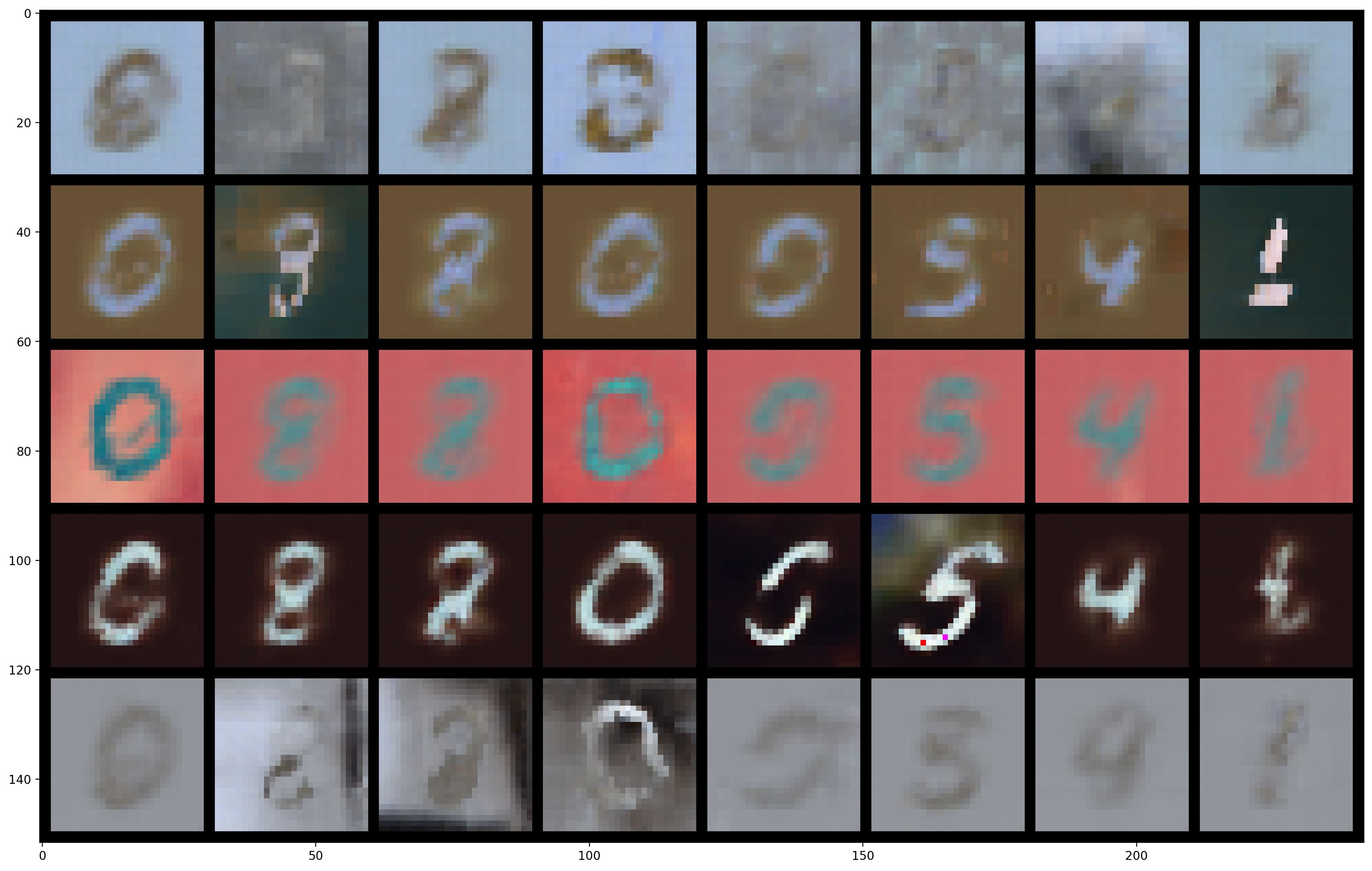}
         \caption*{\gls{MMVAE}}
      
     \end{subfigure}
       \begin{subfigure}{0.25\textwidth}
         \centering
         \includegraphics[width=\linewidth]{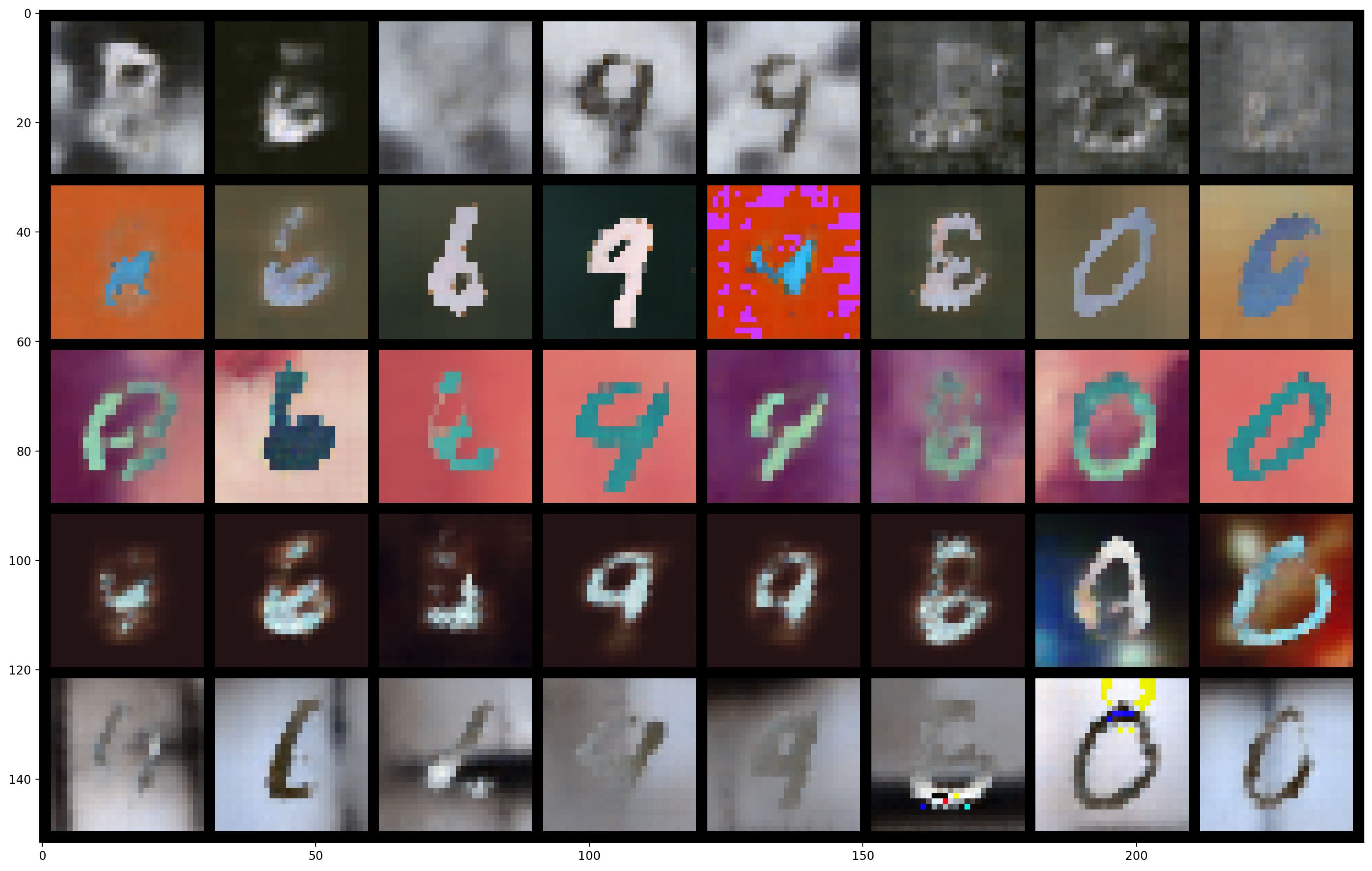}
         \caption*{\gls{MOPOE}}
     \end{subfigure} 
     
      \begin{subfigure}{0.25\textwidth}
         \centering
         \includegraphics[width=\linewidth]{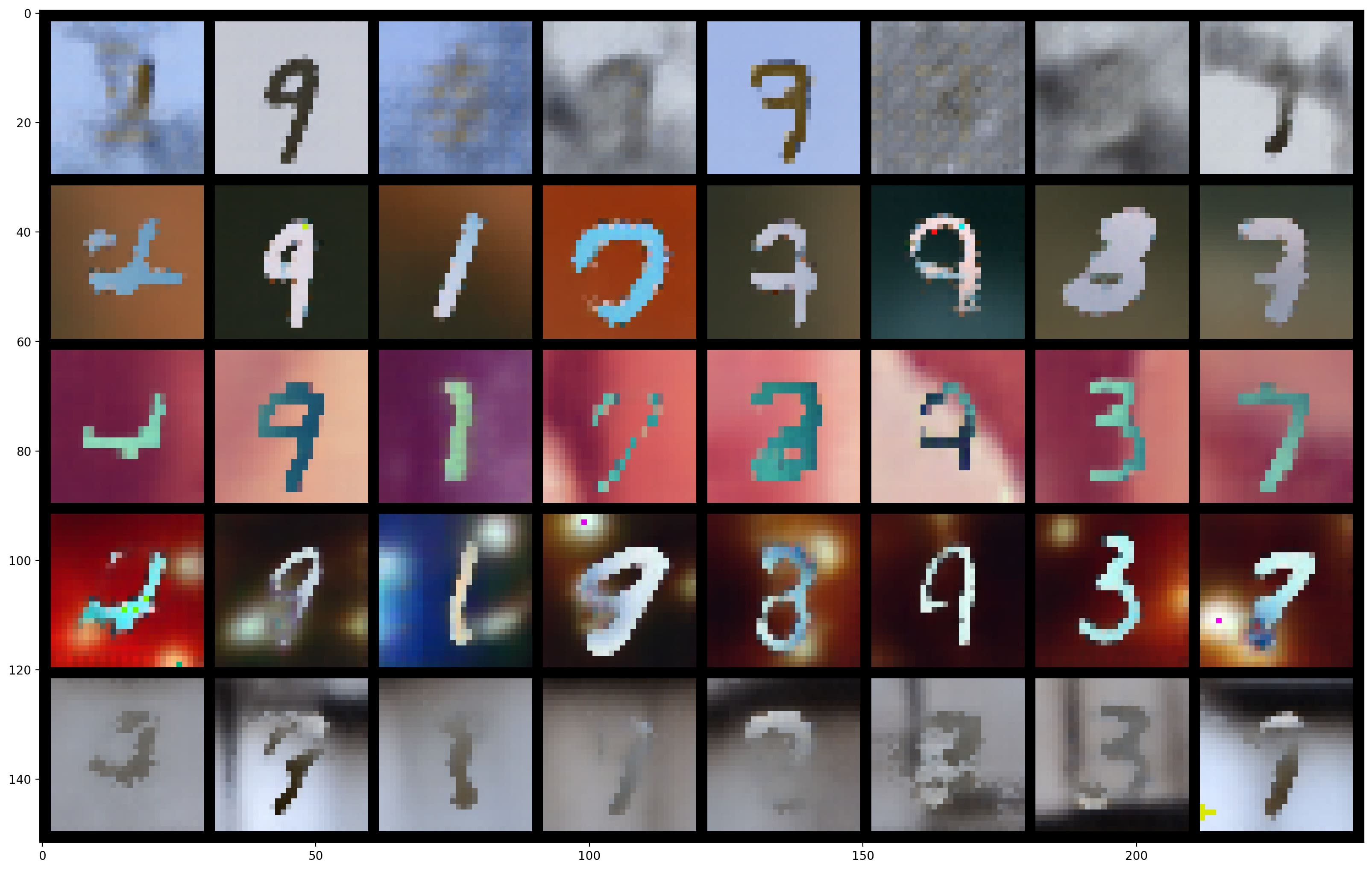}
         \caption*{\gls{NEXUS}}
     
     \end{subfigure}
  \begin{subfigure}{0.25\textwidth}
         \centering
         \includegraphics[width=\linewidth]{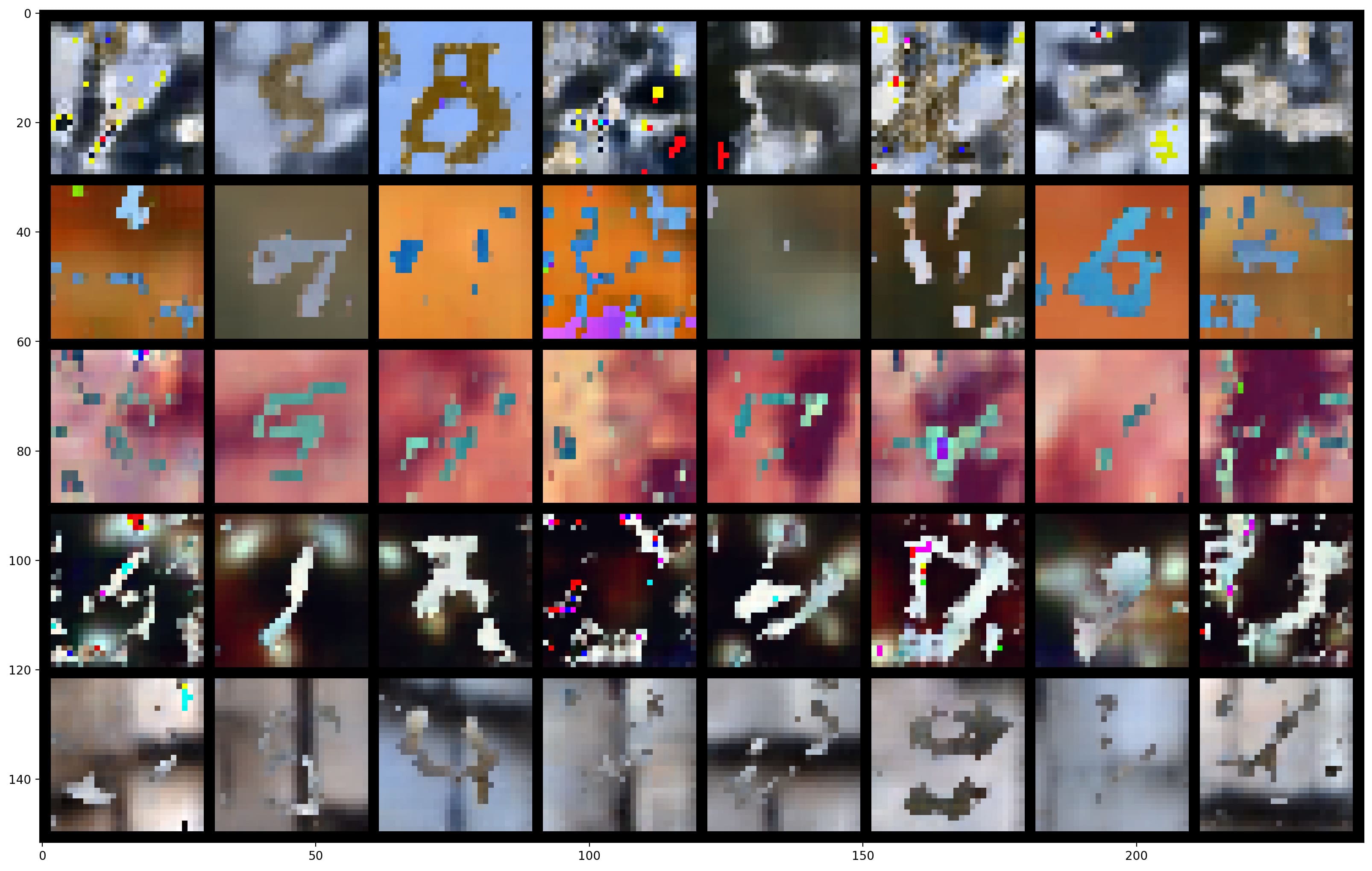}
         \caption*{\gls{MVTCAE}}
   
     \end{subfigure}
  \begin{subfigure}{0.25\textwidth}
         \centering
         \includegraphics[page=1,width=\linewidth]{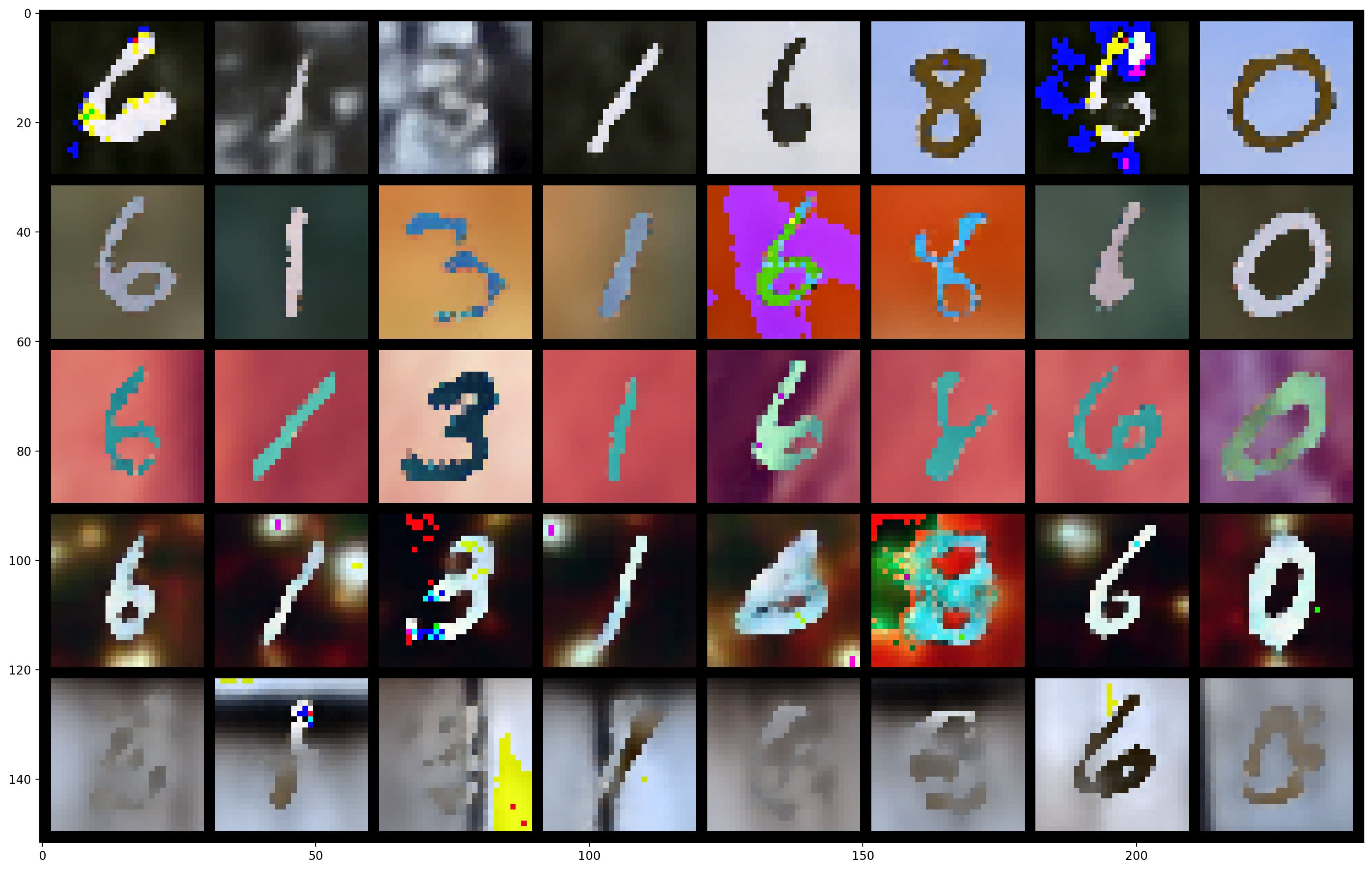}
         \caption*{\textbf{\gls{MLD} (ours) }}
       
     \end{subfigure}
    
       \label{qual_mmnist}
\end{subfigure}
        \caption{Results for \textbf{\polymnist} data-set. \textit{Left}: a comparison of the generative coherence (\% $\uparrow$) and quality in terms of \gls{FID} ($\downarrow$) as a function of the number of inputs. We report the average performance following the leave-one-out strategy (see \ref{apdx:dataset_eval}).  \textit{Right}: are qualitative results for the joint generation of the 5 modalities.   }
        \label{fig:res_mmnist}
\end{figure}

The \textbf{\polymnist} data-set \citep{sutter2021generalized} consists of 5 modalities synthetically generated by using \mnist digits and varying the background images. The homogeneous nature of the modalities is expected to mitigate gradient conflict issues in \gls{VAE}-based models, and consequently reduce modality collapse. However, \gls{MLD} still outperforms all alternatives, as shown \Cref{fig:res_mmnist}. Concerning generation coherence, \gls{MLD} achieves the best performance in all cases with the single exception of a single observed modality. On the qualitative performance side, not only \gls{MLD} is superior to alternatives, but its results are stable when more modalities are considered, a capability that not all competitors share. 

\begin{figure}[hb]
\vspace*{-3mm}
     \centering
     \begin{subfigure}{0.16\textwidth}
         \centering
         \includegraphics[width=\linewidth]{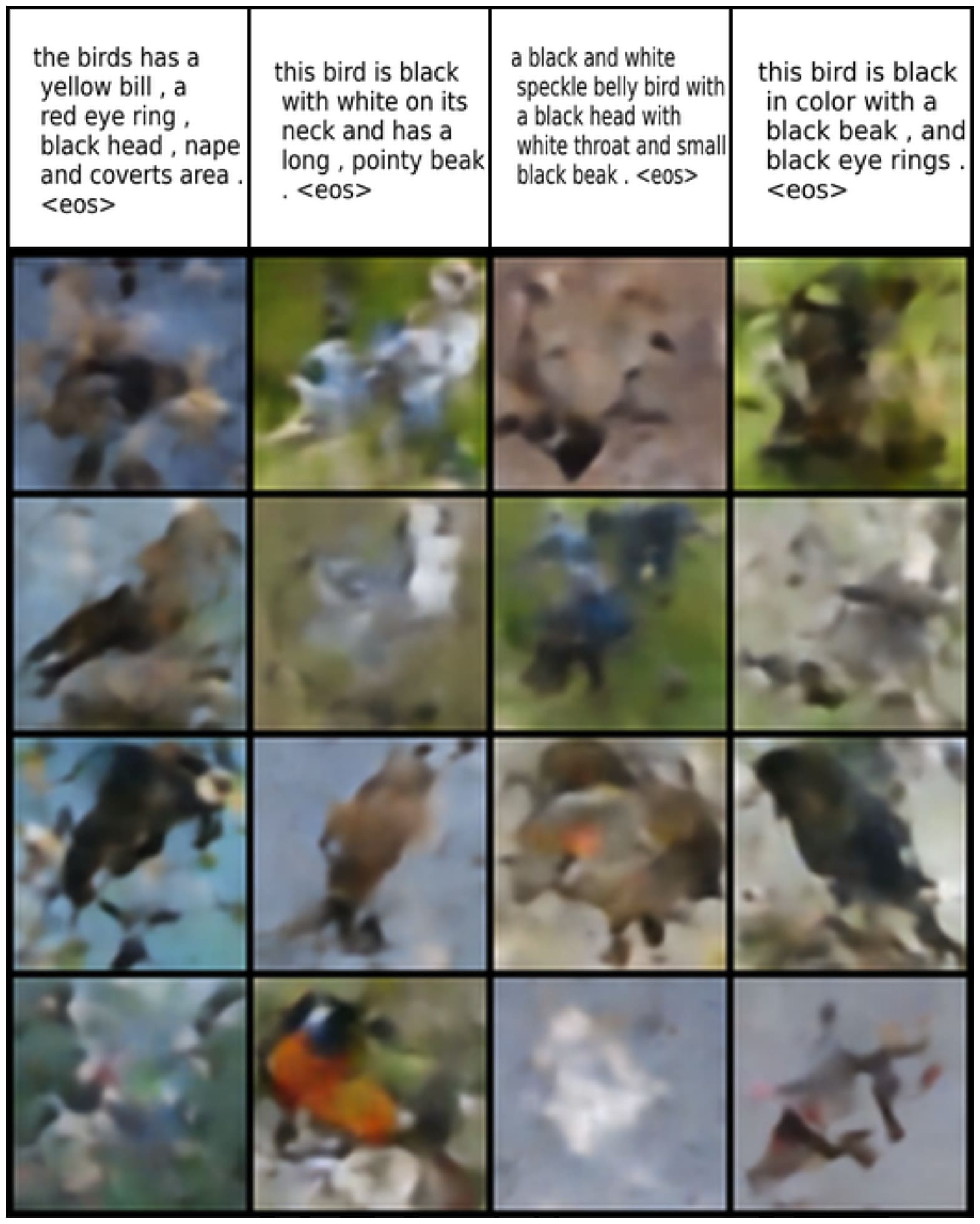}
        \caption*{\gls{MVAE}}
        \end{subfigure}
       \begin{subfigure}{0.16\textwidth}
         \centering
         \includegraphics[width=\linewidth]{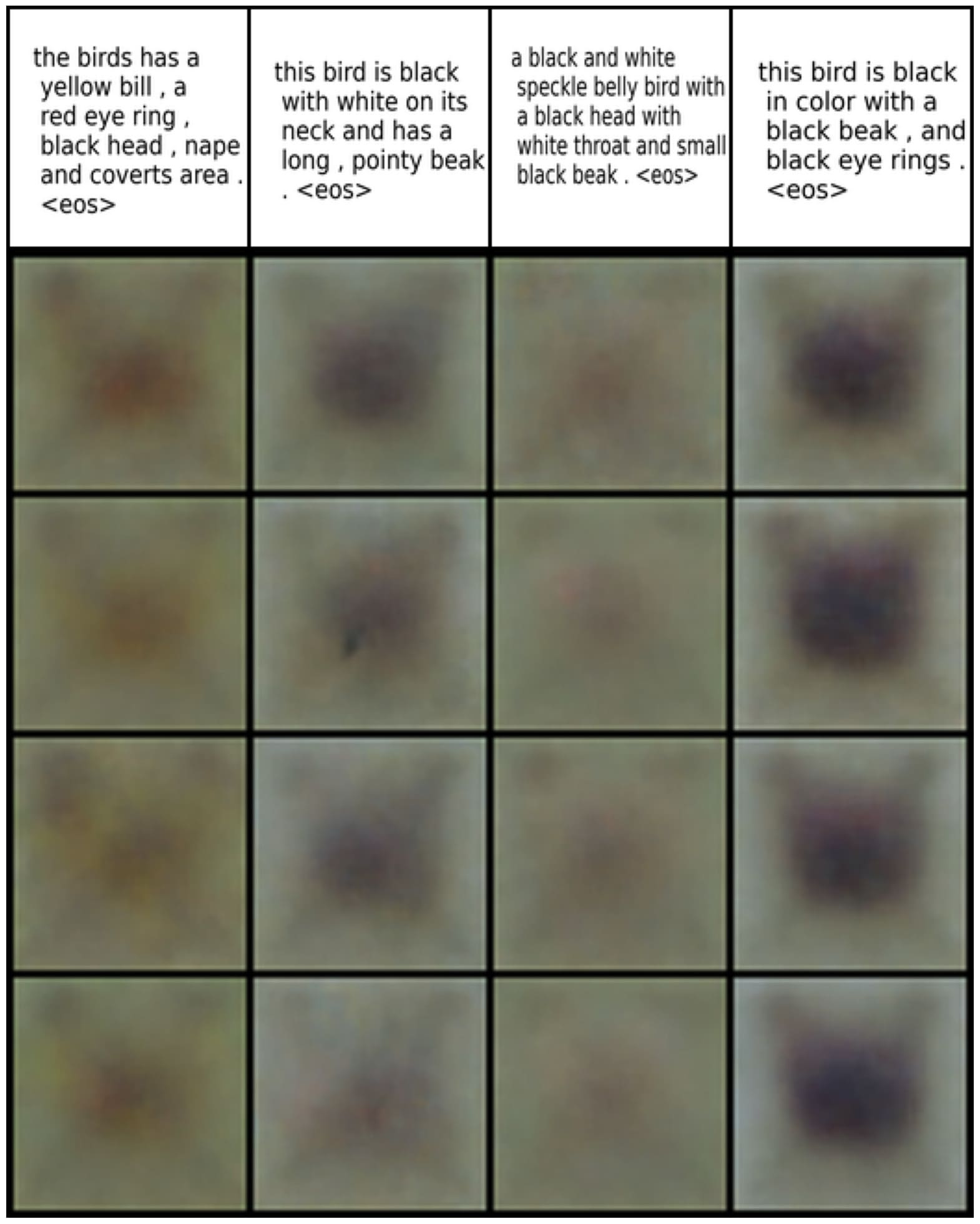}
         \caption*{\gls{MOPOE}}
     \end{subfigure}
     \begin{subfigure}{0.16\textwidth}
         \centering
         \includegraphics[width=\linewidth]{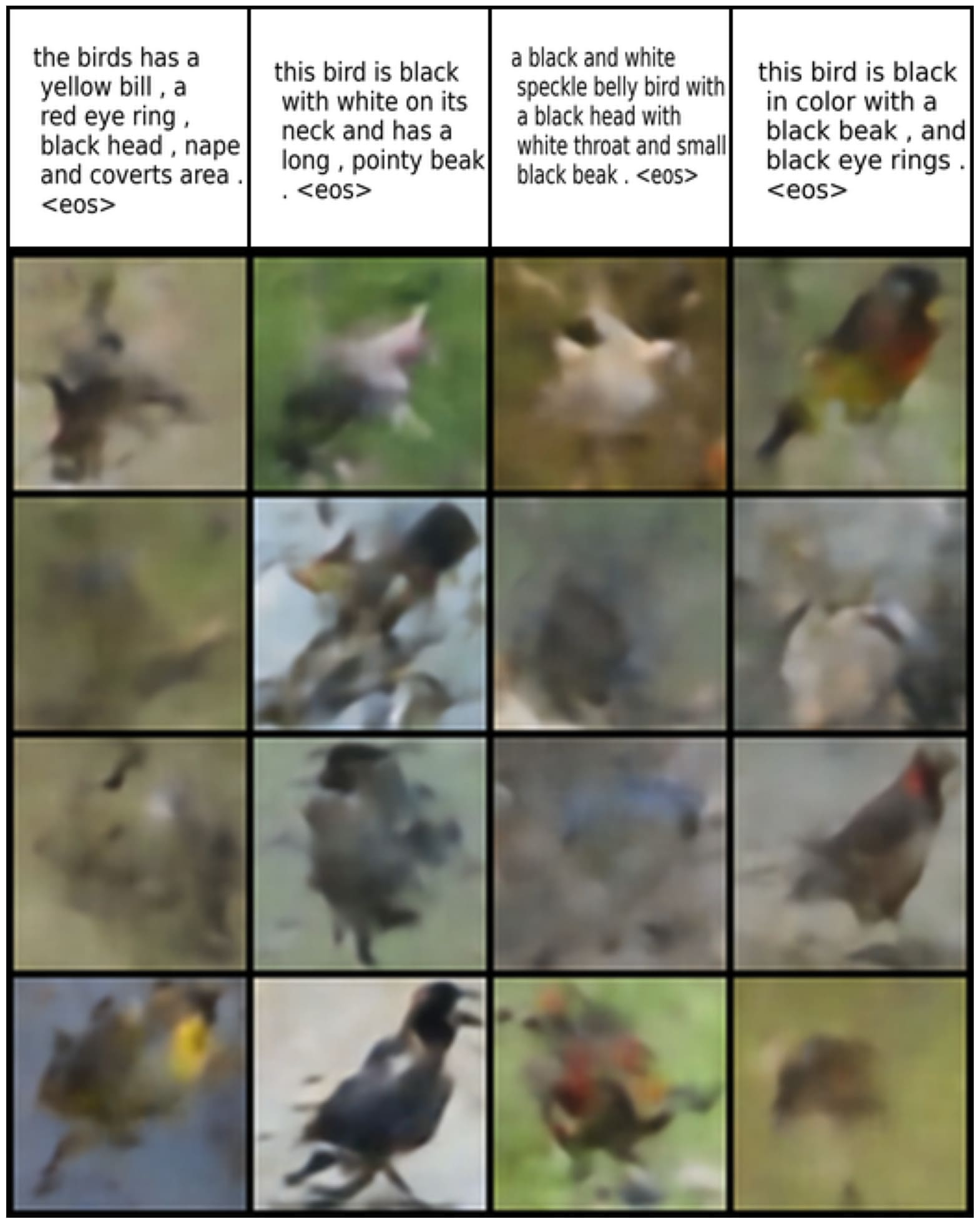}
         \caption*{\gls{MVTCAE}}
     \end{subfigure}
  \begin{subfigure}{0.16\textwidth}
         \centering
         \includegraphics[page=1,width=\linewidth]{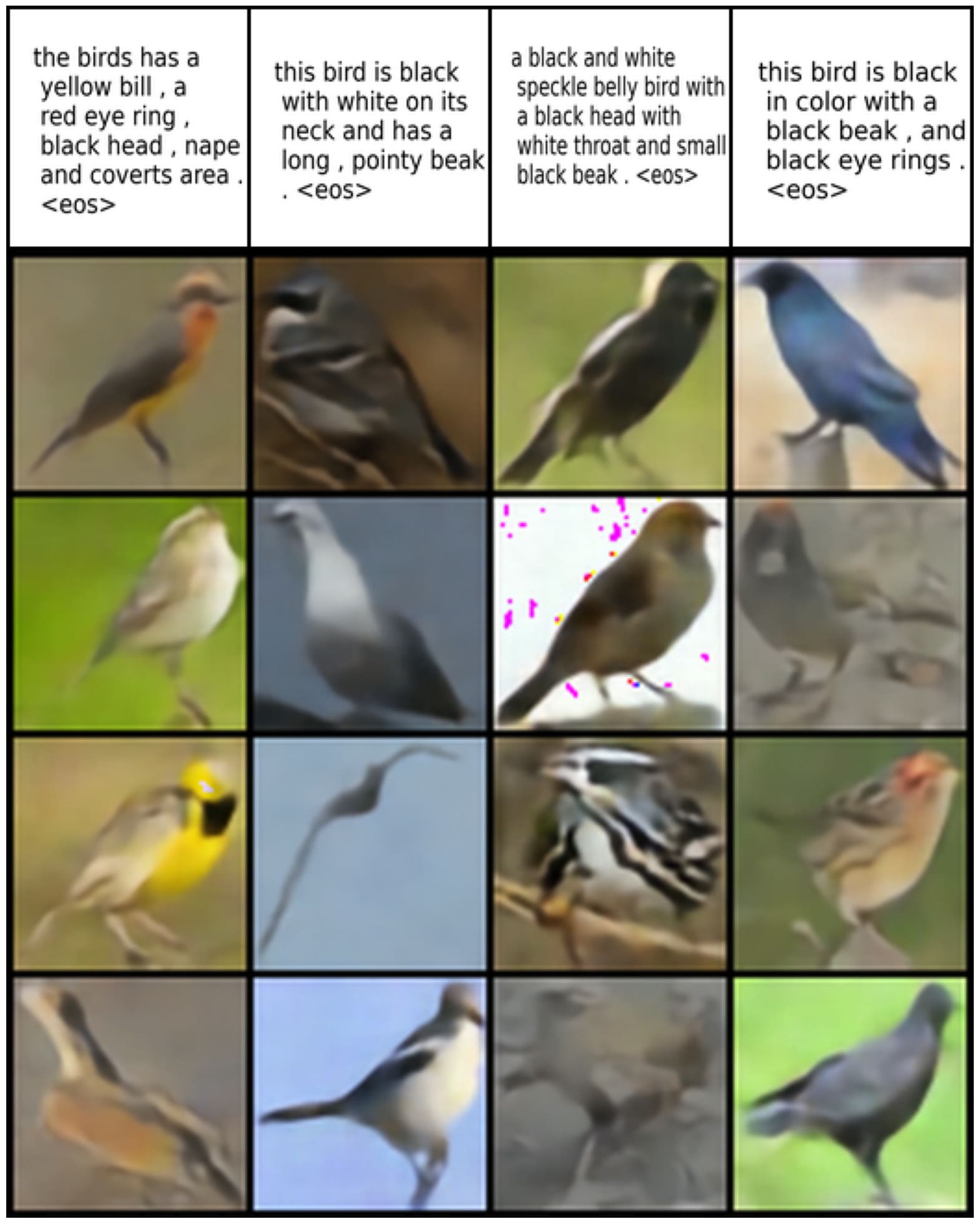}
         \caption*{\textbf{\gls{MLD} (ours) }}
   \end{subfigure}
 \begin{subfigure}{0.16\textwidth}
         \centering
         \includegraphics[page=1,width=\linewidth]{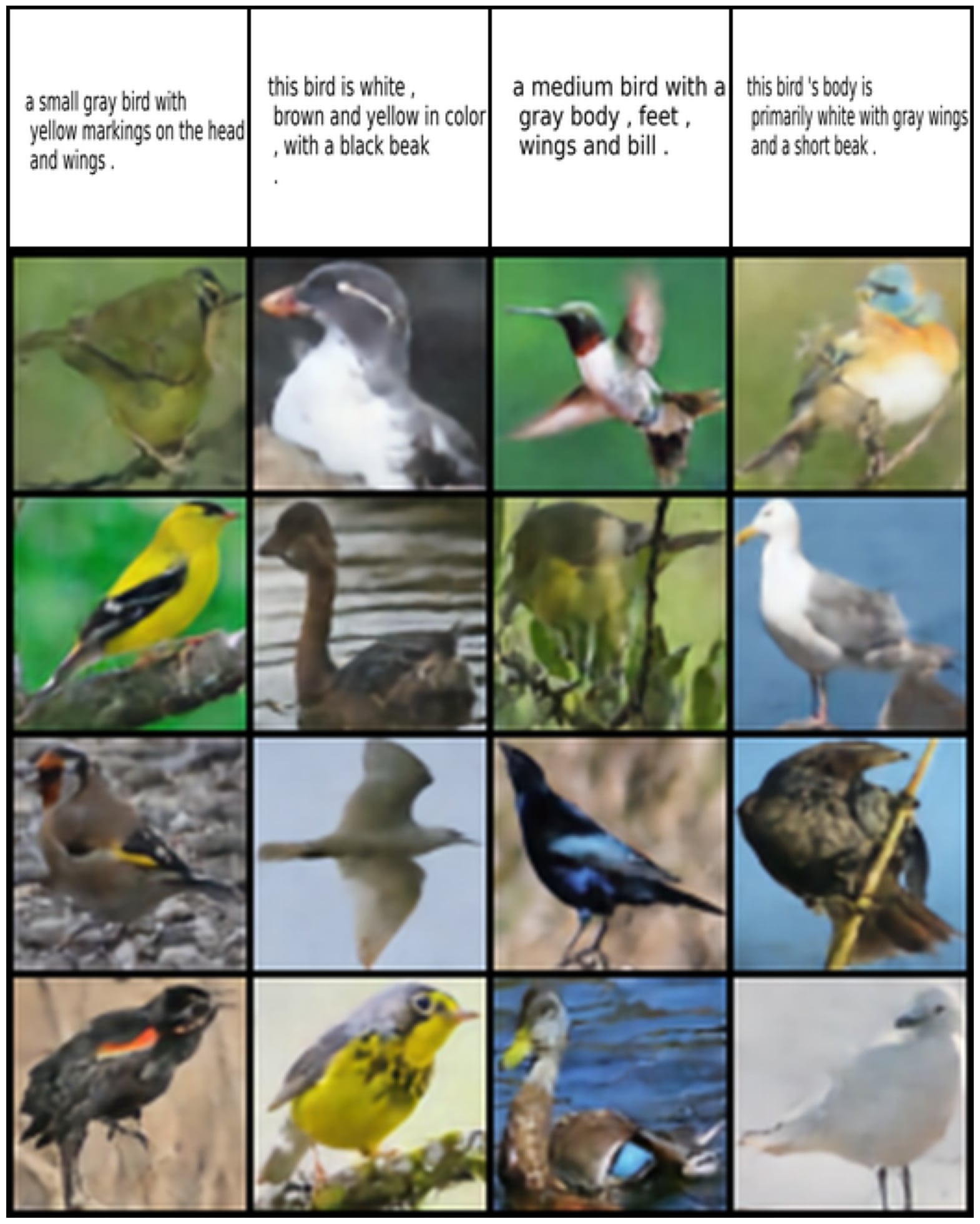}
         
         \caption*{\textbf{\gls{MLD}* (ours)  }}
   \end{subfigure}
        \caption{Qualitative results on \textbf{\cub} data-set. Caption used as condition to generate the bird images. \textbf{MLD}* denotes the version of our method using a powerful image autoencoder. }
       \label{cap_image_cub}
        
\end{figure}

Finally, we explore the Caltech Birds \textbf{\cub} \citep{mmvaeShi2019} data-set, following the same experimentation protocol in \cite{Daunhawer2021} by using real bird images (instead of ResNet-features as in \cite{mmvaeShi2019}).  \Cref{cap_image_cub} presents qualitative results for caption to image conditional generation. 
\gls{MLD} is the only model capable of generating bird images with convincing coherence. Clearly, none of the \gls{VAE}-based methods is able to achieve sufficient caption to image conditional generation quality using the same simple autoencoder architecture. Note that an image autoencoder with larger capacity improves considerably \gls{MLD} generative performance, suggesting that careful engineering applied to modality specific autoencoders is a promising avenue for future work.
We report quantitative results in \ref{apdx:additionnal_res}, where we show generation quality \gls{FID} metric. Due to the unavailability of the labels in this data-set, coherence evaluation as with the previous data-sets is not possible. We then resort to \gls{CLIP-S} \cite{hessel2021clipscore} an image-captioning metric, that, despite its limitations for the considered data-set \cite{kim2022mutual}, shows that \gls{MLD} outperforms competitors.

\section{Conclusion and Limitations}\label{sec:conclusion}
We have presented a new multi-modal generative model, Multimodal Latent Diffusion (\gls{MLD}), to address the well-known coherence--quality tradeoff that is inherent in existing multi-modal \gls{VAE}-based models. \gls{MLD} uses a set of independently trained, uni-modal, deterministic autoencoders. Generative properties of our model stem from a masked diffusion process that operates on latent variables. We also developed a new multi-time training method to learn the conditional score network for multi-modal diffusion. An extensive experimental campaign on various real-life data-sets, provided compelling evidence on the effectiveness of \gls{MLD} for multi-modal generative modeling. In all scenarios, including cases with loosely correlated modalities and high-resolution datasets, \gls{MLD} consistently outperformed the alternatives from the state-of-the-art. 

\bibliography{paper.bib}
\bibliographystyle{iclr2024_conference}

\appendix

\newpage
\section{Appendix}\label{sec:apdx}

\setcounter{section}{0}
\renewcommand{\thesection}{\Alph{section}}
\renewcommand{\theHsection}{appendixsection.\Alph{section}}

\section*{Multi-modal Latent Diffusion --- Supplementary material}

\section{Diffusion in  the multimodal latent space}
\label{apdx:mld_details}

In this section, we provide additional technical details of \gls{MLD}. We first discuss a naive approach based on \textit{In-painting} which uses only unconditional score network for both joint and conditional generation.
We also discuss alternative training scheme based on a work from the caption-text translation literature \cite{bao2023transformer}. Finally, we provide extra technical details for the score network architecture and sampling technique.

\subsection{Modalities Auto-Encoders} 
Each deterministic autoencoders used in the first stage of \gls{MLD} uses a vector latent space with no size constraints. Instead, \gls{VAE}-based models, generally require the latent space of each individual \gls{VAE} to be exactly of the same size, to allow the definition of a joint latent space.

In our approach, before concatenation, the modality-specific latent spaces are \textit{normalized} by element-wise mean and standard deviation. In practice, we use the statistics retrieved from the first training batch, which we found sufficient to gain sufficient statistical confidence.
This operation allows the harmonization of different modality-specific latent spaces and, therefore, facilitate the learning of a joint score network.

\begin{figure}[H]
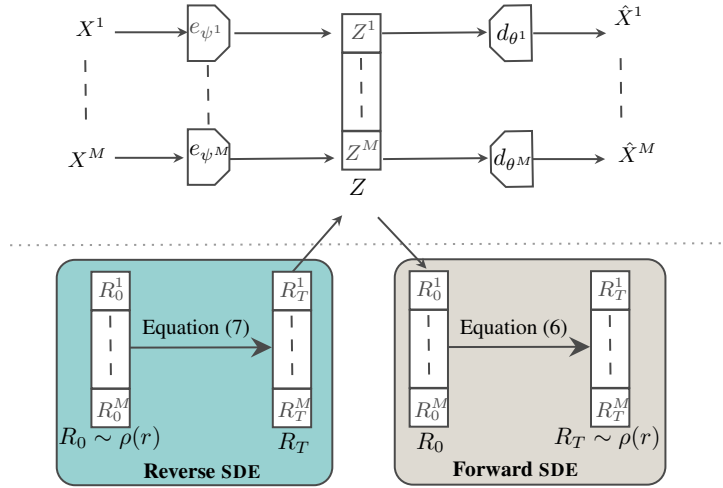

    \centering
    \include{figures/plots/gen_arch}
    \caption{Multi-modal Latent Diffusion. Two-stage model involving: \textbf{  \textit{Top:} }deterministic, modality-specific encoder/decoders, \textbf{ \textit{Bottom:} } score-based diffusion model on the concatenated latent spaces.}
    \label{fig:my_label}
\end{figure}

\subsection{Multi-modal diffusion SDE}

In \Cref{sec:method}, we presented our multi-modal latent diffusion process allowing multi-modal joint and conditional  generation. The role of the  \gls{SDE} is to gradually add noise to the data, perturbing its structure until attaining a noise distribution. 
In this work, we consider \gls{VPSDE} \cite{Song2020}. In this framework we have : $\rho(r) \sim \mathcal{N}(0;I) $,   $\alpha(t) = -\frac{1}{2}\beta(t) $ and $g(t) = \sqrt{\beta(t)}$, where $  \beta(t)  = \beta_{min} + t ( \beta_{max} - \beta_{min} )$. Following \citep{ho2020denoising,Song2020}, we set $\beta_{min} = 0.1 $ and $\beta_{max} = 20$.
With this configuration and by substitution of \Cref{eq:fw_sde_j}, we obtain the following forward \gls{SDE}:

\begin{equation}\label{eq:fw_sde_beta}
    \dd R_t =  -\frac{1}{2}\beta(t) R_t\dd t+ \sqrt{\beta(t)} \dd W_t, \qquad t \in [0,T].
\end{equation}

The corresponding perturbation kernel is given by :
\begin{equation}\label{eq:diffuson_kernel}
    q(r|z,t) = \mathcal{N}(r ; e^{-  \frac{1}{4}  t^2 (\beta_{max} - \beta_{min} ) - \frac{1}{2} t \beta_{min}   } z , (1-e^{-  \frac{1}{2}  t^2 (\beta_{max} - \beta_{min} ) - t \beta_{min}   } )\mathbf{I} ).
\end{equation}

The marginal score $\nabla\log q(R_{t},t)$ is approximated by a score network $ s_\chi (R_{t}, t)$ whose parameters $\chi$ can be optimized by minimizing the \gls{ELBO} in \Cref{elbo_diff_j}, where we found that using the same re-scaling as in \cite{Song2020} is more stable.

The reverse process is described by a different \gls{SDE} (\Cref{eq:bw_sde_j}). When using a variance-preserving \gls{SDE}, \Cref{eq:bw_sde_j} specializes in:

\begin{equation}\label{eq:reverse_sde_beta}
  \dd R_{t}  = \left[ \frac{1}{2} \beta(T-t) R_{t}  + \beta(T-t) \nabla \log q(R_{t},T-t) \right] \dd t + \sqrt{\beta(T-t)} \dd W_{t},   \\  
\end{equation}
With $R_0\sim \rho(r)$ as initial condition and time $t$ flows from $t=0$ to $t=T$.

Once the parametric score network is optimized,  trough the simulation of \Cref{eq:reverse_sde_beta}, sampling $R_T \sim q_\psi(r)$ is possible allowing \textbf{joint generation}. A numerical \gls{SDE} solver can be used to sample $ R_{T}$ which can be fed to the modality specific decoders to jointly sample a set of $\hat{X}=\{d^i_\theta(R^{i}_T)\}_{i= 0}^M$. 
As explained in \Cref{sec:inpainting}, the use of the unconditional score network $ s_\chi (R_t, t)$ allows \textbf{conditional generation} through the approximation described in \cite{Song2020}.

As described in \Cref{algo:MLD_inpaint_cond}, one can generate a set of modalities $A_1$ conditioned on the available set of modalities $A_2$. First, the available modalities are encoded into their respective latent space $z^{A_2}$ , the initial missing part is sampled from the stationary distribution $R^{A_1}_0 \sim \rho(r^{A_1})$, using an \gls{SDE} solver (e.g. Euler-Maruyama), the reverse diffusion \gls{SDE} (in \Cref{eq:reverse_sde_beta}) is discretized using a finite time steps $\Delta t = \nicefrac{T}{N} $, starting from $t = 0$ and iterating until $t\approx T$. At each iteration, the available portion of the latent space is diffused and brought to the same noise level as $R_t^{A_1}$ allowing the use of the unconditional score network. Lastly, the reverse diffusion update is done. This process is repeated until arriving at $t\approx T$ and obtaining $R^{A_1}_T = \hat{Z}^{A_1}$ which can be decoded to recover $\hat{x}^{A_1}$.
Note that the joint generation can be seen as a special case of \Cref{algo:MLD_inpaint_cond} with $A_2 = \emptyset$. We name this first approach \gls{MLD Inpaint} and provide extensive comparison with our method \gls{MLD} in \Cref{apdx:mld_variants}.

\begin{algorithm}[hbt!] 
\DontPrintSemicolon
\SetAlgoLined
\SetNoFillComment
\LinesNotNumbered 
\caption{\gls{MLD Inpaint} conditional generation}\label{algo:MLD_inpaint_cond}
\KwData{$ x^{A_2}= \{ x^i\}_{i \in A_2} $ }  
$z^{A_2} \gets \{ e_{\phi_i}(x^i) \}_{i \in A_2} $   \tcp*{Encode the available modalities $X$ into their latent space} 
$A_1\gets \{1,\dots,M\}\setminus A_2 $ \tcp*{The set of modalities to generate} 
$R_0 \gets \mathcal{C} ( R_0^{A_1}, z^{A_2}),  \qquad  R_0^{A_1} \sim \rho(r^{A_1}) $  \tcp*{Compose the initial state} 
$R \gets R_0$ \\
$\Delta t \gets \nicefrac{T}{N}$ \\
\For{$n =0  \quad \textbf{to} \quad N-1 \quad  $}{
$t' \gets T - n  \ \Delta t  $ \\
$\bar{R}  \sim  q (r|R_0,t') $ \tcp*{Diffuse the available portion of the latent space(\cref{eq:diffuson_kernel})} 
$R \gets m(A_1) \odot R +   ( 1- m(A_1) ) \odot \bar{R} $ \\
$\epsilon \sim \mathcal{N}(0;I)  \quad$  \textbf{if} $ n < (N-1)  $ \textbf{else} $\epsilon= 0$ \\
$\Delta R  \gets  \Delta t \left[ \frac{1}{2} \beta(t') R  + \beta(t') s_\chi(R,t') \right]  + \sqrt{\beta(t') \Delta t } 
 \epsilon   $ \\
$R \gets R +  \Delta R $  \tcp*{The Euler-Maruyama update step }
 }
 $\hat{z}^{A_1} \gets R^{A_1}$ \\
\textbf{Return} $\hat{X}^{A_1} = \{d^i_\theta(\hat{z}^{i} )\}_{i\in A_1}$  
\end{algorithm}

As discussed in \Cref{sec:inpainting}, the approximation enabling the in-painting approach can be efficient in several domains but its generalization to the multi-modal latent space scenario is not trivial. We argue that this is due to the heterogeneity of modalities which induce different latent spaces characteristics. For different modality-specific latent spaces, the loss of information ratio can vary through the diffusion process. We verify this hypothesis through the following experiment.

\paragraph{Latent space robustness against diffusion perturbation:}
We analyse the effect of the forward diffusion perturbation on the latent space through time. We encode the modalities using their respective encoders to obtain their latent space $Z =  [e_{\psi^1}(X^1) \dots e_{\psi^M}(X^M)]$. 
Given a time $t \in [0,T]$, we diffuse the different latent spaces by applying \Cref{eq:diffuson_kernel} to get $R_t \sim  q(r|z,t)$ with $R_t$ being the perturbed version of the latent space at time $t$. We feed the modality specific decoders with the perturbed latent space $\hat{X}_t=\{d^i_\theta(R^{i}_t)\}_{i= 1}^M$, $\hat{X}_t$ being the output modalities generated using the perturbed latent space. To evaluate the information loss induced by the diffusion process on the different modalities, we assess the coherence preservation in the reconstructed modalities $\hat{X}_t$ by computing the coherence (in \%) as done in \Cref{sec:experiments}.

\begin{figure}[H]
     \centering
     \begin{subfigure}{0.3\textwidth}
         \centering
         \includegraphics[page=1,width=\linewidth]{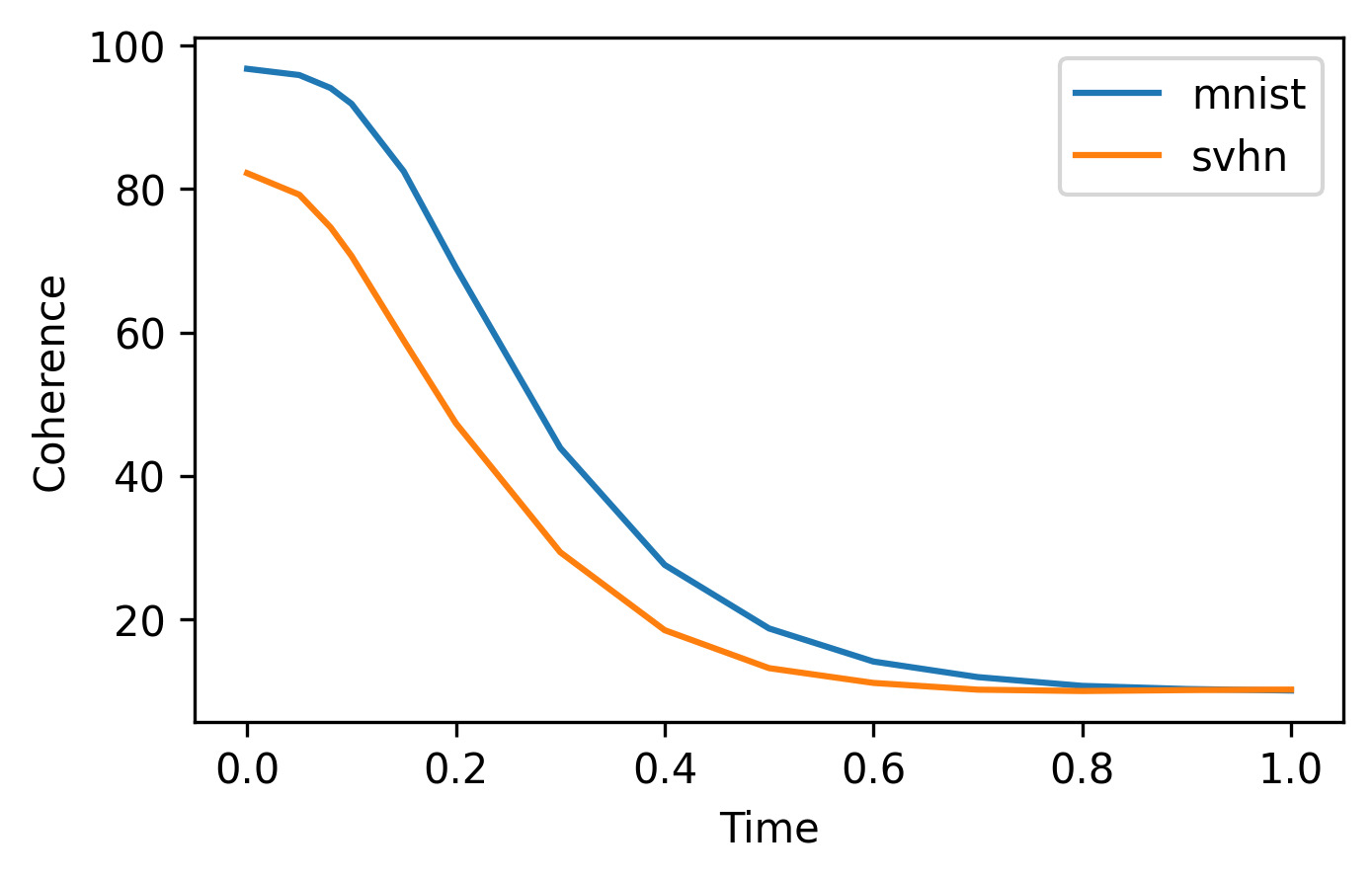}
         \caption{\mnist-\svhn}
     \end{subfigure}
        \begin{subfigure}[b]{0.3\textwidth}
         \centering
         \includegraphics[page=1,width=\linewidth]{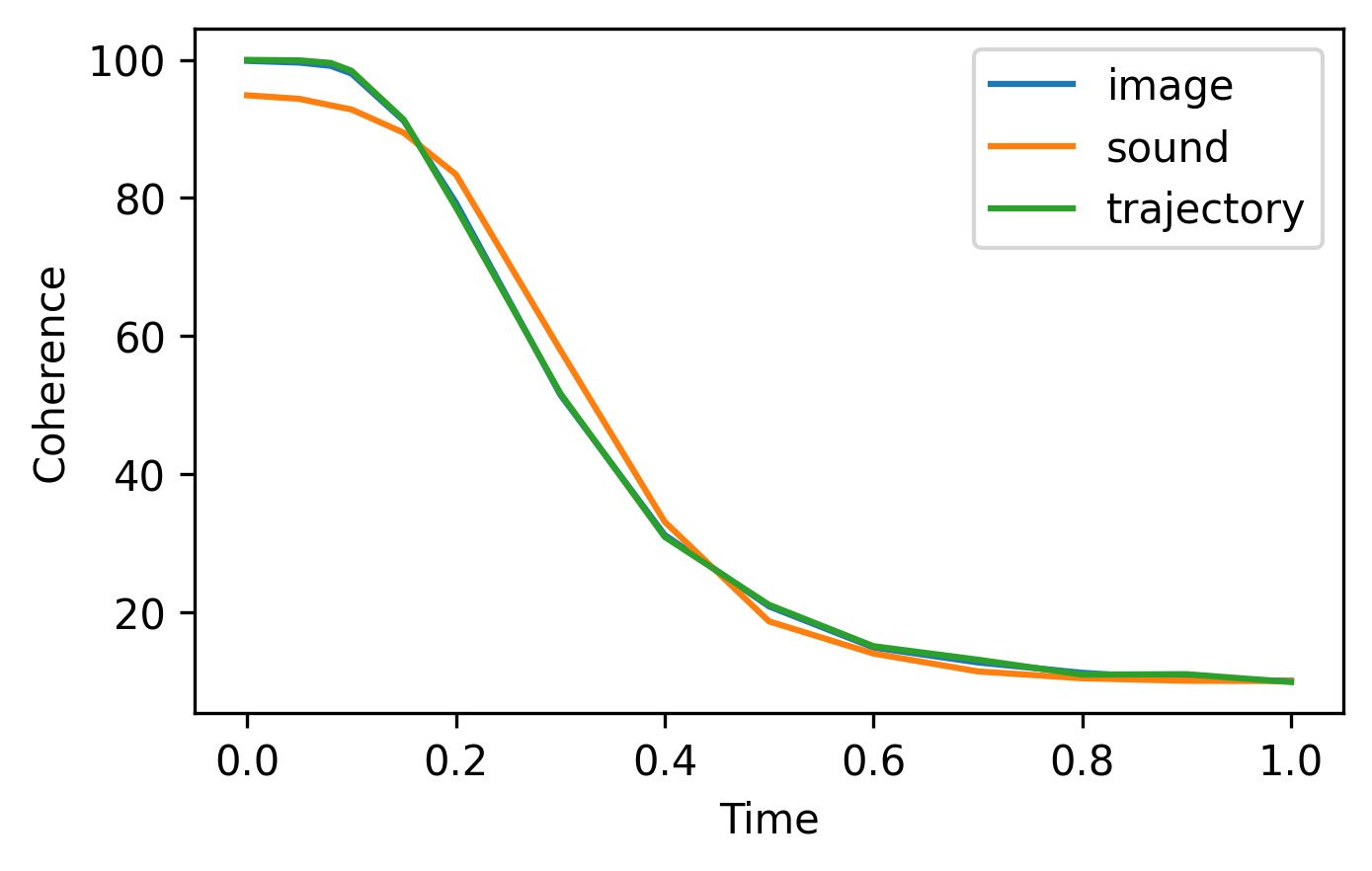}
          \caption{\mhd}
     \end{subfigure}
      \begin{subfigure}[b]{0.3\textwidth}
         \centering
         \includegraphics[page=1,width=\linewidth]{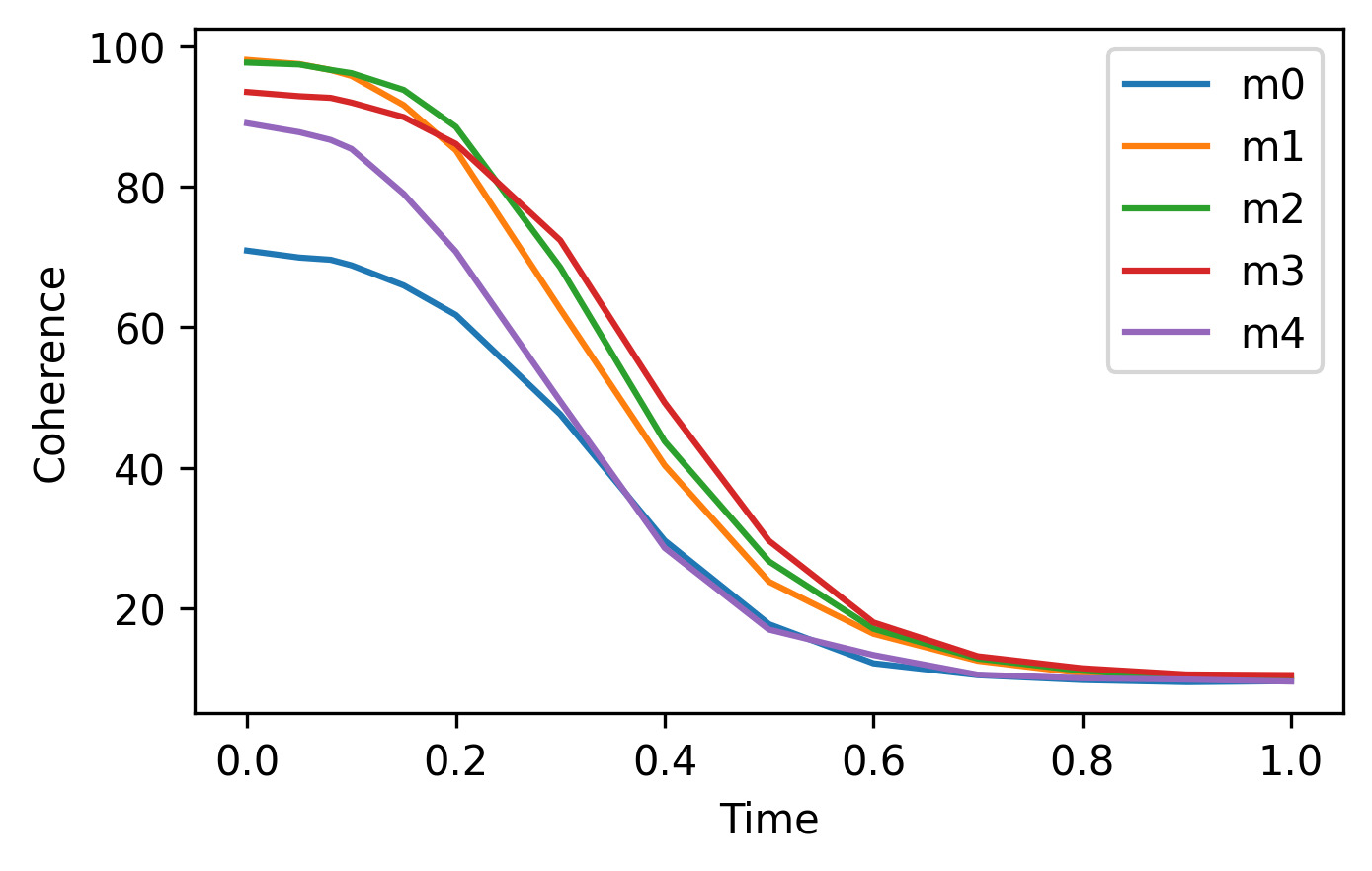}
            \caption{\polymnist}
     \end{subfigure}
   
        \caption{The coherence as a function of the diffusion process time for three datasets.  The diffusion perturbation is applied on the modalities latent space after an element wise normalization.}
        \label{fig:robust_exp}
\end{figure}

We expect to obtain high coherence results for $t \approx 0$, when compared to $t \approx T$, the information in the latent space being more preserved at the beginning of the diffusion process than at the last phase of the froward \gls{SDE} where all dependencies on initial conditions vanish. \Cref{fig:robust_exp} shows the coherence as a function of the diffusion time $t \in [0,1]$ for different modalities across multiple data-sets. We observe that within the same data-set, some modalities stand out with a specific level of robustness (using as a proxy the coherence level) against the diffusion perturbation in comparison with the remaining modalities from the same data-set. For instance, we remark that \svhn is less robust than \mnist which should manifest in an under-performance of \svhn to \mnist conditional generation. An intuition that we verify in \Cref{apdx:mld_variants}.

\subsection{Multi-time Masked Multi-modal SDE}

To learn the score network capable of both conditional and joint generation, we proposed in \Cref{sec:mld_var} a multi-time masked diffusion process. 

\Cref{algo:MLD_training} presents a pseudo-code for the multi time masked training. The masked diffusion process is applied following a randomization with probably $d$. First, a subset of modalities $A_2$ is selected randomly to be the conditioning modalities and $A_1$ the remaining set of modalities to be the diffused modalities. The time $t$ is sampled uniformly from $[0,T]$ and the portion of the latent space corresponding to the subset $A_1$ is diffused accordingly. Using the masking as shown in \Cref{train_line_mask}, the portion of the latent space corresponding to the subset $A_2$ is not diffused and forced to be equal to $R_0^{A_2} = z^{A_2}$. The multi-time vector $\tau$ is constructed. Lastly, the score network is optimized by minimizing a masked loss corresponding to the diffused part of the latent space. With probability $(1- d)$, all the modalities are diffused at the same time and $A_2 =\emptyset$. In order to calibrate the loss, given that the randomization of $A_1$ and $A_2$ can result in diffusing different sizes of the latent space, we re-weight the loss according to the cardinality of the diffused and freezed portions of the latent space:

\begin{equation} \label{eq:weight_eq}
    \Omega(A_1,A_2) = 1 + \frac{dim(A_2)}{dim(A_1)}
\end{equation}
Where $\dim(.) $ is the sum of each latent space cardinality of a given subset of modalities with $dim(\emptyset) = 0 $ .

\begin{algorithm}[hbt!]
\DontPrintSemicolon
\SetAlgoLined
\SetNoFillComment
\LinesNotNumbered 
\caption{\gls{MLD} Masked Multi-time diffusion training step}\label{algo:MLD_training}
\KwData{$ X = \{x^i\}_{i=1}^M $}
\textbf{Param:} $d$ \\
$Z \gets \{ e_{\phi_i}(x^i) \}_{i=0}^M $   \tcp*{Encode the modalities $X$ into their latent space } 
 $A_2 \sim \nu  \quad $ \tcp*{$\nu$ depends on the parameter $d$}
$A_1\gets \{1,\dots,M\}\setminus A_2 $ \\
$t \sim \mathcal{U}[0,T]$ \\

$R  \sim  q (r|Z,t) $ \tcp*{Diffuse the available portion of the latent space(\Cref{eq:diffuson_kernel})}
$R \gets m(A_1) \odot R     + (1- m(A_1) ) \odot Z$ \tcp*{Masked diffusion}  \label{train_line_mask}
$\tau(A_1,t) \gets \left[\mathds{1}(1\in A_1)t,\dots,\mathds{1}(M\in A_1)t\right]$ \tcp*{Construct the multi time vector} 

\textbf{Return} $ \nabla_{\chi} \left\{  \Omega(A_1,A_2) \quad  \norm{m(A_1)\odot \quad \left [  s_{\chi}(R,\tau(A_1,t))-\nabla \log q(R,t|z^{A_2})  \right]     }^2_2   \quad \right\} $

\end{algorithm}

The optimized score network can approximate both the conditional and unconditional true score:
\begin{equation}
    s_\chi(R_t,\tau(A_1,t)) \sim  \nabla\log q(R_t,t\g z^{A_2})).
\end{equation}
The joint generation is a special case of the latter with $A_2 = \emptyset$:
\begin{equation}
    s_\chi(R_t,\tau(A_1,t)) \sim  \nabla\log q(R_t,t) \quad     ,A_1 = \{1,...,M\}
\end{equation}

\Cref{algo:MLD_cond_gen} describes the reverse conditional generation pseudo-code. It's pertinent to compare this algorithm with \Cref{algo:MLD_inpaint_cond}. The main difference resides in the use of the multi-time score network, enabling conditional generation with the multi-time vector playing the role of time information and conditioning signal. On the other hand, in \Cref{algo:MLD_inpaint_cond}, we don't have a conditional score network, therefore we resort to the approximation from \Cref{sec:inpainting}, and use the unconditional score.

\begin{algorithm}[hbt!]
\DontPrintSemicolon
\SetAlgoLined
\SetNoFillComment
\LinesNotNumbered 
\caption{MLD conditional generation.}\label{algo:MLD_cond_gen}
\KwData{$ x^{A_2} \gets \{ x^i\}_{i \in A_2} $ }  
$z^{A_2} \gets \{ e_{\phi_i}(x^i) \}_{i \in A_2} $   \tcp*{Encode the available modalities $X$ into their latent space} 
$A_1\gets \{1,\dots,M\}\setminus A_2 $ \tcp*{The set of modalities to be generated} 
$R_0 \gets \mathcal{C} ( R_0^{A_1}, z^{A_2}),  \qquad  R_0^{A_1} \sim \rho(r^{A_1}) $  \tcp*{Compose the initial latent space} 
$R \gets R_0$\\
$\Delta t \gets \nicefrac{T}{N}$ \\
\For{$n = 0  \quad \textbf{to} \quad N-1 \quad  $}{
$t' \gets T -  n  \ \Delta t $ \\
$\tau(A_1,t') \gets \left[\mathds{1}(1\in A_1)t',\dots,\mathds{1}(M\in A_1)t'\right]$ \tcp*{Construct the multi-time vector}
$\epsilon \sim \mathcal{N}(0;I)  \quad$  \textbf{if} $ n < N \quad  $ \textbf{else} $ \quad\epsilon= 0$ \\
$\Delta R  \gets  \Delta t \left[ \frac{1}{2} \beta(t') R  + \beta(t') s_\chi (R, \tau(A_1,t')) \right]  + \sqrt{\beta(t') \Delta t }  \epsilon   $ \\
$R \gets R +  \Delta R $ \tcp*{The Euler-Maruyama update step }
$R \gets m(A_1) \odot R + (1 -  m(A_1) )\odot R_{0} $  \tcp*{ Update the portion corresponding to the unavailable modalities }
} 
$ \hat{z}^{A_1} = R^{A_1}$ \\
\textbf{Return} $\hat{X}^{A_1} = \{d^i_\theta(\hat{z}^{i} )\}_{i\in A_1}$  
\end{algorithm}

\subsection{Uni-diffuser Training}
The work presented in \cite{bao2023transformer} is specialized for an image-caption application. The approach is based on a multi-modal diffusion model applied to a unified latent embedding, obtained via pre-trained autoencoders, and incorporating pre-trained models (CLIP \cite{radford2021clip} and GPT-2 \cite{radford2019language}). The unified latent space is composed of an image embedding, a CLIP image embedding and a CLIP text embedding. Note that the CLIP model is pre-trained on a pairs of multi-modal data (image-text), which is expected to enhance the generative performance. Since it is not trivial to have a jointly trained encoder similar to CLIP for any type of modality, the evaluation of this model on different modalities across different data-set (e.g. including audio) is not an easy task.

To compare to this work, we adapt the training scheme presented in \cite{bao2023transformer} to our \gls{MLD} method. Instead of applying a masked multi-modal \gls{SDE} for training the score network, every portion of the latent space is diffused according to a different time $t^i \sim \mathcal{U}(0,1)$ and, therefore, the multi-time vector fed to the score network is $\tau(t) = [t^0 \sim \mathcal{U}(0,1),..., t^M \sim \mathcal{U}(0,1) ] $. For fairness, we use the same score network and reverse process sampler as for our \gls{MLD} version with multi-time training, and call this variant \gls{MLD Uni}.

\subsection{Intuitive summary: How does MLD capture modality interactions?}

\gls{MLD} treats the latent spaces of each modality as variables that evolve differently through the diffusion process according to a multi-time vector. 
The masked multi-time training enables the model to learn the score of all the combination of conditionally diffused modalities, using the frozen modalities as the conditioning signal, through a randomized scheme. By learning the score function of the diffused modalities at different time steps, the score model captures the correlation between the modalities. At test time, the diffusion time of each modality is chosen to modulate its influence on the generation, as follows. 

For joint generation the model uses the unconditional score which corresponds to using the same diffusion time for all modalities. Thus, all the modalities influence each other equally. This ensures that modality interaction information is faithful to the one characterizing the observed data distribution. 

The model can also generate modalities conditionally by using the conditional score, by freezing the conditioning modalities during the reverse process. The freezed state is similar to the final state of the revere process where information is not perturbed, thus the influence of the conditioning modalities is maximal. Subsequently, the generated modalities reflect the necessary information from the conditioning modalities and achieve the desired correlation.

\subsection{Technical details}

\paragraph{Sampling schedule:}
\label{sec:repaint_sche}
We use the sampling schedule proposed in \cite{lugmayr2022repaint}, which has shown to improve the coherence of the conditional and joint generation. We use the best parameters suggested by the authors: $N = 250$ time-steps, applied $r = 10$ re-sampling times with jump size $j = 10$. For readability in \cref{algo:MLD_inpaint_cond} and \cref{algo:MLD_cond_gen},  we present pseudo code  with a linear sampling schedule which can be easily adapted to any other schedule.

\begin{figure}[h]
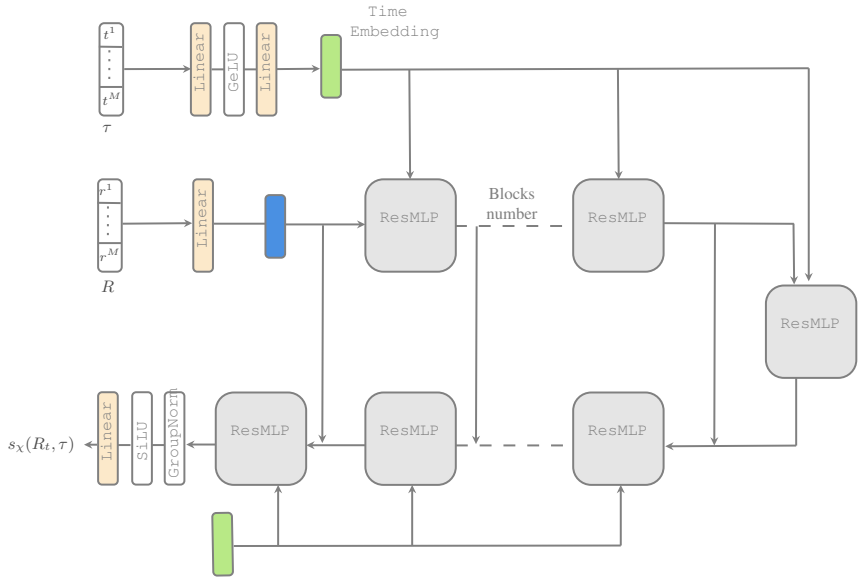

\centering
\include{figures/plots/arch}
\caption{Score network $s_\chi$ architecture used in our \gls{MLD} implementation. Residual \gls{MLP} block architecture is shown in \Cref{fig:resmlp}.}
\label{fig:score_net}
\end{figure}

\paragraph{Training the score network: }
Inspired by the architecture from \citep{dupont22_fucnta}, we use simple Residual \gls{MLP} blocks with skip connections as our score network (see \Cref{fig:score_net}). We fix the \textbf{width} and \textbf{number of blocks} proportionally to the number of the modalities and the latent space size. As in \cite{song2020improved}, we use \gls{EMA} of model parameters with a momentum parameter $m = 0.999$.

\begin{figure}[h]
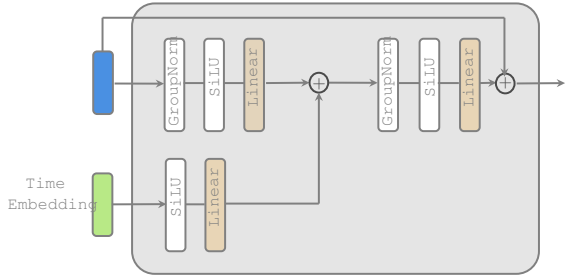

    \centering
    \include{figures/plots/arch_mlpres}
    \caption{Architecture of ResMLP block.}
    \label{fig:resmlp}
\end{figure}

\newpage
\section{MLD ablations study}
\label{apdx:mld_variants}
In this section, we compare \gls{MLD} with two variants presented in \Cref{apdx:mld_details} : \gls{MLD Inpaint}, a naive approach without our proposed \textit{multi-time masked} \gls{SDE}, \gls{MLD Uni} a variant of our method using the same training scheme of \cite{bao2023transformer}. We also analyse the effect of the $d$ randomization parameter on \gls{MLD} performance through ablations study. 
\subsection{MLD and its variants}
\Cref{table:mld_variants_comparison} summarizes the different approaches adopted in each variant. All the considered models share the same deterministic autoencoders trained during the first stage. 

For fairness, our evaluation was done using the same configuration and code basis of \gls{MLD}. This includes: the autoencoder architectures and latent space size (similar to \Cref{sec:experiments}), the same score network (\Cref{fig:score_net}) is used across experiments, with \gls{MLD Inpaint} using the same architecture with one time dimension instead of the multi-time vector.
In all the variants, the joint and conditional generation are conducted using the same reverse sampling schedule described in \Cref{sec:repaint_sche}.

\begin{table}[h]
\caption{\gls{MLD} and its variants ablation study}
\centering
\begin{tabular}{c|c|c|c}
\toprule
Model & Multi-time diffusion & Training & Conditional and joint generation
 \\
\midrule 
\gls{MLD Inpaint} & x & \Cref{elbo_diff_j} & \Cref{algo:MLD_inpaint_cond}  \\
\gls{MLD Uni} & \checkmark & \cite{bao2023transformer} & \Cref{algo:MLD_cond_gen}  \\
\gls{MLD} & \checkmark & \Cref{algo:MLD_training} & \Cref{algo:MLD_cond_gen}  \\
\bottomrule
\end{tabular}
\label{table:mld_variants_comparison}
\end{table}

\paragraph{Results}
In some cases, the \gls{MLD} variants can match the joint generation performance of \gls{MLD} but, overall, they are less efficient and have noticeable weaknesses: \gls{MLD Inpaint} under-performs in conditional generation when considering relatively complex modalities, \gls{MLD Uni} is not able to leverage the presence of multiple modalities to improve cross generation, especially for data-sets with a large number of modalities. On the other hand, \gls{MLD} is able to overcome all these limitations.

\begin{figure}[h]
     \centering

\begin{subfigure}{0.25\textwidth}
         \centering
         \includegraphics[width=\linewidth]{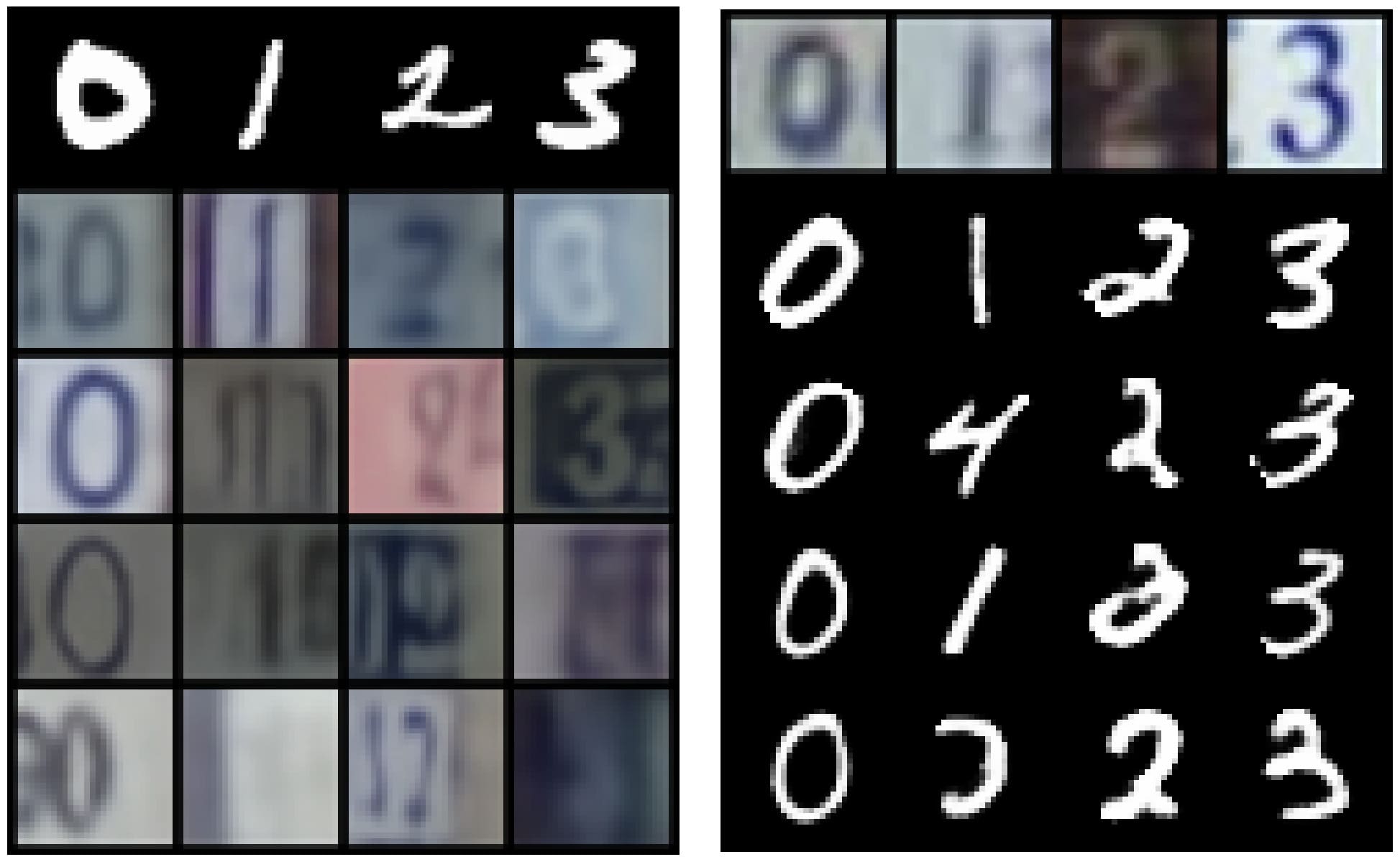}
    \caption*{\gls{MLD Inpaint}}
    \end{subfigure}
\begin{subfigure}{0.25\textwidth}
         \centering
         \includegraphics[width=\linewidth]{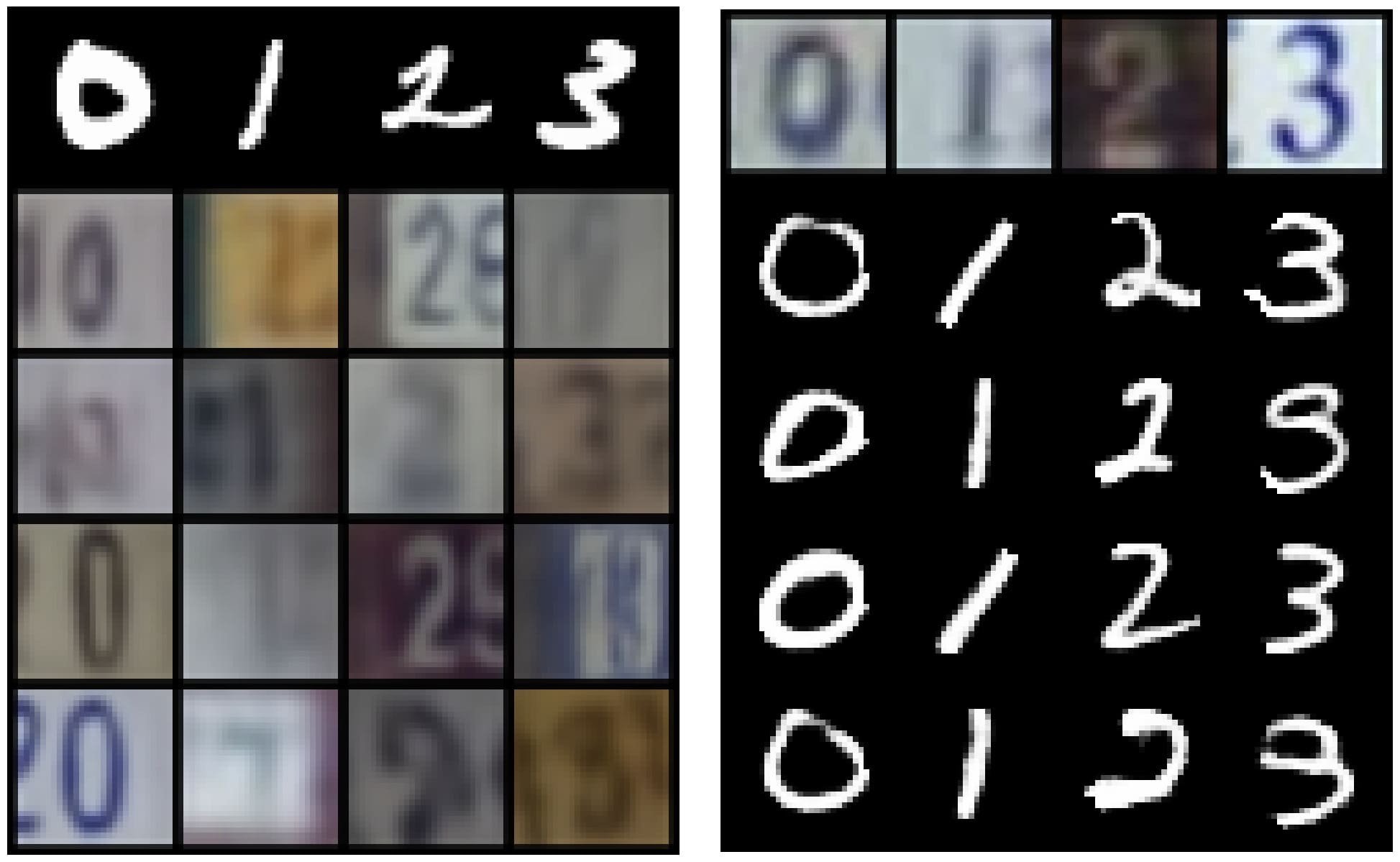}
         \caption*{\gls{MLD Uni}}
\end{subfigure}
  \begin{subfigure}{0.25\textwidth}
         \centering
         \includegraphics[page=1,width=\linewidth]{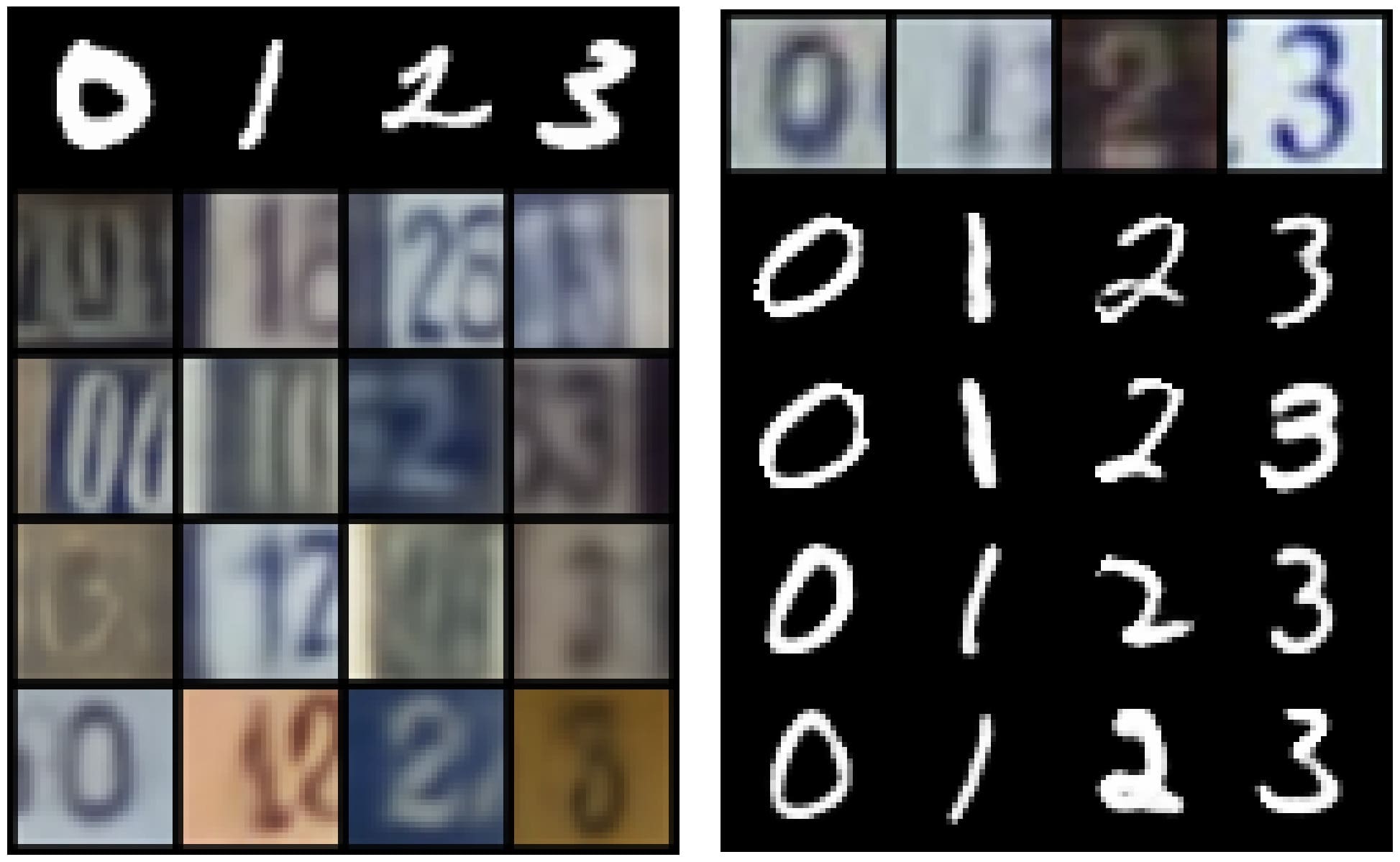}
         \caption*{\textbf{\gls{MLD}}}
     \end{subfigure}
\caption{Qualitative results for \textbf{\mnist-\svhn}. For each model we report: \mnist to \svhn conditional generation in the left,   \svhn to \mnist conditional generation in the right.}
        \label{fig:cond_joint_ms_detail}
\end{figure}

\paragraph{\mnist-\svhn.} 
In \Cref{coh_quality:ms_detailed_mld}, \gls{MLD} achieves the best results and dominates cross generation performance. We observe that \gls{MLD Inpaint} lacks coherence for \svhn to \mnist conditional generation, a results we expected by analysing the experiment in \Cref{fig:robust_exp}. \gls{MLD Uni}, despite the use of a multi-time diffusion process, under-performs our method, which indicates the effectiveness of our masked diffusion process in learning the conditional score network. Since all the models use the same deterministic autoencoders, the observed generative quality performance are relatively similar (See \Cref{fig:cond_joint_ms_detail} for qualitative results ).

\begin{table}[h]
\caption{Generation Coherence and Quality for MNIST-SVHN (M is for MNIST and S for SVHN ). The generation quality is measured in terms of FMD for MNIST and FID for SVHN.\newline}
 \label{coh_quality:ms_detailed_mld}
\centering
\resizebox{\textwidth}{!}{\begin{tabular}{c|c|cc|cc|cc}
\toprule
\multirow{2}{*}{ Models }  & \multicolumn{3}{c}{Coherence (\%$\uparrow$) } &   \multicolumn{4}{c}{Quality ($\downarrow$)} \\
    \cmidrule{2-8}
    & Joint &  M $\rightarrow$ S & S $\rightarrow$ M &
     Joint(M)  &  Joint(S) &M $\rightarrow$ S &  S $\rightarrow$ M  \\
    \midrule

    MLD-Inpaint & $\textbf{85.53}_{ \pm 0.22 } $& $\underline{81.76}_{\pm 0.23 }$ & $63.28_{\pm 1.16} $  &   $\textbf{3.85}_{ \pm 0.02}$ &
$ 60.86_{\pm 1.27}$&$59.86_{\pm1.18}$&
$\textbf{3.55}_{\pm 0.11} $ 
    \\
    MLD-Uni & $82.19_{ \pm 0.97} $& $79.31_{\pm 1.21 }$ & $\underline{72.78}_{\pm 1.81} $&  $4.1_{ \pm 0.17}$ &
$ 57.41_{\pm 1.43}$ &
$\underline{57.84}_{\pm1.57}$&
$4.84_{\pm 0.28} $

    \\
    MLD  & $\underline{85.22}_{ \pm 0.5}$ & $\textbf{83.79}_{\pm 0.62} $&$ \textbf{79.13}_{\pm 0.38}$ & $\underline{3.93}_{ \pm 0.12}$ &
    $\textbf{56.36}_{\pm1.63}$   &
    $\textbf{	57.2}_{\pm1.47}$  &
    $\underline{3.67}_{\pm 0.14} $ 
    \\
    \bottomrule
   
\end{tabular}}
\end{table}

\paragraph{\mhd.} \Cref{coh:mhd_detailed_mld} shows the performance results for the \mhd data-set in terms of generative coherence.
\gls{MLD} achieves the best joint generation coherence and, along with \gls{MLD Uni}, they dominate the cross generation coherence. \gls{MLD Inpaint} shows a lack of coherence when conditioning on the sound modality alone, a predictable result since this is a more difficult configuration, as the sound modality is loosely correlated to other modalities. We also observe that \gls{MLD Inpaint} performs worse than the two other alternatives when conditioned on the trajectory modality, which is the smallest modality in terms of latent size. This indicates another limitation of the naive approach regarding coherent generation when handling different latent spaces sizes, a weakness our method \gls{MLD} overcomes. \Cref{qua:mhd_detailed_mld} presents the qualitative generative performance which are homogeneous across the variants with \gls{MLD}, achieving either the best or second best performance.

\begin{table}[H]
\centering
\caption{Generation Coherence (\%$\uparrow$) for \mhd  (Higher is better). Line above refers to the generated modality
while the observed modalities subset are presented below. \newline}
 \label{coh:mhd_detailed_mld}
\resizebox{\textwidth}{!}{\begin{tabular}{c|c|ccc|ccc|ccc}
\toprule
\multirow{2}{*}{ Models }  & \multirow{2}{*}{ Joint }
& \multicolumn{3}{c}{I (Image)}  & \multicolumn{3}{c}{T (Trajectory)}  & \multicolumn{3}{c}{S (Sound) } 
\\
\cmidrule{3-11}
    &&  T & S  & T,S  & I & S  & I,S & I & T & I,T \\
    \midrule

MLD-Inpaint  &  
$96.88_{ \pm 0.35 }$ &
    $63.9_{\pm 1.7 }$ & 
    $56.52 _{\pm 1.89 }$ &
     $ 95.83_{\pm 0.48}$ &
     
    $\underline{99.58}_{ \pm 0.1  }$  &
    $56.51 _{\pm  1.89 }$  &
    $\underline{99.89}_{\pm0.04}$ &  
    $95.81_{\pm0.25}$ &
    $56.51 _{\pm  1.89 }$  &
    $96.38_{\pm0.35}$   \\
    
MLD-Uni &  
  $\underline{97.69}_{ \pm 0.26 }$ &
    $\textbf{99.91}_{\pm 0.04 }$ &
    $\textbf{89.87}_{\pm 0.38 }$ & 
    $ \textbf{99.92}_{\pm 0.04}$ & 
    $\textbf{99.68}_{ \pm 0.1  }$  & 
    $\textbf{89.78}_{\pm  0.45 }$  & 
    $99.38_{\pm0.31}$ & 
    $\underline{97.54}_{\pm0.2}$ & 
    $\underline{97.65}_{\pm  0.41 }$ 
    & $\underline{97.79}_{\pm 0.41}$ 
    \\
 MLD &  
 $\textbf{98.34}_{ \pm 0.22 }$
 &
    $99.45_{\pm 0.09 }$ &
    $\underline{88.91}_{\pm 0.54 }$ & 
    $ \underline{99.88}_{\pm 0.04}$ & 
    
    $\underline{99.58}_{ \pm 0.03  }$  & 
    $\underline{88.92} _{\pm  0.53 }$  & 
    $\textbf{99.91}_{\pm0.02}$ &  
    
    $\textbf{97.63}_{\pm0.14}$ & 
    $\textbf{97.7} _{\pm  0.34 }$ 
    & $\textbf{98.01}_{\pm 0.21}$  \\
    \bottomrule
   
\end{tabular}}

\end{table}

\begin{table}[H]
\centering
\caption{Generation quality for \mhd . The metrics reported are \gls{FMD} for Image and Trajectory modalities and \gls{FAD} for the sound modalities (Lower is better).  \newline}
 \label{qua:mhd_detailed_mld}
\resizebox{\textwidth}{!}{\begin{tabular}{c|cccc|cccc|cccc}
\toprule
\multirow{2}{*}{ Models }   
& \multicolumn{4}{c}{I (Image)}  & \multicolumn{4}{c}{T (Trajectory)}  & \multicolumn{4}{c}{S (Sound) } 
\\
\cmidrule{2-13}
   & Joint & T & S  & T,S  &  Joint & I & S  & I,S & Joint & I & T & I,T   \\
    \midrule
      MLD-Inpaint &   
    $5.35_{ \pm 1.35 }$   &
    $6.23_{ \pm 1.13 }$ &
    $\underline{4.76}_{ \pm 0.68 }$  &
    $3.53_{ \pm 0.36 }$  &

    $\textbf{1.59}_{ \pm 0.12}$   & 
    $\textbf{0.6}_{ \pm 0.05}$ &
    $\textbf{1.81}_{ \pm 0.13 }$  &
    $\textbf{0.54}_{ \pm 0.06 }$  &

    $2.41_{ \pm 0.07}$ &
    $2.5_{ \pm 0.04}$ &
    $2.52_{ \pm 0.02 }$  &
    $2.49_{ \pm 0.05 }$  
   
    \\
    MLD-Uni &   
     $\textbf{7.91}_{ \pm 2.2 }$   &
    $\textbf{1.65}_{ \pm 0.33 }$ &
    $6.29_{ \pm 1.38 }$  &
    $\underline{3.06}_{ \pm 0.54 }$  &

    $\underline{2.53}_{ \pm 0.5}$   & 
    $1.18_{ \pm 0.26}$ &
    $3.18_{ \pm 0.77 }$  &
    $2.84_{ \pm 1.14 }$

    & $\textbf{2.11}_{ \pm 0.08 }$
    & $\textbf{2.25}_{ \pm 0.05 }$ 
     & $\textbf{2.1}_{ \pm 0.0 }$ 
       & $\textbf{	2.15}_{ \pm 0.01 }$ 
       
      \\
    MLD &   
    $\underline{7.98}_{ \pm 1.41 }$ &
    $\underline{1.7}_{ \pm 0.14 }$ &
    $\textbf{4.54}_{ \pm 0.45 }$ & 
    $\textbf{1.84}_{ \pm 0.27 }$

    & $3.18_{ \pm 0.18 }$ 
      & $\underline{0.83}_{ \pm 0.03 }$ 
     & $\underline{2.07}_{ \pm 0.26 }$ 
       & $\underline{0.6}_{ \pm 0.05}$

        & $ \underline{2.39}_{ \pm 0.1 } $
        & $\underline{2.31}_{ \pm 0.07 }$ 
     & $\underline{2.33}_{ \pm 0.11 }$ 
       & $\underline{	2.29}_{ \pm 0.06 }$ 
       
       \\

    \bottomrule
   
\end{tabular}}

\end{table}

\begin{figure} [h]
     \centering
    \begin{subfigure}{0.25\textwidth}
     \begin{subfigure}{1\textwidth}
         \centering
         \includegraphics[page=1,width=\linewidth]{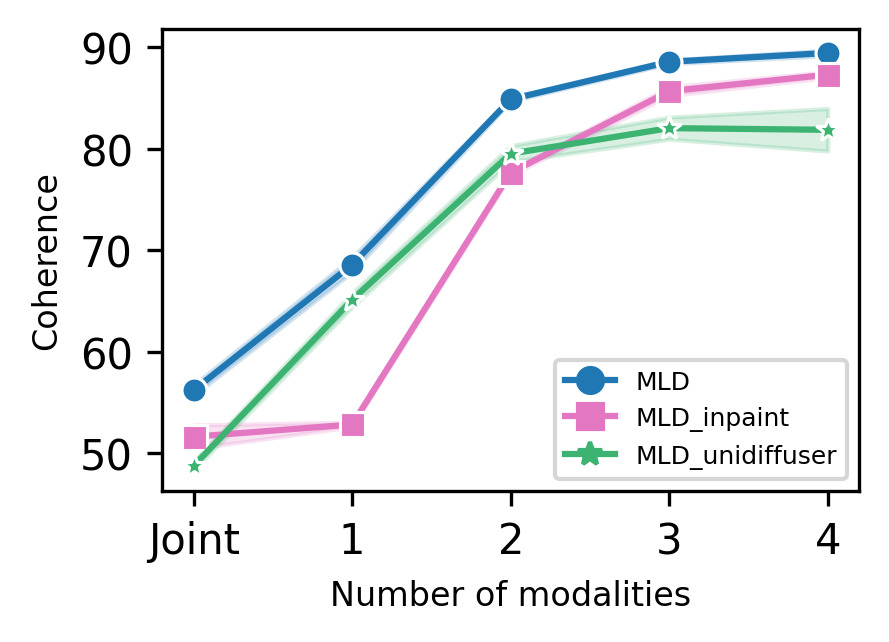}
     \end{subfigure}
     \begin{subfigure}{1\textwidth}
         \centering
         \includegraphics[width=\linewidth]{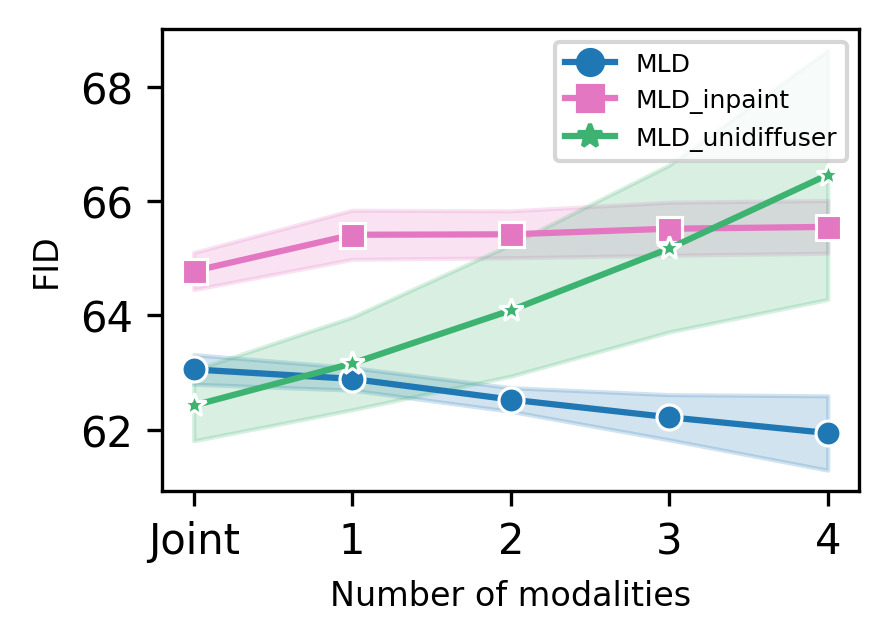}
         
     \end{subfigure}
\label{mmnist_quantitavie_mld}
\end{subfigure}
\begin{subfigure}{0.74\textwidth}
    \centering 
      \begin{subfigure}{0.29\textwidth}
         \centering
         \includegraphics[width=\linewidth]{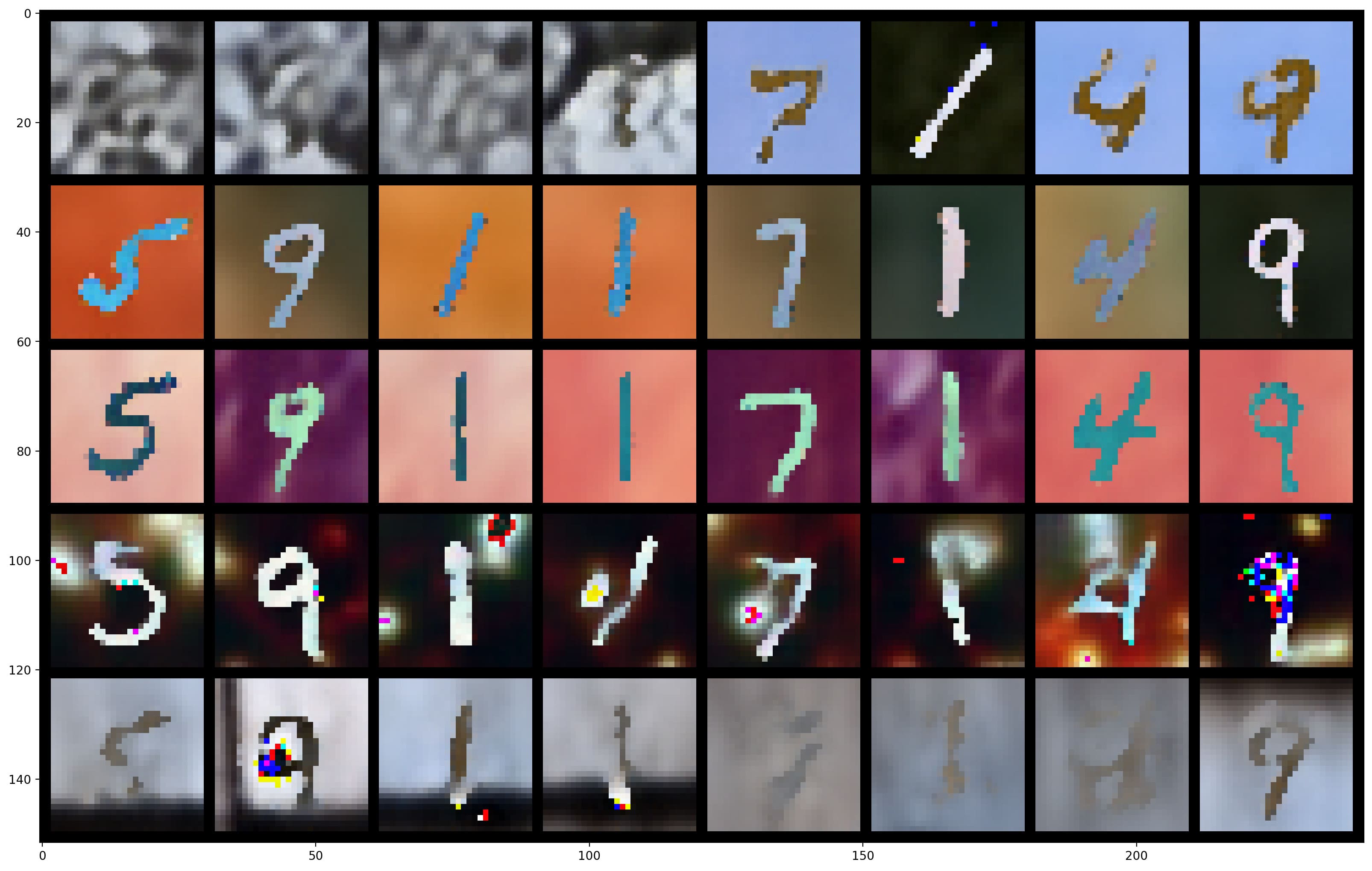}
         \caption*{\gls{MLD Inpaint}}
     \end{subfigure}
        \begin{subfigure}{0.29\textwidth}
         \centering
         \includegraphics[width=\linewidth]{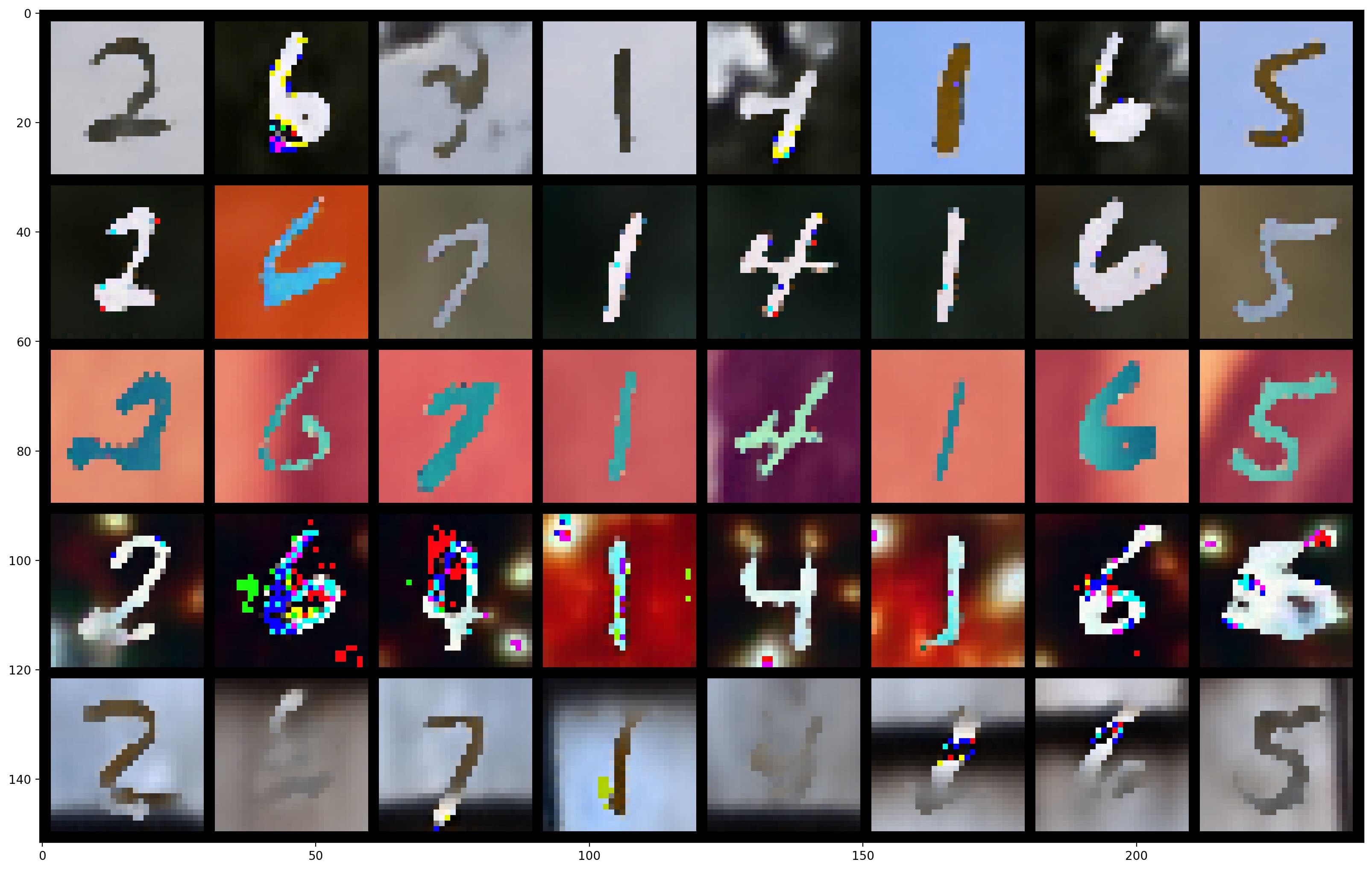}
         \caption*{\gls{MLD Uni}}
   
     \end{subfigure}
    
  \begin{subfigure}{0.29\textwidth}
         \centering
         \includegraphics[page=1,width=\linewidth]{figures/datasets/mmnist/quali/MLD_sampling}
         \caption*{\textbf{\gls{MLD} }}
       
     \end{subfigure}
       \label{qual_mmnist_mld}
\end{subfigure}
        \caption{Results for \textbf{\polymnist} data-set. \textit{Left}: a comparison of the generative coherence (\% $\uparrow$) and quality in terms of \gls{FID} ($\downarrow$)) as a function of the number of modality input. We report the average performance following the leave-one-out strategy (see \Cref{apdx:dataset_eval}).  \textit{Right}: are qualitative results for the joint generation of the 5 modalities.   }
        \label{fig:res_mmnist_mld}
\end{figure}

\paragraph{\polymnist.} In \Cref{fig:res_mmnist_mld}, we remark the superiority of \gls{MLD} in both generative coherence and quality. {MLD-Uni} is not able to leverage the presence of a large number of modalities in conditional generation coherence. Interestingly, an increase in the number of input modalities impacts negatively the performance of \gls{MLD Uni}.

\begin{figure}[h]
     \centering
     \begin{subfigure}{0.19\textwidth}
         \centering
         \includegraphics[width=\linewidth]{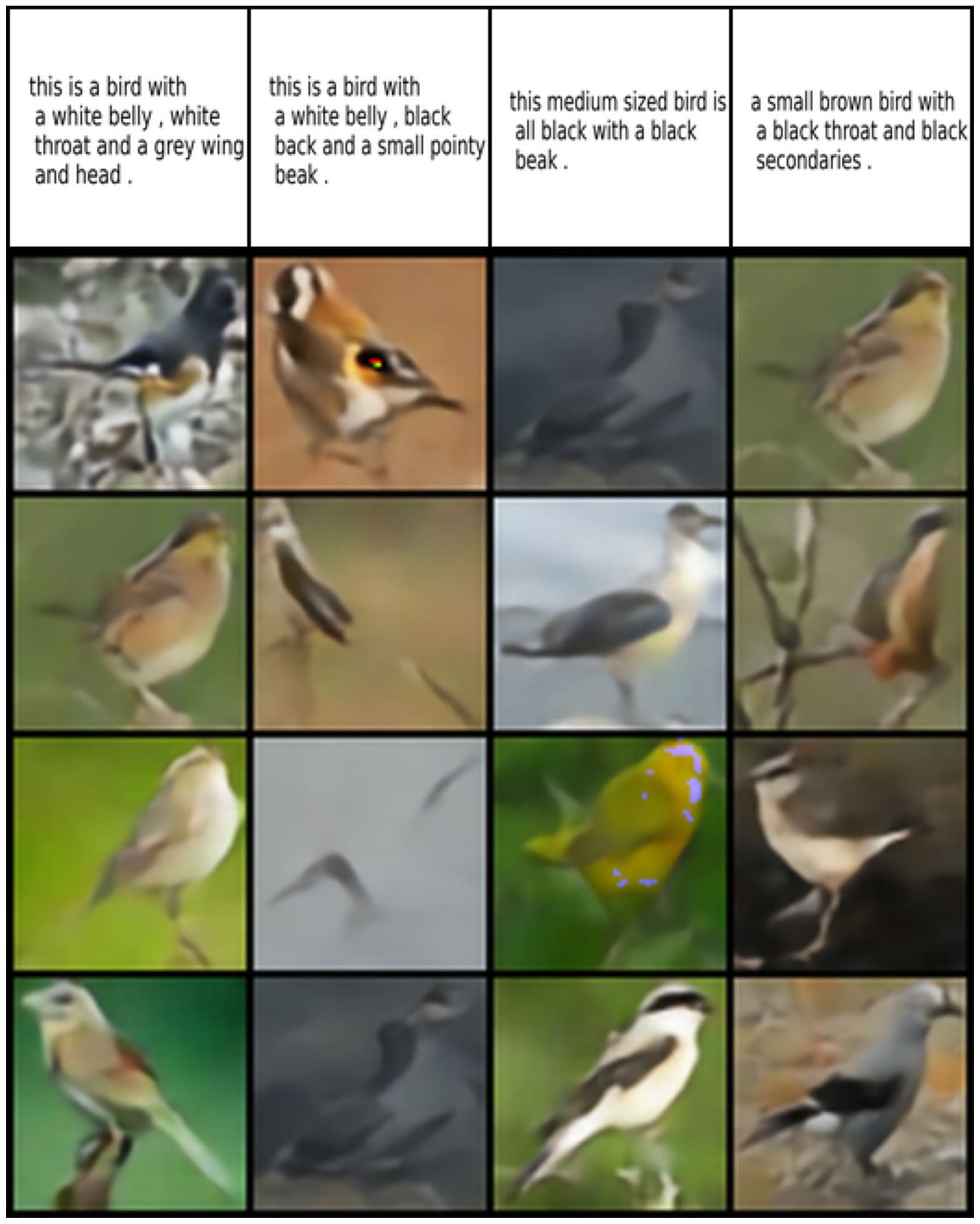}
         \caption*{\gls{MLD Inpaint}} 
     \end{subfigure}
       \begin{subfigure}{0.19\textwidth}
         \centering
         \includegraphics[width=\linewidth]{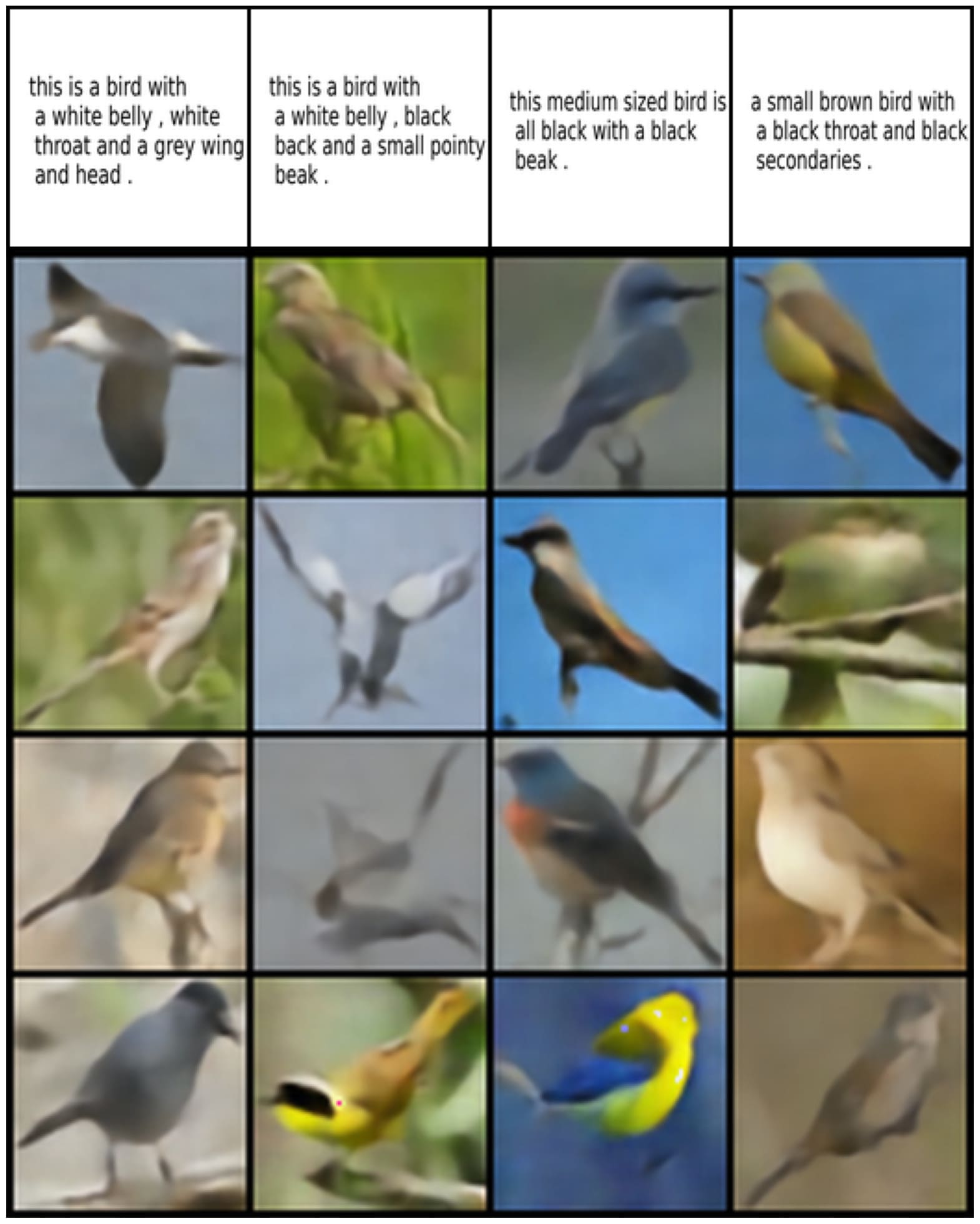}
         \caption*{\gls{MLD Uni}}
     \end{subfigure}
  \begin{subfigure}{0.19\textwidth}
         \centering
         \includegraphics[page=1,width=\linewidth]{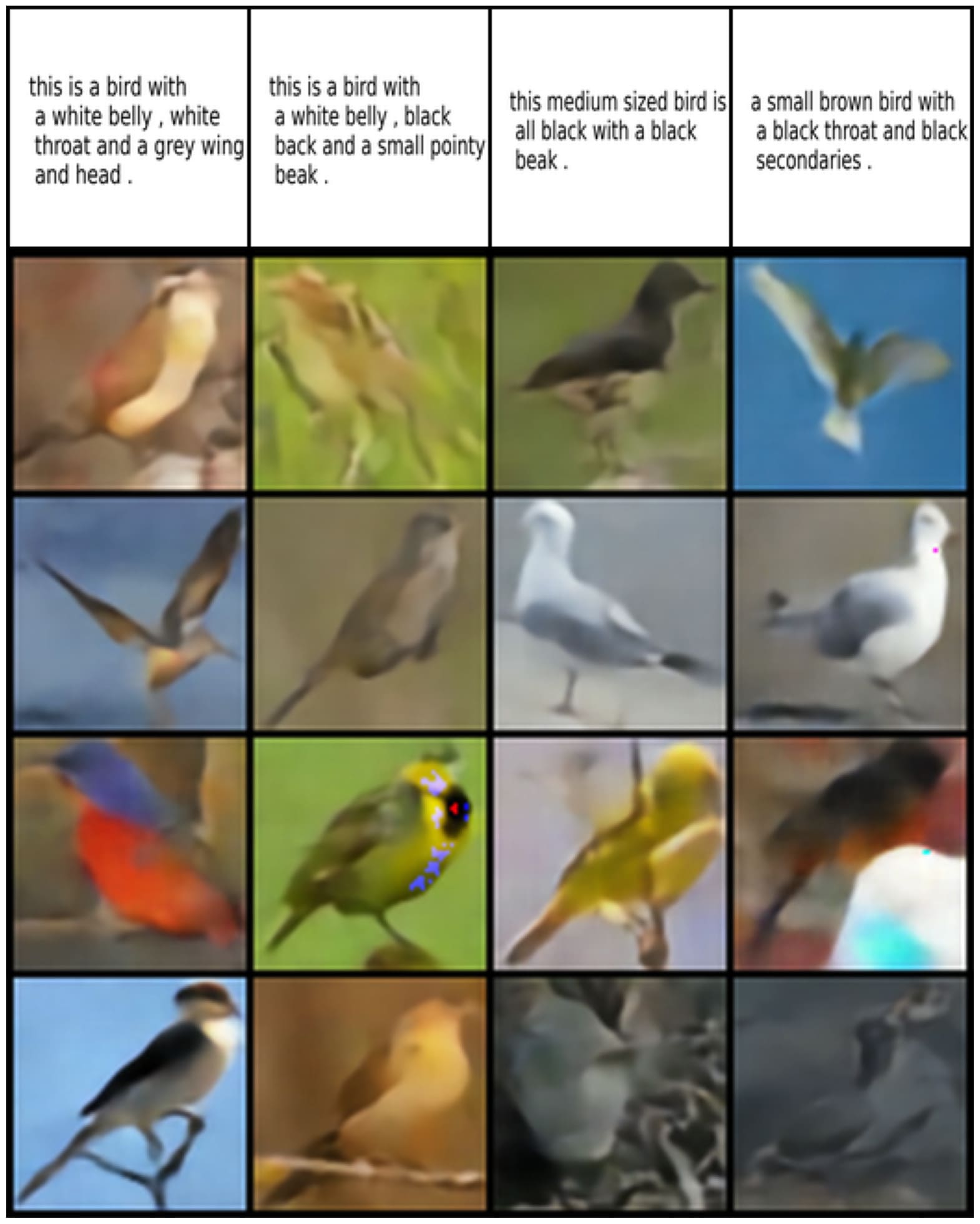}
         \caption*{\textbf{\gls{MLD} }}
   \end{subfigure}

        \caption{Qualitative results on \textbf{\cub} data-set.
        Caption used as condition to generate the bird images.}
       \label{cap_image_cub_mld}
        
\end{figure}

\paragraph{\cub.} \Cref{cap_image_cub_mld} shows qualitative results for caption to image conditional generation. All the variants are based on the same first stage autoencoders, and the generative performance in terms of quality are comparable.

\subsection{Randomization $d$-ablations study}

\begin{figure}[H]
     \centering
         \begin{subfigure}[b]{0.28\textwidth}
         \centering
         \includegraphics[page=1,width=\linewidth]{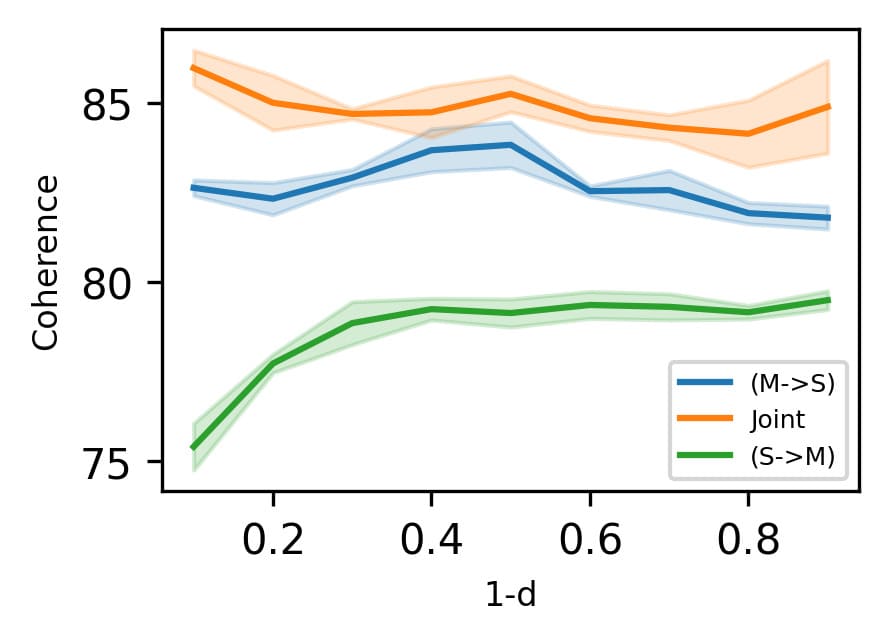}
            \caption{\mnist-\svhn: Coherence (\%$\uparrow$)}
     \end{subfigure}
    \begin{subfigure}[b]{0.28\textwidth}
         \centering
    \includegraphics[page=1,width=\linewidth]{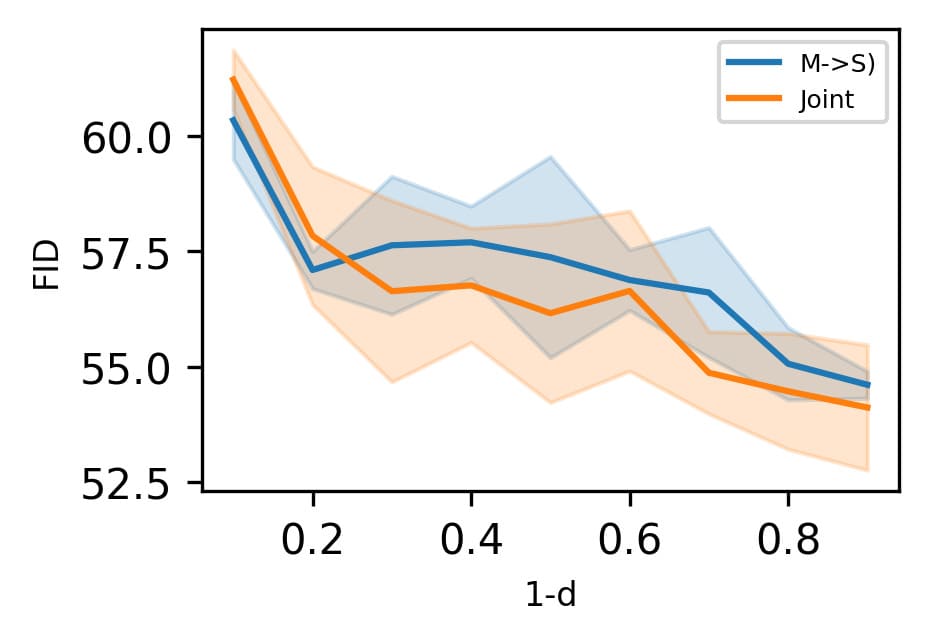}
           \caption{\svhn:\gls{FID} ($\downarrow$)}
     \end{subfigure}
     \begin{subfigure}{0.28\textwidth}
         \centering
         \includegraphics[page=1,width=\linewidth]{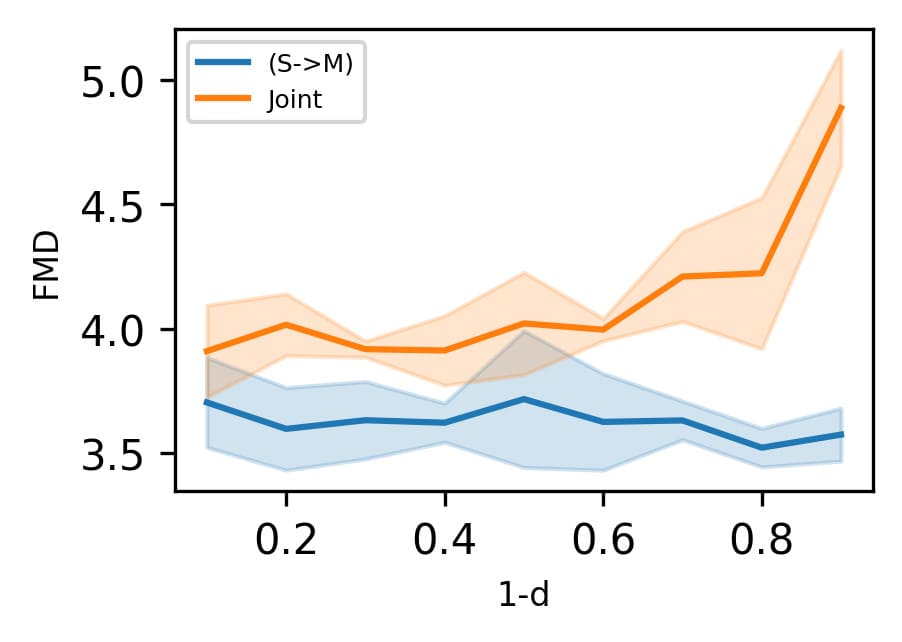}
            \caption{\mnist:\gls{FMD}($\downarrow$)}
     \end{subfigure}
    \caption{The randomization parameter $d$ ablations study on \textbf{\mnist-\svhn}. }
        \label{fig:ablationd}
\end{figure}

The $d$ parameter controls the randomization of the \textit{multi-time masked diffusion process} during training in \Cref{algo:MLD_training}. With probability $d$, the concatenated latent space corresponding to all the modalities is diffused at the same time. With probability $(1-d)$, a portion of the latent space corresponding to a random subset of the modalities is not diffused and freezed during the training step.
To study the parameter $d$ and its effect on the performance of our \gls{MLD} model, we use $d \in \{0.1, .., 0.9\} $.
\Cref{fig:ablationd} shows the $d$-ablations study results on the \textbf{\mnist-\svhn} dataset. 
 We report the performance results averaged over 5 independent seeds as a function of the probability ($1-d$) :
        \textbf{Left:} the conditional and joint coherence for \textbf{\mnist-\svhn} dataset.
        \textbf{Middle:} the quality performance in terms of \gls{FID} for \svhn generation.
         \textbf{Right:} the quality performance in terms of \gls{FMD} for \mnist generation.

We observe that higher value for $1-d$ thus greater probability of applying the \textit{multi-time masked diffusion}, improves the \svhn to \mnist conditional generation coherence. This confirms that the masked multi-time  training enables better conditional generation. Overall, on the \textbf{\mnist-\svhn} dataset, \gls{MLD} shows weak sensibility to the $d$ parameter whenever the value of $d \in [0.2,0.7]$. 

\newpage
\section{Datasets and evaluation protocol}
\label{apdx:dataset_eval}

\subsection{Datasets description}

\begin{figure}[h]
     \centering
     \begin{subfigure}[b]{0.3\linewidth}
      \centering
         \begin{subfigure}{1.0\linewidth}
             \centering
             \includegraphics[width=1.4in]{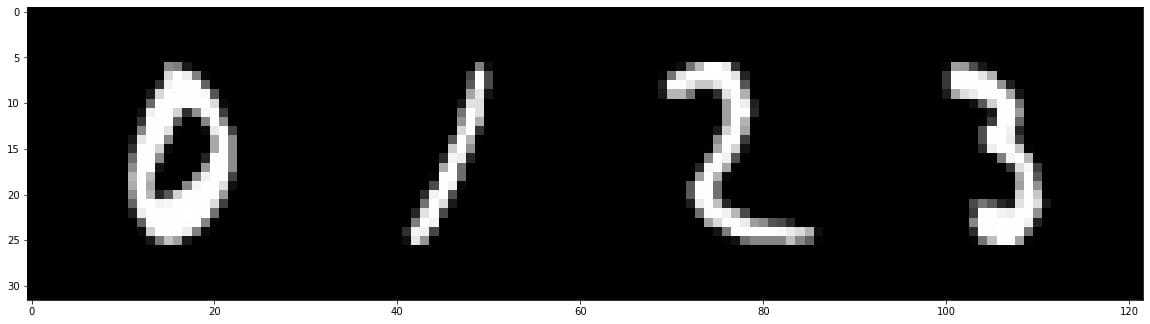}
             \caption*{\mnist}
         \end{subfigure}
         \begin{subfigure}{1.0\linewidth}
             \centering
             \includegraphics[width=1.4in]{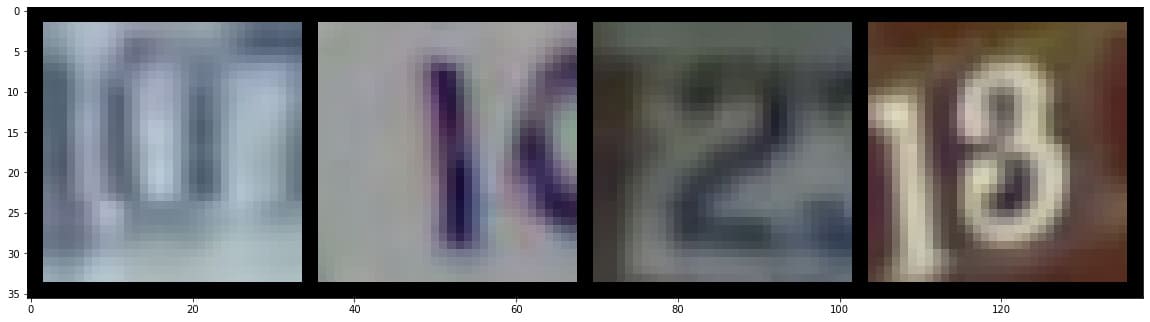}
             \caption*{\svhn}
         \end{subfigure}
        \caption{\textbf{\mnist-\svhn} }\label{fig:ms_dataset}
    \end{subfigure}
     \begin{subfigure}[b]{0.3\linewidth}
      \centering
         \begin{subfigure}{1.0\linewidth}
             \centering
             \includegraphics[width=1.4in]{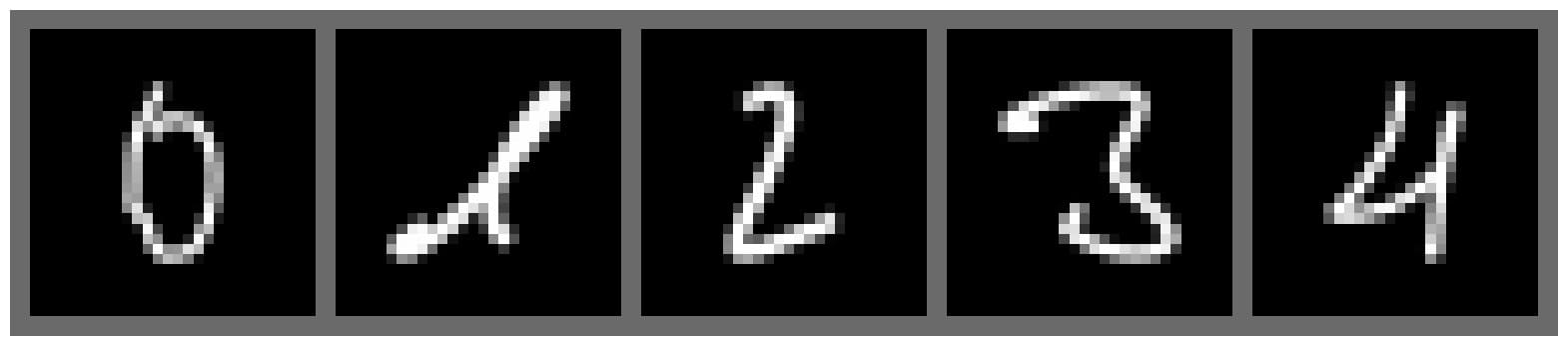}
             \label{fig:mjd_image}
             \caption*{Image}
         \end{subfigure}
         \begin{subfigure}{1.0\linewidth}
             \centering
             \includegraphics[width=1.4in]{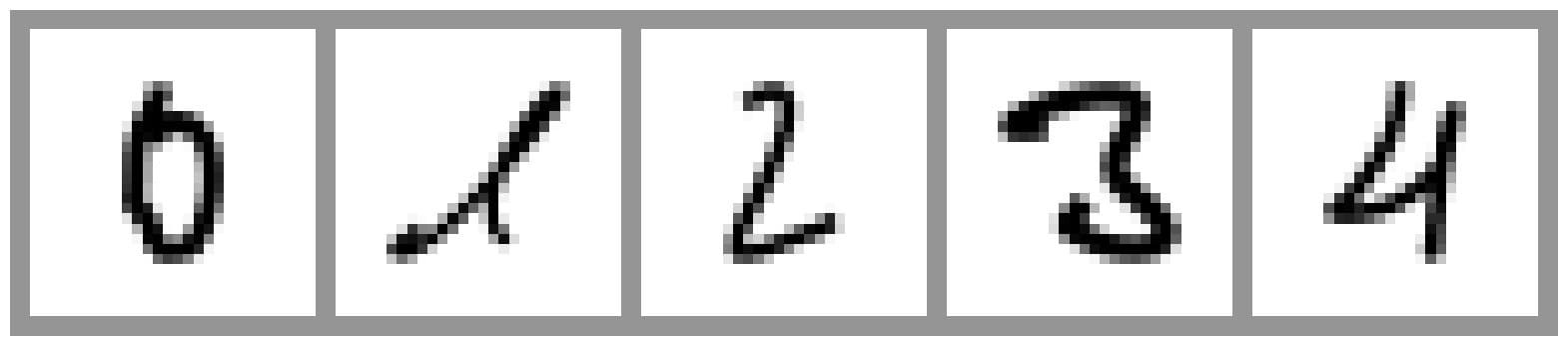}
             \label{fig:mhd_traj}
             \caption*{Trajectory}
         \end{subfigure}
        \begin{subfigure}{1.0\linewidth}
             \centering
             \includegraphics[width=1.4in]{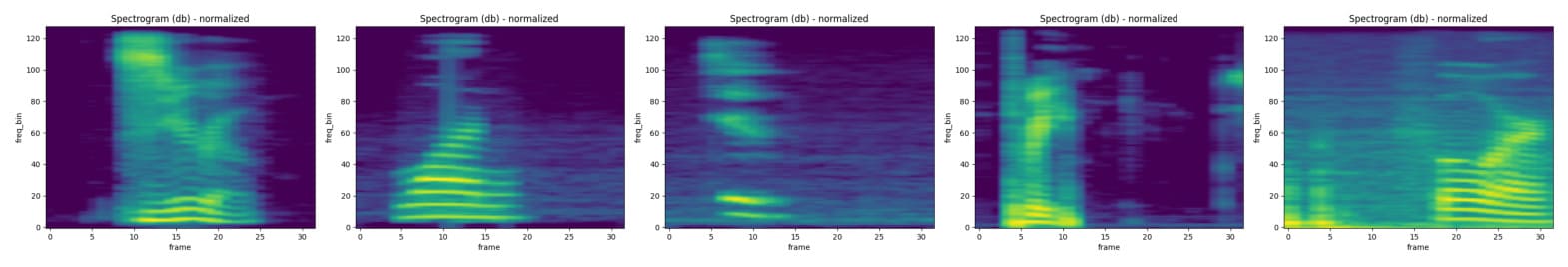}
             \label{fig:mhd_sound}
             \caption*{Sound Mel-Spectogram}
         \end{subfigure}
        \caption{\textbf{\mhd}}
    \end{subfigure}
    \begin{subfigure}[b]{0.3\linewidth}
     \centering
         \begin{subfigure}{1.0\linewidth}
             \centering
             \includegraphics[width=1.4in]{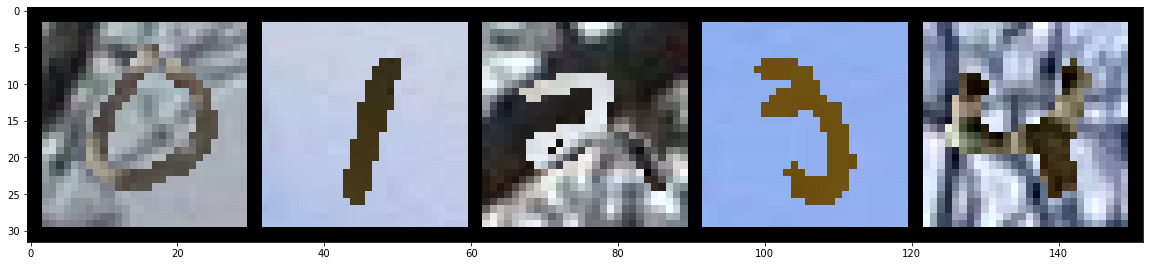}
             \label{fig:mm0}
             
         \end{subfigure}
          \begin{subfigure}{1.0\linewidth}
             \centering
             \includegraphics[width=1.4in]{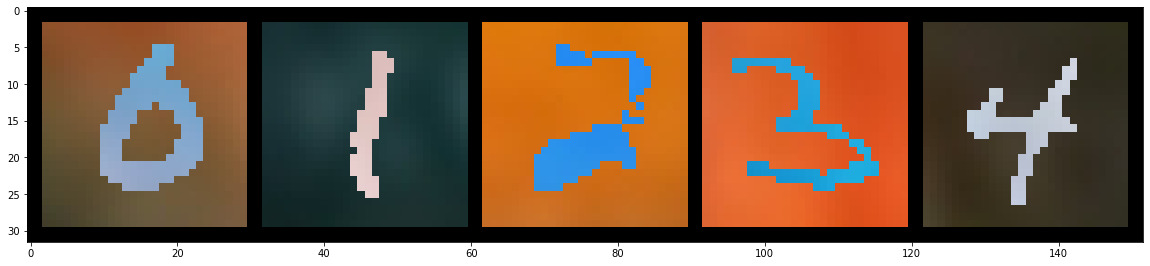}
             \label{fig:mm1} 
          
         \end{subfigure}
          \begin{subfigure}{1.0\linewidth}
             \centering
             \includegraphics[width=1.4in]{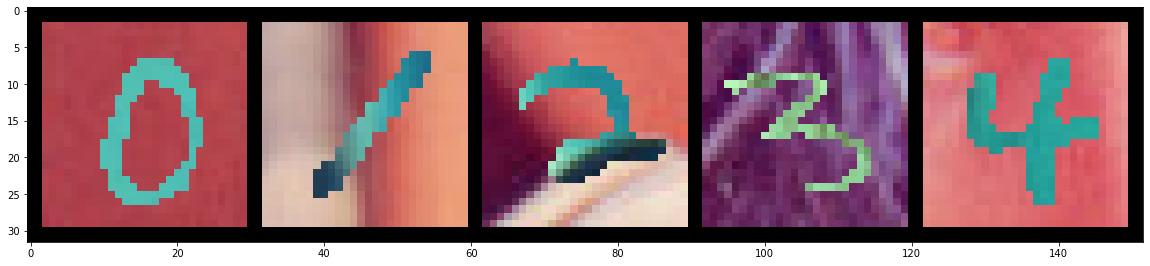}
             \label{fig:mm2}
             
         \end{subfigure}
           \begin{subfigure}{1.0\linewidth}
             \centering
             \includegraphics[width=1.4in]{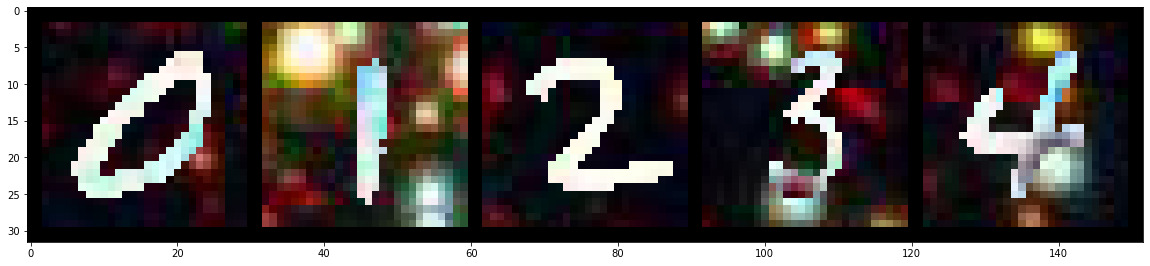}
             \label{fig:mm3}
         \end{subfigure}
           \begin{subfigure}{1.0\linewidth}
             \centering
             \includegraphics[width=1.4in]{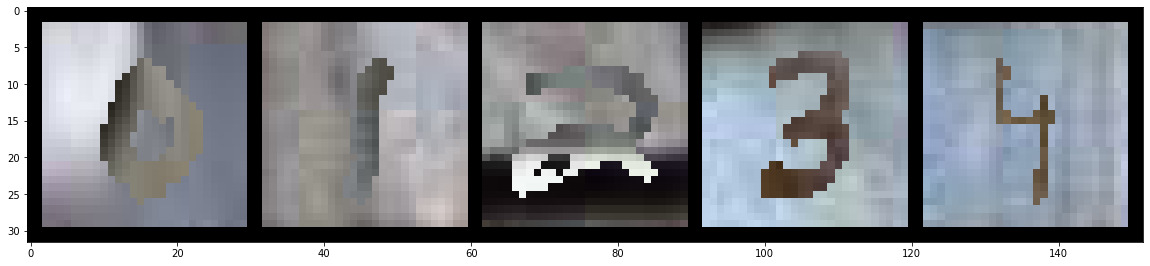}
             \label{fig:mm4}
             
         \end{subfigure}
            \caption{\textbf{ \polymnist}} \label{fig:datasets_mmnist}
            
    \end{subfigure}
    
       \begin{subfigure}{0.5\linewidth}
             \centering
             \includegraphics[width=1.4in]{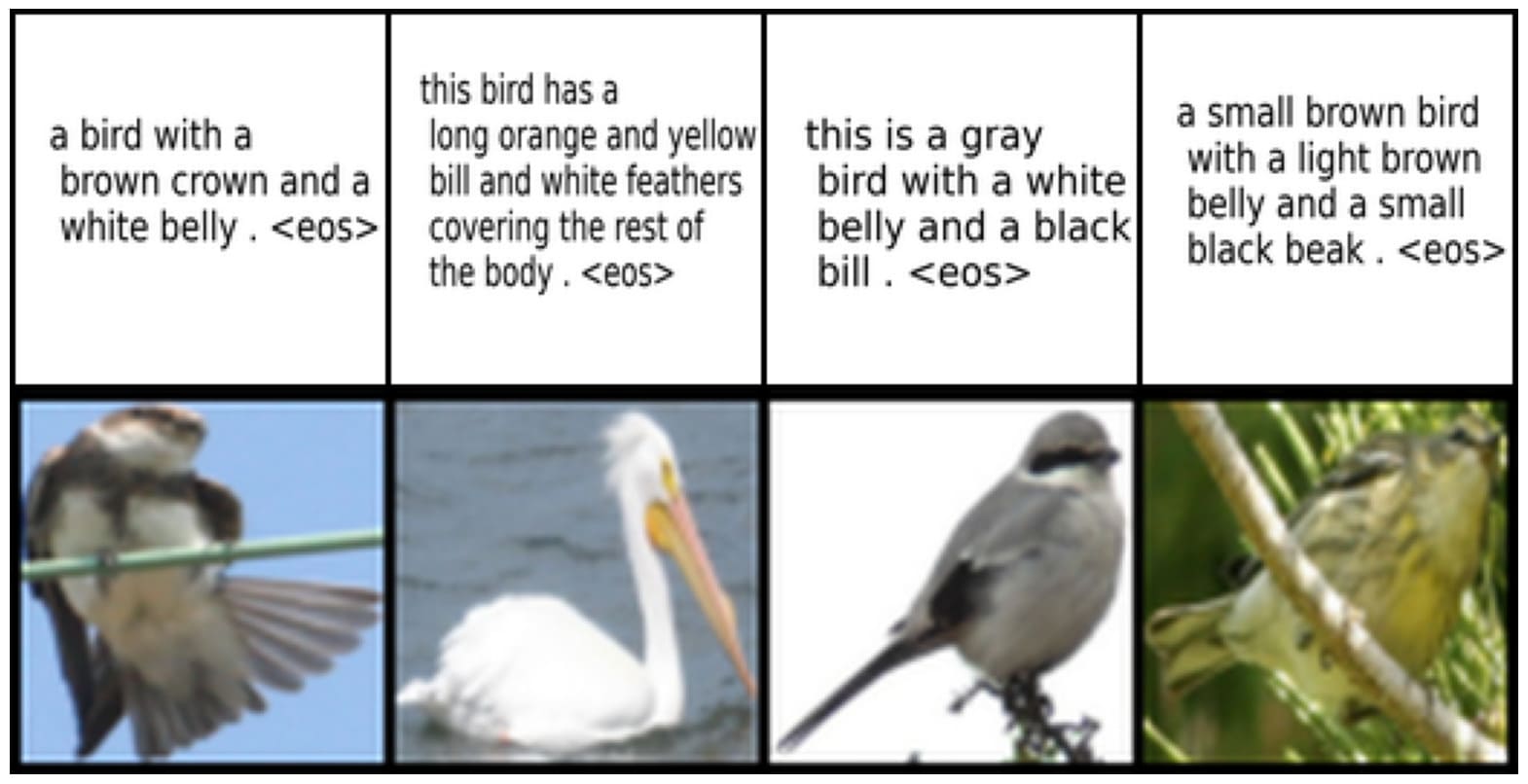}
             \caption{\textbf{ \cub}}
             \label{fig:cub_d}
             
         \end{subfigure}
        \caption{Illustrative example of the Datasets used for the evaluation }\label{fig:datasets}
\end{figure}

\textbf{\mnist-\svhn}  \cite{mmvaeShi2019} is constructed using pairs of \mnist and \svhn, sharing the same digit class (See \Cref{fig:ms_dataset}). Each instance of a digit class (in either dataset) is randomly paired with 20 instances of the same digit class from the other data-set. \svhn modality samples are obtained from house numbers in Google Street View images, characterized by a variety of colors, shapes and  angles. A high number of \svhn samples are noisy and can contain different digits within the same sample due to the imperfect cropping of the original full house number image. One challenge of this data-set for multi-modal generative models is to learn to extract digit number and reconstruct a coherent \mnist modality.

\textbf{\mhd}  \cite{Vasco2022} is composed of 3 modalities: synthetically generated images and motion trajectories of handwritten digits associated with their speech sounds. The images are gray scale $1 \cross 28 \cross 28$ and the handwriting trajectory are represented by a $1 \cross 200$ vector. The spoken digits sound is $1s$ long audio processed as Mel-Spectrograms constructed with a hopping window of $512$ ms with $128$ Mel Bins  resulting in a $1 \cross 128 \cross 32$ representation. This benchmark is the closest to a real world multi-modal sensors scenario because of the presence of three completely different modalities, the audio modality representing a complex data type. Therefore, similar to \svhn, the conditional generation of sound to coherent images or trajectories represents a challenging use case.

\textbf{\polymnist} \cite{sutter2021generalized} is an extended version of the \mnist data-set to 5 modalities. Each modality is constructed using a randomly set of \mnist digits with an overlay over a random crop from a modality specific, 3 channel image background. This synthetic generated data-set allows the evaluation of the scalability of multi-modal generative models to large number of modalities. Although this data-set is composed of only images, the different modality-specific background having different textures, results in different levels of difficulty. In \Cref{fig:datasets_mmnist}, the digits numbers are more difficult to distinguish in modality 1 and 5 than in the remaining modalities.

\textbf{\cub} \cite{mmvaeShi2019} is comprised of bird images and their associated text captions. The work in \cite{mmvaeShi2019} used a simplified version based on pre-computed ResNet-features. We follow \cite{Daunhawer2021} and conduct all our experiments on the real image data instead. Each image from the 11,788 photos of birds from Caltech-Birds \cite{wah2011caltech} are resized to $ 3 \cross 64 \cross 64 $ image size and coupled with 10 textual descriptions of the respective bird (See \Cref{fig:cub_d}).

\subsection{Evaluation metrics}
Multimodal generative models are evaluated in terms of generative coherence and quality.

\subsubsection{Generation Coherence}
\label{apdx:eval_coh}
 We measure \textit{coherence} by verifying that generated data (both for joint and conditional generations) share the same information across modalities. Following \cite{mmvaeShi2019, sutter2021generalized, tcmvaehwang21,Vasco2022,Daunhawer2021}, we consider the class label of the modalities as the shared information and use pre-trained classifiers to extract the label information form the generated samples and compare it across modalities.

For \textbf{\mnist-\svhn} , \textbf{\mhd}  and \textbf{\polymnist}, the shared semantic information is the digit class number. Single modality classifiers are trained to classify the digit number of a given modality sample. To compute the conditional generation of modality $m$ with a subset of modalities $A$, we feed the modality specific pre-trained classifier $\mathbf{C}_m$ with the conditional generated sample $\hat{X}^m$. The predicted label class is compared to the ground truth label $y_{X^{A}}$ which is the label of modalities of the subset $X^{A}$. For $N$ samples, the matching rate average establishes the coherence. For all the experiments, $N$ is equal to the length of the test-set.
\begin{equation}
    \textit{Coherence}( \hat{X}^m | X^{A}  ) = \frac{1}{N} \sum_1^N \mathds{1}_{ \{ \mathbf{C}_m ( \hat{X}^m) = y_{X^{A}} \}}  
\end{equation}

The \textbf{joint generation coherence} is measured by feeding the generated samples of each modality to their specific trained classifier. The rate with which all classifiers output the same predicted digit label for $N$ generations is considered as the joint generation coherence. 

\textbf{The leave one out coherence:} is the conditional generation coherence using all the possible subsets excluding the generated modality: $\textit{Coherence}( \hat{X}^m | X^A  )$ with $A= \{1,..,M \} \setminus m $ ).
Due to the large number of modalities in \textbf{\polymnist}, similar to  \cite{ sutter2021generalized, tcmvaehwang21,Daunhawer2021}  we compute the average \textbf{leave one out coherence} conditional coherence as a function of the input modalities subset size.

Due to the unavailability of labels in the \textbf{\cub} data-set, we use \gls{CLIP-S} \cite{hessel2021clipscore} a state of the art metric for image captioning evaluation.


\subsubsection{Generation Quality}
For each modality, we consider the following metrics: 
\begin{itemize}
    \item \textbf{RGB Images}: \gls{FID} \cite{fid} is the  state-of-the-art standard metric to evaluate image generation quality of generative models.
    \item \textbf{Audio}: \gls{FAD} \cite{fad},  is state-of-the-art standard metric in the evaluation of audio generation. \gls{FAD} performs well in terms of robustness against noise and is consistent with human judgments \cite{audiomet}. Similar to \gls{FID}, a Fréchet distance is computed but VGGish (audio classifer model) embeddings are used instead.
    \item \textbf{Other modalities} For other modality types, we derive \gls{FMD} (Fréchet Modality Distance), a similar metric to  \gls{FID} and \gls{FAD}. We compute the \textbf{Fréchet distance} between the statistics retrieved from the activations of the modality specific pre-trained classifiers used for coherence evaluation. \gls{FMD} is used to evaluate the generative quality of \mnist modality in \textbf{\mnist-\svhn} and image and trajectory modalities in \textbf{\mhd} data-set.
\end{itemize}
For conditional generation, we compute the quality metric (\gls{FID},\gls{FAD} or \gls{FMD}) using the conditionally generated modality and the real data. For joint generation, we use the randomly generated modality and randomly selected same number of samples from the real data.

 For \textbf{\cub}, we use $10000$ samples to evaluate the generation quality in terms of \gls{FID}.
In the remaining experiments, we use $5000$ samples to evaluate the performance in terms of \gls{FID}, \gls{FAD} or \gls{FMD}.

\newpage
\section{Implementation details}
\label{apdx:implementation}

We report in this section the implementation details for each benchmark. We used the same unified code-base for all the baselines, using the \textit{PyTorch} framework. The \gls{VAE} implementation is adapted from the official code whenever it's available (\gls{MVAE}, \gls{MMVAE} and \gls{MOPOE} as in \footnote{\url{https://github.com/thomassutter/MoPoE}}, \gls{MVTCAE} \footnote{\url{https://github.com/gr8joo/MVTCAE}} and \gls{NEXUS}\footnote{\url{https://github.com/miguelsvasco/nexus_pytorch}} ). For fairness, \gls{MLD} and all the \gls{VAE}-based models use the same autoencoder architecture. We use the best hyper-parameters suggested by the authors. Across all the data-sets, we use the \textit{Adam optimizer} \cite{kingma2014adam} for training.

\subsection{MLD}
\gls{MLD} uses the same autoencoders architecture used for \gls{VAE}-based models, except that these are deterministic autoencoders. The autoencoders are trained using the same reconstruction loss term as for the \gls{VAE}-based models. \Cref{table:mld_ae_implementation} and \Cref{table:mld_score_hyper} summarize the hyper-parameters used during the two phases of \gls{MLD} training.
Note that for the image modality in the \cub dataset, to overcome over-fitting in training the deterministic autoencoder, data augmentation was necessary (we used \textit{TrivialAugmentWide} from the Torchvision library). 

\begin{table}[h]
\caption{\gls{MLD}: The deterministic autoencoders hyper-parameters}
\centering
\begin{tabular}{l|ccccccccc}
\toprule
Dataset & Modality & Latent space & Batch size & Lr & Epochs & Weight decay    \\
\midrule 
\multirow{2}{*}{ \textbf{\mnist-\svhn} }
& \mnist & 16 &  \multirow{2}{*}{ 128 } & \multirow{2}{*}{ 1e-3 } & \multirow{2}{*}{ 150 }   \\
& \svhn & 64 &  \\\midrule
\multirow{3}{*}{ \textbf{\mhd} }
& Image & 64 &\multirow{3}{*}{ 64 } & \multirow{3}{*}{ 1e-3 } & \multirow{3}{*}{ 500 }\\
& Trajectory & 16   \\
& Sound & 128   \\ \midrule
\multirow{1}{*}{ \textbf{\polymnist} }
& All modalities & 160 & 128 & 1e-3  &300 \\ \midrule
\multirow{2}{*}{ \textbf{\cub} } 
& Caption & 32 & \multirow{2}{*}{ 128 } & 1e-3 & 500 \\
& Image  & 64 &  & 1e-4 &  300 & 1e-6   \\
\midrule
\multirow{3}{*}{ \textbf{CelebAMask-HQ} }
& Image & 256 &\multirow{3}{*}{ 64 } & \multirow{3}{*}{ 1e-3 } & \multirow{3}{*}{ 200 }\\
& Mask & 128   \\
& Attributes & 32  \\

\bottomrule
\end{tabular}
\label{table:mld_ae_implementation}
\end{table}

\begin{table}[h]
\caption{\gls{MLD}: The score network hyper-parameters}
\centering
\begin{tabular}{l|ccccccccc}
\toprule
Dataset & $ d $ &  Blocks & Width &Time embed & Batch size & Lr & Epochs  \\
\midrule 
\textbf{\mnist-\svhn}& 0.5  & 2 & 512 &256& 128 &  \multirow{4}{*}{ 1e-4 } & 150   \\
\textbf{\mhd}& 0.3  & 2 & 1024 & 512 &128&  & 3000   \\
\textbf{\polymnist}& 0.5  & 2 & 1536 &512 &256&  & 3000   \\
\textbf{\cub}& 0.7 & 2 & 1024 & 512 &64&  & 3000   \\
\textbf{CelebAMask-HQ}& 0.5 & 2 & 1536 & 512 &64&  & 3000   \\
\bottomrule
\end{tabular}
\label{table:mld_score_hyper}
\end{table}

\subsection{VAE-based models}
For \textbf{\mnist-\svhn}, we follow \cite{sutter2021generalized,mmvaeShi2019} and use the same autoencoder architecture and pre-trained classifier.The latent space size is set to $20$, $\beta = 5.0 $. For \gls{MVTCAE} $\alpha = \frac{5}{6} $. For both modalities, the likelihood is estimated using Laplace distribution.
For \gls{NEXUS}, we use the same modalities latent space sizes as in \gls{MLD}, the joint \gls{NEXUS} latent space is set to $20$, $\beta_i = 1.0 $ and $\beta_c = 5.0$. We train all the \gls{VAE}-models for $150$ epochs with $256$ batch size and learning rate of $1e-3$.

For \textbf{\mhd}, we reuse the autoencoders architecture and pre-trained classifier of \cite{Vasco2022}. We adopt the hyper-parameters of \cite{Vasco2022} to train \gls{NEXUS} model with the same settings, besides discarding the label modality. For the remaining \gls{VAE}-based models, the latent space size is set to $128$, $\beta = 1.0 $ and $\alpha = \frac{5}{6} $ for \gls{MVTCAE}. For all the modalities, \gls{MSE} is used to compute the reconstruction loss, similar to \cite{Vasco2022}. These models are trained for $600$ epochs with  $128$ batch size and learning rate of $1e-3$.

For \textbf{\polymnist}, we use the same autoencoders architecture and pretrained classifier used by \cite{sutter2021generalized, tcmvaehwang21}.
We set the latent space size to $512$, $\beta = 2.5 $ and $\alpha = \frac{5}{6} $ for \gls{MVTCAE}. For all the modalities, the likelihood is estimated using Laplace distribution. For \gls{NEXUS}, we use the same modality latent space size as in \gls{MLD}, the joint \gls{NEXUS} latent space to $64$, $\beta_i = 1.0 $ and $\beta_c = 2.5$. We train all the models for $300$ epochs with $256$ batch size and learning rate of $1e-3$.

For \textbf{\cub}, we use the same autoencoders architecture and implementation settings as in \cite{Daunhawer2021}. Laplace and one-hot categorical distributions are used to estimate likelihoods of the image and caption modalities respectively. The latent space size is set to $64$, $\beta = 9.0 $ for \gls{MVAE}, \gls{MVTCAE} and \gls{MOPOE} and $\beta =1$ for \gls{MMVAE}. We set $\alpha = \frac{5}{6} $ for \gls{MVTCAE}.
For \gls{NEXUS}, we use the same modalities latent space sizes as in \gls{MLD}, the joint \gls{NEXUS} latent space is set to $64$, $\beta_i = 1.0 $ and $\beta_c = 1$. We train all the models for $150$ epochs with $64$ batch size, with learning rate of $5e-4$ for \gls{MVAE}, \gls{MVTCAE} and \gls{MOPOE} and $1e-3$ for the remaining models.

Finally, note that in the official implementation of \cite{sutter2021generalized} and \cite{tcmvaehwang21}, for the \textbf{\polymnist} and \textbf{\mnist-\svhn} data-sets, the classifiers were used for evaluation using dropout. In our implementation, we make sure to deactivate dropout during evaluation step.

\subsection{MLD with powerfull autoencoder}

\begin{figure}[H]
     \centering
     \begin{subfigure}{0.30\textwidth}
         \centering
         \includegraphics[width=\linewidth]{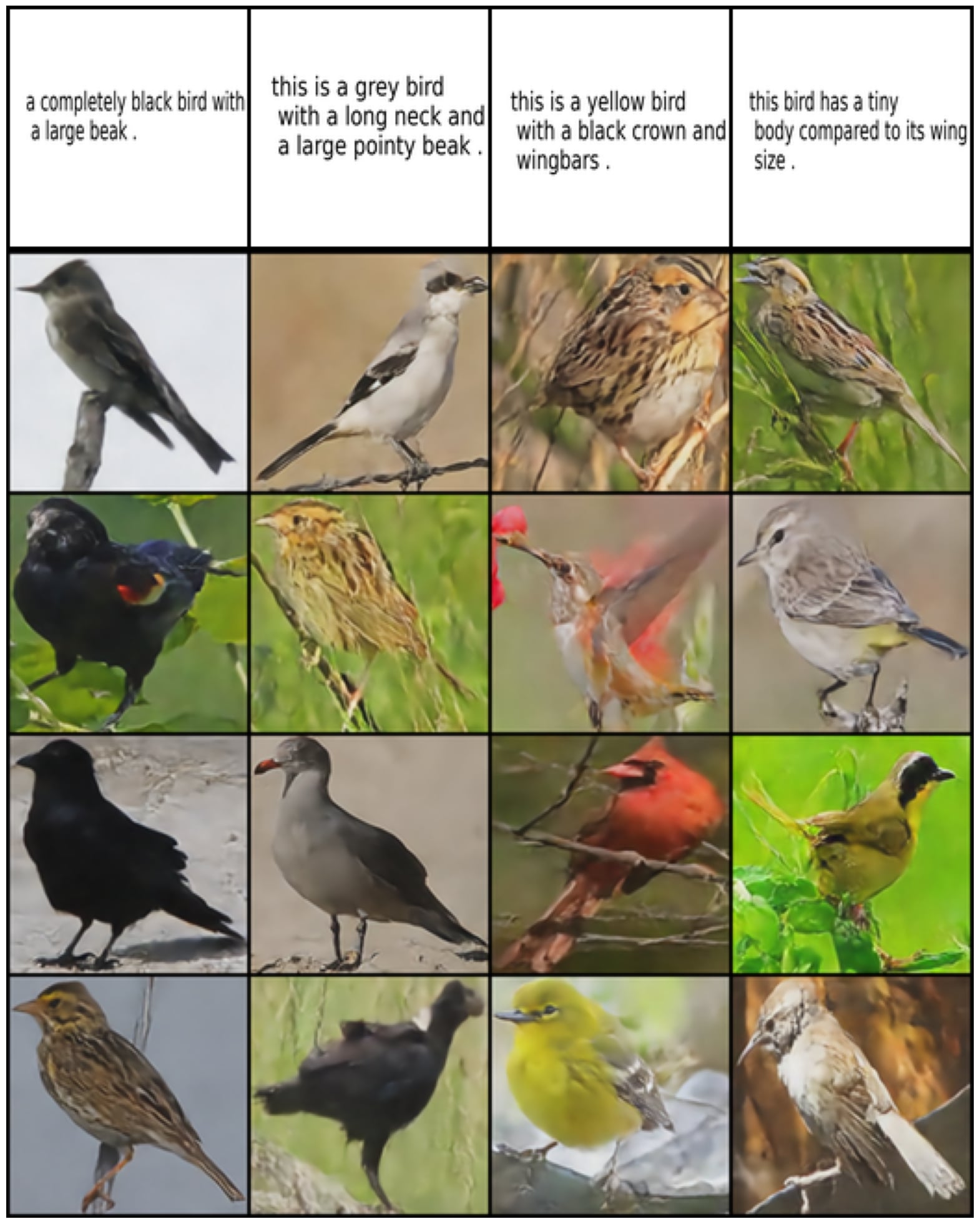}
          \caption{ Conditional Generation: Caption used as condition to generate the bird images.}
    \end{subfigure}
  \begin{subfigure}{0.69\textwidth}
         \centering
         \includegraphics[width=\linewidth]{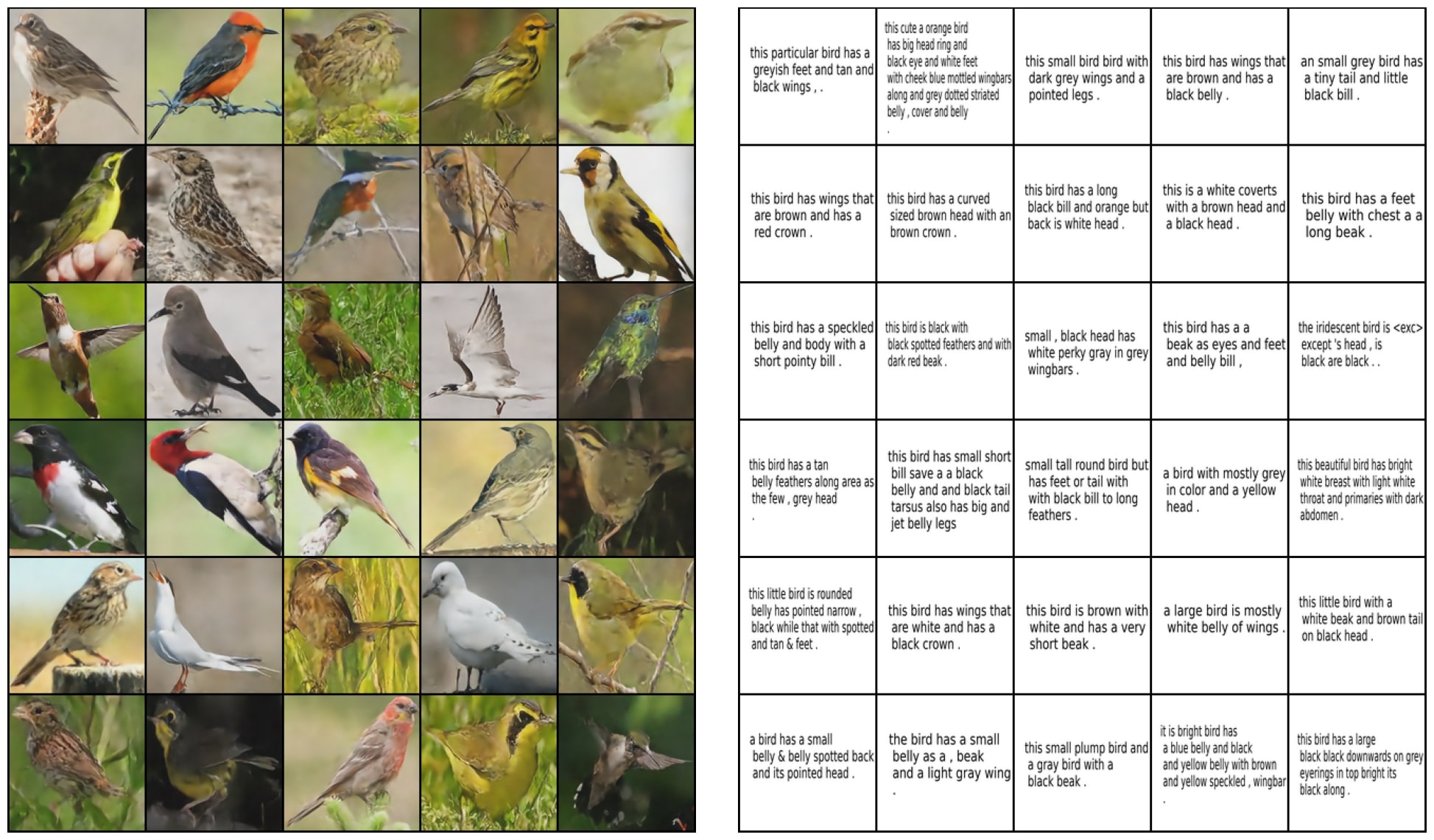}
         \caption{Joint generation: Images and captions are generated simultaneously.}
        
\end{subfigure}
    
        \caption{Qualitative results of \textbf{\gls{MLD}*} on \textbf{CUB} data-set with $128 $x$128$ resolution with powerful image autoencoder.
        }
       \label{cub128}
\end{figure}

Here we provide more detail about the CUB experiment using 
 more powerful autoencoder denoted \gls{MLD}* in \Cref{cap_image_cub}. We use an architecture similar to \cite{Rombach_2022_CVPR} adapted to (64X64) resolution images. We modified the autoencoder architecture to be deterministic and train the model with a simple Mean square error loss. We kept the same configuration of the CUB experiment described in the previous experiment on the same dataset including the text autoencoder, score network and hyper-parameters. 
We also perform experiments with the same settings on (128X128) resolution images. We included the qualitative results in \cref{cub128}.

\subsection{Computation Resources}
In our experiments, we used 4 A100 GPUs, for a total of roughly 4 months of experiments.

\newpage
\section{Additional results}
\label{apdx:additionnal_res}
In this section, we report detailed results for all of our experiments, including standard deviation and additional qualitative samples, for all the data-sets and all the methods we compared in our work.

\subsection{MNIST-SVHN}

\subsubsection{Self reconstruction}

In \Cref{self_coh_quality:ms}  we report results about \textit{self-coherence}, which we use to support the arguments from \Cref{sec:motivation}. This metric is used to measure the loss of information due to latent collapse, by showing the ability of all competing models to reconstruct an arbitrary modality given the same modality or a set thereof as an input. For our \gls{MLD} model, the self-reconstruction is done without using the diffusion model component: the modality is encoded using its deterministic encoder and the decoder is fed with the latent space to get the reconstruction. 

We observe that \gls{VAE} based models fail at reconstructing \svhn given \svhn. This is especially more visible for product of experts based models (\gls{MVAE} and \gls{MVTCAE}. In \gls{MLD}, the deterministic autoencoders do not suffer from such weakness and achieve overall the best performance. 

\begin{figure}[h]
     \centering
    \begin{subfigure}{0.4\textwidth}
         \centering
         \includegraphics[width=\linewidth]{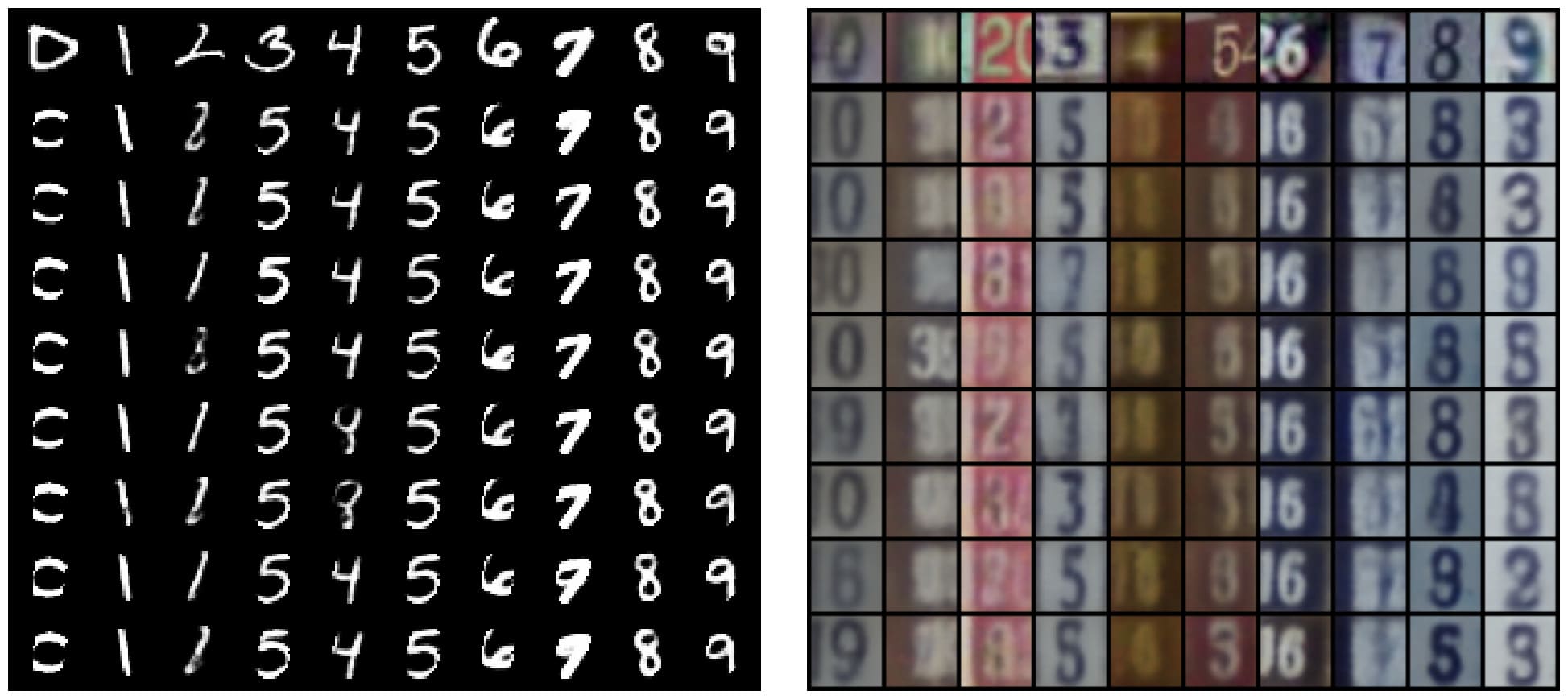}
         \caption*{\gls{MVAE}}
     \end{subfigure}
    \begin{subfigure}[b]{0.4\textwidth}
         \centering
         \includegraphics[width=\linewidth]{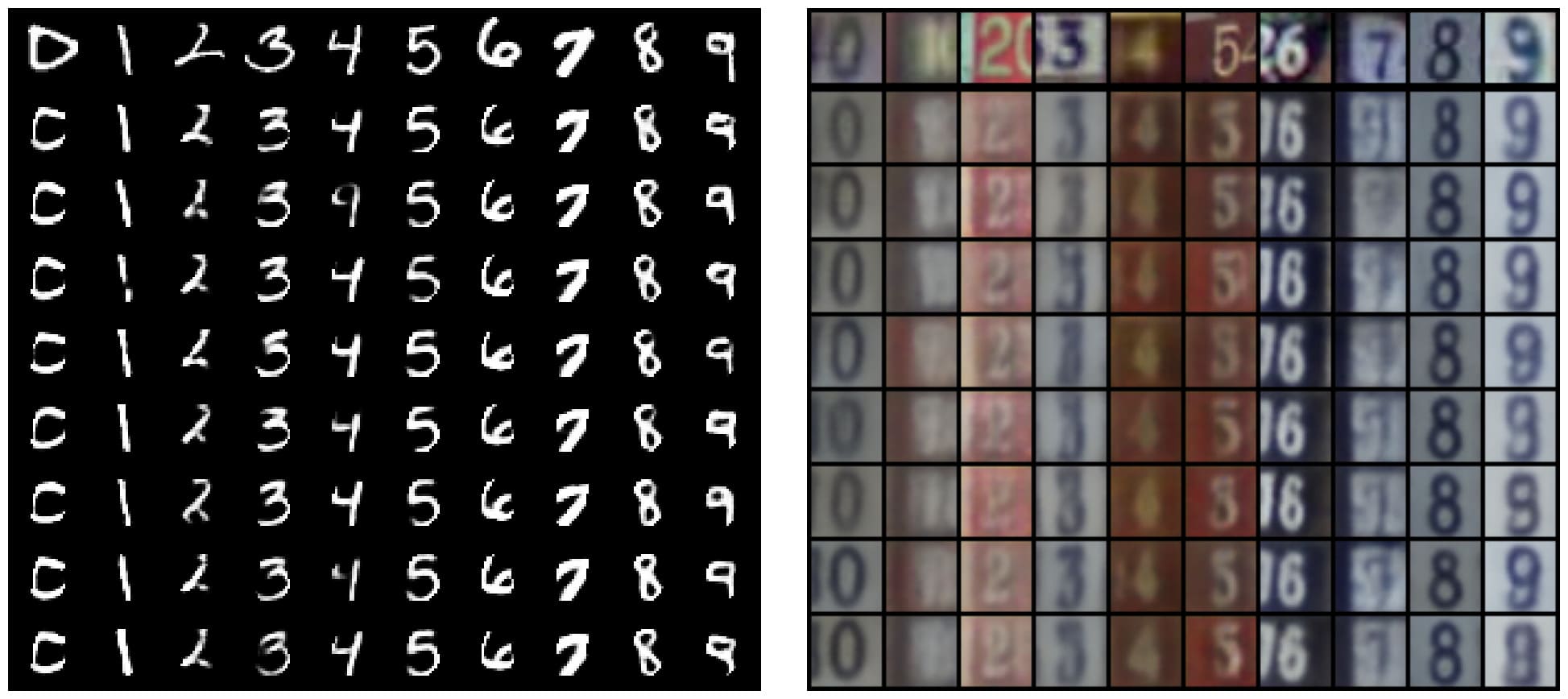}
         \caption*{\gls{MMVAE}}
     \end{subfigure}
     
      \begin{subfigure}{0.4\textwidth}
         \centering
         \includegraphics[width=\linewidth]{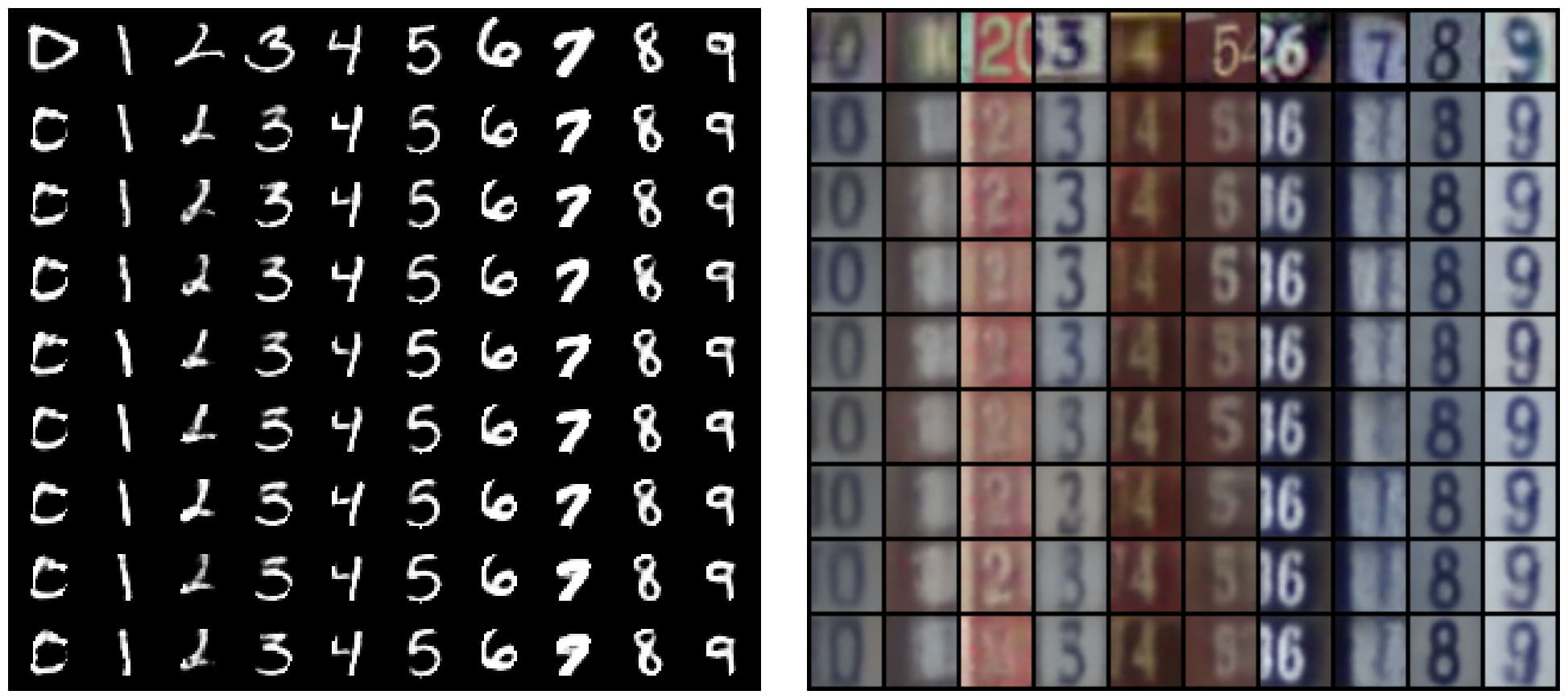}
         \caption*{\gls{MOPOE}}
     \end{subfigure}
     \begin{subfigure}{0.4\textwidth}
         \centering
         \includegraphics[width=\linewidth]{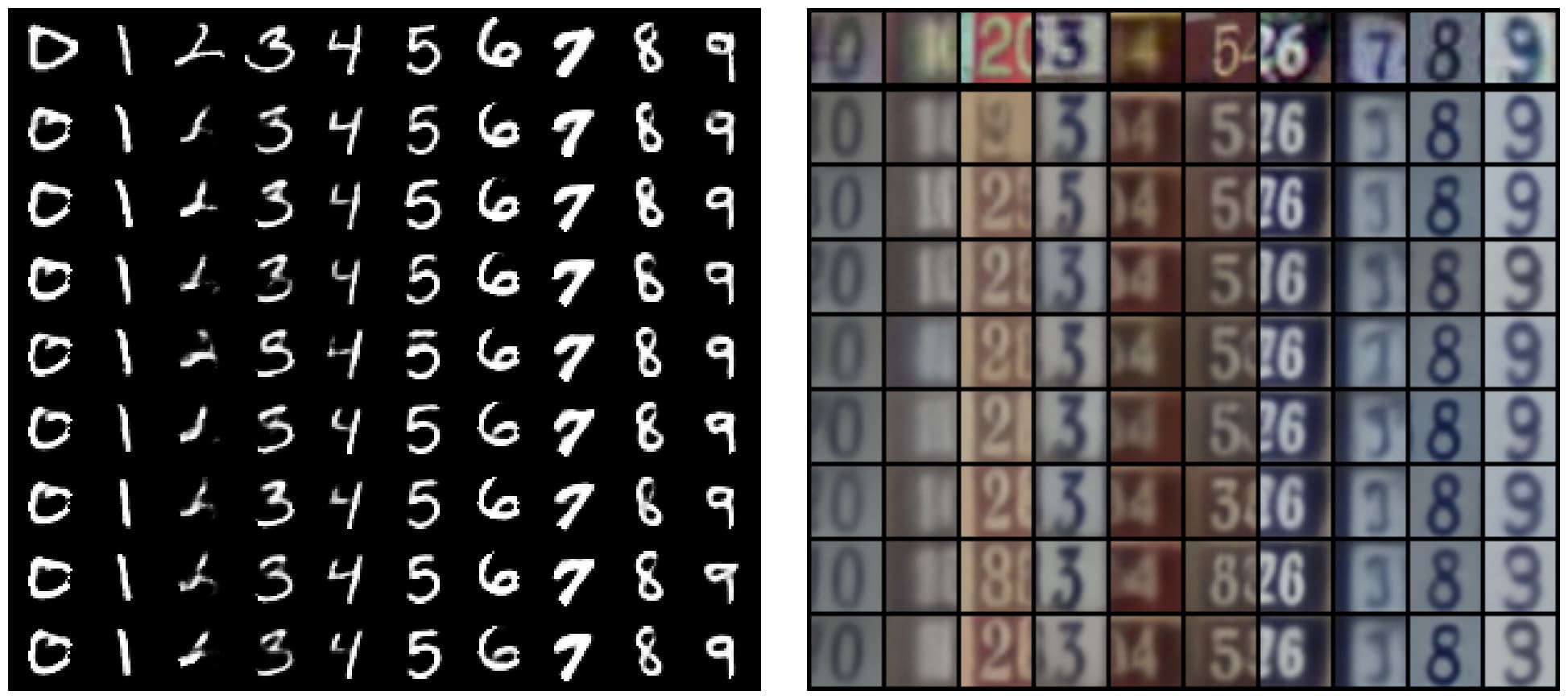}
         \caption*{\gls{NEXUS}}
     \end{subfigure}

    \begin{subfigure}{0.4\textwidth}
         \centering
         \includegraphics[width=\linewidth]{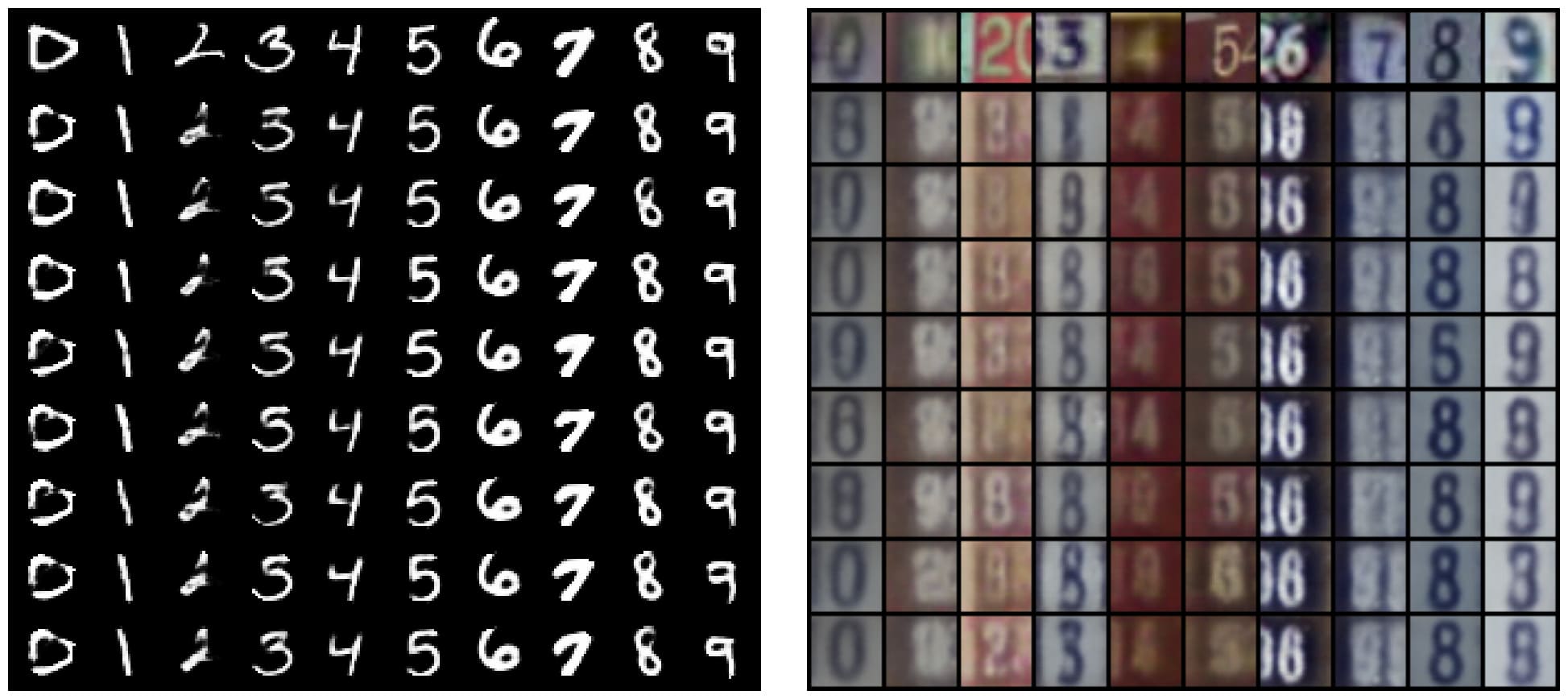}
         \caption*{\gls{MVTCAE}}
     \end{subfigure}
    \begin{subfigure}{0.4\textwidth}
         \centering
         \includegraphics[page=1,width=\linewidth]{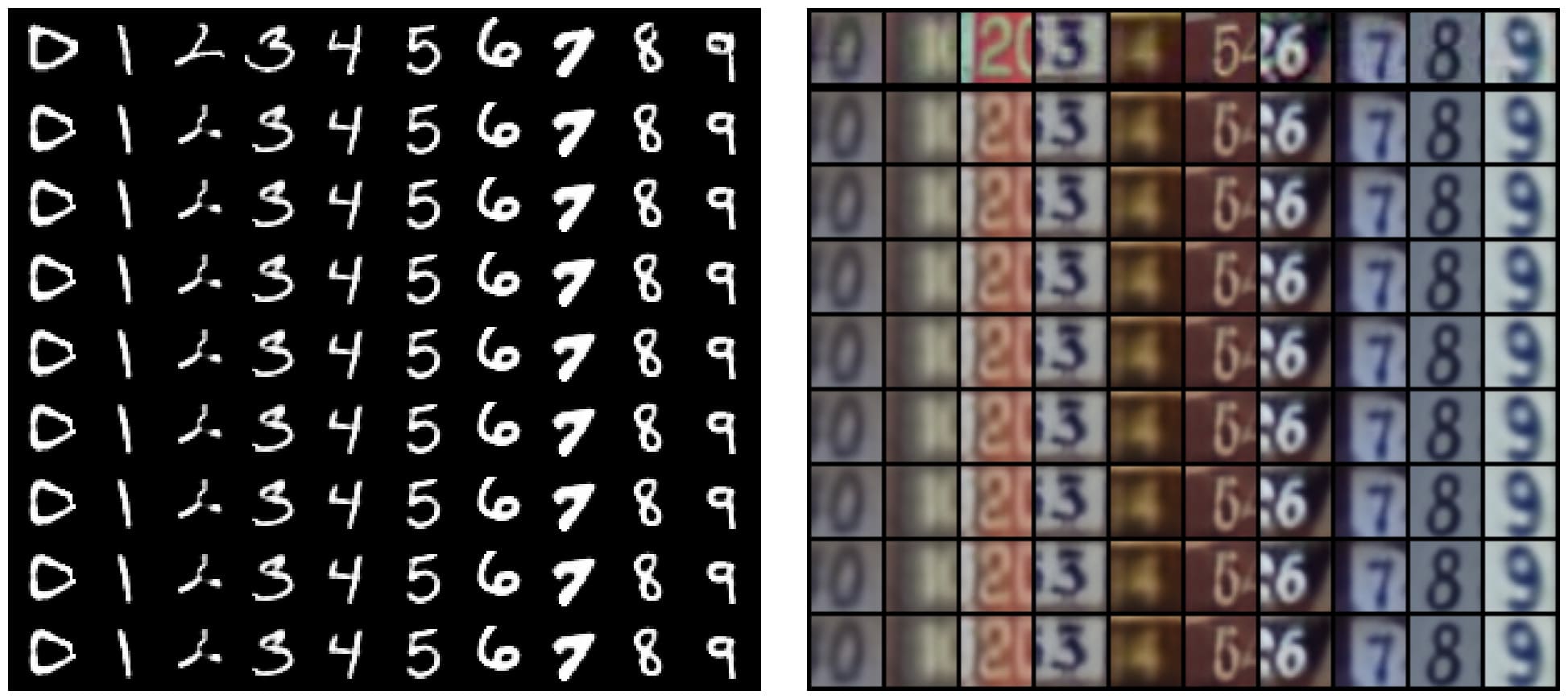}
         \caption*{ \textbf{\gls{MLD} (ours) }  }
     \end{subfigure}
     
        \caption{Self-generation qualitative results for \textbf{\mnist-\svhn}. For each model we report: \mnist to \mnist conditional generation in the left,   \svhn to \svhn conditional generation in the right. \newline }
        \label{fig:self_recon}
\end{figure}

\Cref{fig:self_recon} shows qualitative results for the self-generation. We remark that some samples generated using \gls{VAE}-based models, the digits differs from the ones in the input sample, indicating information loss due to the latent collapse. For example, in the case of \gls{MVAE}, generation of the \mnist digit 3, in \gls{MVTCAE} generation of the \svhn digit 2.

\begin{table}[H]
\centering

\caption{Self-generation coherence and quality for \textbf{\mnist-\svhn} ( M :\mnist, S: \svhn ). The generation quality is measured in terms of \gls{FMD}  for \mnist and \gls{FID} for \svhn.\newline}
 \label{self_coh_quality:ms}
 
\resizebox{\textwidth}{!}{\begin{tabular}{c|cccc|cccc}
\toprule
\multirow{2}{*}{ Models }  & \multicolumn{4}{c}{Coherence (\%$\uparrow$) } &   \multicolumn{4}{c}{Quality ($\downarrow$)} \\
    \cmidrule{2-9}
    & M $\rightarrow$ M & M,S $\rightarrow$ M &S $\rightarrow$ S &  M,S $\rightarrow$ M &  M $\rightarrow$ M & M,S $\rightarrow$ M &S $\rightarrow$ S &  M,S $\rightarrow$ M   \\
    \midrule
     \gls{MVAE} &  
     $86.92_{ \pm 0.8 }$ &
     $88.03_{\pm 0.78 }$  &
     $40.62_{\pm 0.99}$ & 
     $68.01_{ \pm 1.29}$ &
     
      $10.75_{ \pm 1.04 }$ &
     $10.79_{\pm 1.02 }$  &
     $60.22_{\pm 1.01}$ & 
     $59.0_{ \pm 0.6}$ 
     \\
    \gls{MMVAE}  &  $87.22_{ \pm 1.87 }$ &
     $77.35_{\pm 4.19 }$  &
     $67.31_{\pm 6.93}$ & 
     $39.44_{ \pm 3.43}$ &
       $12.15_{ \pm 1.25 }$ &
     $20.24_{\pm 1.04}$  &
     $58.1_{\pm 3.14}$ & 
     $171.42_{ \pm 4.55}$

    \\

    \gls{MOPOE} & 
    $89.95_{ \pm 0.84 }$ &
     $91.71_{\pm 0.77 }$  &
     $67.26_{\pm 0.8}$ & 
     $\underline{83.58}_{ \pm 0.44}$ &
     
    $9.39_{ \pm 0.76 }$ &
     $10.1_{\pm 0.73 }$  &
     $53.19_{\pm1.06}$ & 
     $57.34_{ \pm 1.35}$ 
        \\
   \gls{NEXUS} &     $92.63_{ \pm 0.45 }$ &
     $93.59_{\pm 0.4 }$  &
     $\underline{68.31}_{\pm 0.46}$ & 
     $83.13_{ \pm 0.58}$ &
     
     $4.92_{ \pm 0.61 }$ &
     $5.16_{\pm 0.59 }$  &
     $85.67_{\pm 2.74}$ & 
     $97.86_{ \pm 2.86}$

   \\
    \gls{MVTCAE} &     $\underline{94.33}_{ \pm 0.18 }$ &
     $\underline{95.18}_{\pm 0.19 }$  &
     $47.47_{\pm 0.76}$ & 
     $\textbf{86.6}_{ \pm 0.23}$ &
     
     $\underline{4.67}_{ \pm 0.35 }$ &
     $\underline{4.94}_{\pm 0.37 }$  &
     $\underline{52.29}_{\pm 1.17}$ & 
     $\underline{53.55}_{ \pm 1.19}$ 
      \\

    \midrule
    \gls{MLD}  &  $\textbf{96.73}_{ \pm 0.0 }$ &
     $\textbf{96.73}{\pm 0.0 }$  &
     $\textbf{82.19}_{\pm 0.0}$ & 
     $82.19_{ \pm 0.0}$ &
     $\textbf{2.25}_{ \pm 0.03 }$ &
     $\textbf{2.25}{\pm 0.03 }$  &
     $\textbf{48.47}_{\pm 0.63}$ & 
     $\textbf{48.47}_{ \pm 0.63}$ 
    \\
    \bottomrule
\end{tabular}}

\end{table}

\subsubsection{Detailed results}

\begin{table}[h]
\centering
\caption{Generative Coherence for \textbf{\mnist-\svhn}. We report the detailed version of \Cref{coh_quality:ms} with standard deviation for 5 independent runs with different seeds.\newline}
 \label{coh_quality:ms_detailed}
\resizebox{\textwidth}{!}{\begin{tabular}{c|c|cc|cc|cc}
\toprule
\multirow{2}{*}{ Models }  & \multicolumn{3}{c}{Coherence (\%$\uparrow$) } &   \multicolumn{4}{c}{Quality ($\downarrow$)} \\
    \cmidrule{2-8}
    & Joint &  M $\rightarrow$ S & S $\rightarrow$ M &
     Joint(M)  &  Joint(S) &M $\rightarrow$ S &  S $\rightarrow$ M  \\
    \midrule
     \gls{MVAE} &  
     $38.19_{ \pm 2.27 }$ &
     $48.21_{\pm 2.56 }$  &
     $ 28.57_{\pm 1.46}$ & 
     $13.34_{ \pm 0.93}$ &
     $68.0_{\pm 0.99}$   & 
     $ 68.9_{\pm 1.84}  $ &
     $ 13.66_{\pm 0.95} $ 
    
     \\
    \gls{MMVAE} &  $37.82_{ \pm 1.19 }$ & $11.72_{\pm 0.33 }$ & $ 67.55_{\pm 9.22}$ & $25.89_{ \pm 0.46}$ &
    $ 146.82_{\pm 4.76}  $& $393.33_{\pm 4.86}$  &
    $ 53.37_{\pm 1.87} $ 
 
    \\
    \gls{MOPOE} & $ 39.93_{ \pm 1.54 }$ & $12.27_{\pm 0.68 }$ & $68.82_{\pm 0.39}$ &$20.11_{ \pm 0.96}$ &
    $ 129.2_{\pm 6.33}  $&$373.73_{\pm26.42}$&
    $ 43.34_{\pm 1.72} $
        \\

    \gls{NEXUS} &  $ 40.0_{ \pm 2.74 }$ & $16.68_{\pm 5.93 }$ & $70.67_{\pm 0.77}$ &  $13.84_{ \pm 1.41}$ &
        $ 98.13_{\pm 5.9}  $ &$281.28_{\pm16.07}$&
 $ 53.41_{\pm 1.54} $
   \\

    \gls{MVTCAE} & $48.78_{ \pm 1 } $& $\underline{81.97}_{\pm 0.32 }$ & $49.78_{\pm 0.88} $&$12.98_{ \pm 0.68}$ &
    $ \textbf{52.92}_{\pm 1.39}  $&$69.48_{\pm1.64}$ &
    $ 13.55_{\pm 0.8} $ 

     \\ 
\gls{MMVAEplus} & $17.64_{ \pm 4.12 }$ &$13.23_{ \pm 4.96 }$ & $29.69_{ \pm 5.08 }$ 
&$26.60_{ \pm 2.58 }$ & $121.77_{ \pm 37.77 }$ 
& $240.90_{ \pm 85.74 }$
&$35.11_{ \pm 4.25 }$
\\
  \gls{MMVAEplus}(K=10) & $41.59_{ \pm 4.89 }$ & $55.3_{ \pm 9.89 }$
& $56.41_{ \pm 5.37 }$ & $19.05_{ \pm 1.10 }$ & $67.13_{ \pm 4.58 }$ & $75.9_{ \pm 12.91 }$ & $18.16_{ \pm 2.20 }$
\\
    \midrule
    \gls{MLD}  & $\underline{85.22}_{ \pm 0.5}$ & $\textbf{83.79}_{\pm 0.62} $&$ \textbf{79.13}_{\pm 0.38}$ & $\underline{3.93}_{ \pm 0.12}$ &
    $\underline{56.36}_{\pm1.63}$   &
    $\textbf{	57.2}_{\pm1.47}$  &
    $\underline{3.67}_{\pm 0.14} $ 
    \\
    
    \bottomrule
\end{tabular}}
\end{table}

\begin{figure}[H]
     \centering
       \centering
     \begin{subfigure}{0.40\textwidth}
         \centering
         \includegraphics[width=\linewidth]{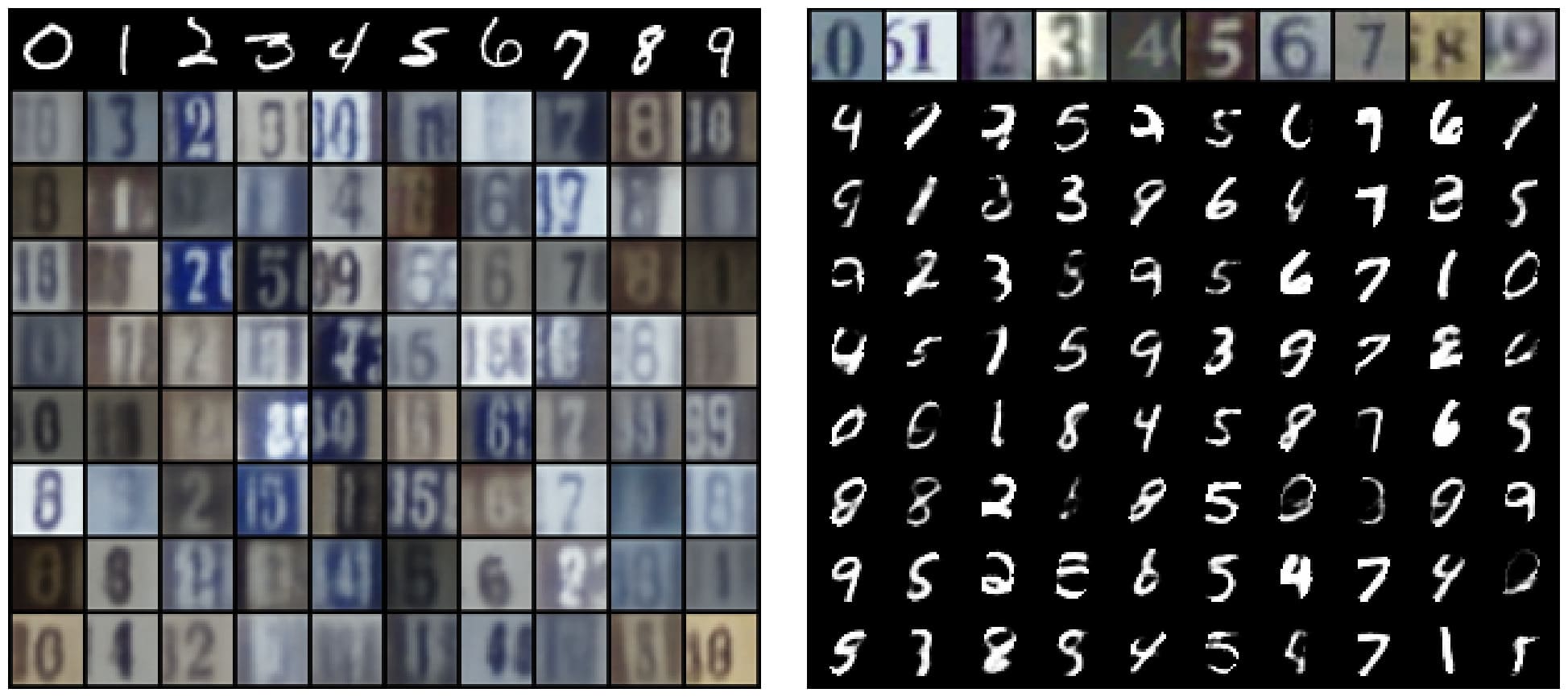}
         \caption*{\gls{MVAE}}
    
     \end{subfigure}
     \begin{subfigure}{0.40\textwidth}
         \centering
         \includegraphics[width=\linewidth]{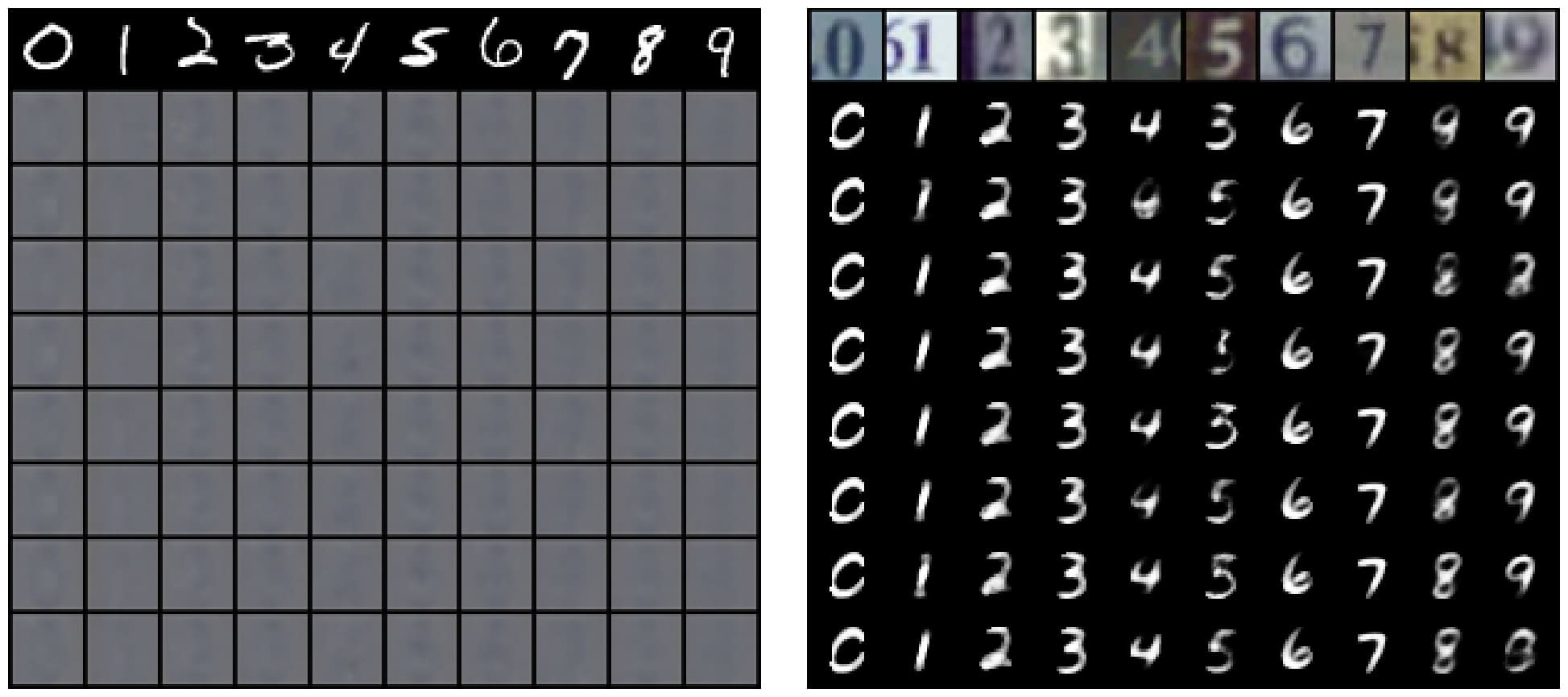}
         \caption*{\gls{MMVAE}}
      
     \end{subfigure}
     
       \begin{subfigure}{0.40\textwidth}
         \centering
         \includegraphics[width=\linewidth]{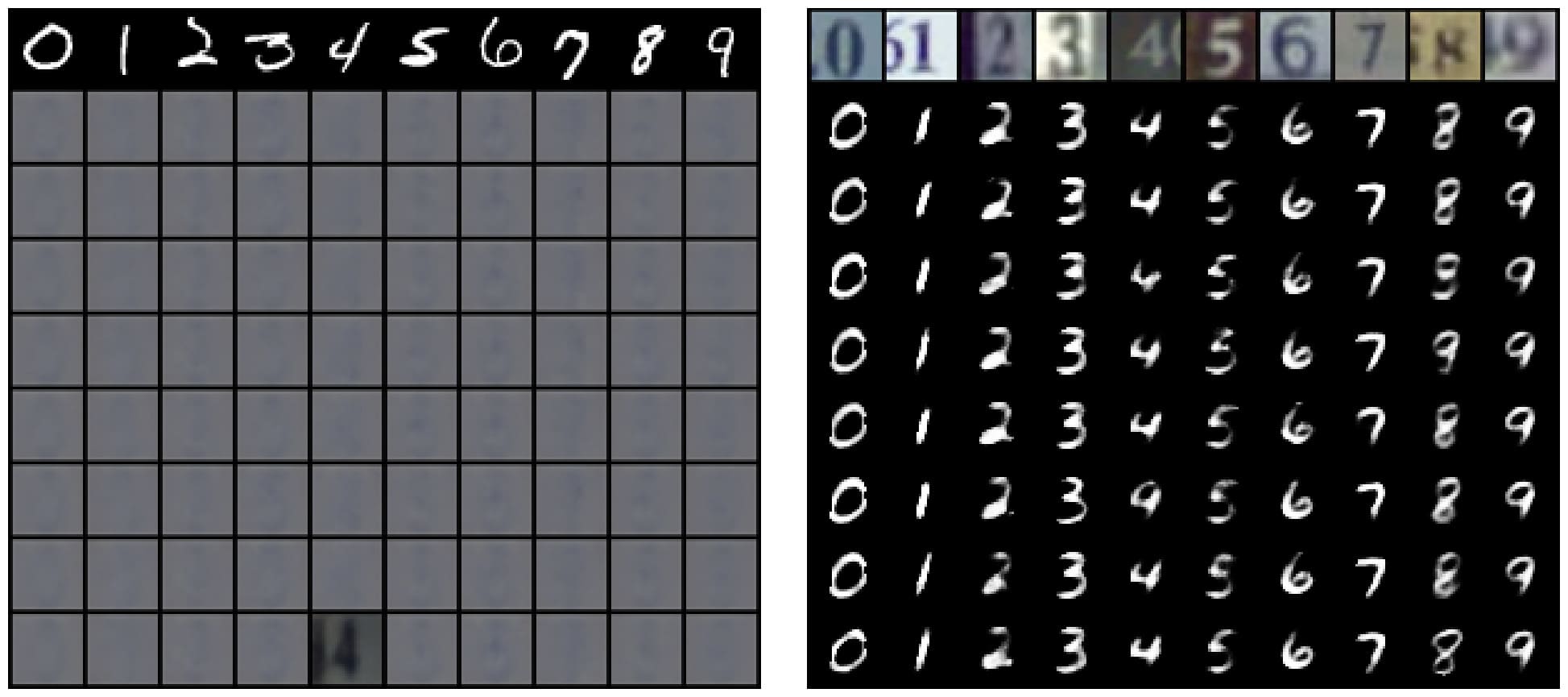}
         \caption*{\gls{MOPOE}}
     
     \end{subfigure}
      \begin{subfigure}{0.40\textwidth}
         \centering
         \includegraphics[width=\linewidth]{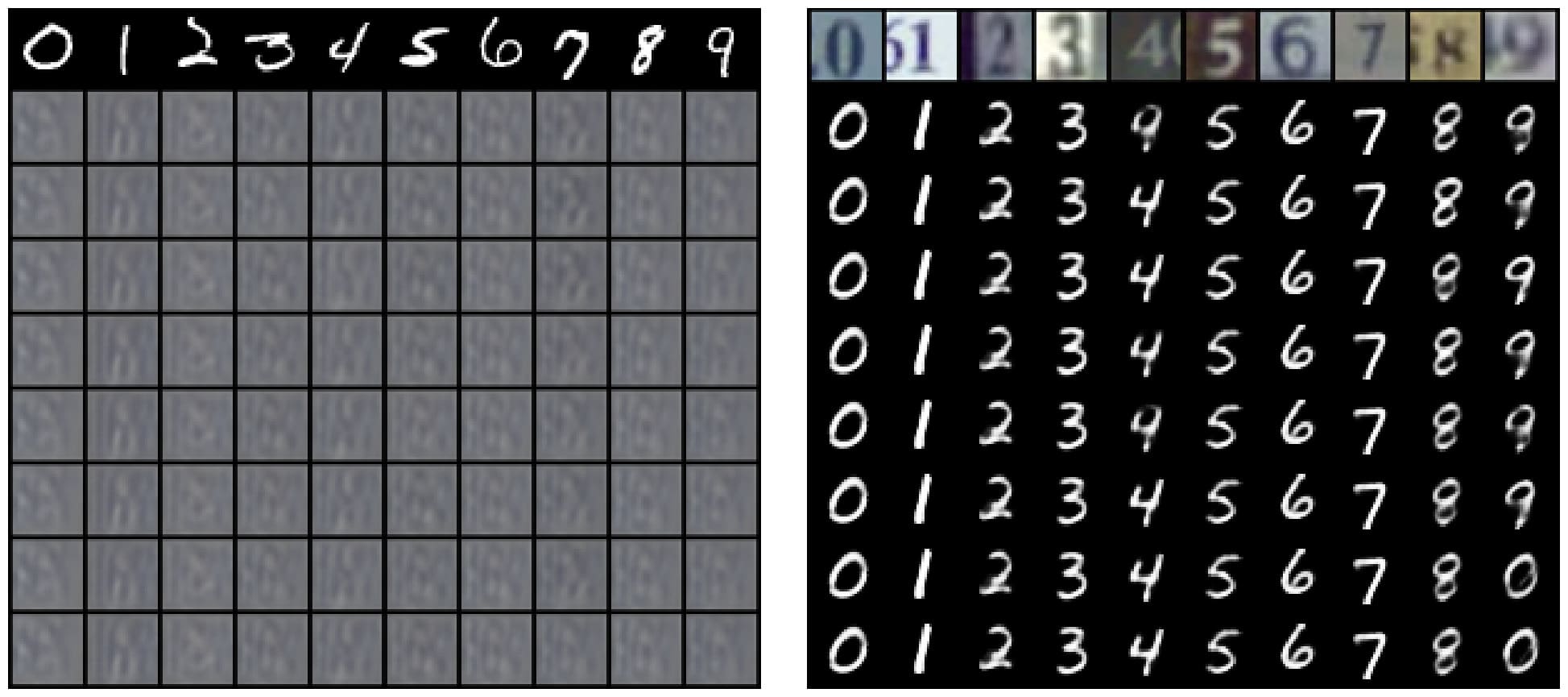}
         \caption*{\gls{NEXUS}}
   
     \end{subfigure}
     
  \begin{subfigure}{0.40\textwidth}
         \centering
         \includegraphics[width=\linewidth]{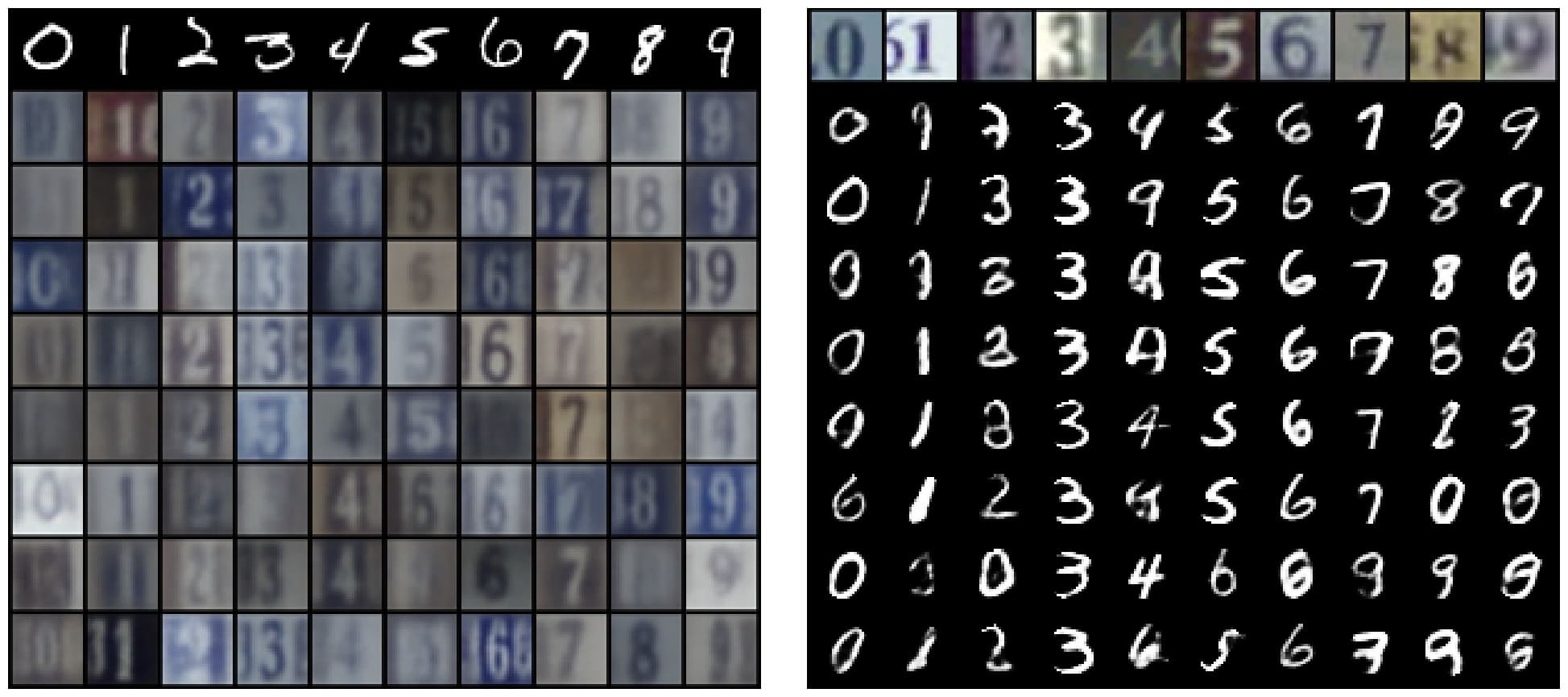}
         \caption*{\gls{MVTCAE}}
      
     \end{subfigure}
\begin{subfigure}{0.40\textwidth}
         \centering
         \includegraphics[width=\linewidth]{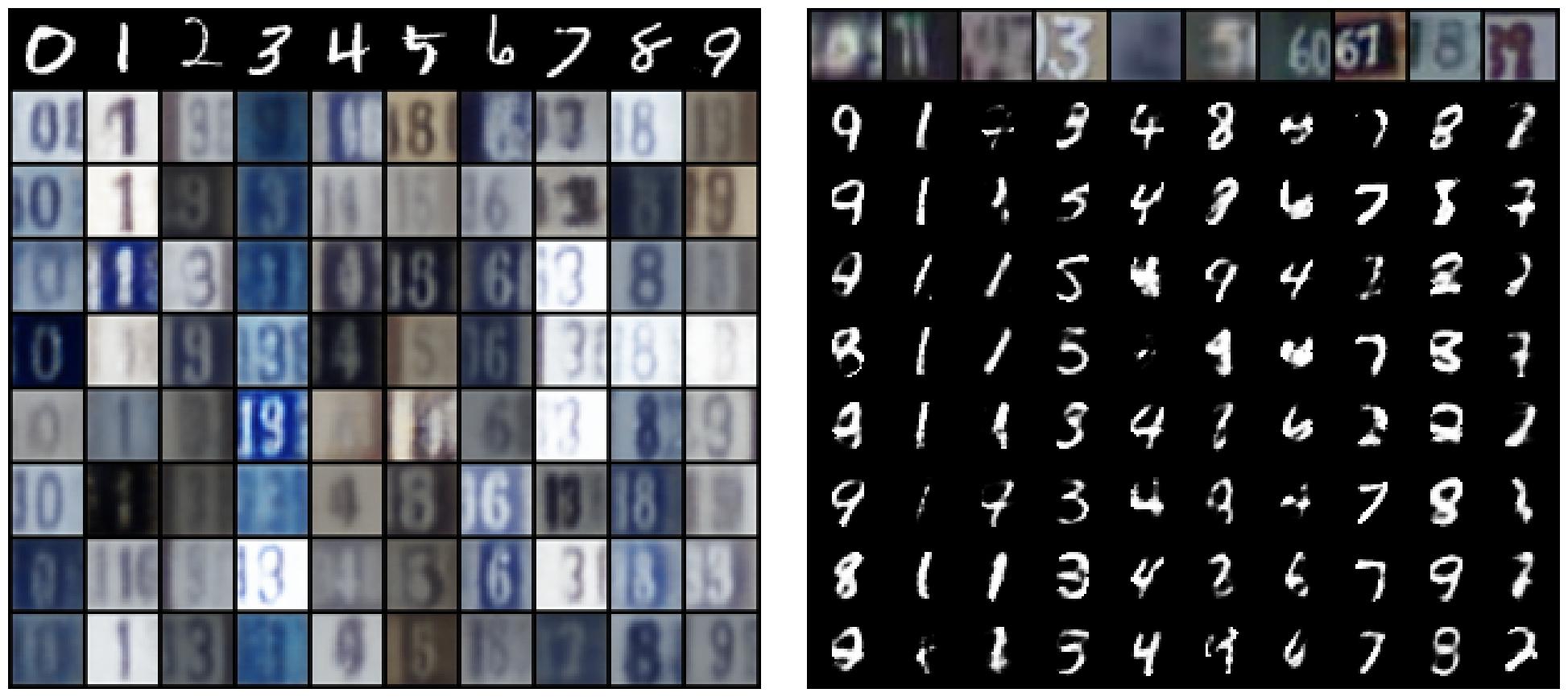}
         \caption*{\gls{MMVAEplus}(K=10)}
      
     \end{subfigure}
     
  \begin{subfigure}{0.40\textwidth}
         \centering
         \includegraphics[page=1,width=\linewidth]{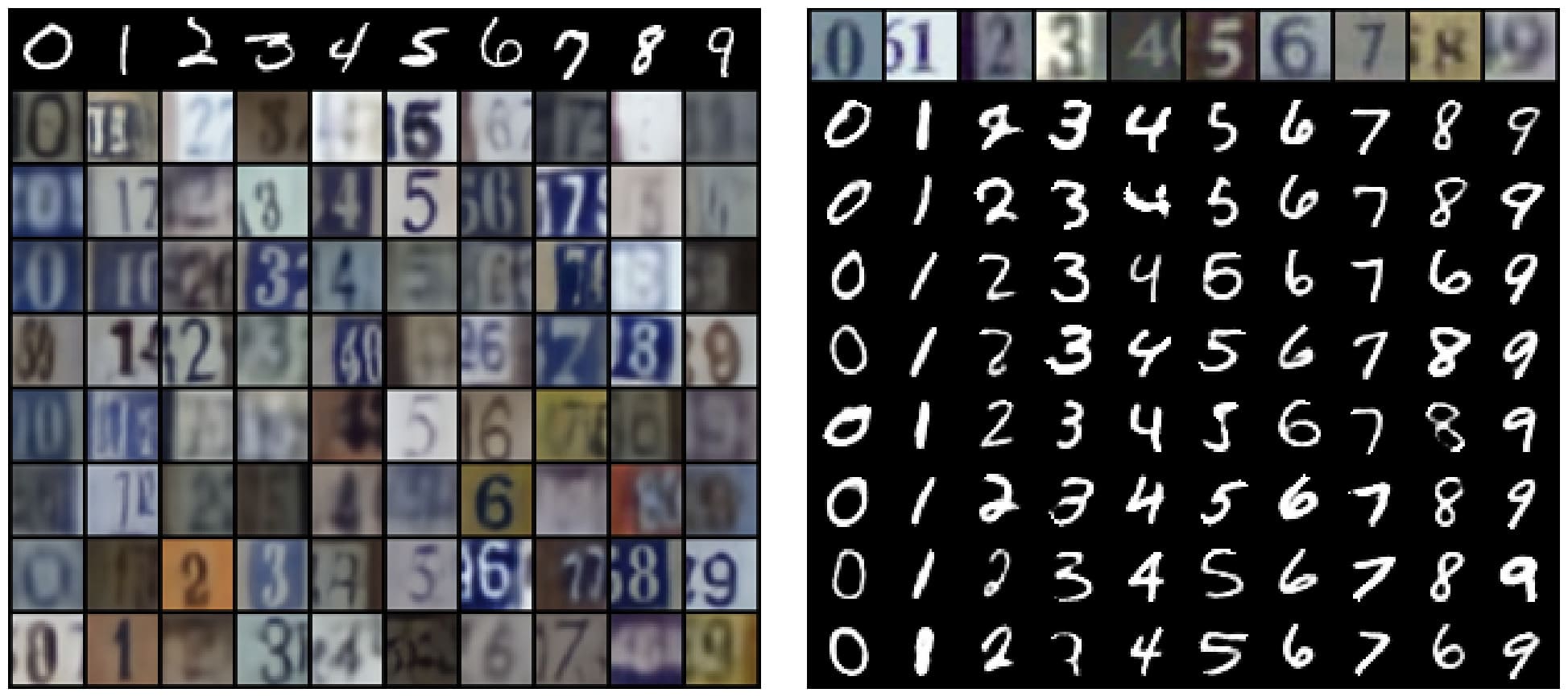}
         \caption*{\textbf{\gls{MLD} (ours) }  }
     
     \end{subfigure}
     
        \caption{Additional qualitative results for \textbf{\mnist-\svhn}. For each model we report: \mnist to \svhn conditional generation in the left,   \svhn to \mnist conditional generation in the right. }
        \label{fig:cond_ms_detailed}
\end{figure}

\newpage

\begin{figure}[h]
     \centering
     \begin{subfigure}{0.40\textwidth}
         \centering
         \includegraphics[width=\linewidth]{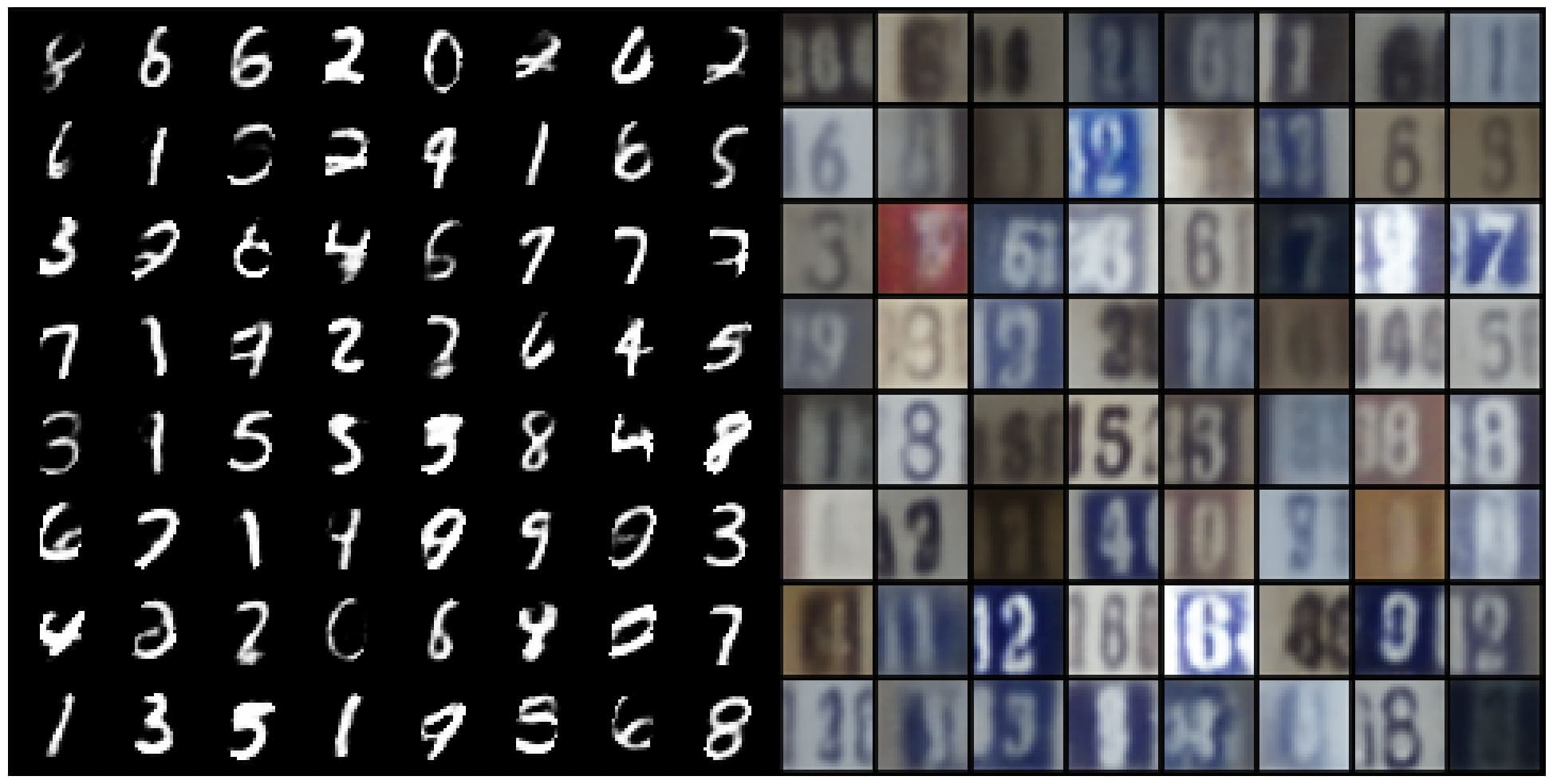}
  
     \caption*{\gls{MVAE}}
     \end{subfigure}
     \begin{subfigure}{0.40\textwidth}
         \centering
         \includegraphics[width=\linewidth]{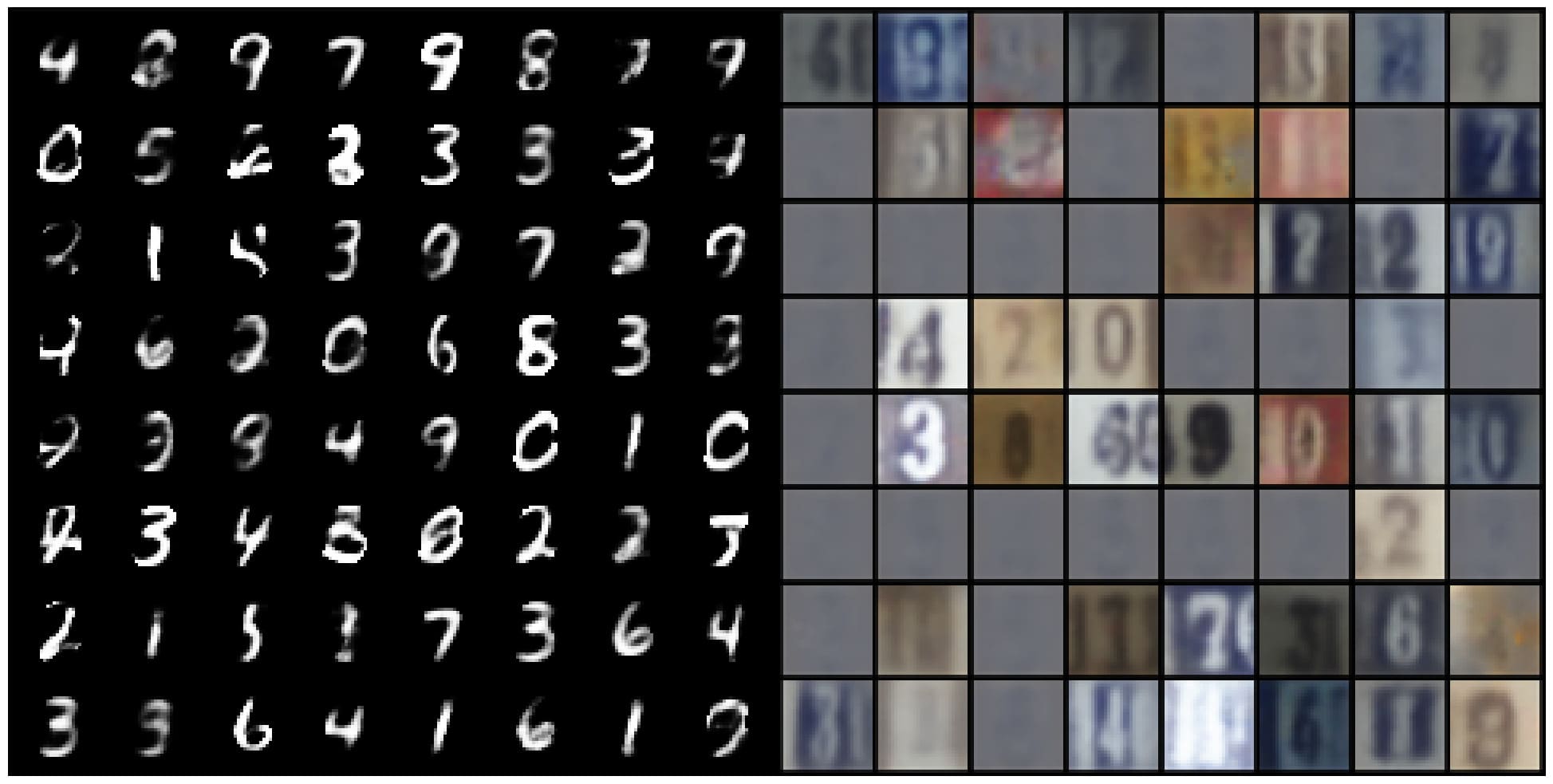}
        \caption*{\gls{MMVAE}}
     \end{subfigure}
     
       \begin{subfigure}{0.40\textwidth}
         \centering
         \includegraphics[width=\linewidth]{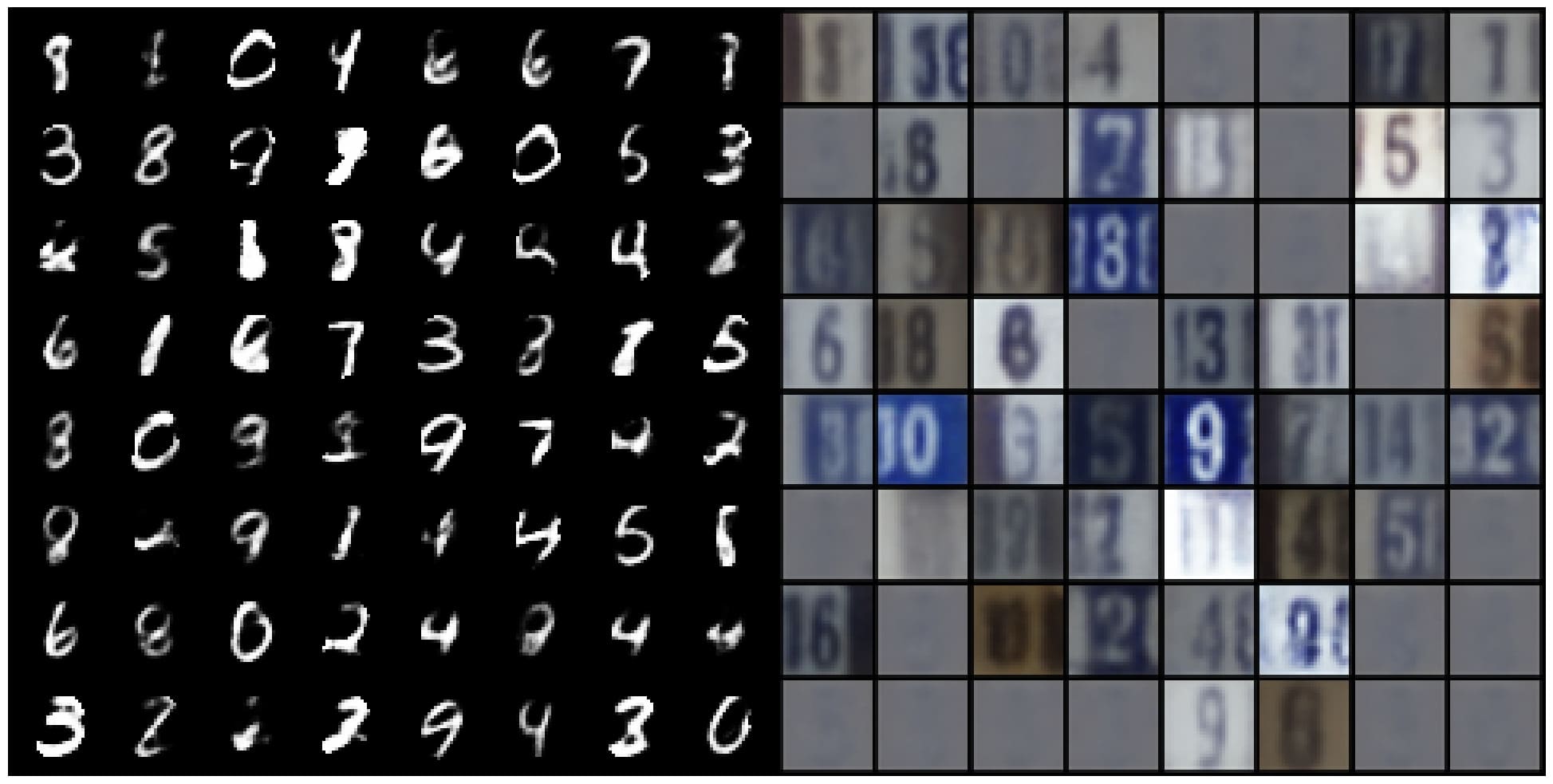}
  
            \caption*{\gls{MOPOE}}
     \end{subfigure}
      \begin{subfigure}{0.40\textwidth}
         \centering
         \includegraphics[width=\linewidth]{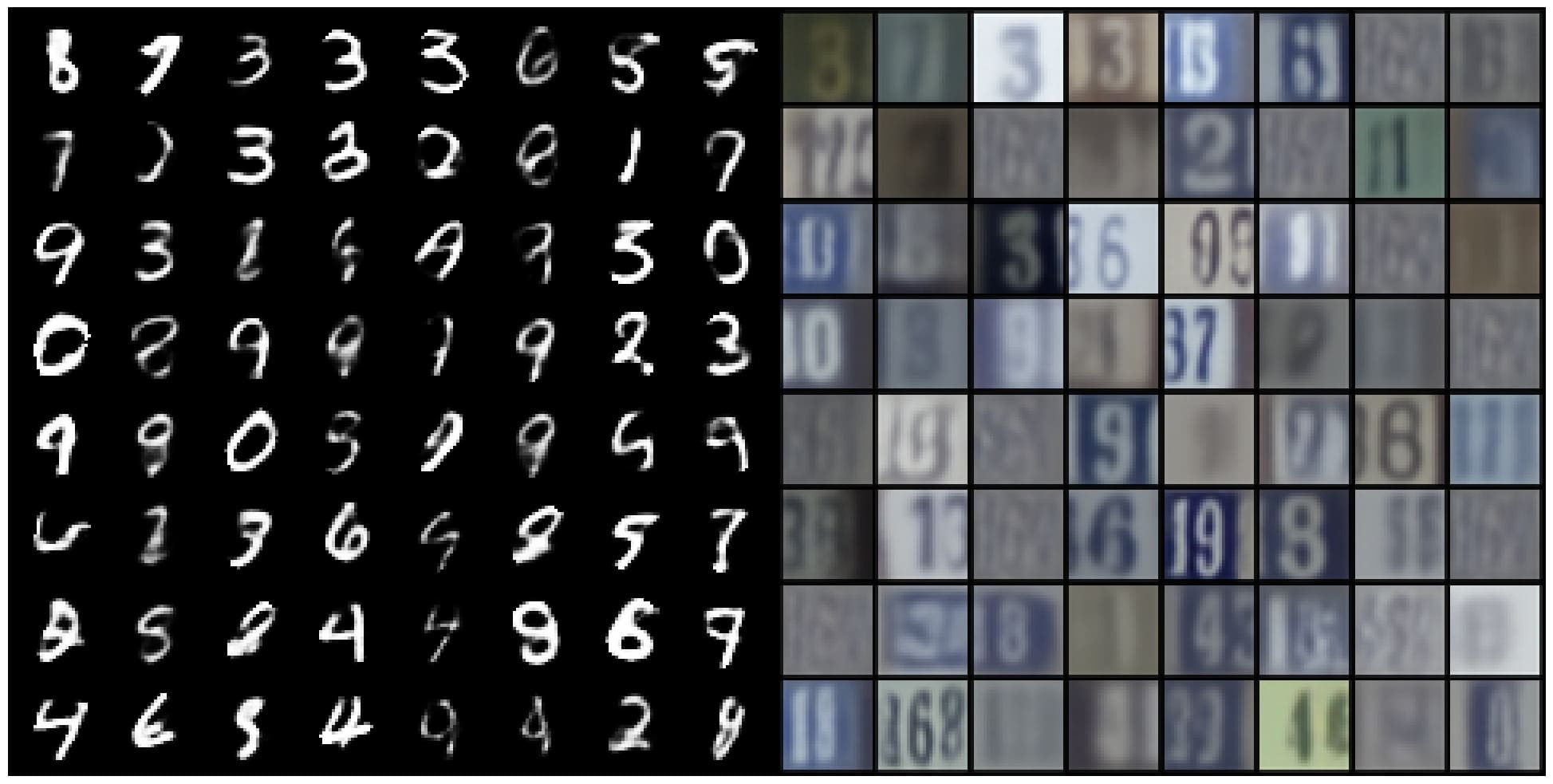}
         \caption*{\gls{NEXUS}}
     \end{subfigure}
     
  \begin{subfigure}{0.40\textwidth}
         \centering
         \includegraphics[width=\linewidth]{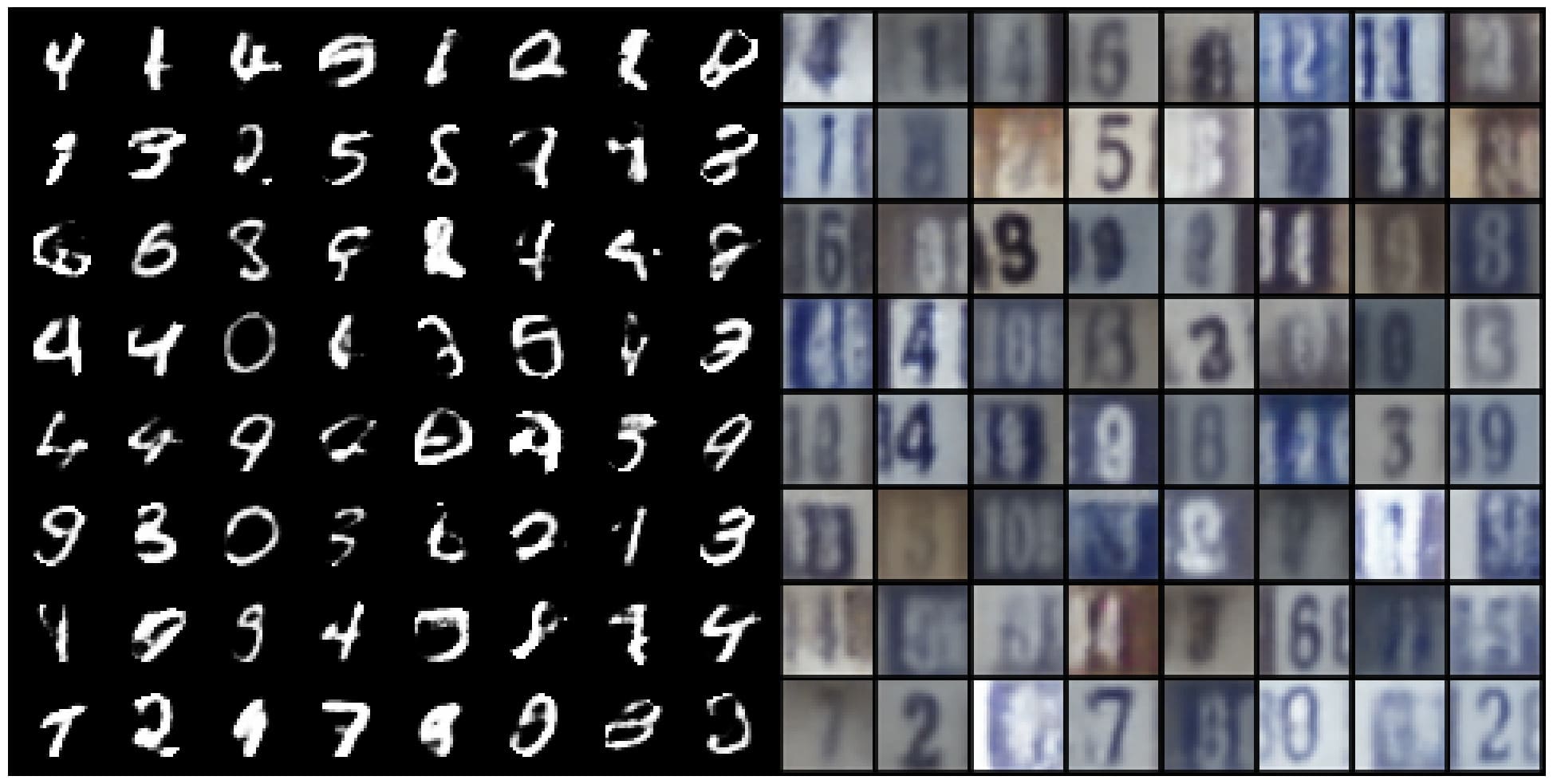}
         \caption*{\gls{MVTCAE}}
        
     \end{subfigure}
      \begin{subfigure}{0.40\textwidth}
         \centering
         \includegraphics[page=1,width=\linewidth]{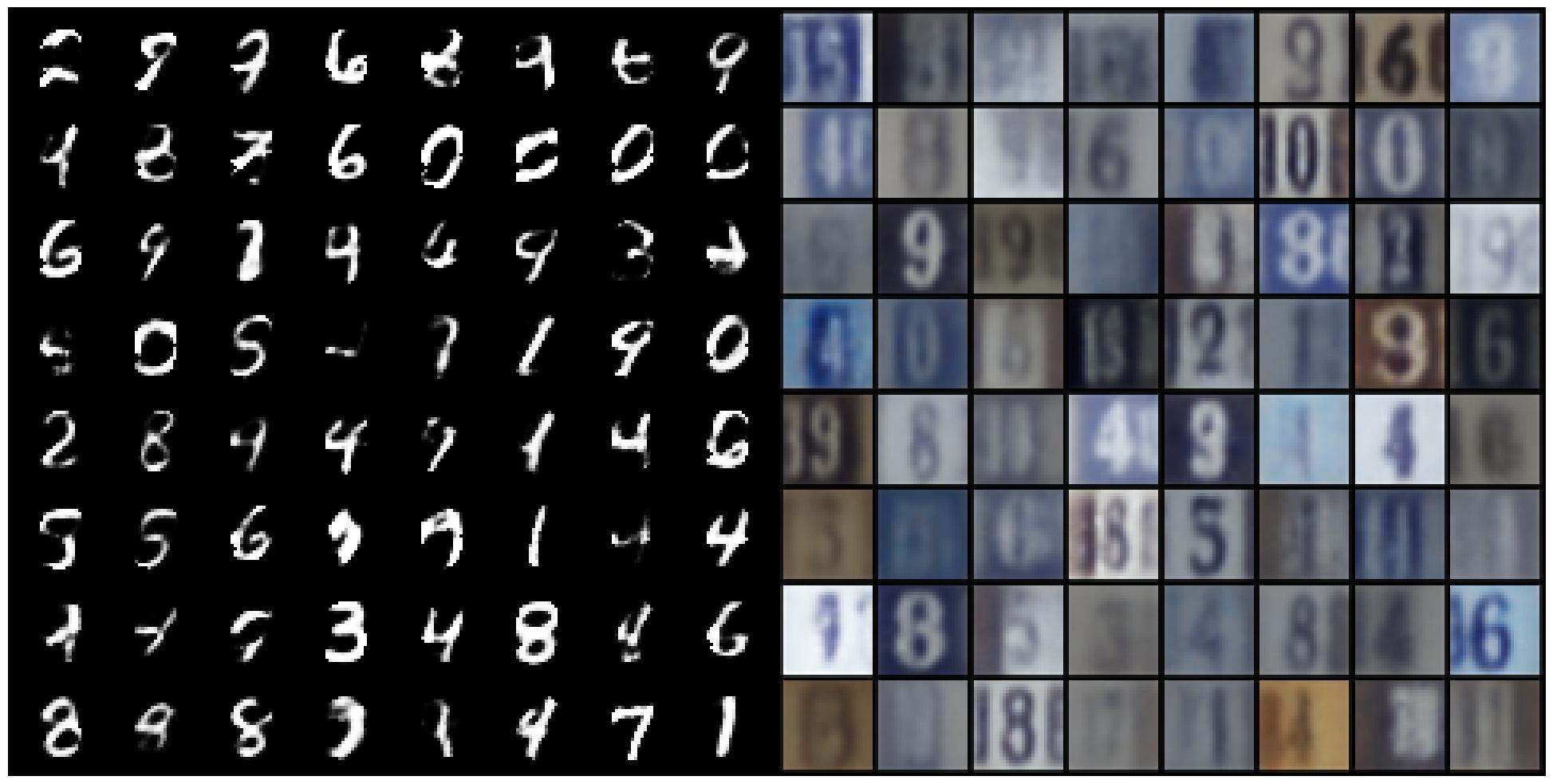}
         \caption*{\gls{MMVAEplus}(K=10)  }
         
     \end{subfigure}
     
  \begin{subfigure}{0.40\textwidth}
         \centering
         \includegraphics[page=1,width=\linewidth]{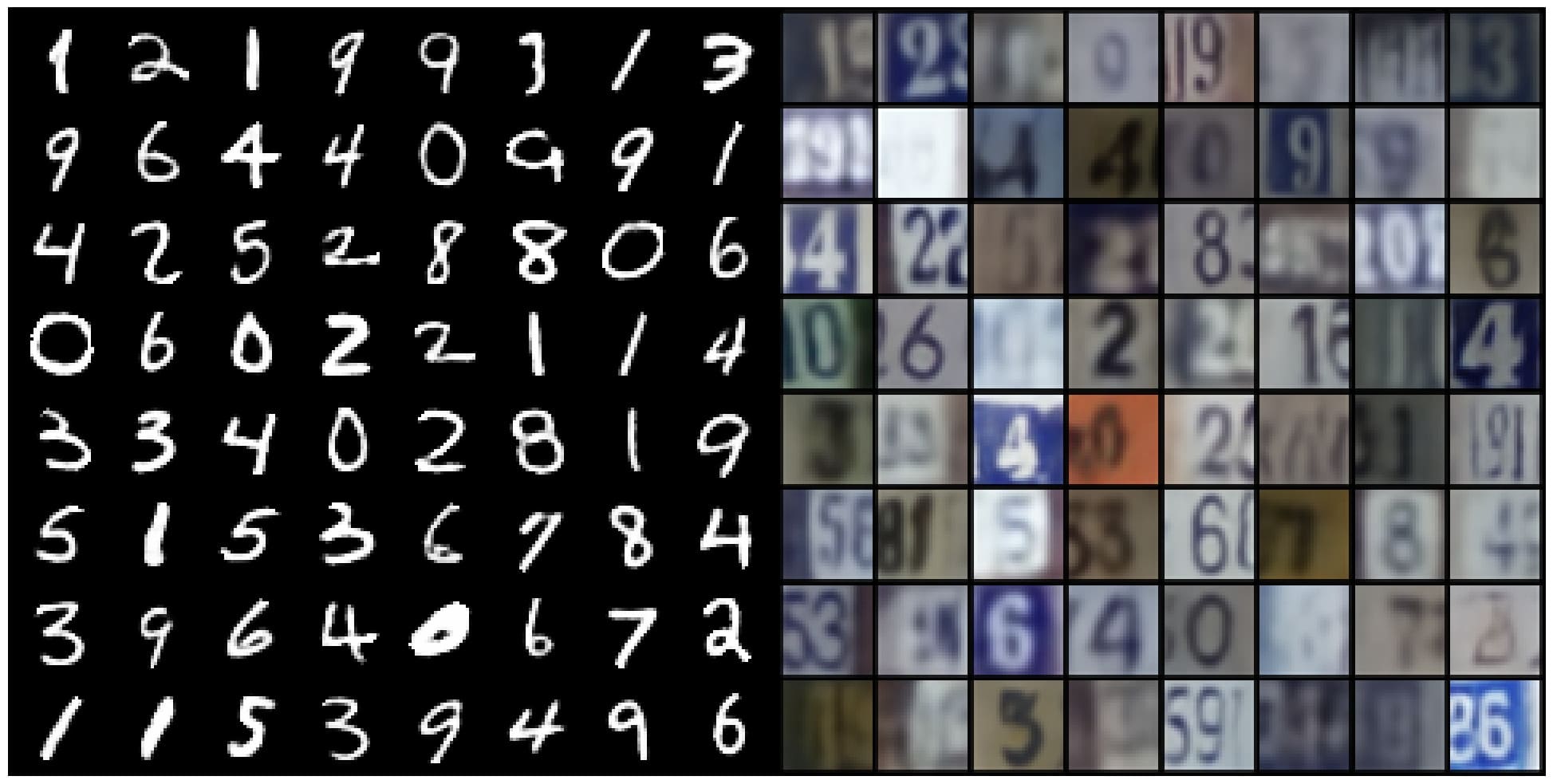}
         \caption*{\textbf{\gls{MLD} (ours) }  }
         
     \end{subfigure}
        \caption{Qualitative results for \textbf{\mnist-\svhn} joint generation.}
        \label{qualitative:ms_joint_detailed}
\end{figure}

\subsection{\mhd}

\begin{table}[H]
\centering
\caption{ Generative Coherence for \textbf{\mhd}. We report the detailed version of \Cref{coh:mhd} with standard deviation for 5 independent runs with different seeds.\newline}
\resizebox{\textwidth}{!}{\begin{tabular}{c|c|ccc|ccc|ccc}
\toprule
\multirow{2}{*}{ Models }  & \multirow{2}{*}{ Joint }
& \multicolumn{3}{c}{I (Image)}  & \multicolumn{3}{c}{T (Trajectory)}  & \multicolumn{3}{c}{S (Sound) } 
\\
\cmidrule{3-11}
    &&  T & S  & T,S  & I & S  & I,S & I & T & I,T \\
    \midrule
 \gls{MVAE}   &$37.77_{ \pm 3.32 }$ &
    
    $11.68_{\pm 0.35 }$ &
    $26.46 _{\pm 1.84 }$ & 
    $ 28.4_{\pm 1.47}$ & 
    
    $95.55_{\pm1.39}$  &
    $26.66 _{\pm 1.72 }$ &
     $96.58_{\pm1.06}$ &  
    
    $58.87_{\pm4.89}$ & 
    $10.39 _{\pm  0.42 }$  & 
    $58.16_{\pm5.24}$    \\

  \gls{MMVAE}   &    
    $34.78_{ \pm 0.83 }$ &
    
    $\textbf{99.7}_{\pm 0.03 }$ &
    $69.69 _{\pm 1.66}$ & 
    $ 84.74_{\pm 0.95}$ & 
    $\underline{99.3}_{ \pm 0.07 }$  &
    $85.46 _{\pm 1.57 }$  & 
    $92.39_{\pm0.95}$ &  
    
    $49.95_{\pm0.79}$ & 
    $50.14_{\pm  0.89 }$  & 
    $50.17_{\pm0.99}$   \\

    \gls{MOPOE}   &    $48.84_{ \pm 0.36 }$ &
    $\underline{99.64}_{\pm 0.08 }$ & $68.67 _{\pm 2.07 }$ & $ \underline{99.69}_{\pm 0.04}$ & 
    $99.28_{ \pm 0.08  }$  & $\underline{87.42}_{\pm  0.41 }$  & $99.35_{\pm0.04}$ &  
    $50.73_{\pm3.72}$ & $51.5 _{\pm  3.52 }$  & $56.97_{\pm6.34}$   \\

     \gls{NEXUS} &  $26.56_{ \pm 1.71 }$ &
    $94.58_{\pm 0.34 }$ & $\underline{83.1}_{\pm 0.74 }$ & $ 95.27_{\pm 0.52}$ & 
    $88.51_{ \pm 0.64  }$  & $76.82 _{\pm  3.63 }$  & $93.27_{\pm0.91}$ &  
    $70.06_{\pm2.83}$ & $75.84 _{\pm  2.53 }$  & $89.48_{\pm3.24}$ 
    \\

      \gls{MVTCAE} &  $42.28_{ \pm 1.12 }$ 
     &
    $99.54_{\pm 0.07 }$ & $72.05 _{\pm 0.95 }$ & $ 99.63_{\pm 0.05}$ & 
    $99.22_{ \pm 0.08  }$  & $72.03 _{\pm  0.48 }$  & $\underline{99.39}_{\pm0.02}$ &  
    $\underline{92.58}_{\pm0.47}$ & $\underline{93.07}_{\pm  0.36 }$  & $\underline{94.78}_{\pm0.25}$  \\

      \gls{MMVAEplus} &  $41.67_{ \pm 2.3 }$ 
     &
    $98.05_{\pm 0.19 }$ & $84.16_{\pm 	0.57 }$ & $ 91.88_{\pm }$ & 
    $97.47_{ \pm 0.89  }$  & $81.16_{\pm 2.24  }$  & $89.31_{\pm 1.54}$ &  
    $64.34_{\pm	4.46}$ & 
    $65.42_{\pm  5.42 }$  &
    $64.88_{\pm 4.93}$  \\

   \gls{MMVAEplus}(k=10) &  $42.60_{ \pm 2.5 }$ 
     &
    $99.44_{\pm 0.07 }$ & $\textbf{89.75} _{\pm 	0.75 }$ & $ 94.7_{\pm 0.72}$ & 
    $99.44_{ \pm 0.18  }$  
    & $\textbf{89.58}_{\pm 0.4  }$  & $95.01_{\pm0.30}$ &  
    $87.15_{\pm	2.81}$ & 
    $87.99_{\pm  2.55 }$  &
    $87.57_{\pm 2.09 }$  \\

    \midrule
 \gls{MLD} &  
 $\textbf{98.34}_{ \pm 0.22 }$
 &
    $99.45_{\pm 0.09 }$ &
    $\underline{88.91}_{\pm 0.54 }$ & 
    $ \textbf{99.88}_{\pm 0.04}$ & 
    
    $\textbf{99.58}_{ \pm 0.03  }$  & 
    $\underline{88.92} _{\pm  0.53 }$  & 
    $\textbf{99.91}_{\pm0.02}$ &  
    
    $\textbf{97.63}_{\pm0.14}$ & 
    $\textbf{97.7} _{\pm  0.34 }$ 
    & $\textbf{98.01}_{\pm 0.21}$  \\
    \bottomrule
   
\end{tabular}}
 \label{coh:mhd_detailed}

\end{table}

\begin{table}[h]
\centering
\caption{ Generative quality for \textbf{\mhd}. We report the detailed version of \Cref{qua:mhd} with standard deviation for 5 independent runs with different seeds.\newline}
\resizebox{\textwidth}{!}{\begin{tabular}{c|cccc|cccc|cccc}
\toprule
\multirow{2}{*}{ Models } & \multicolumn{4}{c}{I (Image)}  & \multicolumn{4}{c}{T (Trajectory)}  & \multicolumn{4}{c}{S (Sound) } 
\\
\cmidrule{2-13}
    & Joint & T & S  & T,S &  Joint &  I & S  & I,S & Joint  & I & T & I,T  \\
    \midrule
  \gls{MVAE}  & $\underline{94.9}_{ \pm7.37 }$   &  $93.73_{ \pm 5.44 }$ &
    $92.55_{ \pm 7.37 }$  &
    $91.08_{ \pm 10.24 }$  &

    $39.51_{ \pm 6.04 }$   &
    $20.42_{ \pm 4.42 }$ &
    $38.77_{ \pm 6.29 }$  &
    $19.25_{ \pm 4.26 }$  &
    
    $14.14_{ \pm 0.25 }$ &
  $\underline{14.13}_{ \pm 0.19 }$ &
    $14.08_{ \pm 0.24 }$  &
    $14.17_{ \pm 4.26 }$

    \\
 \gls{MMVAE} & $224.01_{ \pm12.58 }$   &  $22.6_{ \pm 4.3 }$ &
    $789.12_{ \pm 12.58 }$  &
    $170.41_{ \pm 8.06 }$  &
    
    $16.52_{ \pm1.17 }$   & 
    $\textbf{0.5}_{ \pm 0.05 }$ &
    $30.39_{ \pm 1.38 }$  &
    $6.07_{ \pm 0.37 }$  &

 $22.8_{ \pm0.39 }$    &
    $22.61_{ \pm 0.75 }$ &
    $23.72_{ \pm 0.86 }$  &
    $23.01_{ \pm 0.67 }$

\\
  \gls{MOPOE} &  $147.81_{ \pm10.37 }$   &  $16.29_{ \pm 0.85 }$ &
    $838.38_{ \pm 10.84 }$  &
    $15.89_{ \pm 1.96 }$  &
   
 $\underline{13.92}_{ \pm0.96 }$   & 
$\underline{0.52}_{ \pm 0.12 }$ &
    $33.38_{ \pm 1.14 }$  &
    $\textbf{0.53}_{ \pm 0.1 }$  &
   
       $18.53_{ \pm0.27 }$&
    $24.11_{ \pm0.4 }$ &
    $24.1_{ \pm 0.41 }$  &
    $23.93_{ \pm 0.87 }$  

    \\
   \gls{NEXUS} &  $281.76_{ \pm12.69 }$   &      $116.65_{ \pm 9.99 }$ &
    $282.34_{ \pm 12.69 }$  &
    $117.24_{ \pm 8.53 }$  &
    
      $18.59_{ \pm2.16 }$   &
    $6.67_{ \pm 0.23 }$ &
    $33.01_{ \pm 3.41 }$  &
    $7.54_{ \pm 0.29 }$  &
  
       $\underline{13.99}_{ \pm0.9}$   &
    $19.52_{ \pm 0.14 }$ &
    $18.71_{ \pm 0.24 }$  &
    $16.3_{ \pm 0.59 }$

    \\

   \gls{MVTCAE} &    $121.85_{ \pm 3.44 }$   &   $\underline{5.34}_{ \pm 0.33 }$ &
    $\underline{54.57}_{ \pm 7.79 }$  &
    $\underline{3.16}_{ \pm 0.26 }$  &

        $19.49_{ \pm 0.67}$   & 
    $0.62_{ \pm 0.1}$ &
    $\underline{13.65}_{ \pm 1.24 }$  &
    $0.75_{ \pm 0.13 }$  &

        $15.88_{ \pm 0.19}$   &
        $14.22_{ \pm 0.27}$ &
    $\underline{14.02}_{ \pm 0.14 }$  &
    $\underline{13.96}_{ \pm 0.28}$
    \\

   \gls{MMVAEplus} &    
   
   $97.19_{ \pm 12.37 }$   &  
   $2.80_{ \pm 0.42 }$ &
    $128.56_{ \pm 4.47 }$  &
    $ 114.3_{ \pm 11.4 }$  &

        $22.37_{ \pm 1.87}$   & 
    $1.21_{ \pm 0.22}$ &
    $21.74_{ \pm 3.49 }$  &
    $15.2_{ \pm 1.15 }$  &

        $16.12_{ \pm 0.40}$   &
        $17.31_{ \pm 0.62}$ &
    $17.92_{ \pm 0.19 }$  &
    $17.56_{ \pm 0.48}$
    
    \\

     \gls{MMVAEplus}(K=10) &    
   
   $85.98_{ \pm 1.25 }$   &  
   $1.83_{ \pm 0.26 }$ &
    $70.72_{ \pm 1.76 }$  &
    $62.43_{ \pm 3.4  }$  &

        $21.10_{ \pm 1.25}$   & 
    $1.38_{ \pm 0.34}$ &
    $8.52_{ \pm 0.79 }$  &
    $7.22_{ \pm 1.6 }$  &

   $14.58_{ \pm 0.47 }$   &  
   $14.33_{ \pm 0.51 }$ &
    $14.34_{ \pm 0.42 }$  &
    $14.32_{ \pm 0.6 }$  
    
    \\
    \midrule
      
   \textbf{\gls{MLD} (ours) }&     $\textbf{7.98}_{ \pm 1.41 }$ & $\textbf{1.7}_{ \pm 0.14 }$ &
    $\textbf{4.54}_{ \pm 0.45 }$ & 
    $\textbf{1.84}_{ \pm 0.27 }$  &
 
      $\textbf{3.18}_{ \pm 0.18 }$ 
      & $0.83_{ \pm 0.03 }$ 
     & $\textbf{2.07}_{ \pm 0.26 }$ 
       & $\underline{0.6}_{ \pm 0.05}$

 & $ \textbf{2.39}_{ \pm 0.1 } $
        & $\textbf{2.31}_{ \pm 0.07 }$ 
     & $\textbf{2.33}_{ \pm 0.11 }$ 
       & $\textbf{	2.29}_{ \pm 0.06 }$ 
      
       \\

    \bottomrule
   
\end{tabular}}

 \label{qua:mhd_detailed}

\end{table}

\begin{figure}[h]
     \centering
       \centering
     \begin{subfigure}{0.25\textwidth}
         \centering
         \includegraphics[width=\linewidth]{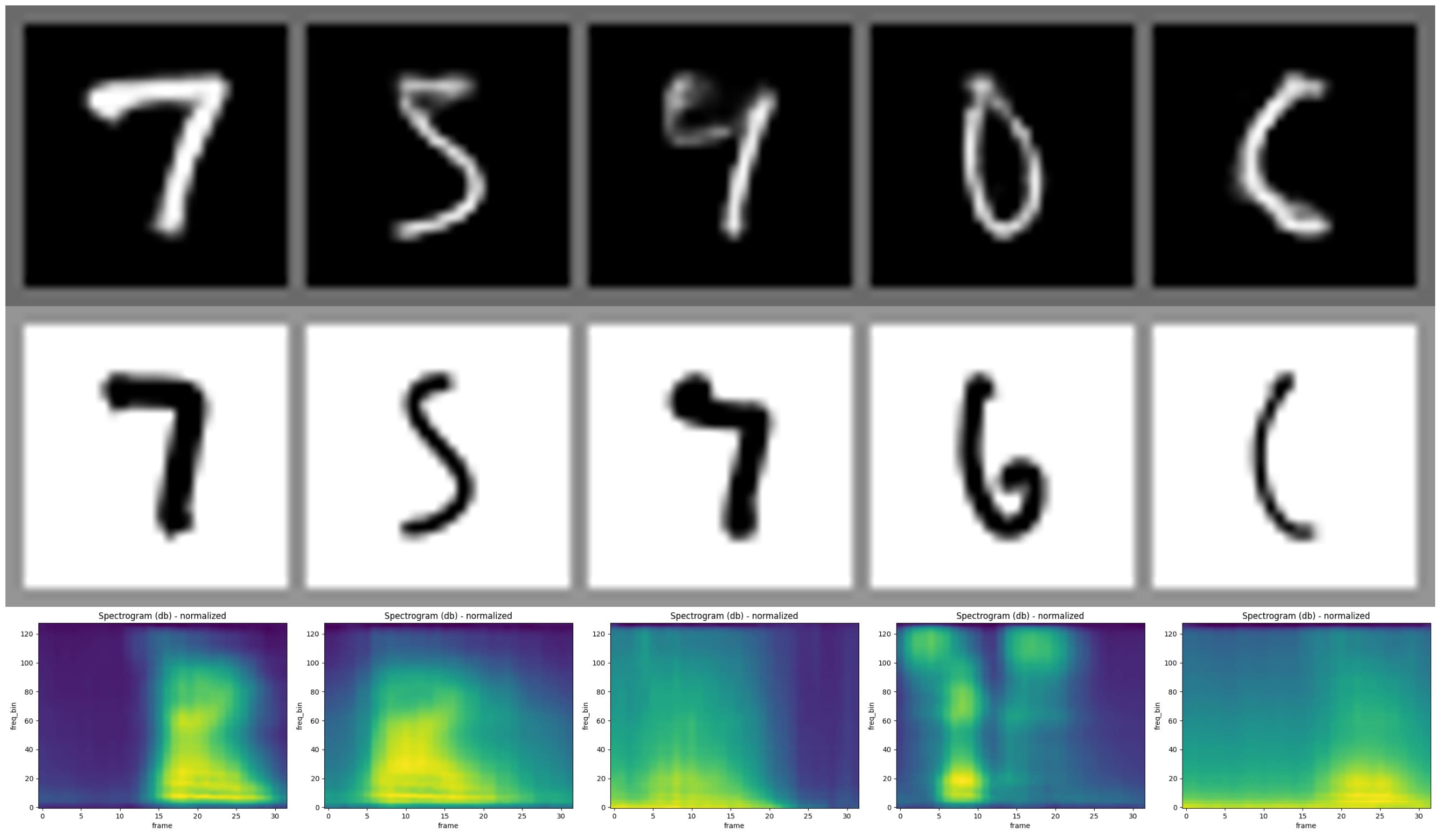}
  \caption*{\gls{MVAE}}
     \end{subfigure}
     \begin{subfigure}{0.25\textwidth}
         \centering
         \includegraphics[width=\linewidth]{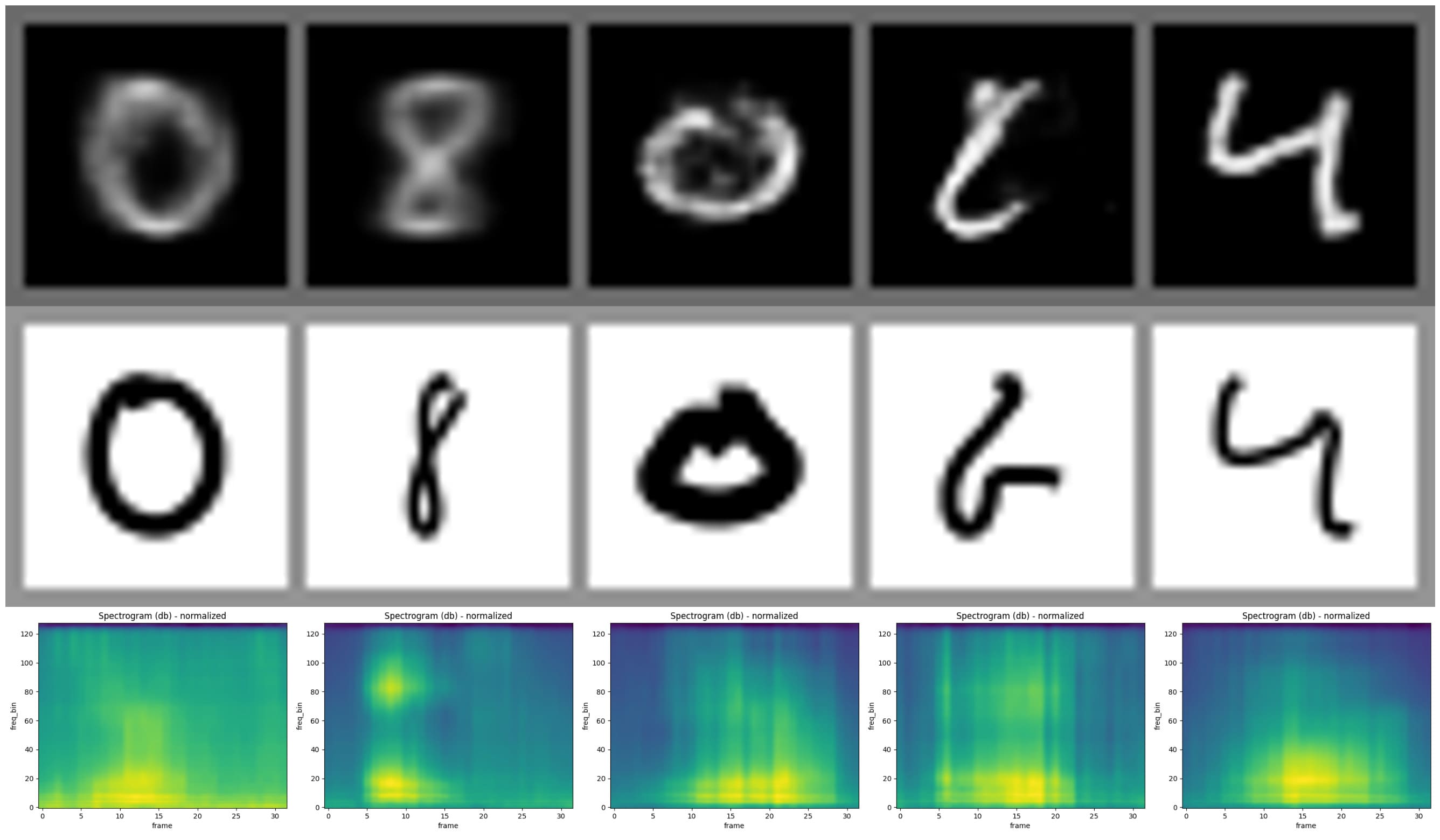}
   \caption*{\gls{MMVAE}}
     \end{subfigure}
       \begin{subfigure}{0.25\textwidth}
         \centering
         \includegraphics[width=\linewidth]{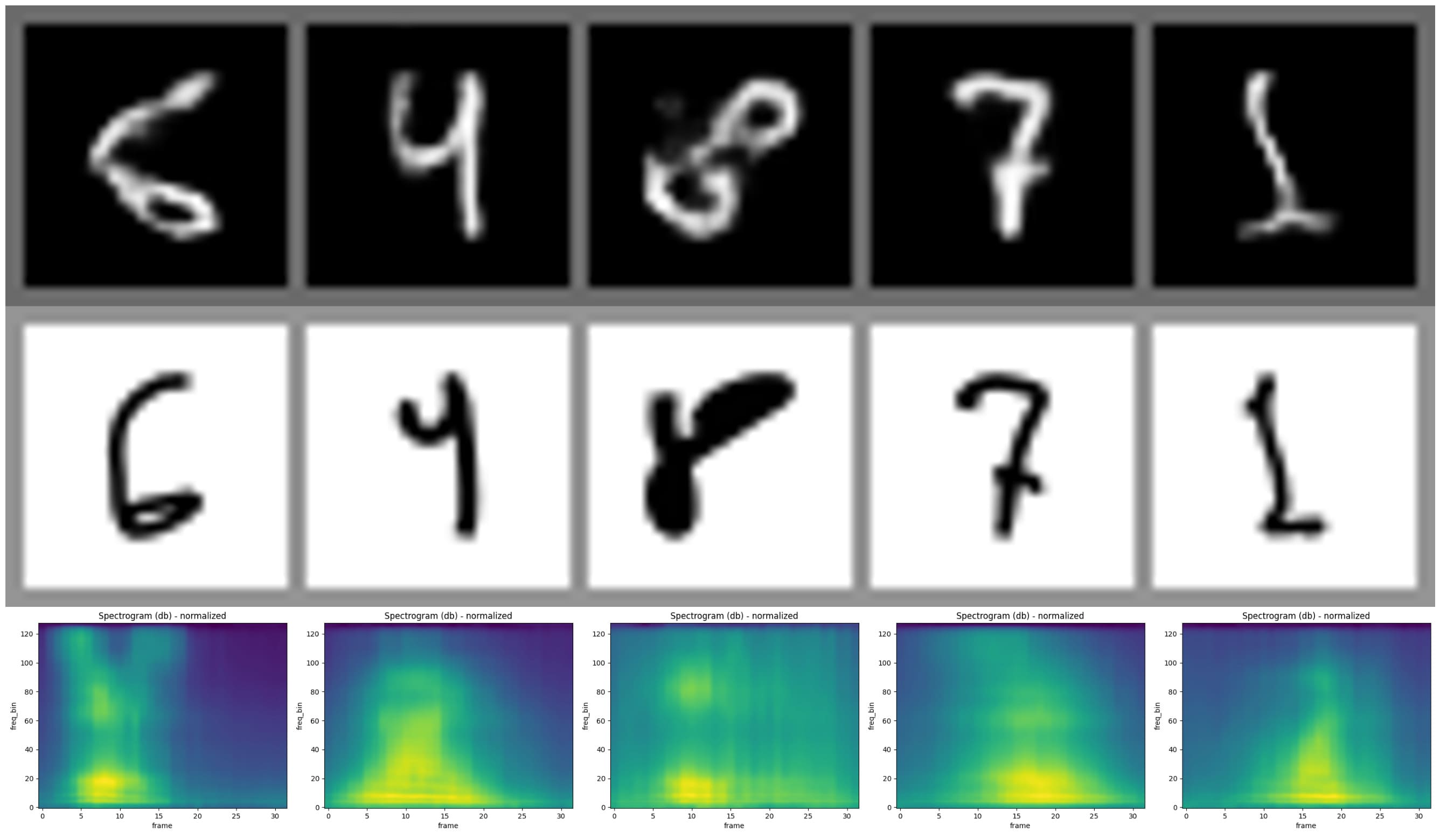}
         \caption*{\gls{MOPOE}}
     \end{subfigure}
     
      \begin{subfigure}{0.25\textwidth}
         \centering
         \includegraphics[width=\linewidth]{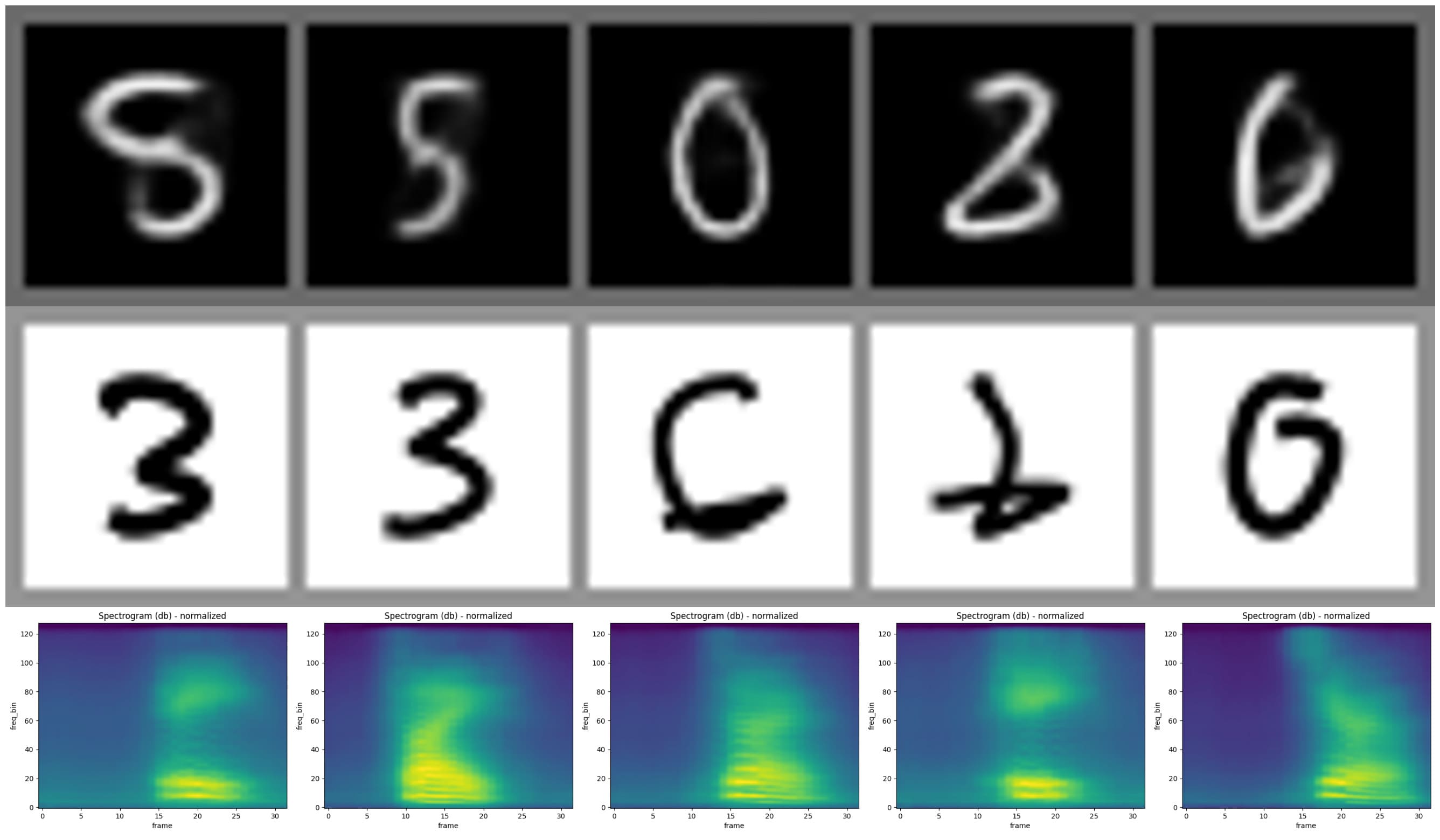}
         \caption*{\gls{NEXUS}}
     \end{subfigure}
  \begin{subfigure}{0.25\textwidth}
         \centering
         \includegraphics[width=\linewidth]{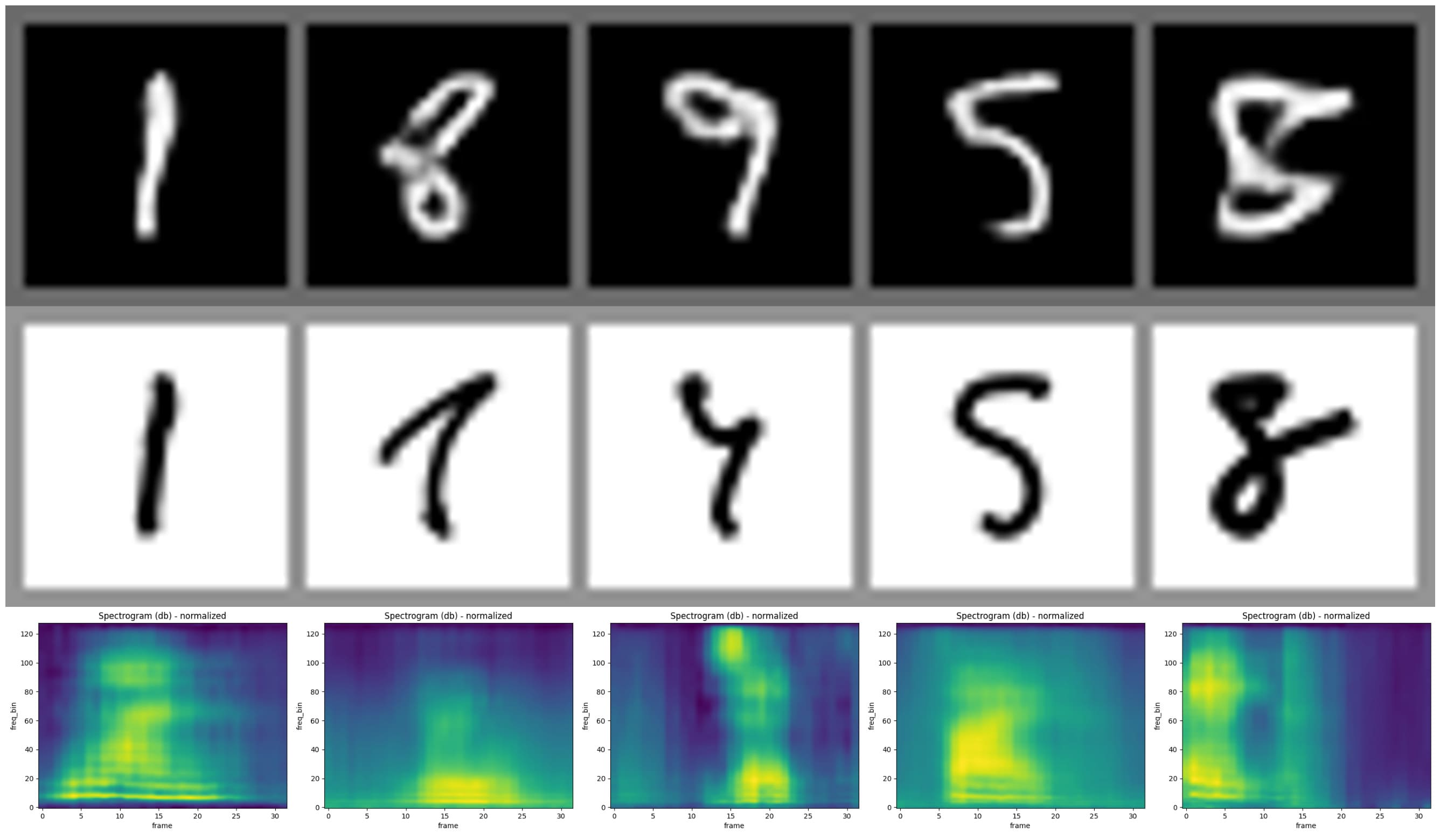}
         \caption*{\gls{MVTCAE}}
     
     \end{subfigure}
    \begin{subfigure}{0.25\textwidth}
         \centering
         \includegraphics[page=1,width=\linewidth]{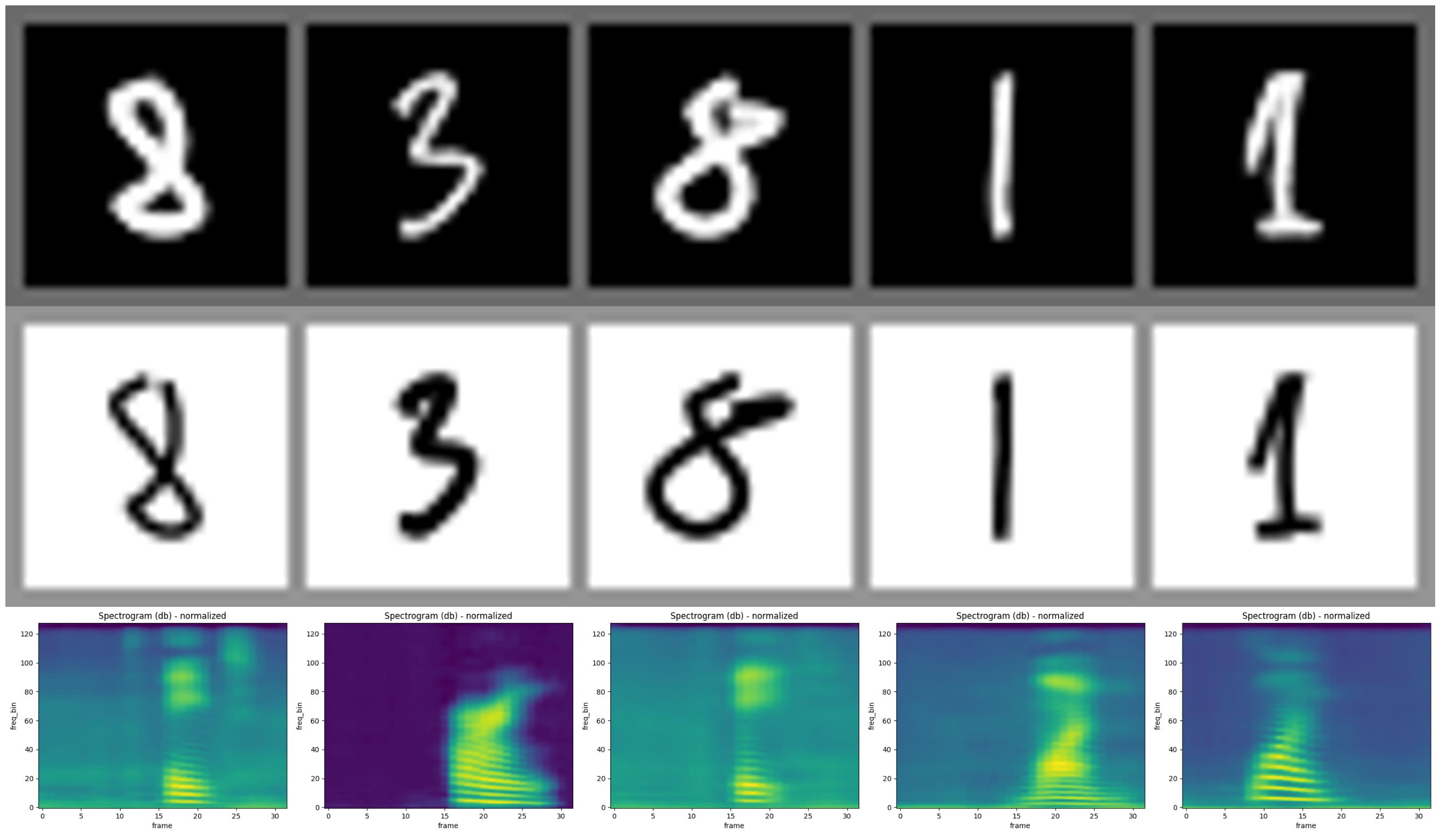}
         \caption*{\gls{MLD Inpaint} }
     \end{subfigure}
     
  \begin{subfigure}{0.25\textwidth}
         \centering
         \includegraphics[page=1,width=\linewidth]{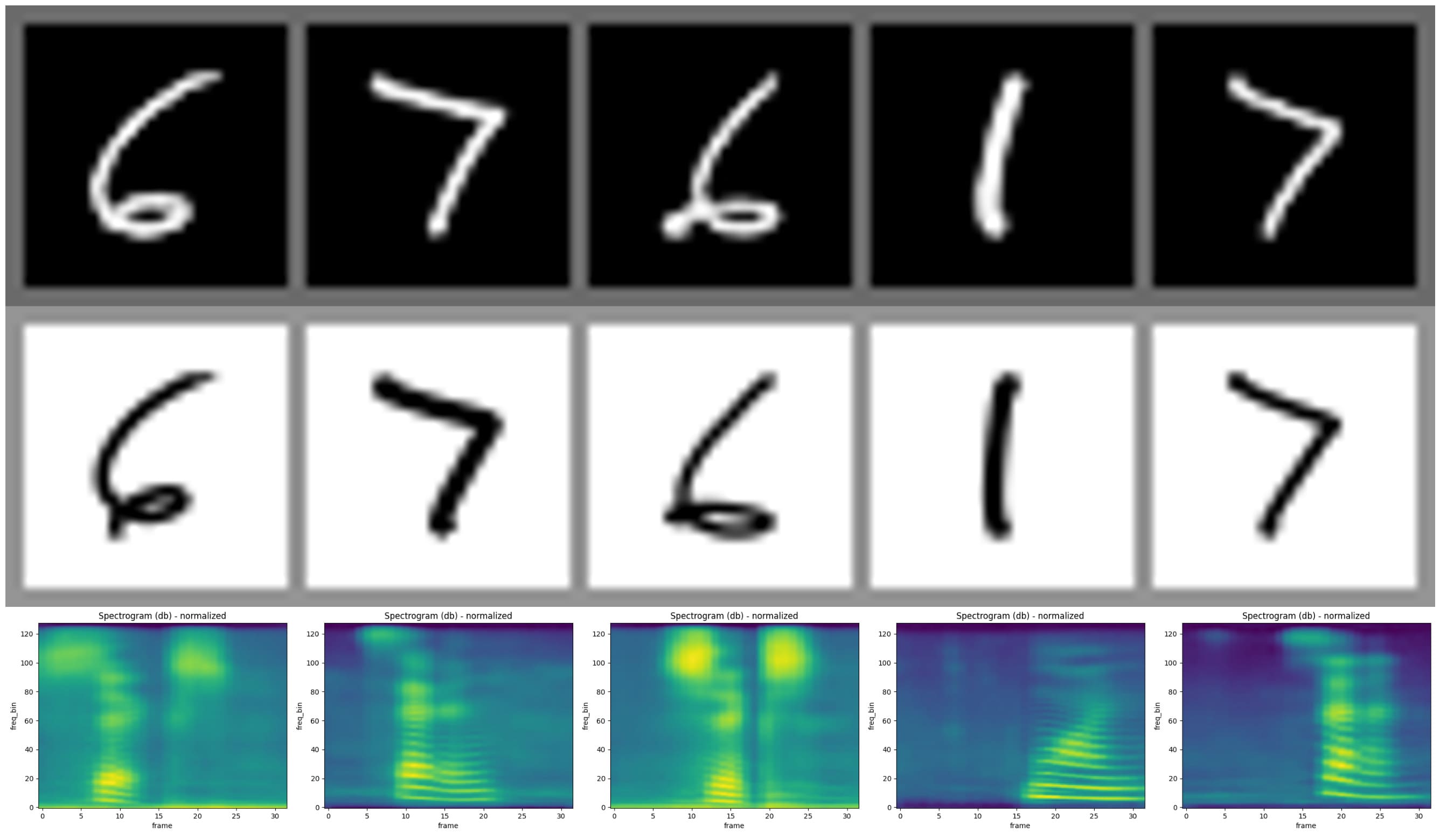}
         \caption*{\gls{MLD Uni} }
     \end{subfigure}
  \begin{subfigure}{0.25\textwidth}
         \centering
         \includegraphics[page=1,width=\linewidth]{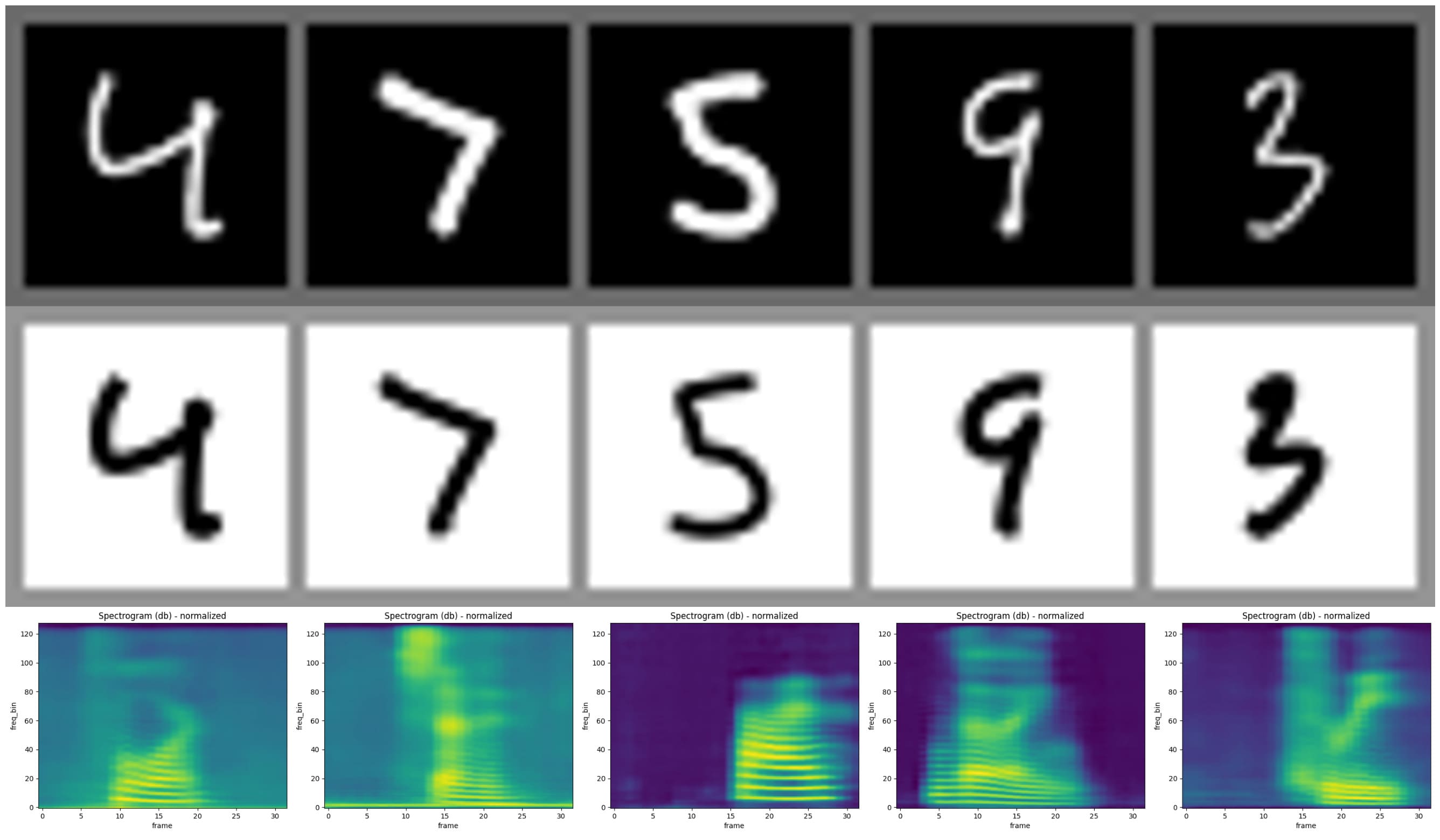}
         \caption*{ \textbf{\gls{MLD} (ours)} }
     \end{subfigure}
     
        \caption{Joint generation qualitative results for \textbf{\mhd}. The three modalities are randomly generated simultaneously (\textbf{Top row}: image,\textbf{Middle row}: trajectory vector converted into image, \textbf{Bottom row}: sound Mel-Spectogram ).   }
        \label{fig:joint_mhd}
\end{figure}

\begin{figure}[H]
     \centering
       \centering
\begin{subfigure}{0.20\textwidth}
         \centering
\includegraphics[width=\linewidth]{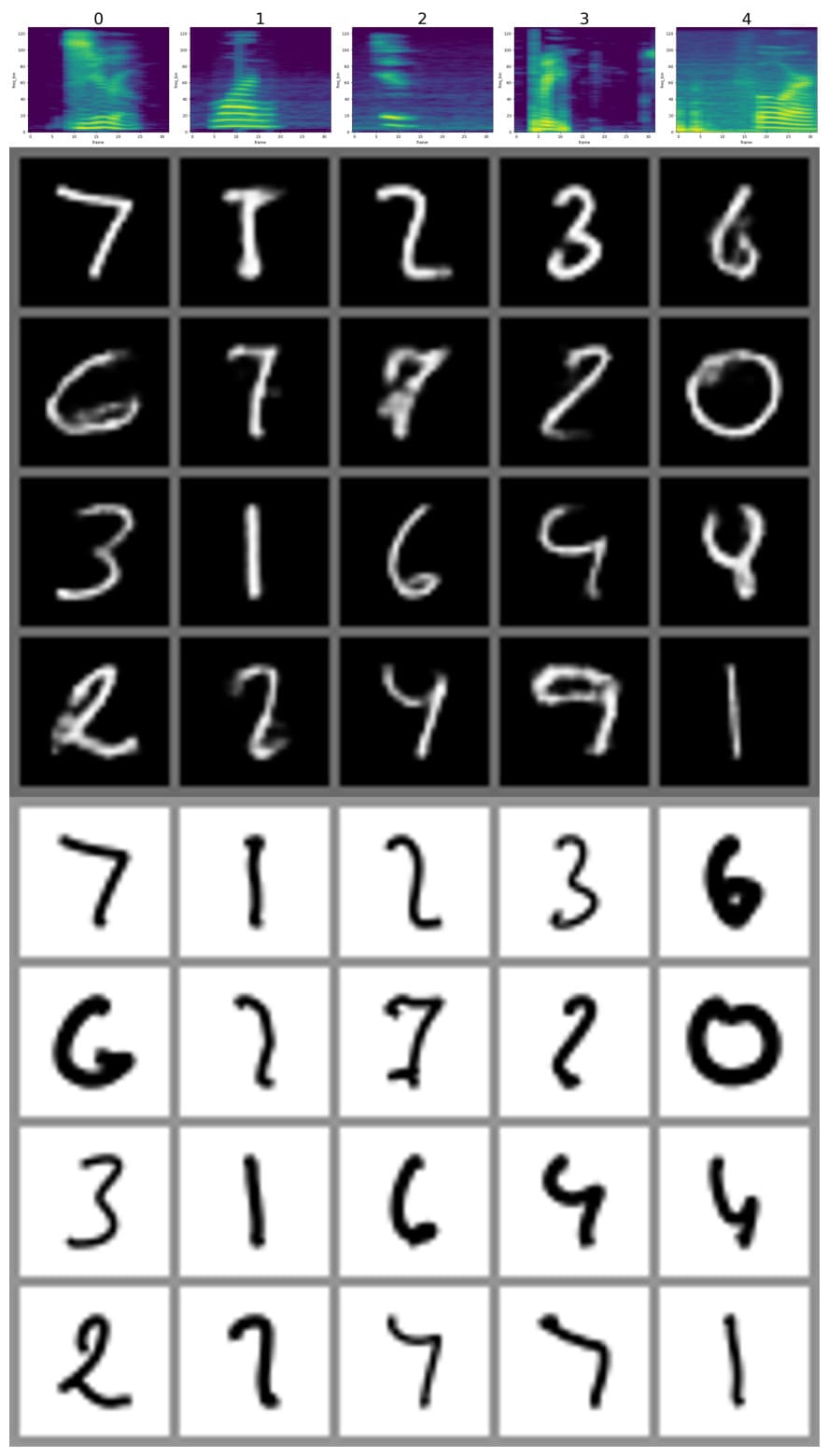}
  \caption*{\gls{MVAE}}
\end{subfigure}
\begin{subfigure}{0.20\textwidth}
         \centering
         \includegraphics[width=\linewidth]{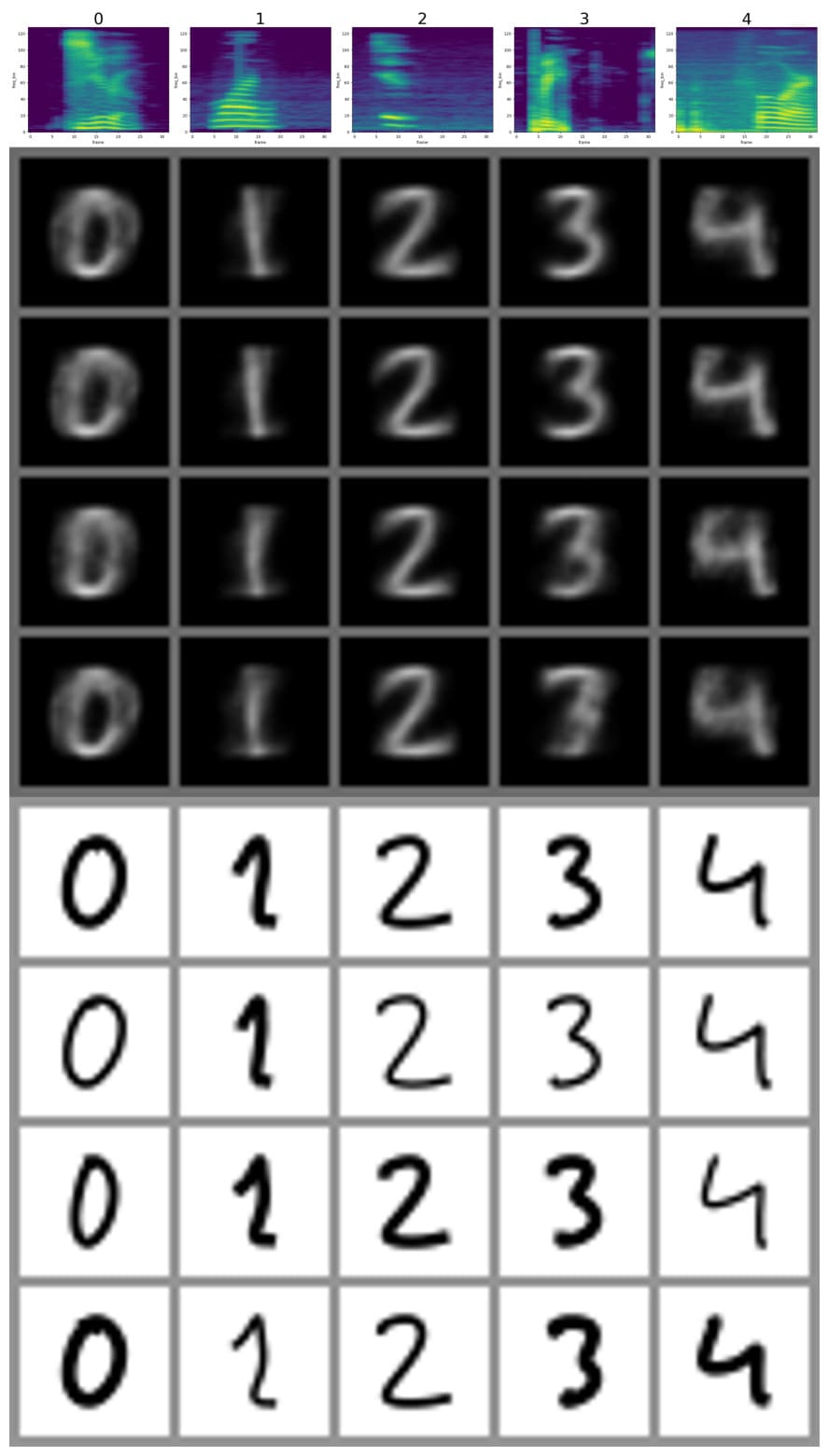}
   \caption*{\gls{MMVAE}}
\end{subfigure}
\begin{subfigure}{0.20\textwidth}
         \centering
         \includegraphics[width=\linewidth]{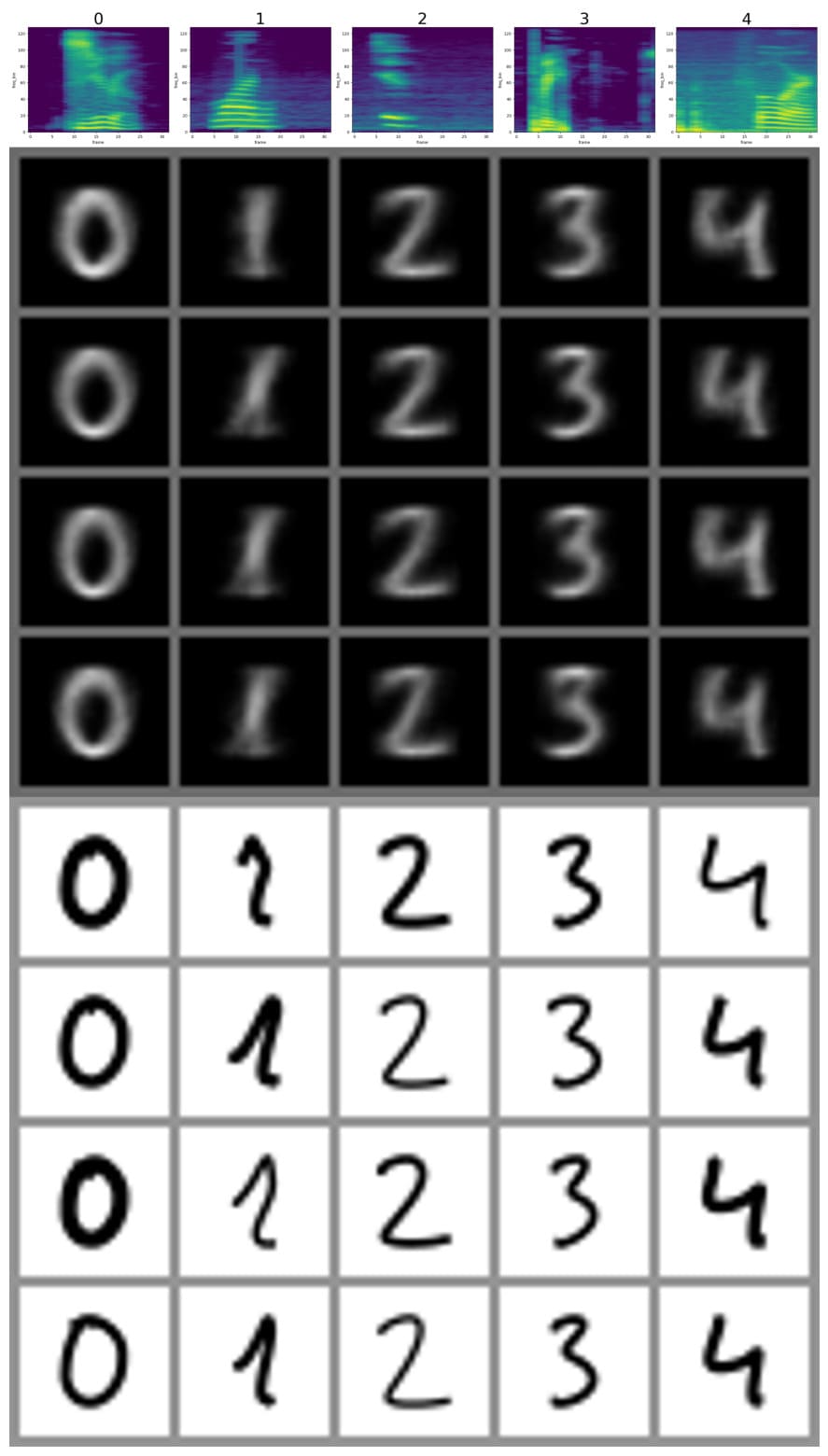}
         \caption*{\gls{MOPOE}}
\end{subfigure}
\begin{subfigure}{0.20\textwidth}
         \centering
         \includegraphics[width=\linewidth]{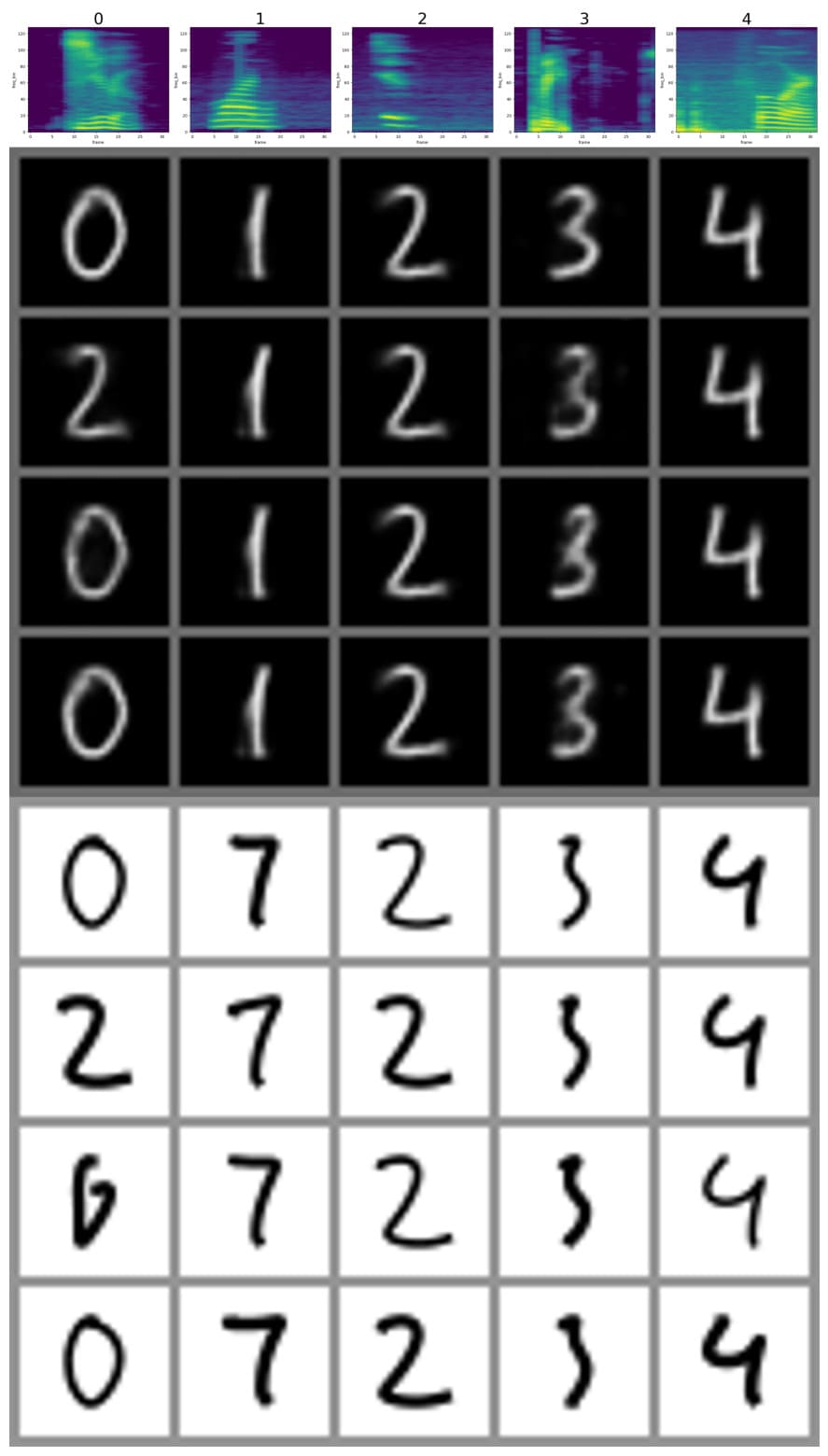}
         \caption*{\gls{NEXUS}}
     \end{subfigure}
     
\begin{subfigure}{0.20\textwidth}
         \centering
         \includegraphics[width=\linewidth]{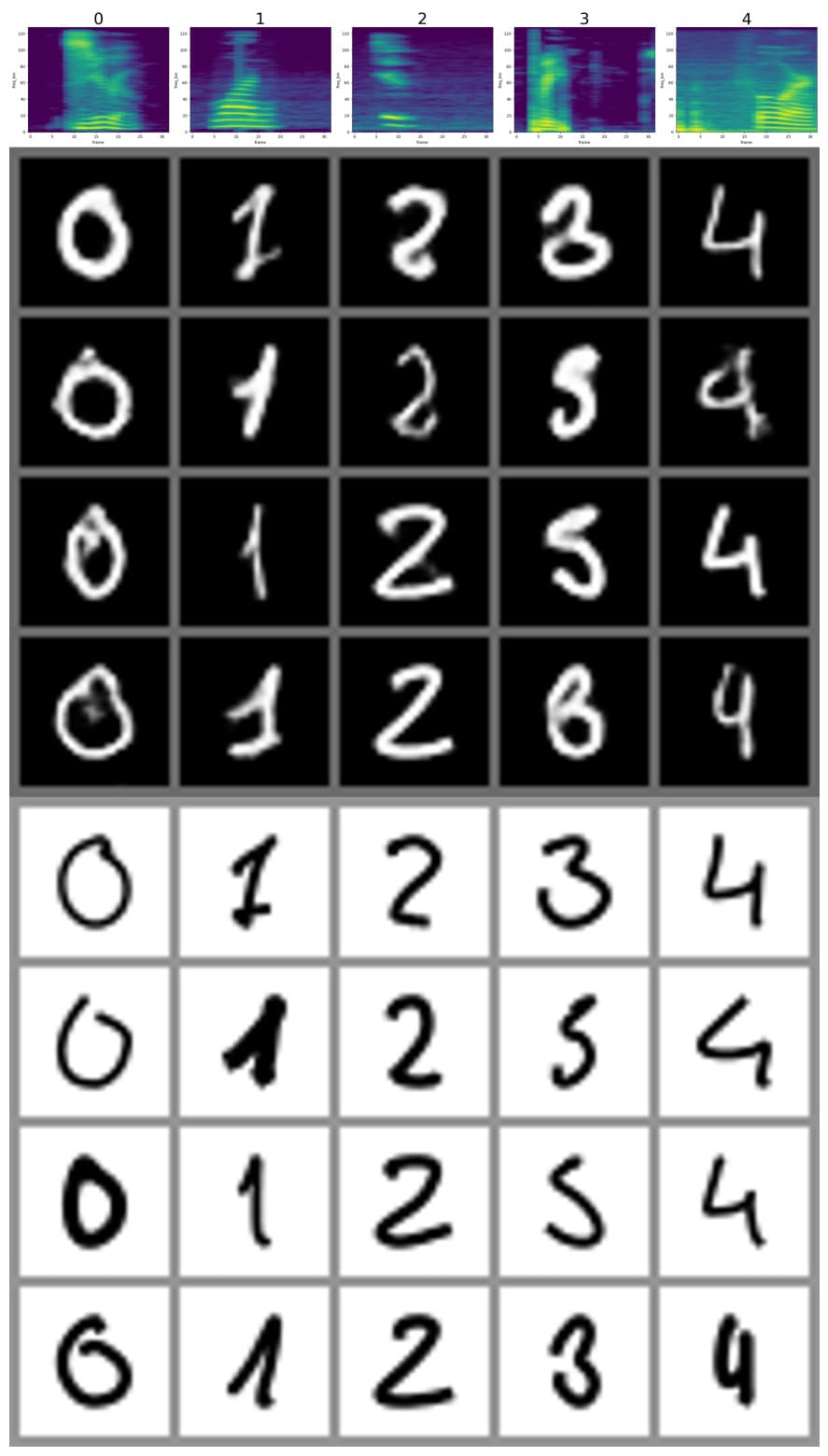}
         \caption*{\gls{MVTCAE}}
\end{subfigure}
  \begin{subfigure}{0.20\textwidth}
         \centering
         \includegraphics[page=1,width=\linewidth]{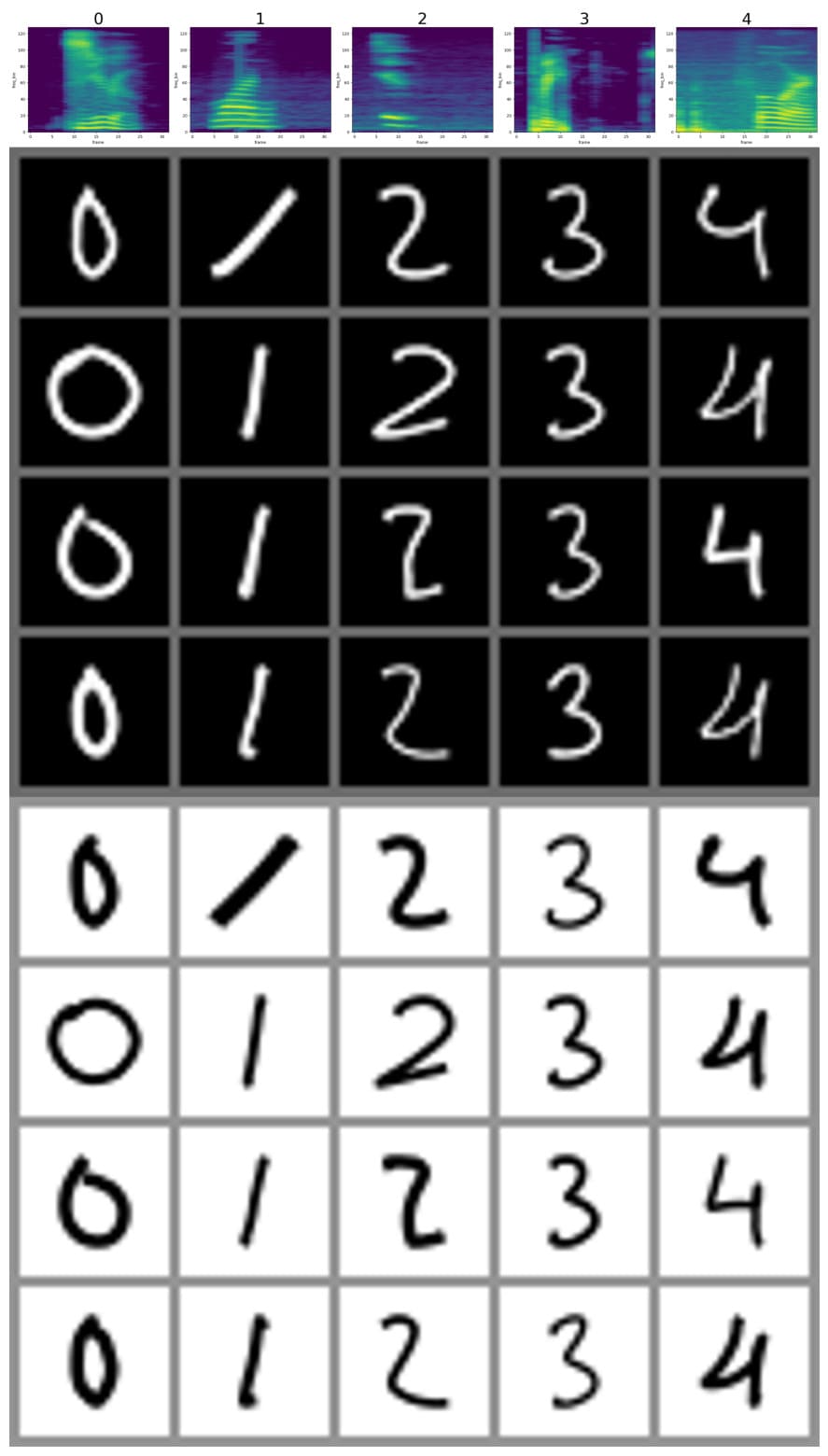}
         \caption*{ \textbf{\gls{MLD} (ours)} }
     \end{subfigure}
     
        \caption{ Sound to image and trajectory conditional generation qualitative results for \textbf{\mhd}. For each model we report: In the \textbf{Top row}, the sound mel-spectograms of the digits \{0,1,2,3,4\} from the left to the right, in \textbf{the rows below}, the generated images and trajectories samples.   }
        \label{fig:cond_mhd}
\end{figure}

\subsection{POLYMNIST}

\begin{table}[H]

\caption{Generation Coherence (\%) for \textbf{\polymnist} (Higher is better) used for the plots in \Cref{fig:res_mmnist} and \Cref{fig:res_mmnist_mld}. We report the average \textit{leave one out coherence} as a function of the number of observed modalities. \textit{Joint} refers to random generation of the 5 modalities simultaneously.
\newline
}  
\centering
\resizebox{0.8\textwidth}{!}{\begin{tabular}{c|c|cccc}
\toprule
\multirow{2}{*}{ Models }    &  \multicolumn{5}{c}{Coherence (\%$\uparrow$)}\\
    \cmidrule{2-6}
     &Joint  & 1 &  2 & 3 & 4   \\
    \midrule
    \gls{MVAE} &  $4.0_{ \pm 1.49 }$ & $37.51_{\pm 3.16 }$ & $ 48.06_{\pm 3.55}$ & $53.19_{ \pm3.37  }$  & $56.09 _{\pm  3.31 }$ \\
    \gls{MMVAE} &  $25.8_{ \pm 1.43 }$ & $\textbf{75.15}_{\pm 2.54}$ & $ 75.14_{\pm2.47}$ & $75.09_{ \pm2.6  }$  & $75.09 _{\pm  2.58 }$    \\

   \gls{MOPOE} & $ 17.32_{ \pm 2.47 }$ &
   $\underline{69.37}_{\pm 1.85}$ &
   $\underline{81.29}_{\pm 2.34}$ &
   $85.26_{ \pm 2.36  }$  & 
   $86.7 _{\pm 2.39 }$      \\
 
    \gls{NEXUS} &$ 18.24_{ \pm 0.89 }$ &
    $  60.61_{ \pm 2.51 }$ &
    $ 72.14_{ \pm 2.79 }$ & 
    $ 76.81_{ \pm 2.75 }$ &
    $ 78.92_{ \pm 2.64}$ \\

    \gls{MVTCAE} & $0.21_{\pm 0.05 }$ & 
    $57.66_{\pm 1.06 }$ &
    $78.44_{\pm1.31} $&
    $\underline{85.97}_{\pm 1.43} $&
    $\underline{88.81}_{ \pm 1.49  }$ \\

    \gls{MMVAEplus} &  $26.28_{\pm 2.19 }$ & 
    $54.74_{\pm 0.5 }$ &
    $54.06_{\pm0.33} $&
    $55.2_{\pm 1.32} $&
    $53.17_{ \pm0.75 }$ \\

    \gls{MMVAEplus} (K=10)&  $14.53_{\pm 4.94 }$ & 
    $58.93_{\pm 6.3 }$ &
    $59.42_{\pm8.8} $&
    $60.77_{\pm 8.03} $&
    $58.24_{ \pm 7.42  }$ \\

    \midrule
    \gls{MLD Inpaint} & $\underline{51.65}_{\pm 1.16 }$ &  $52.85_{\pm 0.23 }$ &
    $77.65_{\pm0.24} $& 
    $85.66_{\pm 0.43} $&
    $87.29_{ \pm 0.29  }$ \\

    \gls{MLD Uni} & 
    $48.79_{\pm 0.43 }$ &  
    $65.12_{\pm 0.7 }$ &
    $79.52_{\pm0.8} $&
    $82.03_{\pm 1.19} $&
    $81.86_{ \pm 2.09  }$ 
    \\

    \gls{MLD} & $\textbf{56.23}_{ \pm 0.52}$ & $68.58_{\pm 0.72} $&$ \textbf{84.87}_{\pm 0.19}$ & $\textbf{88.56}_{ \pm 0.12} $ &$ \textbf{89.43}_{\pm 0.27}  $ \\
    
    \bottomrule
\end{tabular}}
\label{tab:mmnist_detail}
\end{table}

\begin{table}[H]
\caption{Generation quality (\gls{FID} $\downarrow$) for \textbf{\polymnist} (lower is better) used for the plots in \Cref{fig:res_mmnist} and \Cref{fig:res_mmnist_mld}. Similar to \Cref{tab:mmnist_detail}, we report the average \textit{leave one out} \gls{FID} as a function of the number of observed modalities. \textit{Joint} refers to random generation quality of the 5 modalities simultaneously.
\newline
}  
\centering
\resizebox{0.8\textwidth}{!}{\begin{tabular}{c|c|cccc}
\toprule
\multirow{1}{*}{ Models }    &  \multicolumn{5}{c}{Quality ($\downarrow$)}\\
    \cmidrule{2-6}
     &Joint  & 1 &  2 & 3 & 4   \\
    \midrule
    \gls{MVAE} &  $108.74_{ \pm 2.73 }$ & $108.06_{\pm 2.79 }$ & $ 108.05_{\pm 2.73}$ & $108.14_{ \pm2.71  }$  & $108.18_{\pm  2.85 }$ \\

   \gls{MMVAE} &  $165.74_{ \pm 5.4}$ & $208.16_{\pm 10.41 }$ & $ 207.5_{\pm10.57}$ & $207.35_{ \pm10.59 }$  & $207.38_{\pm10.58 }$    \\
   \gls{MOPOE} & $ 113.77_{ \pm 1.62 }$ & $173.87_{\pm 7.34 }$ & $185.06_{\pm 10.21}$ & $191.72_{ \pm 11.26  }$  & $196.17
 _{\pm 11.66}$      \\
    \gls{NEXUS} & $ 91.66_{ \pm 2.93 }$ & $207.14_{\pm 7.71 }$ & $205.54_{\pm 8.6}$ & $204.46_{ \pm 9.08  }$  & $202.43
 _{\pm 9.49}$     \\
    \gls{MVTCAE} &
    $106.55_{\pm 3.83  } $ & $78.3_{\pm2.35} $  &   $85.55_{\pm2.51 }$&   $92.73_{\pm2.65 }$&   $99.13_{\pm2.72 }$\\

     \gls{MMVAEplus} &  $168.88_{\pm 0.12 }$ & 
    $165.67_{\pm 0.14 }$ &
    $166.5_{\pm0.18} $&
    $165.53_{\pm 0.55} $&
    $165.3_{ \pm0.33 }$ \\

    \gls{MMVAEplus} (K=10)&  $156.55_{\pm 3.58 }$ & 
    $154.42_{\pm 2.73 }$ &
    $153.1_{\pm3.01} $&
    $153.06_{\pm 2.88} $&
    $154.9_{ \pm 2.9  }$ \\
    
    \midrule
    \gls{MLD Inpaint} &   $64.78_{\pm0.33 }$&
    $65.41_{\pm 0.43  } $ & $65.42_{\pm0.41} $  &   $65.52_{\pm0.46 }$&   $\underline{65.55}_{\pm0.46 }$\\
    
    \gls{MLD Uni} &   $\textbf{62.42}_{\pm0.62 }$&
    $\underline{63.16}_{\pm 0.81  } $ & $\underline{64.09}_{\pm1.15} $  &   $\underline{65.17}_{\pm1.46 }$&   $66.46_{\pm2.18 }$\\
     
    \gls{MLD} & 
    $\underline{63.05}_{\pm0.26}$   &   $ \textbf{62.89}_{\pm 0.2}$ &
     $\textbf{62.53}_{\pm0.21}$   &   $\textbf{62.22}_{\pm 0.39}$ &   $\textbf{61.94}_{\pm0.65}$ \\
    \bottomrule
\end{tabular}}

\end{table}

\begin{figure} [H]
     \centering
     \begin{subfigure}{0.19\textwidth}
        \begin{subfigure}{1\textwidth}
         \centering
         \includegraphics[page=1,width=\linewidth]{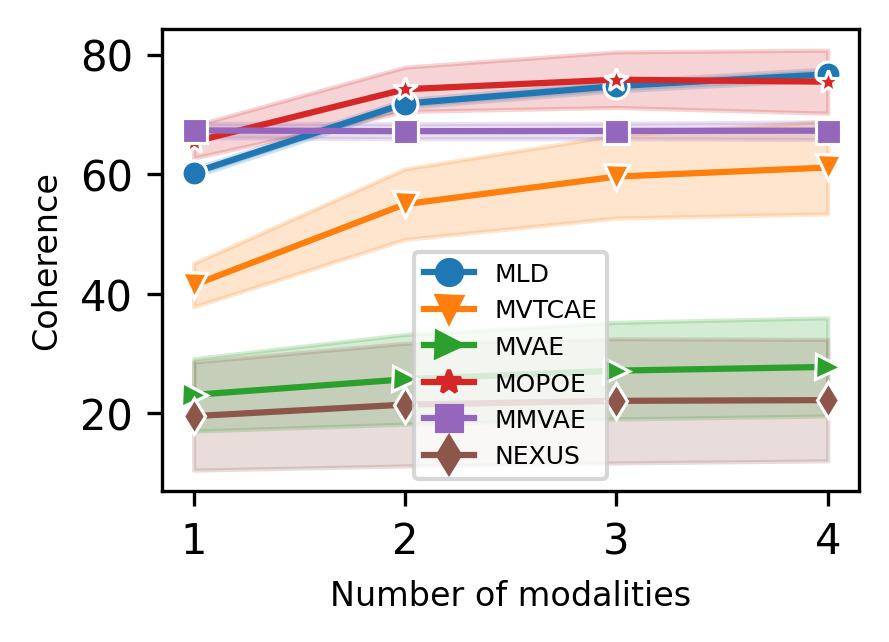}

        \end{subfigure} 
          \begin{subfigure}{1\textwidth}
         \centering
         \includegraphics[page=1,width=\linewidth]{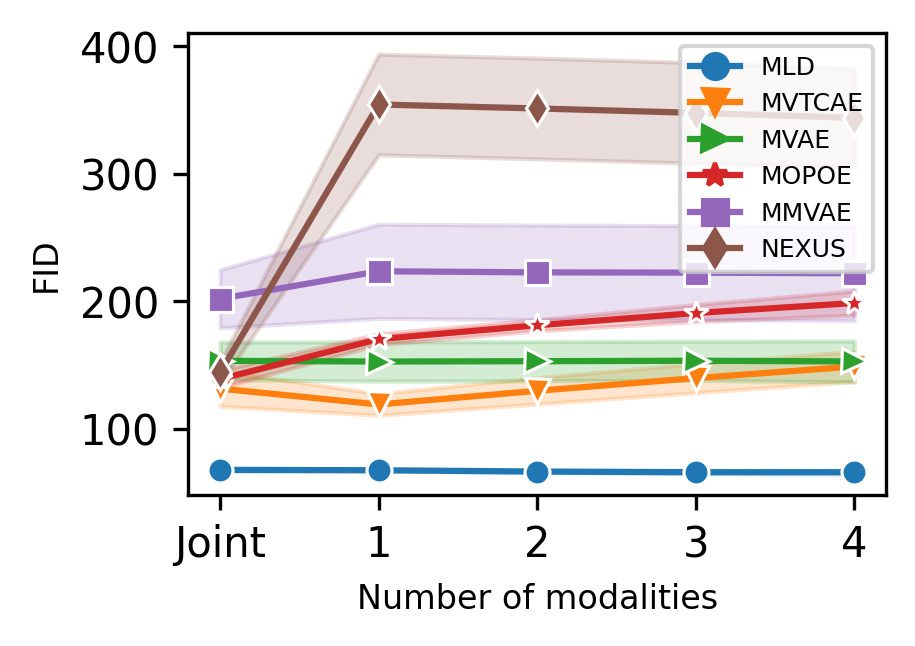}

        \end{subfigure} 
        \caption{$X^0$}
     \end{subfigure} 
   \begin{subfigure}{0.19\textwidth}
        \begin{subfigure}{1\textwidth}
         \centering
         \includegraphics[page=1,width=\linewidth]{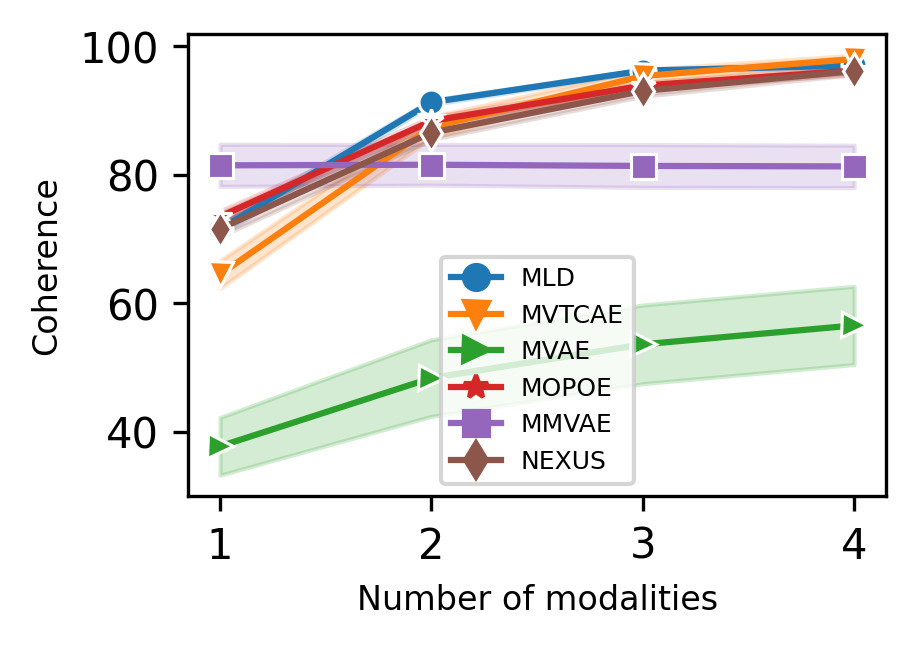}

        \end{subfigure} 
          \begin{subfigure}{1\textwidth}
         \centering
         \includegraphics[page=1,width=\linewidth]{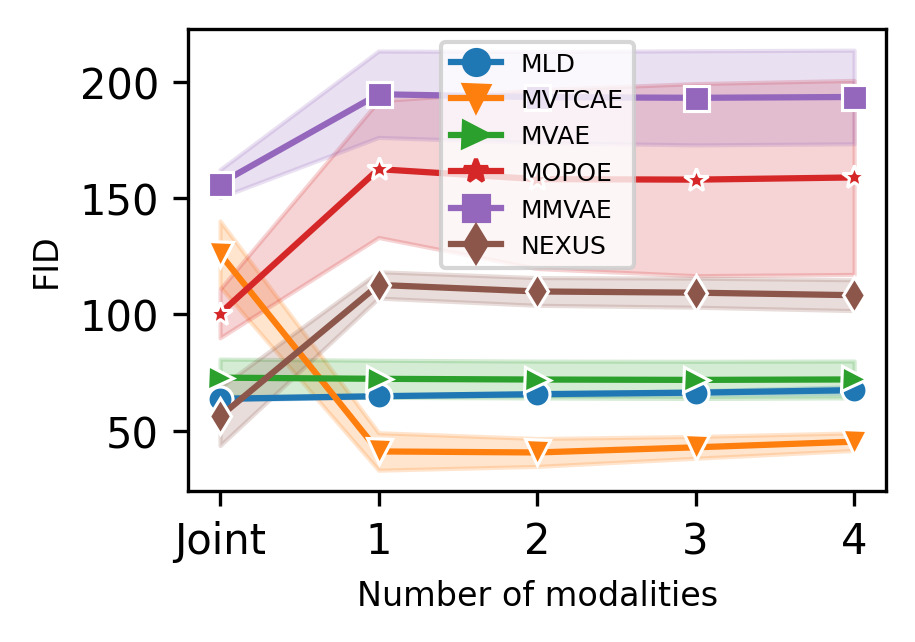}
   
        \end{subfigure} 

        \caption{$X^1$}
     \end{subfigure}
     \begin{subfigure}{0.19\textwidth}
        \begin{subfigure}{1\textwidth}
         \centering
         \includegraphics[page=1,width=\linewidth]{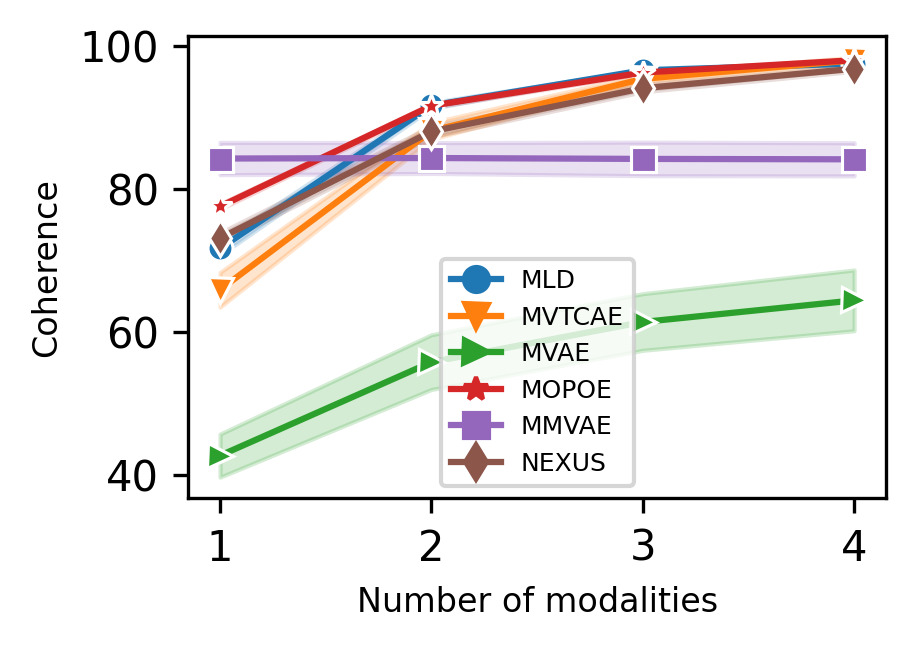}

        \end{subfigure} 
          \begin{subfigure}{1\textwidth}
         \centering
         \includegraphics[page=1,width=\linewidth]{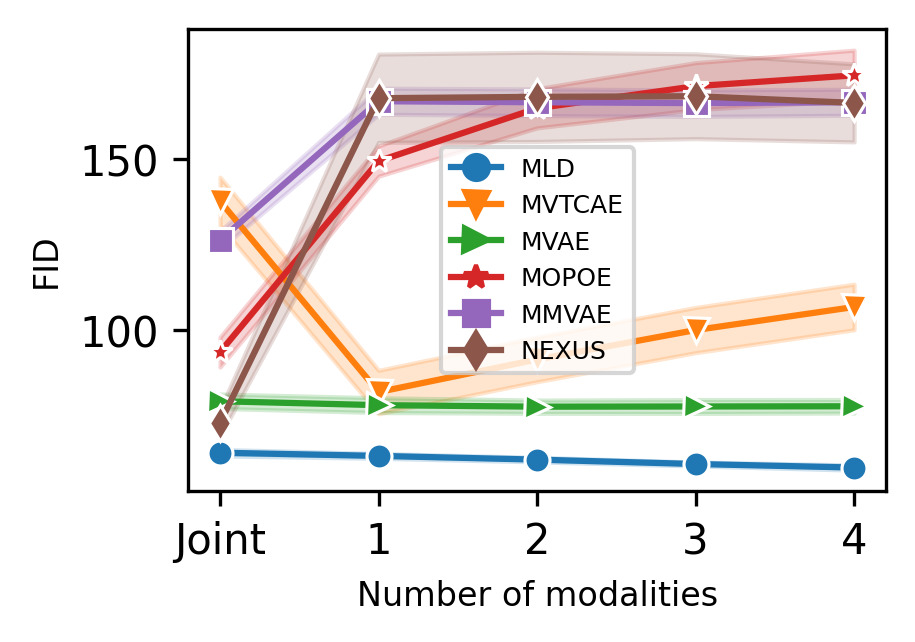}

        \end{subfigure} 
        \caption{$X^2$}
     \end{subfigure}
      \begin{subfigure}{0.19\textwidth}
        \begin{subfigure}{1\textwidth}
         \centering
         \includegraphics[page=1,width=\linewidth]{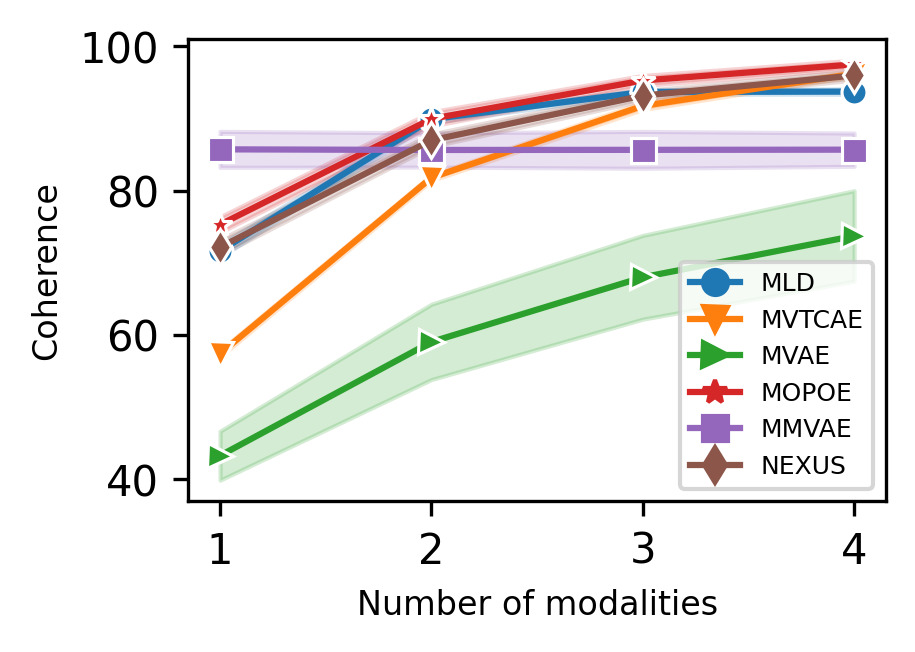}
    
        \end{subfigure} 
          \begin{subfigure}{1\textwidth}
         \centering
         \includegraphics[page=1,width=\linewidth]{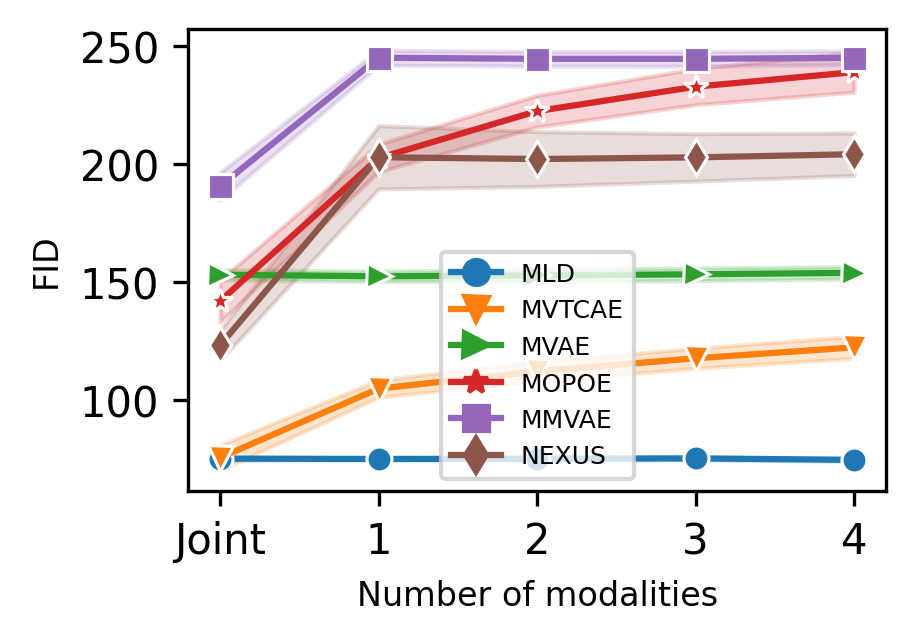}
    
        \end{subfigure} 
        \caption{$X^3$}
     \end{subfigure}
     \begin{subfigure}{0.19\textwidth}
        \begin{subfigure}{1\textwidth}
         \centering
         \includegraphics[page=1,width=\linewidth]{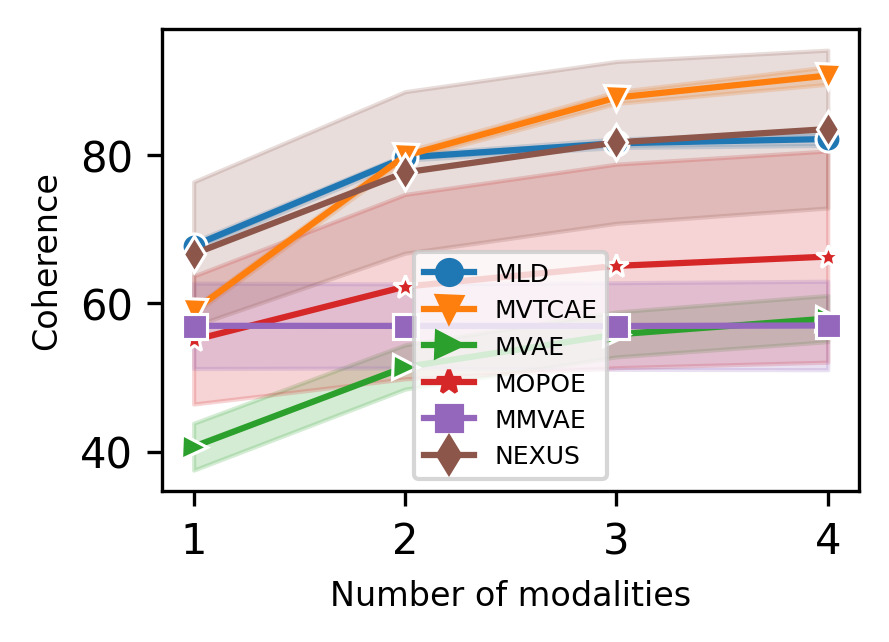}
         
        \end{subfigure} 
          \begin{subfigure}{1\textwidth}
         \centering
         \includegraphics[page=1,width=\linewidth]{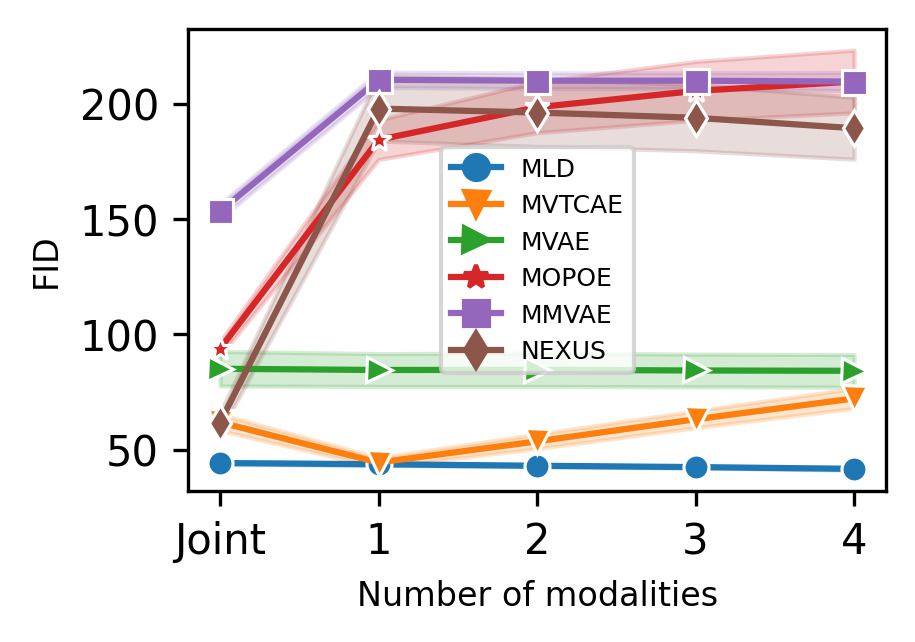}
        
        \end{subfigure} 
        \caption{$X^4$}
     \end{subfigure}
\caption{ \textbf{Top:} Generation Coherence (\%) for \textbf{\polymnist} (Higher is better). \textbf{Bottom:} Generation quality (\gls{FID}) (Lower is better). We report the average \textit{leave one out} performance as a function of the number of observed modalities for each modality $X^i$. \textit{Joint} refers to random generation of the 5 modalities simultaneously.}
        \label{res:mmnist_mld_var_mod}
\end{figure}

\begin{figure} [H]
     \centering
     \begin{subfigure}{0.19\textwidth}
        \begin{subfigure}{1\textwidth}
         \centering
         \includegraphics[page=1,width=\linewidth]{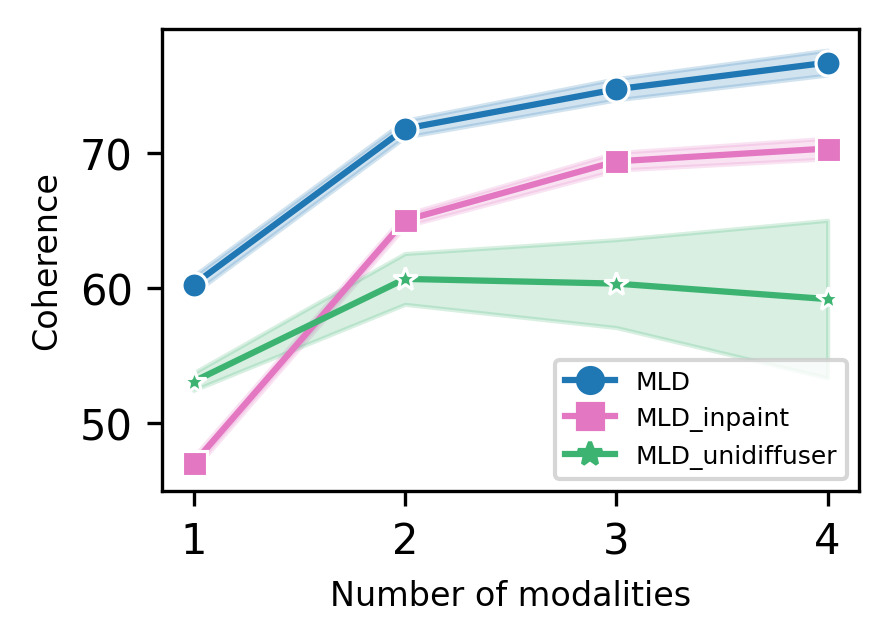}
     
        \end{subfigure} 
          \begin{subfigure}{1\textwidth}
         \centering
         \includegraphics[page=1,width=\linewidth]{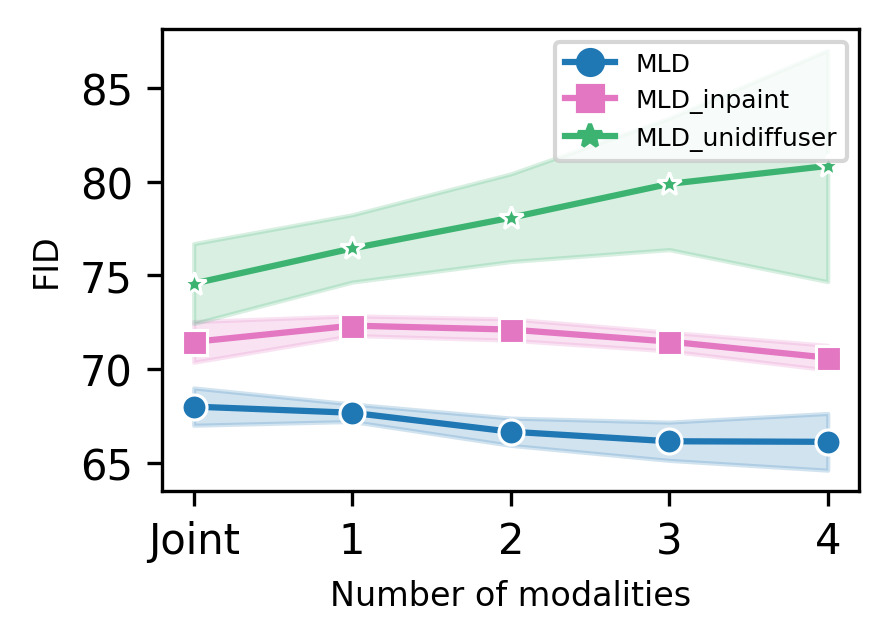}

        \end{subfigure} 
        \caption{$X^0$}
     \end{subfigure} 
   \begin{subfigure}{0.19\textwidth}
        \begin{subfigure}{1\textwidth}
         \centering
         \includegraphics[page=1,width=\linewidth]{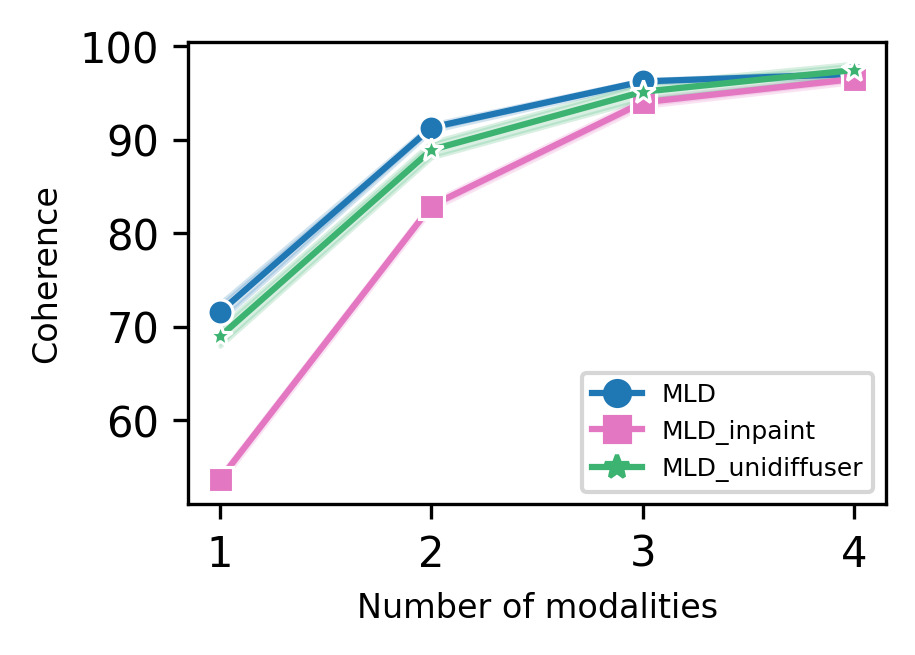}
   
        \end{subfigure} 
          \begin{subfigure}{1\textwidth}
         \centering
         \includegraphics[page=1,width=\linewidth]{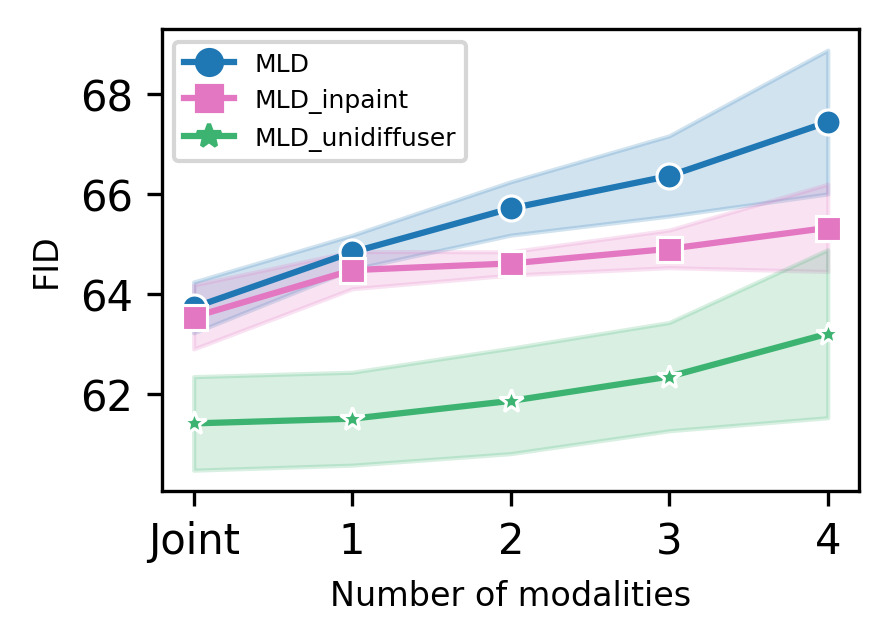}
   
        \end{subfigure} 

        \caption{$X^1$}
     \end{subfigure}
     \begin{subfigure}{0.19\textwidth}
        \begin{subfigure}{1\textwidth}
         \centering
         \includegraphics[page=1,width=\linewidth]{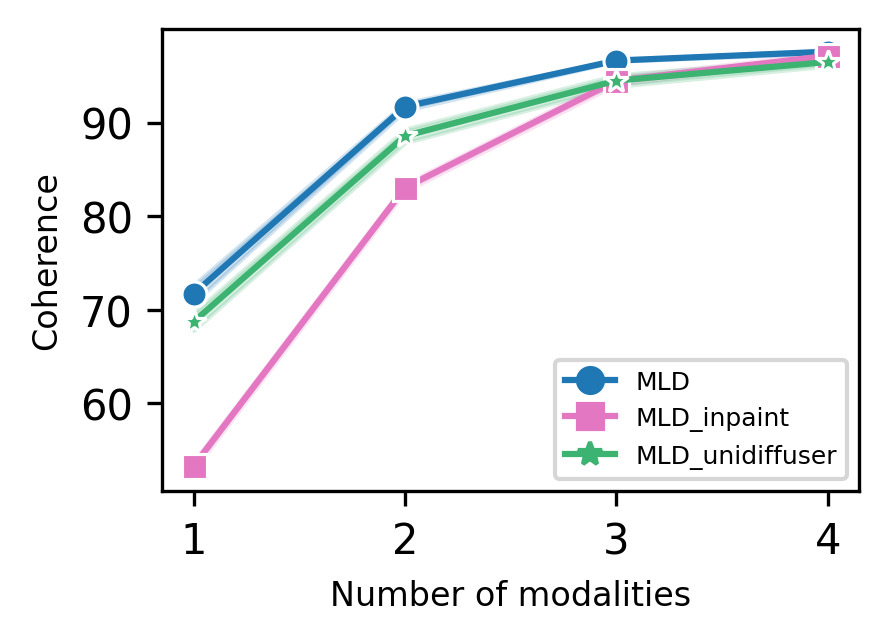}

        \end{subfigure} 
          \begin{subfigure}{1\textwidth}
         \centering
         \includegraphics[page=1,width=\linewidth]{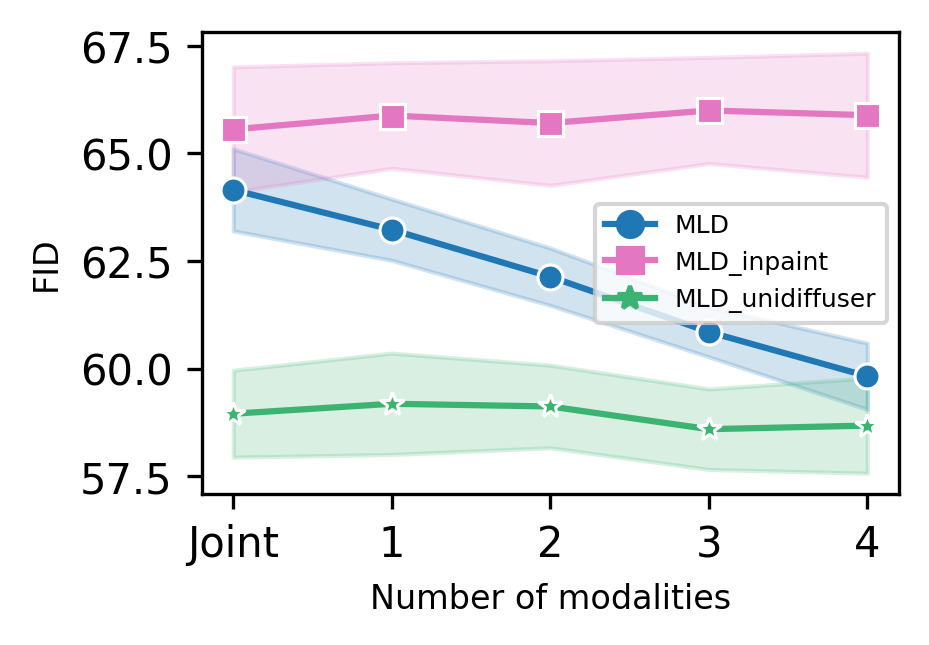}

        \end{subfigure} 
        \caption{$X^2$}
     \end{subfigure}
      \begin{subfigure}{0.19\textwidth}
        \begin{subfigure}{1\textwidth}
         \centering
         \includegraphics[page=1,width=\linewidth]{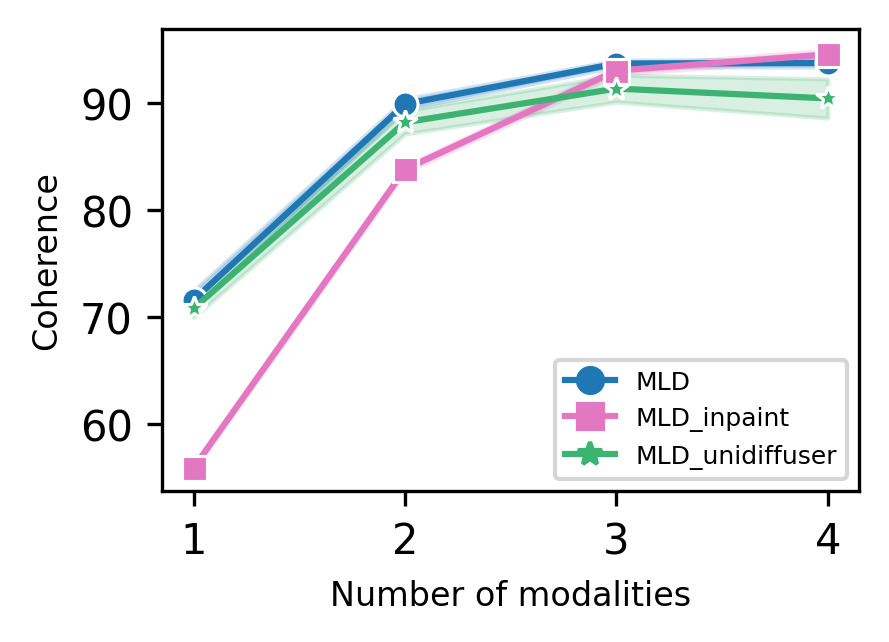}
 
        \end{subfigure} 
          \begin{subfigure}{1\textwidth}
         \centering
         \includegraphics[page=1,width=\linewidth]{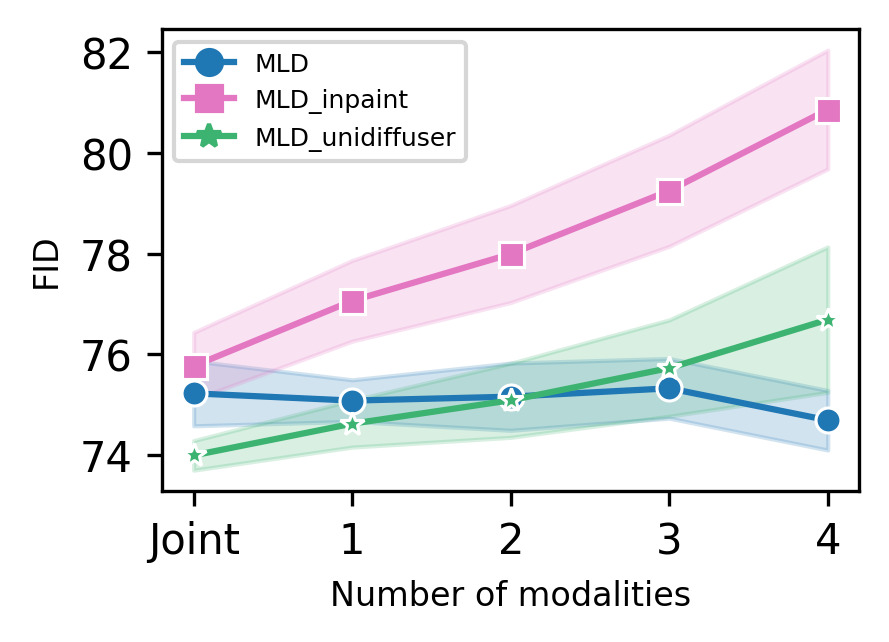}

        \end{subfigure} 
        \caption{$X^3$}
     \end{subfigure}
     \begin{subfigure}{0.19\textwidth}
        \begin{subfigure}{1\textwidth}
         \centering
         \includegraphics[page=1,width=\linewidth]{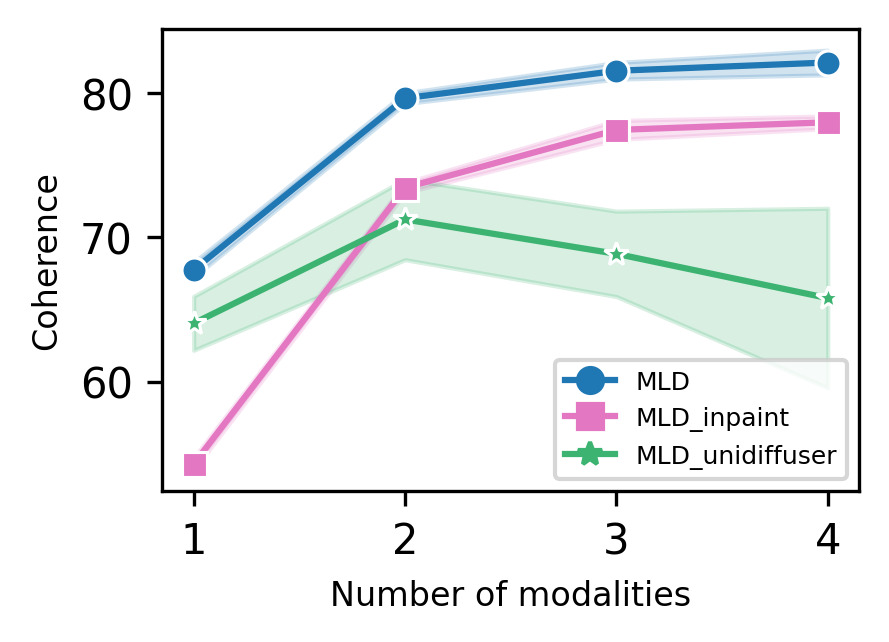}
 
        \end{subfigure} 
          \begin{subfigure}{1\textwidth}
         \centering
         \includegraphics[page=1,width=\linewidth]{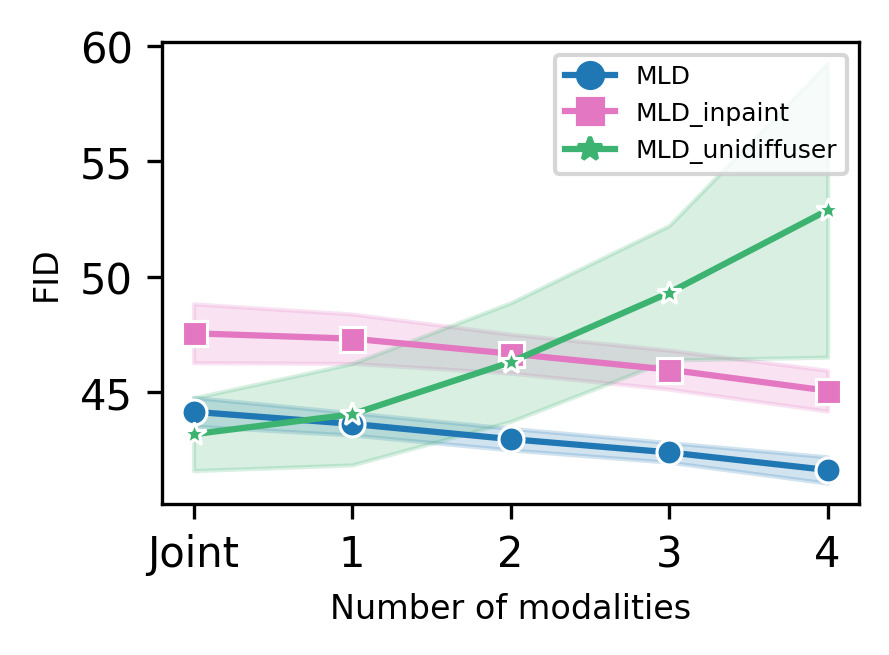}
   
        \end{subfigure} 
        \caption{$X^4$}
     \end{subfigure}
        \caption{\textbf{Top:} Generation Coherence (\%) for \textbf{\polymnist} (Higher is better).\textbf{Bottom:} Generation quality (\gls{FID}) (Lower is better).
        We report the average \textit{leave one out} performance as a function of the number of observed modalities for each modality $X^i$. \textit{Joint} refers to random generation of the 5 modalities simultaneously.}
        \label{res:mmnist_mod}
\end{figure}

\begin{figure}[H]
     \centering
       \centering
     \begin{subfigure}{0.2\textwidth}
         \centering
         \includegraphics[width=\linewidth]{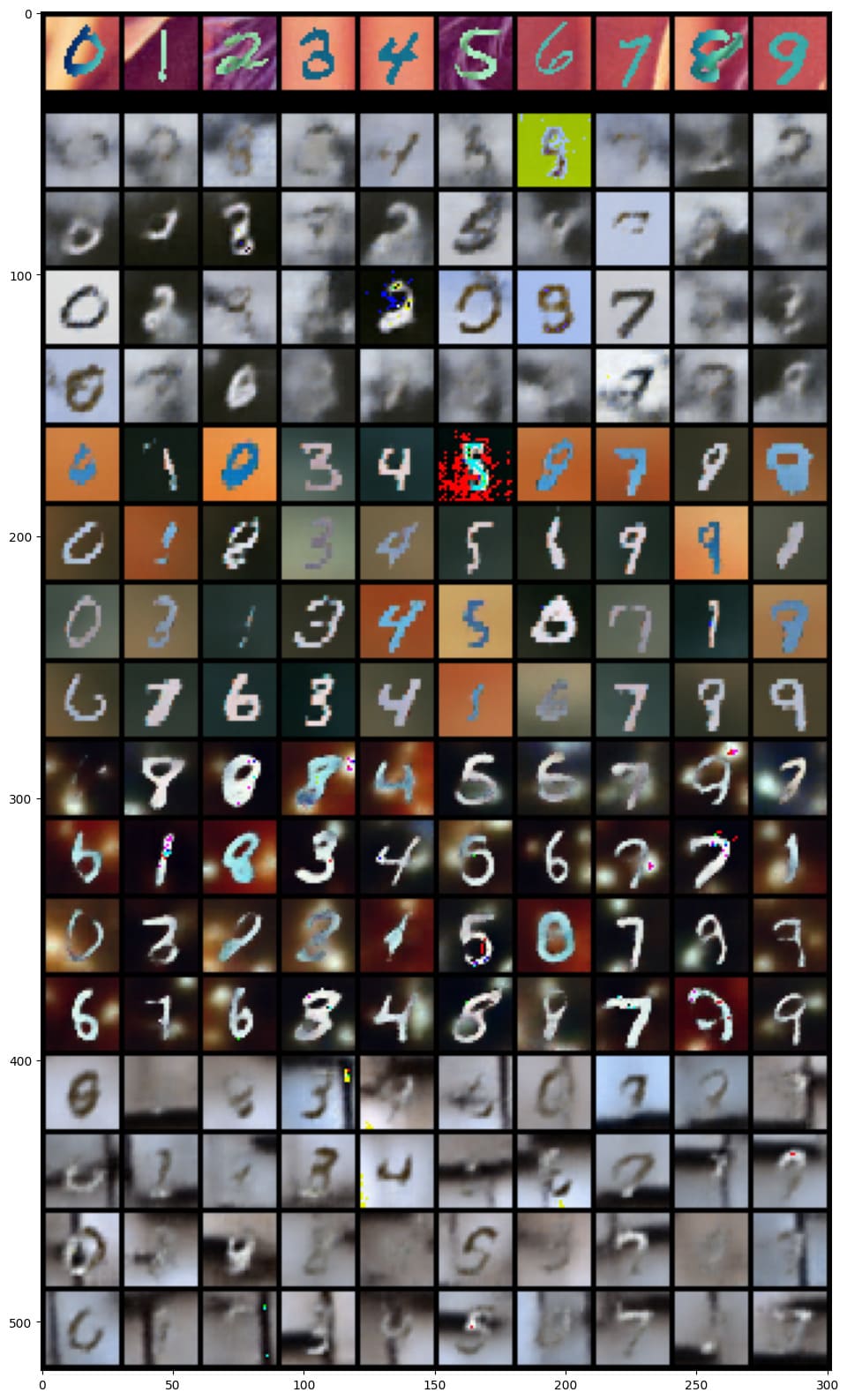}
  \caption*{\gls{MVAE}}
     \end{subfigure}
     \begin{subfigure}{0.2\textwidth}
         \centering
         \includegraphics[width=\linewidth]{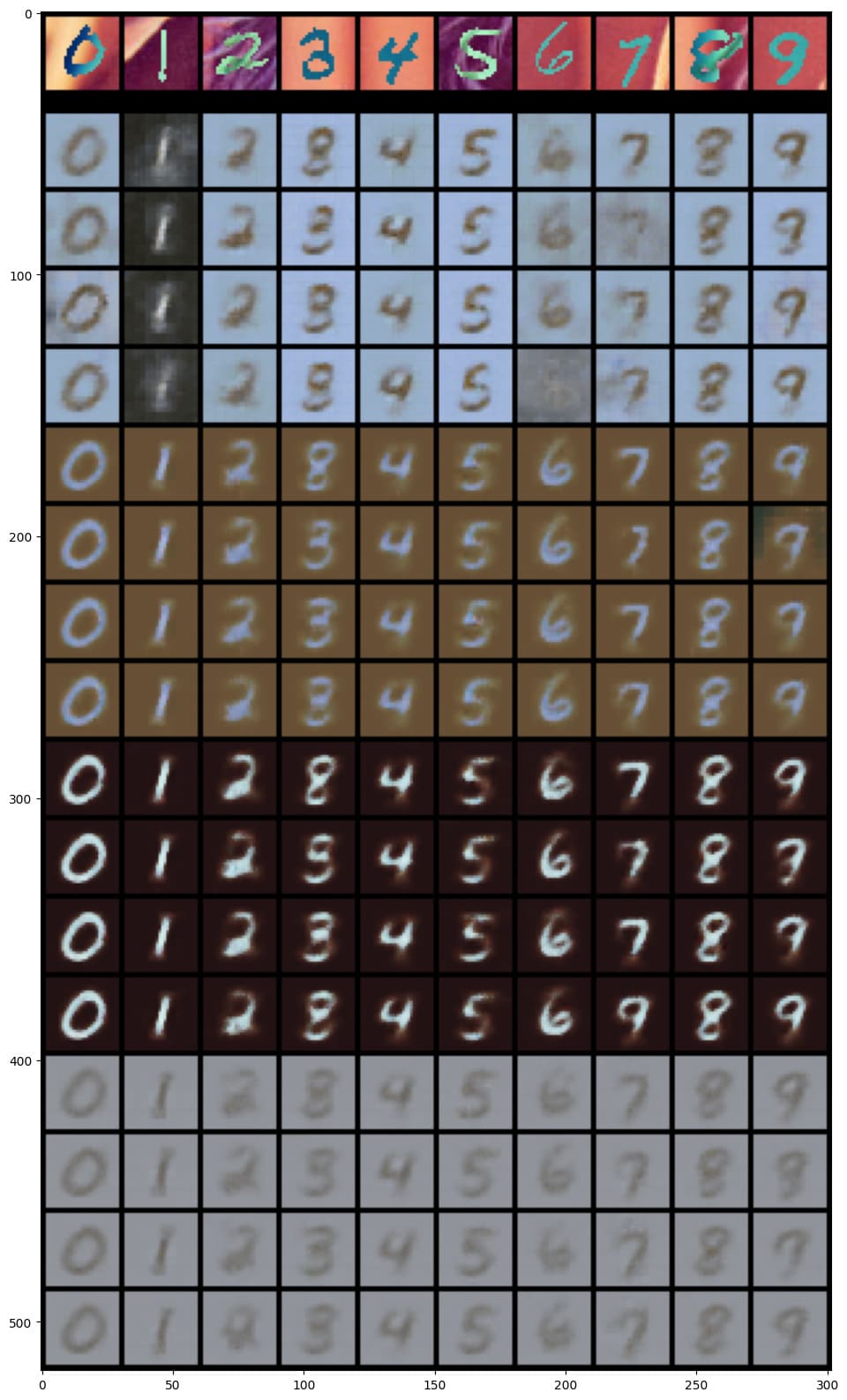}
   \caption*{\gls{MMVAE}}
     \end{subfigure}
       \begin{subfigure}{0.2\textwidth}
         \centering
         \includegraphics[width=\linewidth]{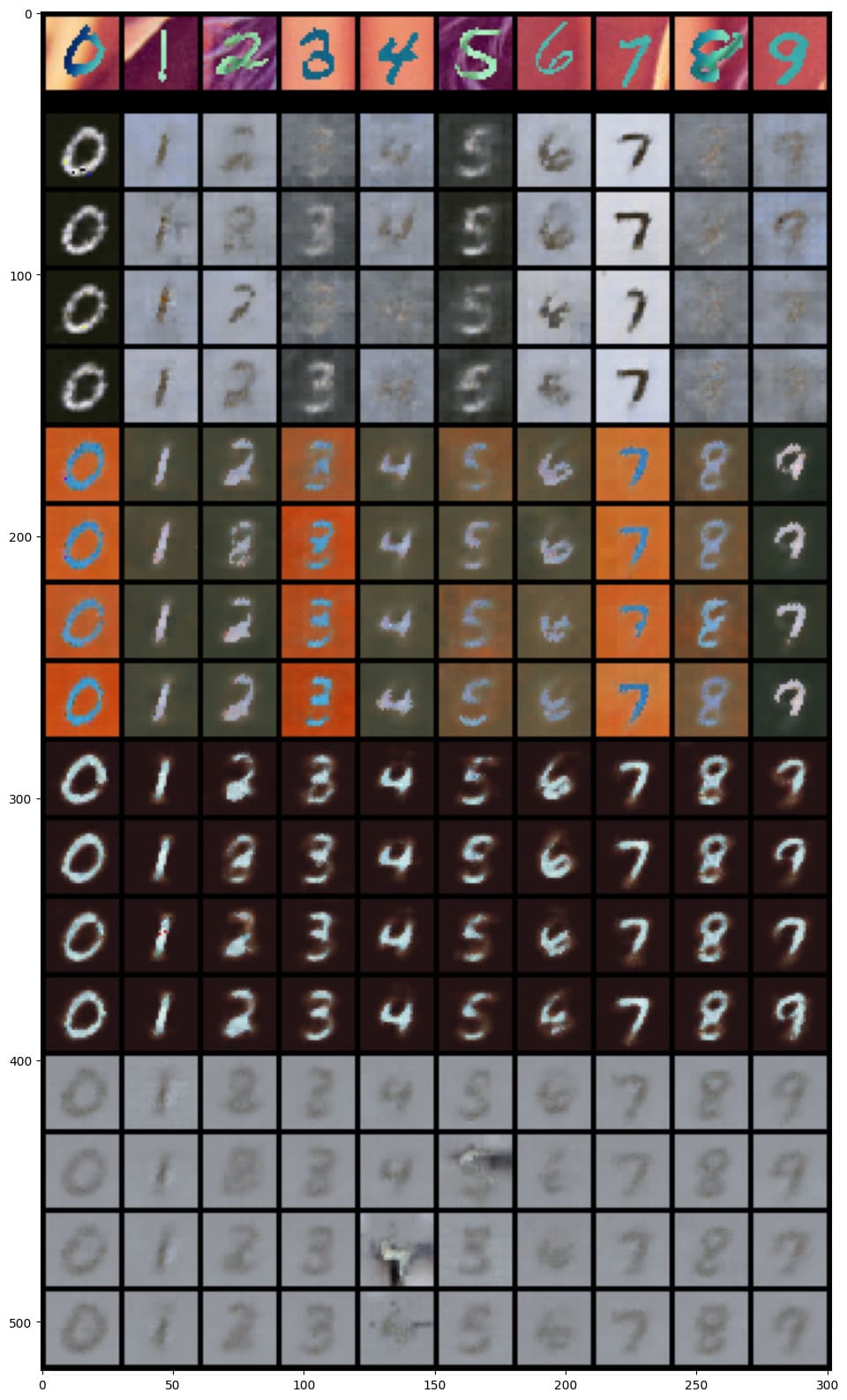}
         \caption*{\gls{MOPOE}}

     \end{subfigure}
      \begin{subfigure}{0.2\textwidth}
         \centering
         \includegraphics[width=\linewidth]{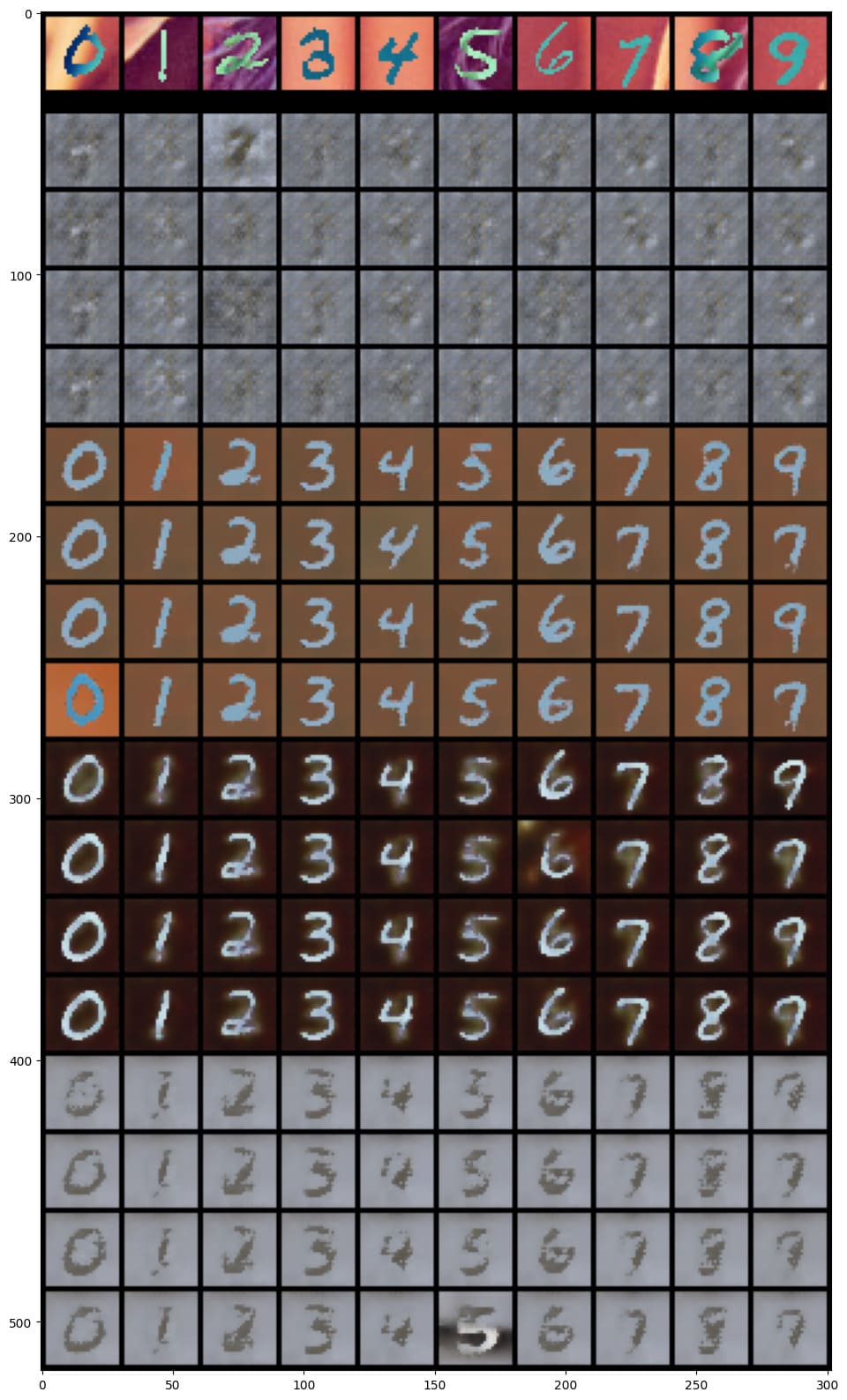}
         \caption*{\gls{NEXUS}}
      
     \end{subfigure}
     
  \begin{subfigure}{0.2\textwidth}
         \centering
         \includegraphics[width=\linewidth]{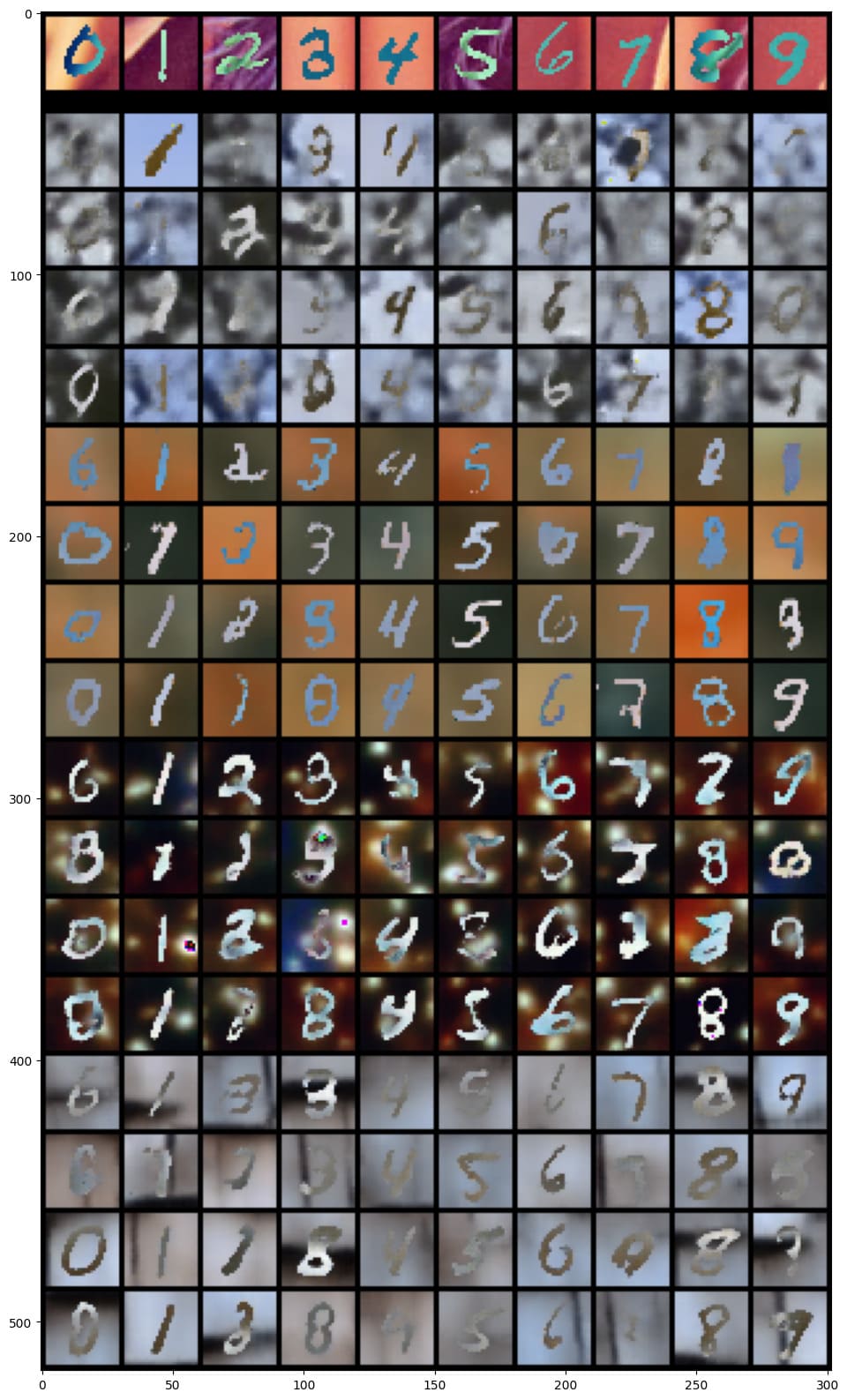}
         \caption*{\gls{MVTCAE}}
     
     \end{subfigure}
    \begin{subfigure}{0.20\textwidth}
         \centering
         \includegraphics[page=1,width=\linewidth]{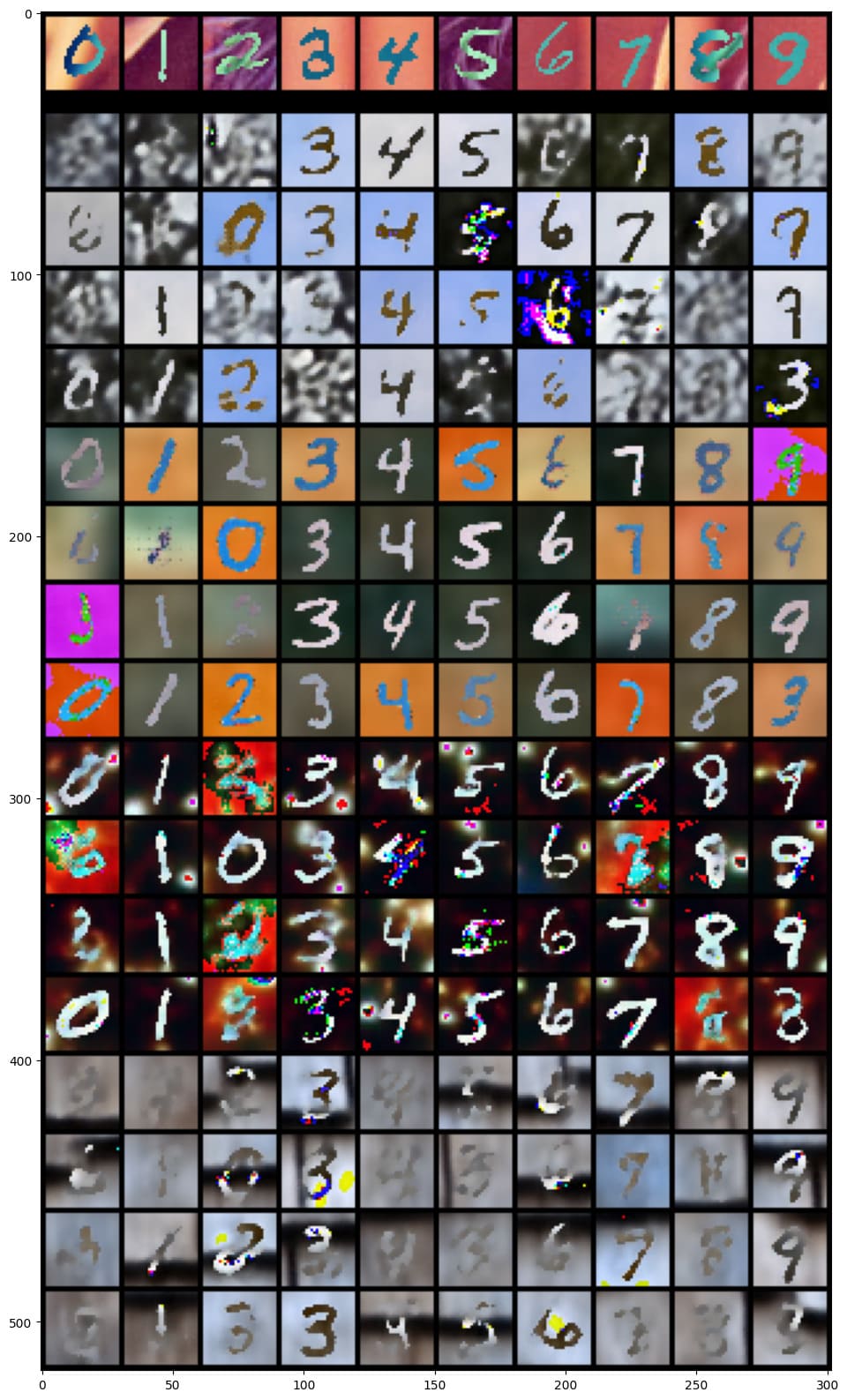}
         \caption*{\gls{MLD Inpaint} }
    
     \end{subfigure}
  \begin{subfigure}{0.20\textwidth}
         \centering
         \includegraphics[page=1,width=\linewidth]{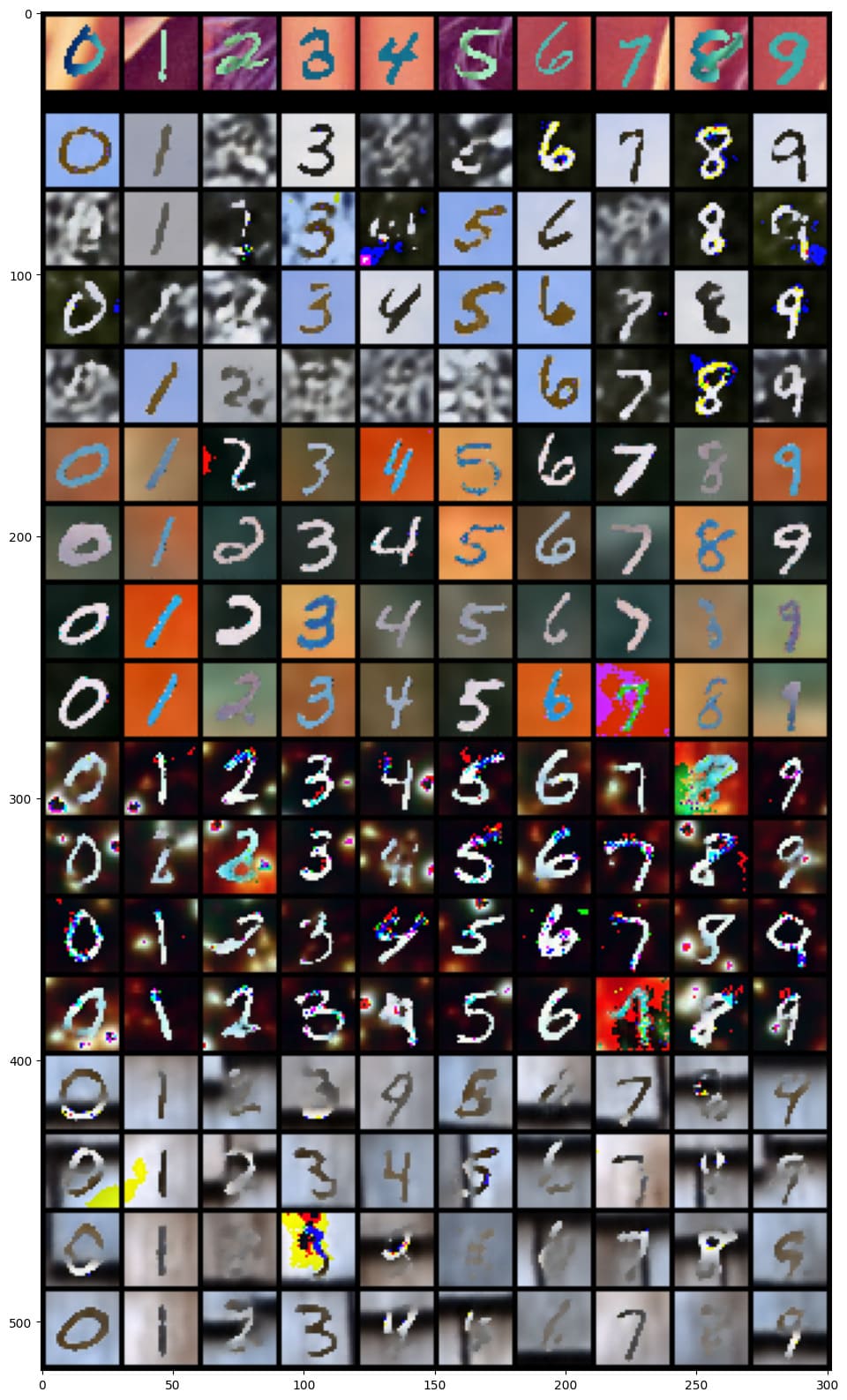}
         \caption*{\gls{MLD Uni} }
     
     \end{subfigure}
  \begin{subfigure}{0.20\textwidth}
         \centering
         \includegraphics[page=1,width=\linewidth]{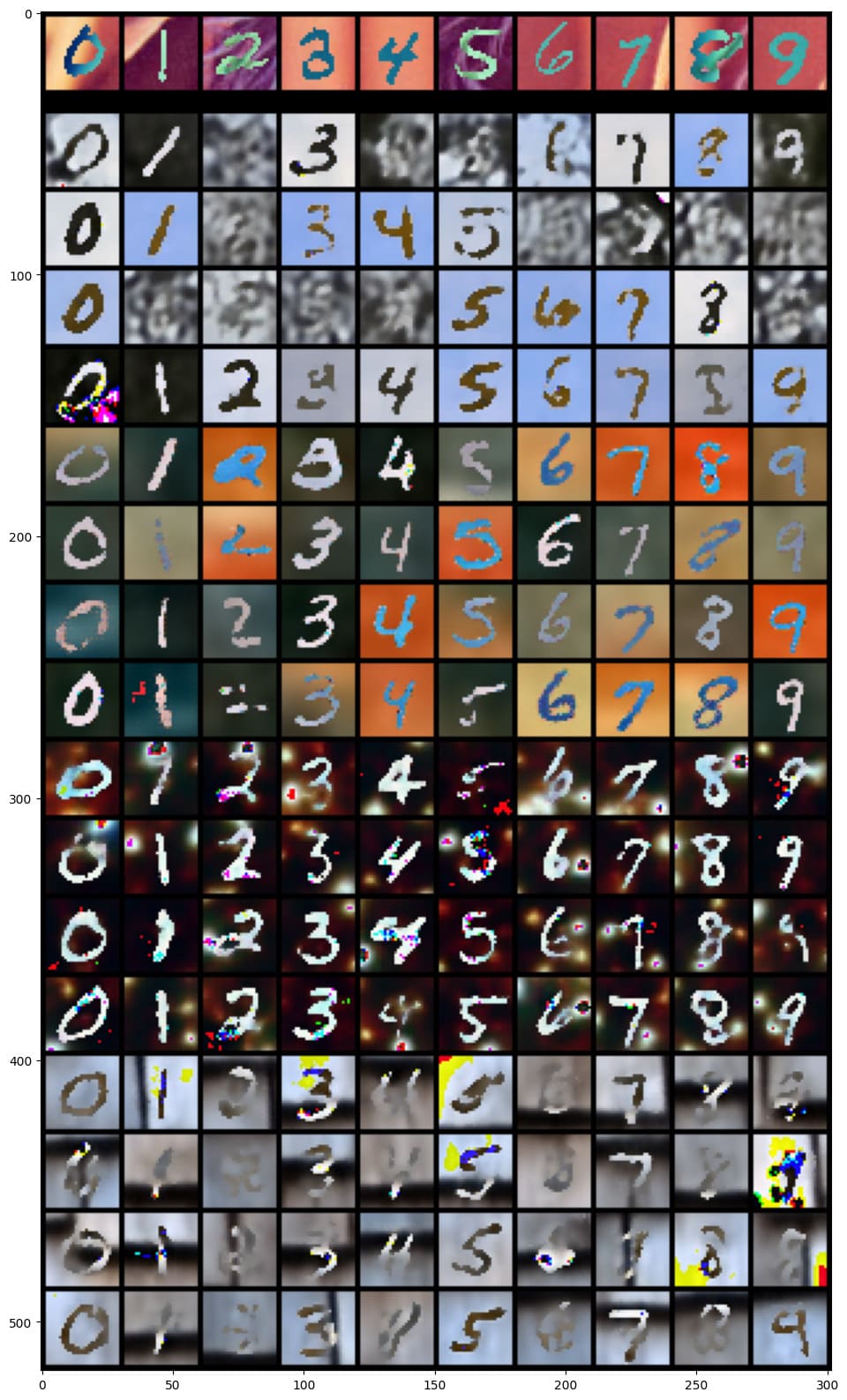}
         \caption*{ \textbf{\gls{MLD} (ours)} }

     \end{subfigure}
     
        \caption{Conditional generation qualitative results for \textbf{\polymnist} . The modality $X^2$ (dirst row) is used as the condition to generate the 4 remaining modalities(The rows below).  }
        \label{fig:cond_mmnist_2}
\end{figure}

\begin{figure}[h]
     \centering
       \centering
     \begin{subfigure}{0.2\textwidth}
         \centering
         \includegraphics[width=\linewidth]{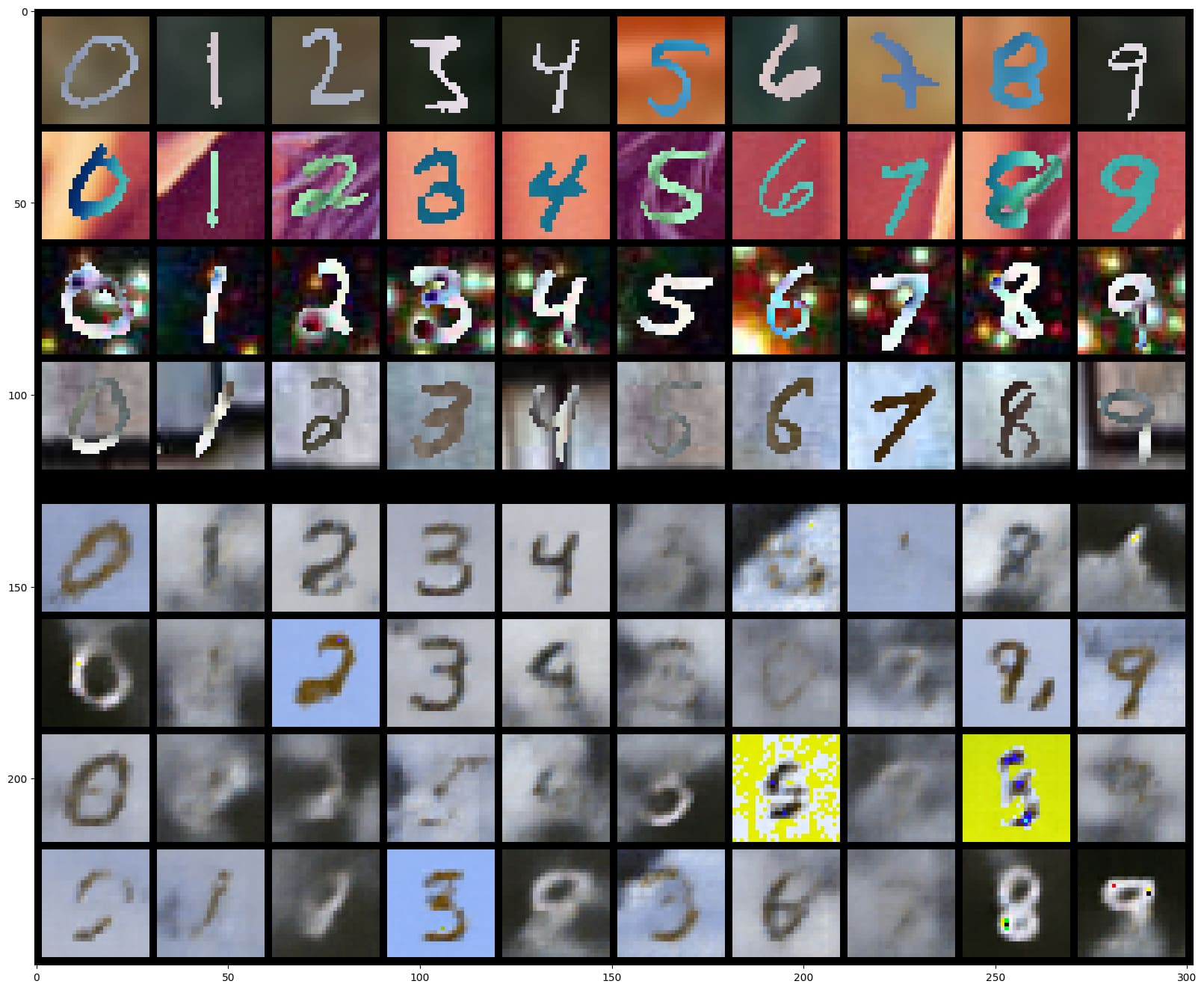}
         \caption*{\gls{MVAE}}

     \end{subfigure}
     \begin{subfigure}{0.2\textwidth}
         \centering
         \includegraphics[width=\linewidth]{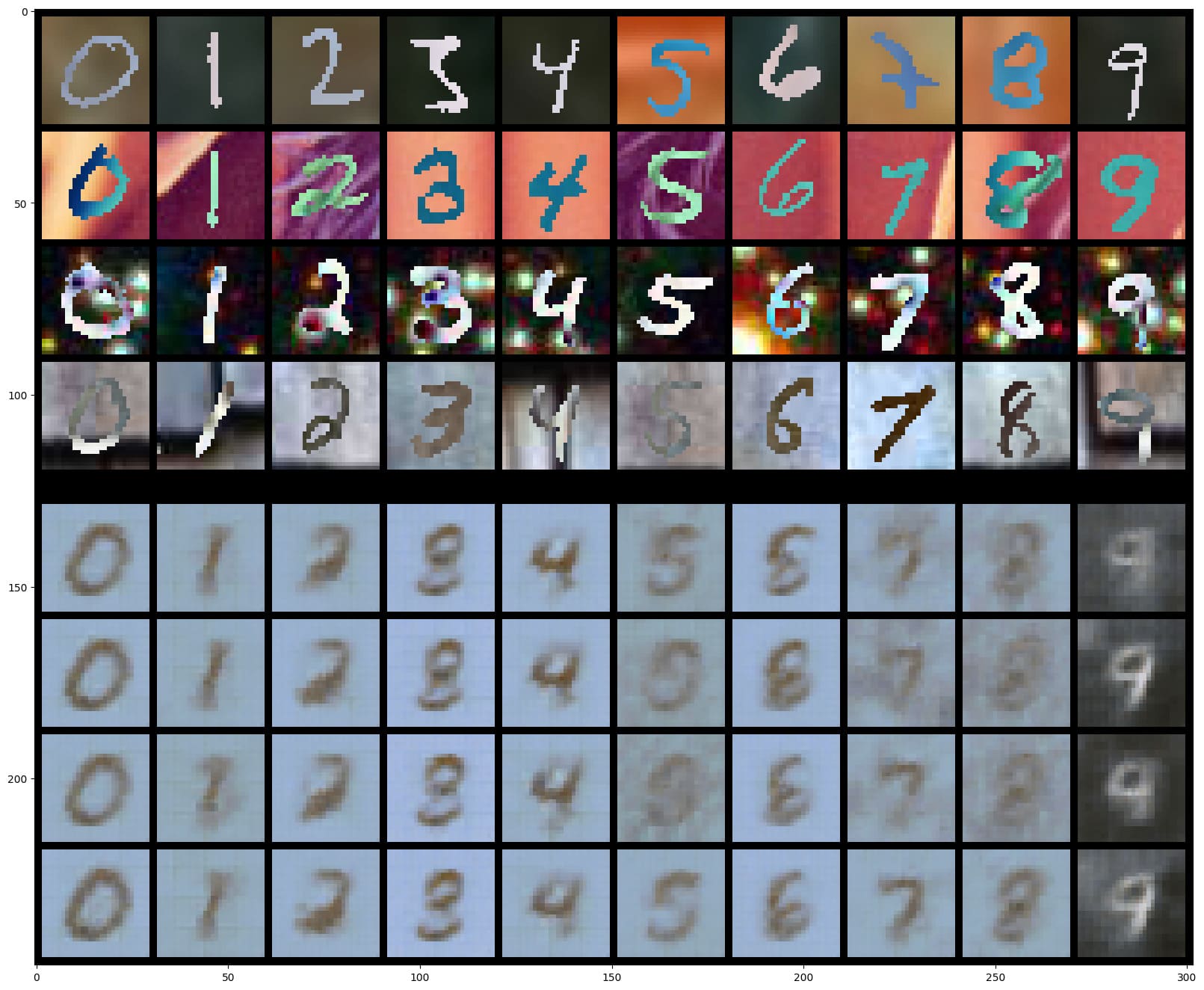}
         \caption*{\gls{MMVAE}}

     \end{subfigure}
       \begin{subfigure}{0.2\textwidth}
         \centering
         \includegraphics[width=\linewidth]{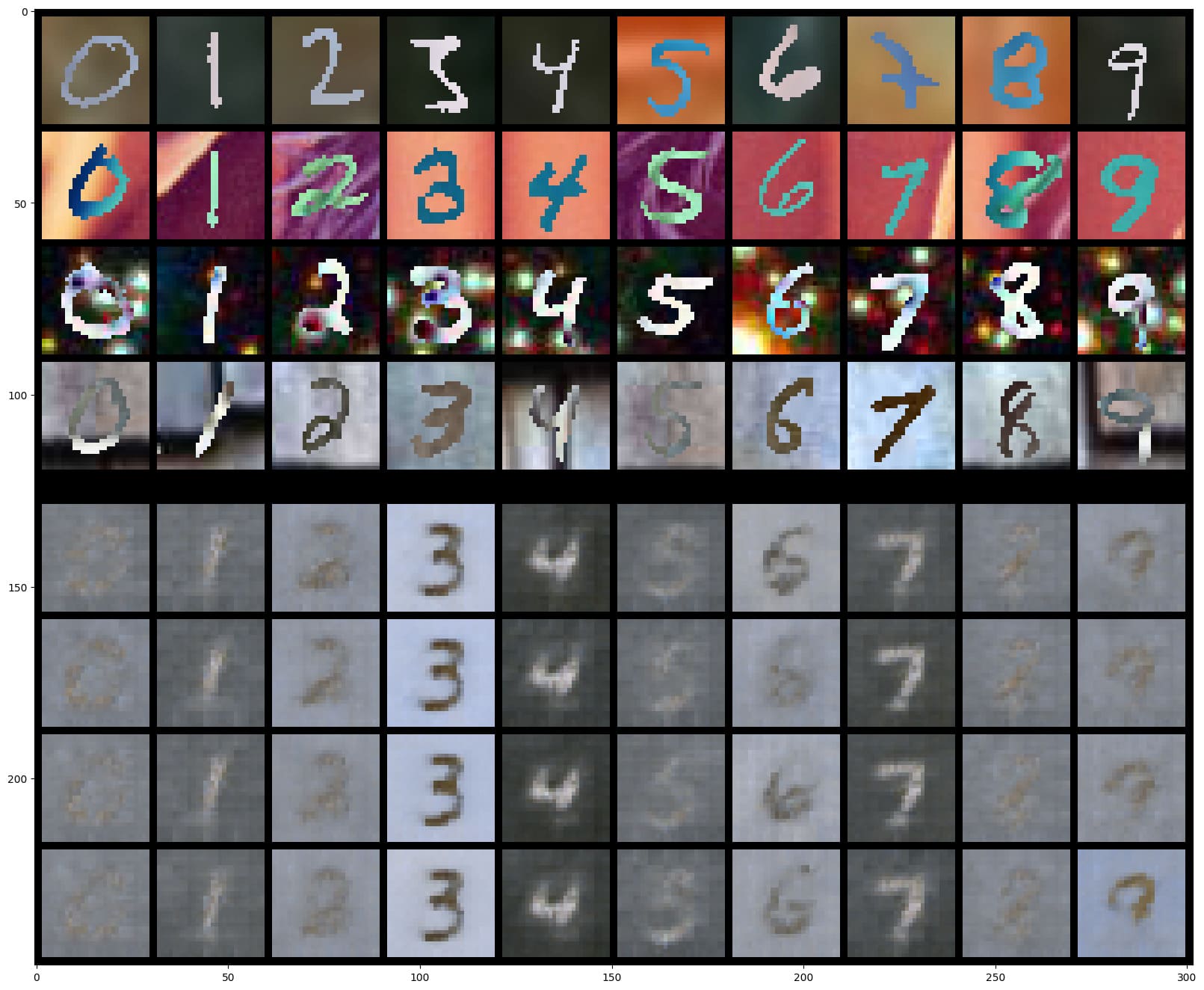}
         \caption*{\gls{MOPOE}}
    
     \end{subfigure}
      \begin{subfigure}{0.2\textwidth}
         \centering
         \includegraphics[width=\linewidth]{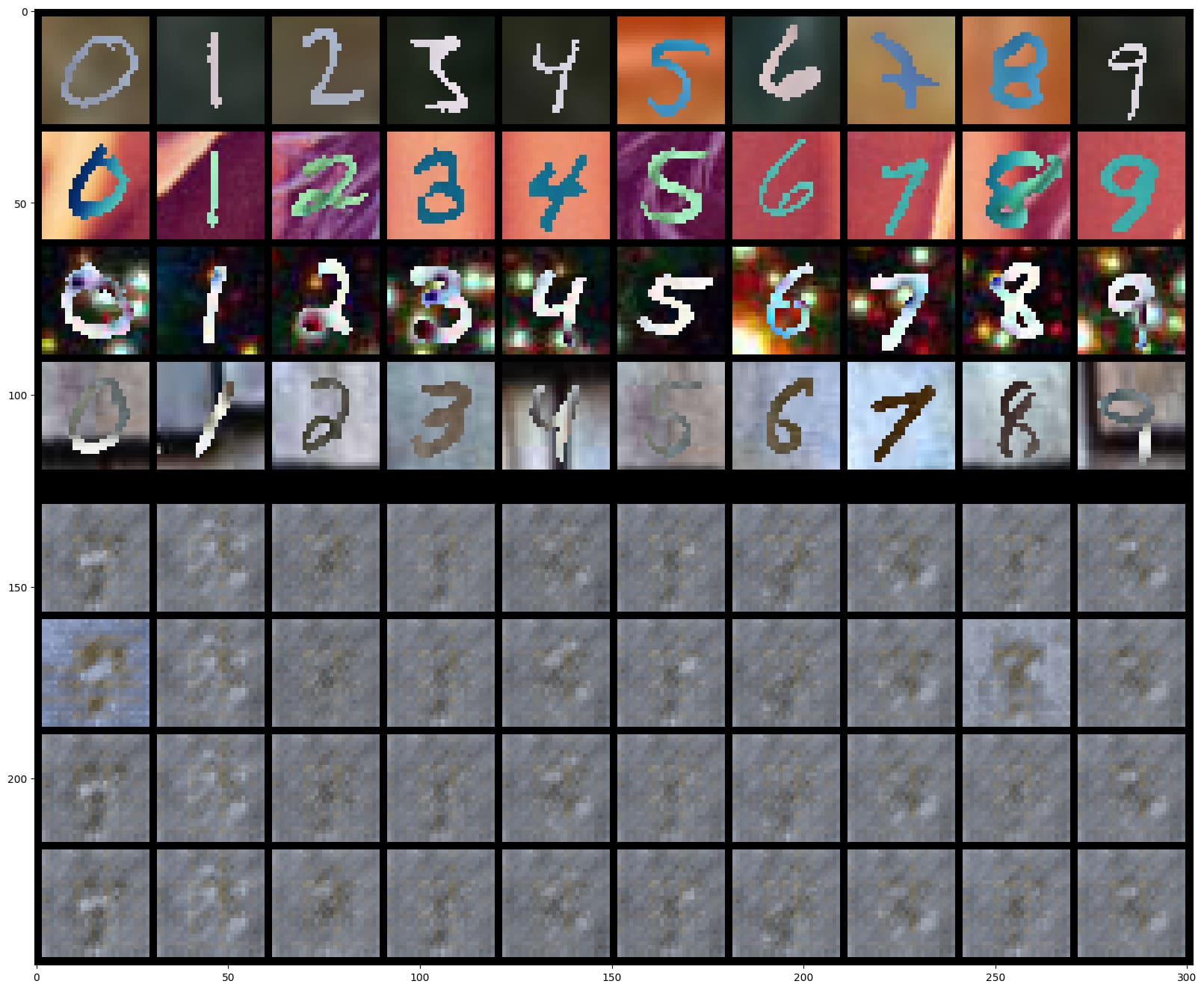}
         \caption*{\gls{NEXUS}}

     \end{subfigure}
     
  \begin{subfigure}{0.2\textwidth}
         \centering
         \includegraphics[width=\linewidth]{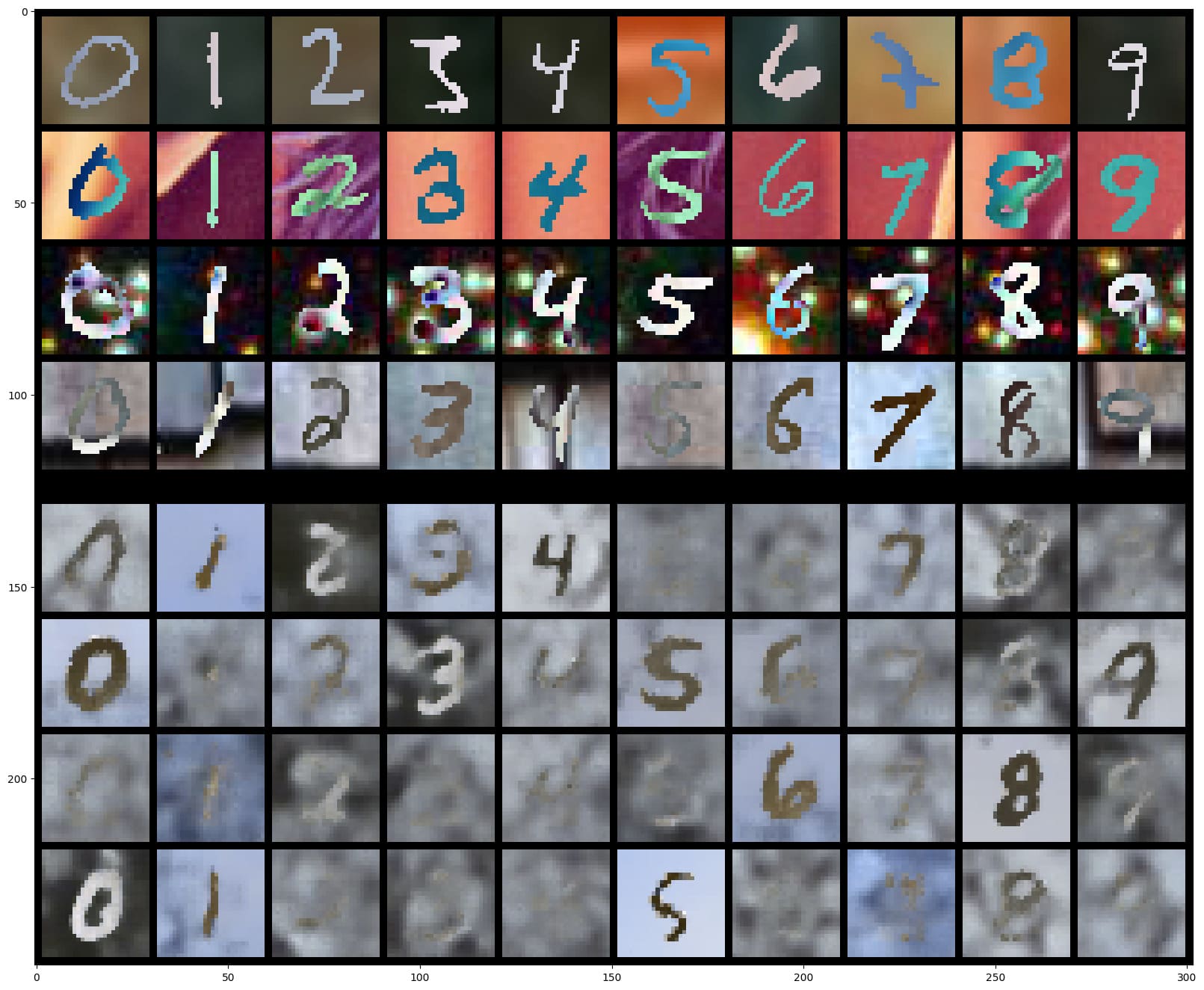}
         \caption*{\gls{MVTCAE}}
 
     \end{subfigure}
    \begin{subfigure}{0.2\textwidth}
         \centering
         \includegraphics[page=1,width=\linewidth]{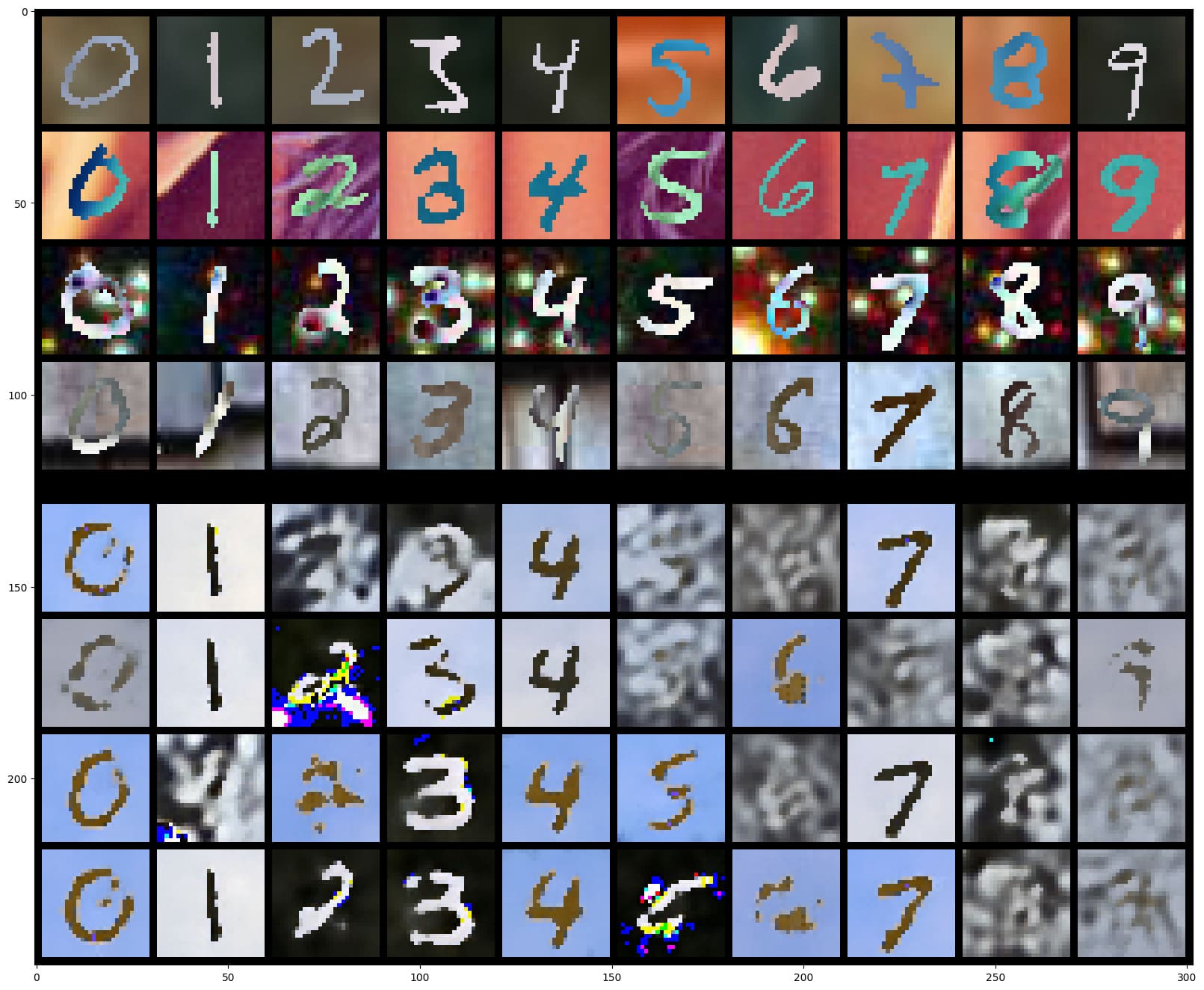}
         \caption*{\gls{MLD Inpaint}  }
     \end{subfigure}
  \begin{subfigure}{0.2\textwidth}
         \centering
         \includegraphics[page=1,width=\linewidth]{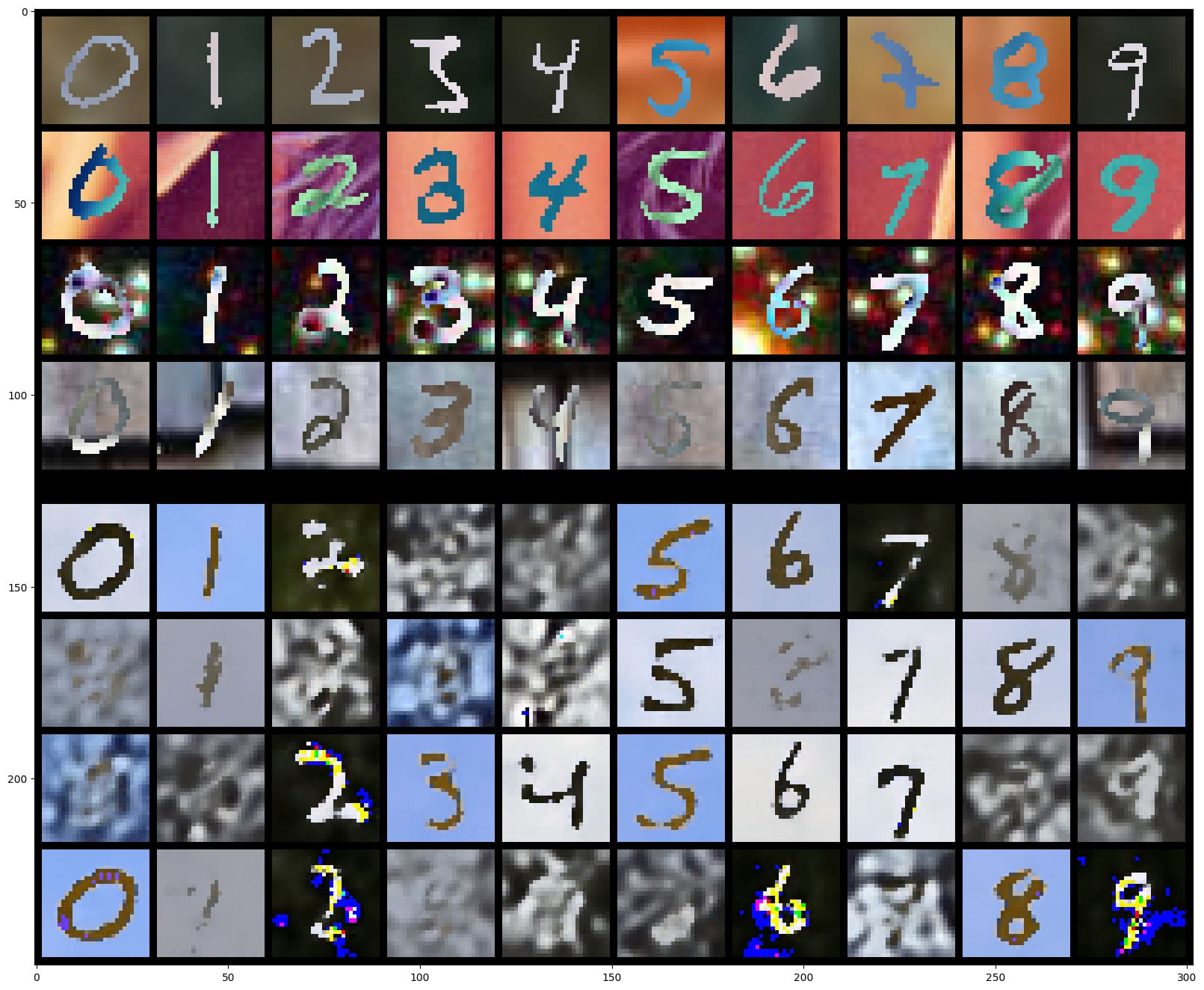}
         \caption*{\gls{MLD Uni}}
     \end{subfigure}
  \begin{subfigure}{0.2\textwidth}
         \centering
         \includegraphics[page=1,width=\linewidth]{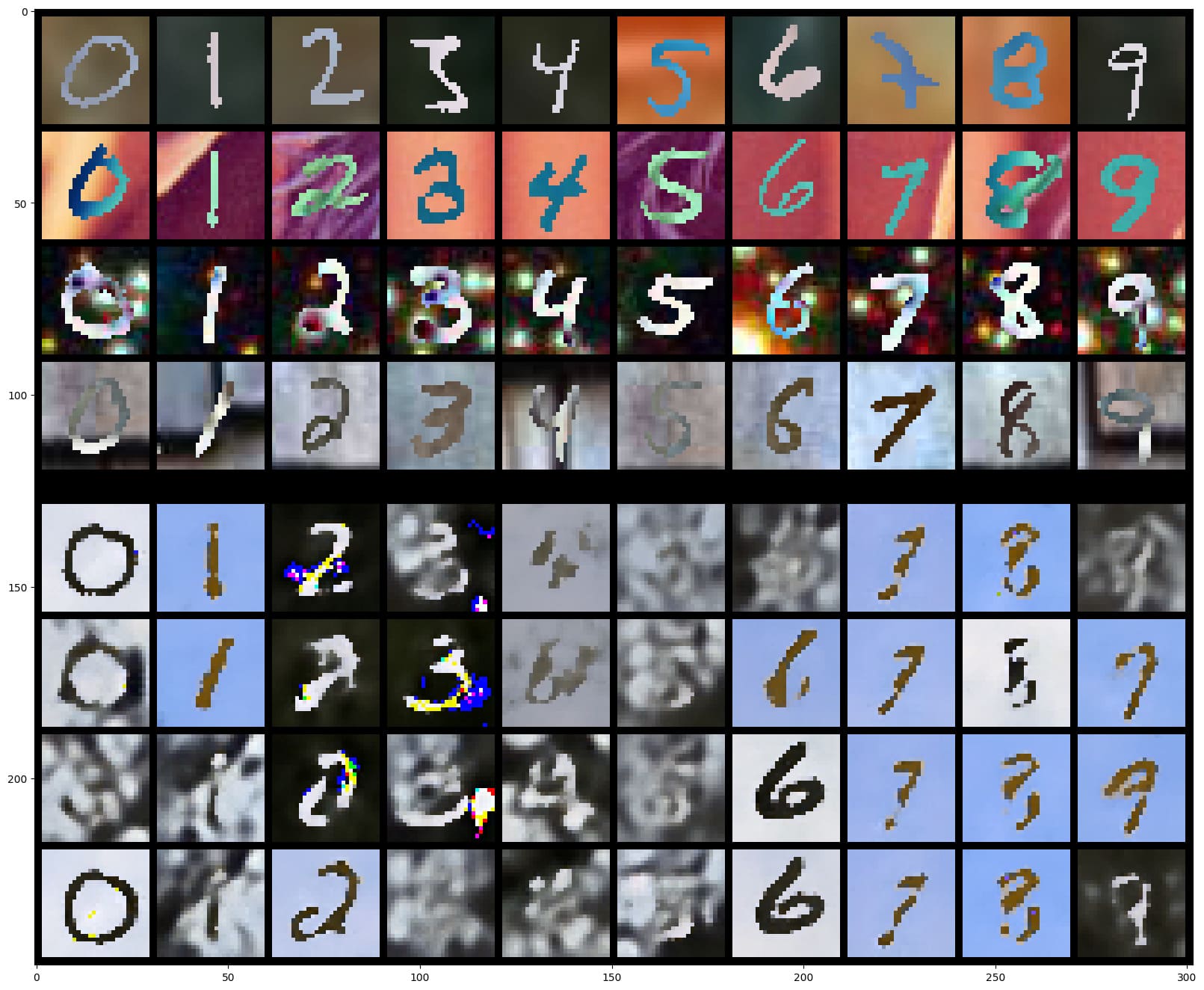}
         \caption*{\textbf{\gls{MLD} (ours)}}
     \end{subfigure}
        \caption{Conditional generation qualitative results for \textbf{\polymnist}. The subset of modalities $X^1 ,X^2,X^3,X^4 $ (First 4 rows) are used as the condition to generate the modality $X^0$ (The rows below).  }
        \label{fig:cond_mmnist_1234}
\end{figure}
\newpage

\subsubsection{Additional experiments with \cite{palumbo2023mmvae} architecture}

\begin{figure} [H]  
\centering
\begin{subfigure}{0.30\textwidth}
    \includegraphics[width=\linewidth]{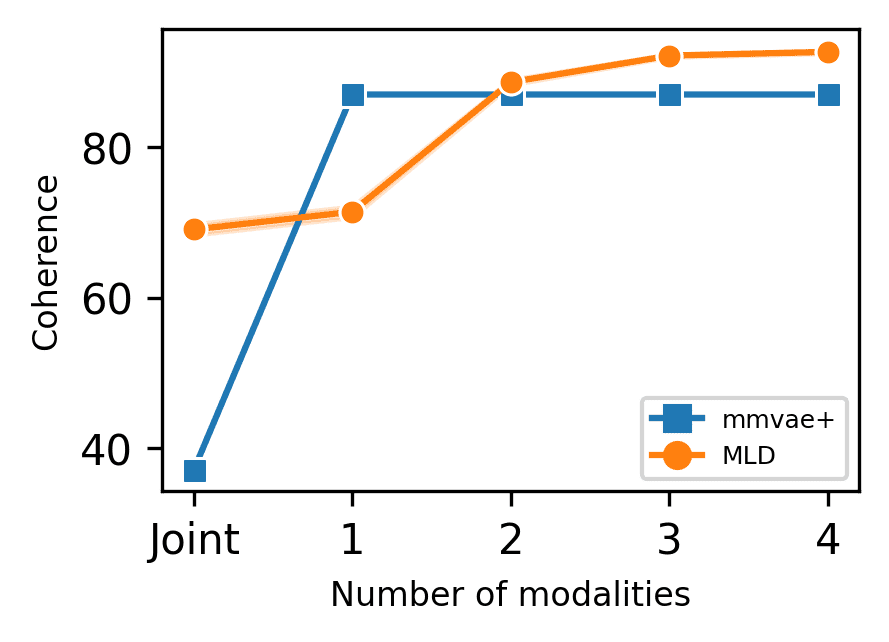}
\end{subfigure}
\begin{subfigure}{0.30\textwidth}
    \includegraphics[width=\linewidth]{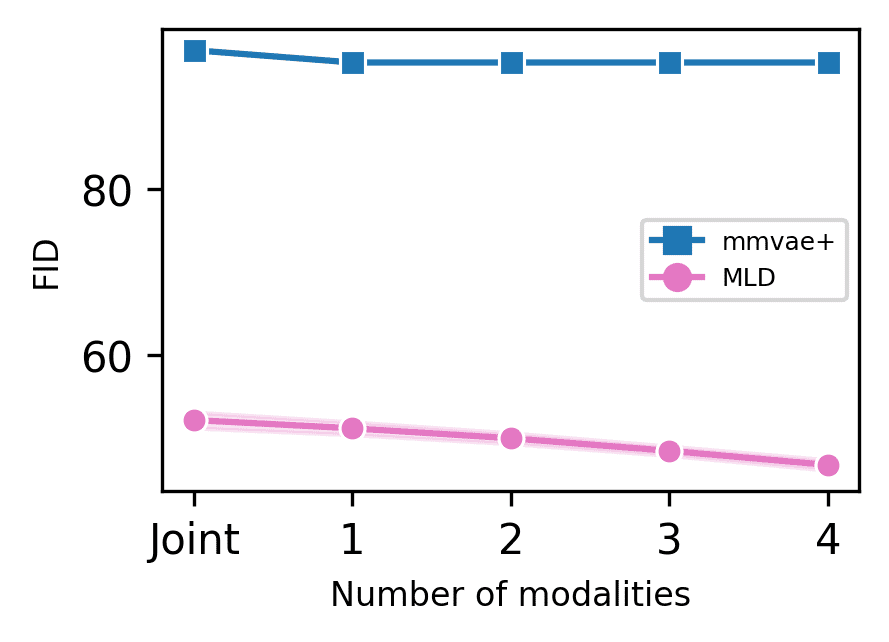}
\end{subfigure}

\label{fig:mmvaeresnet}  
\caption{Results for POLYMNIST data-set. Left: a comparison of the generative coherence (
 $ \uparrow $ ) and quality in terms of FID ($ \downarrow $) as a function of the number of inputs.}
\end{figure}

In our experiments on \polymnist, we used the same architecture as in \cite{sutter2021generalized} \cite{tcmvaehwang21} to ensure a fair settings for all the baselines. In \cite{palumbo2023mmvae}, the experiments on \polymnist are conducted using a different autoencoder architecture which is based on Resnets instead of a a sequence of convolutions layers based autoencoder. We investigate in this section, the performance of \gls{MMVAEplus} and our \gls{MLD} using this architecture. For \gls{MMVAEplus}, we keep the same settings as in \cite{palumbo2023mmvae} including the autoencoder architecture, latent size, and importance sampling K=10 with doubly reparameterized gradient estimator (DReG). For \gls{MLD}, we use the same autoencoder architecture with latent size equal to 160. In \Cref{fig:mmvaeresnet}, we observe that while the new architecture autoencoder enhance the \gls{MMVAEplus} performance, our \gls{MLD} performance is improved as well. Similarly to previous results, \gls{MLD} achieves simultaneously the best generative coherence and quality.

\subsection{CUB}

\begin{table}[H]
\tiny
\centering
\begin{tabular}{c|c|cc|cc}
\toprule
\multirow{2}{*}{ Models }  & \multicolumn{3}{c}{ Coherence ( $ \uparrow $ ) }  &  \multicolumn{2}{c}{Quality ( $ \downarrow $ ) }  \\
    \cmidrule{2-6}
    &  Joint &Image $\rightarrow$ Caption &  Caption $\rightarrow$ Image & Joint $\rightarrow$ Image & Caption $\rightarrow$ Image \\
    \midrule
    
    \gls{MVAE} & $0.66$&$\textbf{0.70}$ &$0.64$ & 
    $158.91$ &  $158.88$
\\
  \gls{MMVAE} & $0.66$&$0.69$ &$0.62$ & 
    $277.8$ &  $212.57$   \\
    \gls{MOPOE} & $0.64$&$0.68$ &$0.55$ & 
    $279.78$ &  $179.04$ \\
    
   \gls{NEXUS} & $0.65$&$0.69$ &$0.59$ & 
    $147.96$ &  $262.9$ \\

   \gls{MVTCAE} &$0.65$&$\textbf{0.70}$ &$0.65$ & 
    $155.75$ &  $168.17$ \\
     \gls{MMVAEplus} &$0.61$&$0.68$ &$0.65$ & 
    $188.63$ &  $247.44$
    \\

     \gls{MMVAEplus}(K=10) &  $0.63$&$0.68$ &$0.62$ & 
    $172.21$ &  $178.88$  
   \\
    \midrule
   \gls{MLD Inpaint} & $\textbf{0.69}$&${0.69}$ &${0.68}$ & 
    ${69.16}$ &  ${68.33}$
    \\
  \gls{MLD Uni}  &$ \textbf{0.69}$&${0.69}$ &$\textbf{0.69}$ & 
    $64.09$ &  $\textbf{61.92}$ \\

    \gls{MLD}  &$ \textbf{0.69}$&$0.69$ &$\textbf{0.69}$ & 
    $\textbf{63.47}$ &  $62.62$ \\
 \midrule
 \gls{MLD}*  &$ \textbf{0.70}$&
 $0.69$ &
 $\textbf{0.69}$ & 
    $\textbf{22.19}$ & 
    $\textbf{22.50}$ \\
 
    \bottomrule

\end{tabular}
\caption{Generation Coherence (\gls{CLIP-S}  : Higher is better ) and Quality (\gls{FID} $\downarrow$ Lower is better ) for CUB dataset. \textbf{\gls{MLD}*} denotes the version of our method using a more powerful image autoencoder.}
\label{coh_qua:cub}

 \end{table}

\begin{figure}[H]
     \centering
     \begin{subfigure}{0.45\textwidth}
         \centering
         \includegraphics[width=\linewidth]{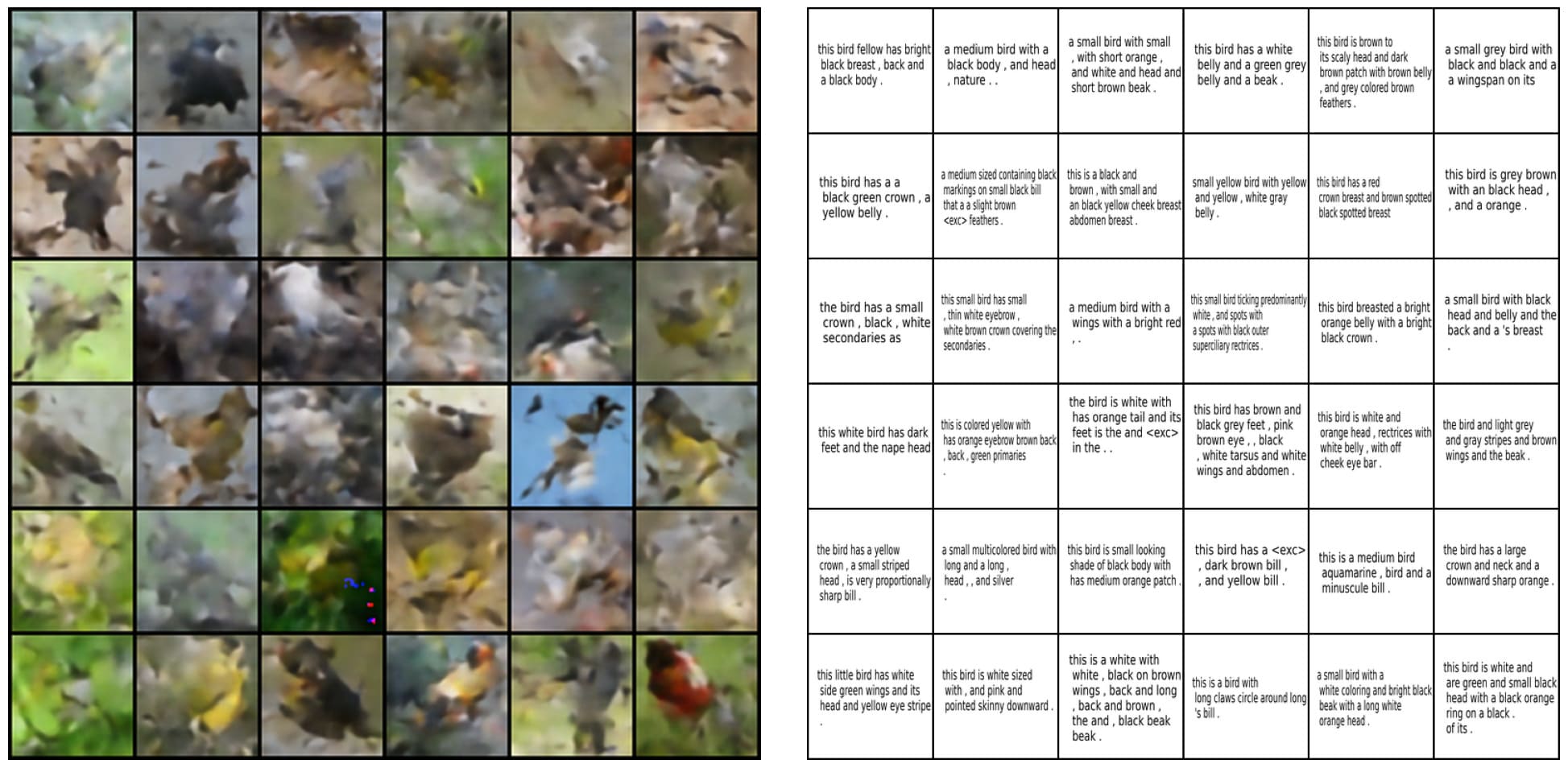}
        \caption*{\gls{MVAE}}
     \end{subfigure}
     \begin{subfigure}{0.45\textwidth}
         \centering
         \includegraphics[width=\linewidth]{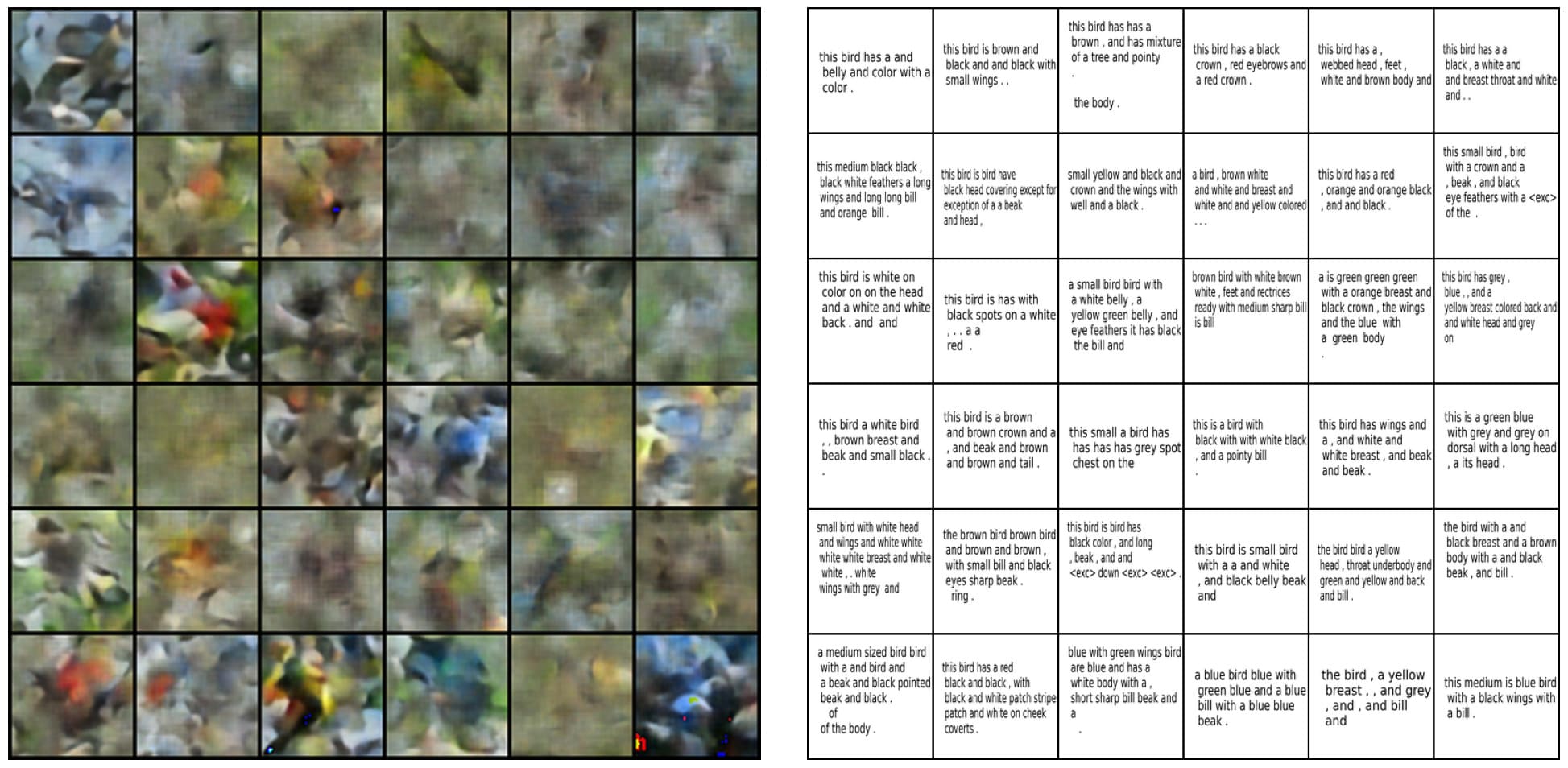}
         \caption*{\gls{MMVAE}}
     \end{subfigure}
     
       \begin{subfigure}{0.45\textwidth}
         \centering
         \includegraphics[width=\linewidth]{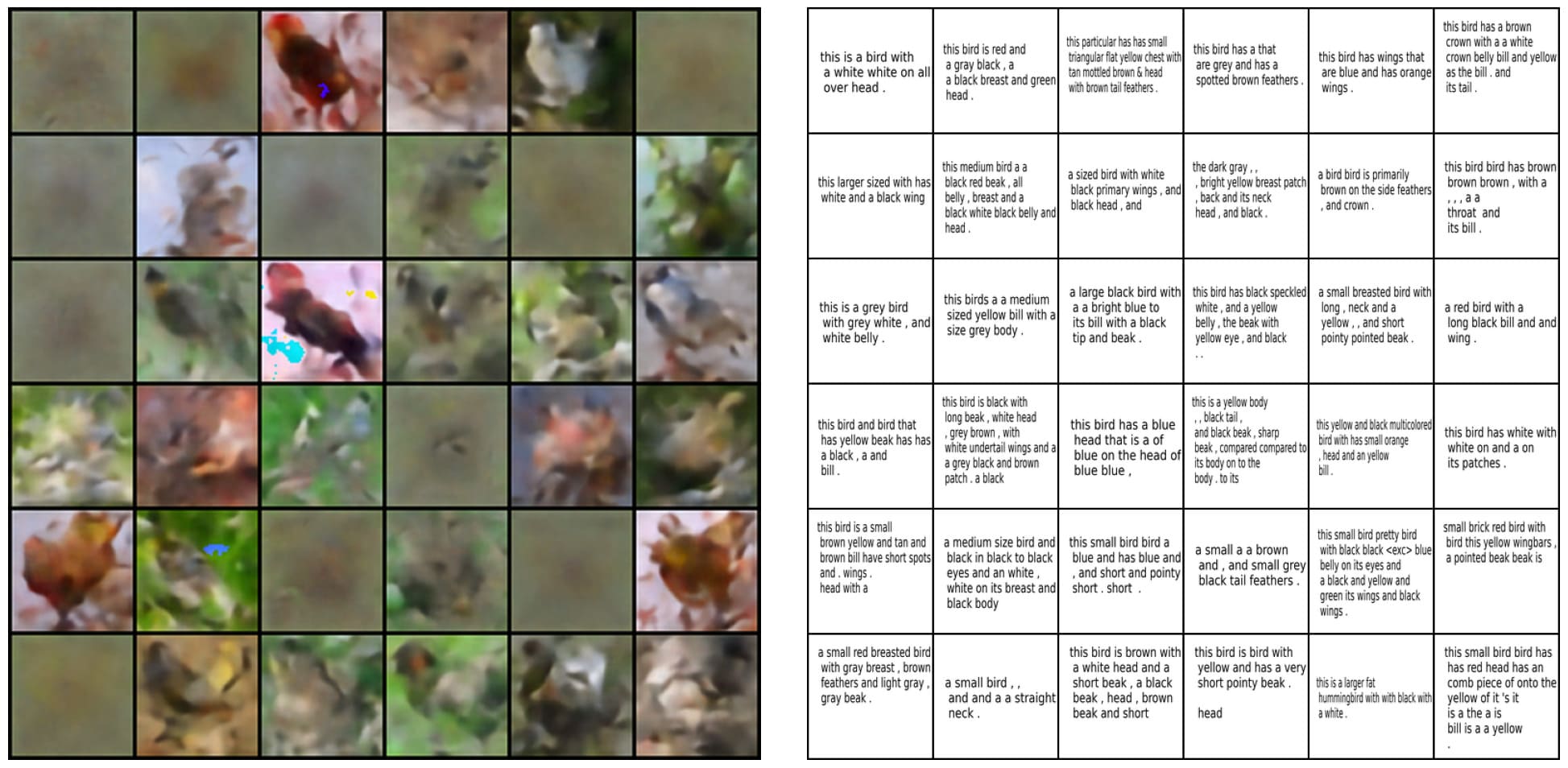}
         \caption*{\gls{MOPOE}}
     \end{subfigure}
      \begin{subfigure}{0.45\textwidth}
         \centering
         \includegraphics[width=\linewidth]{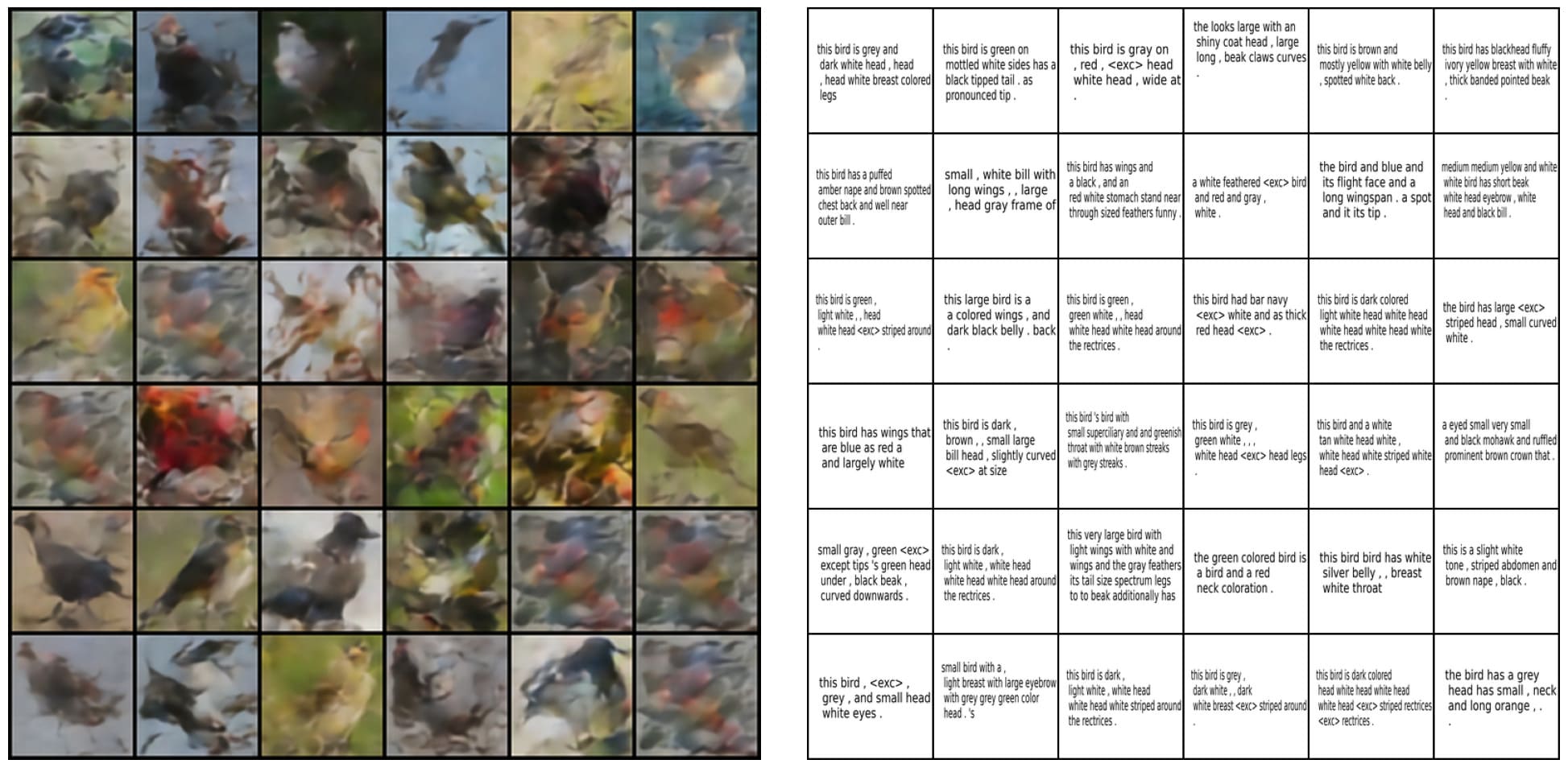}
         \caption*{\gls{NEXUS}}
     \end{subfigure}
     
     \begin{subfigure}{0.45\textwidth}
         \centering
         \includegraphics[width=\linewidth]{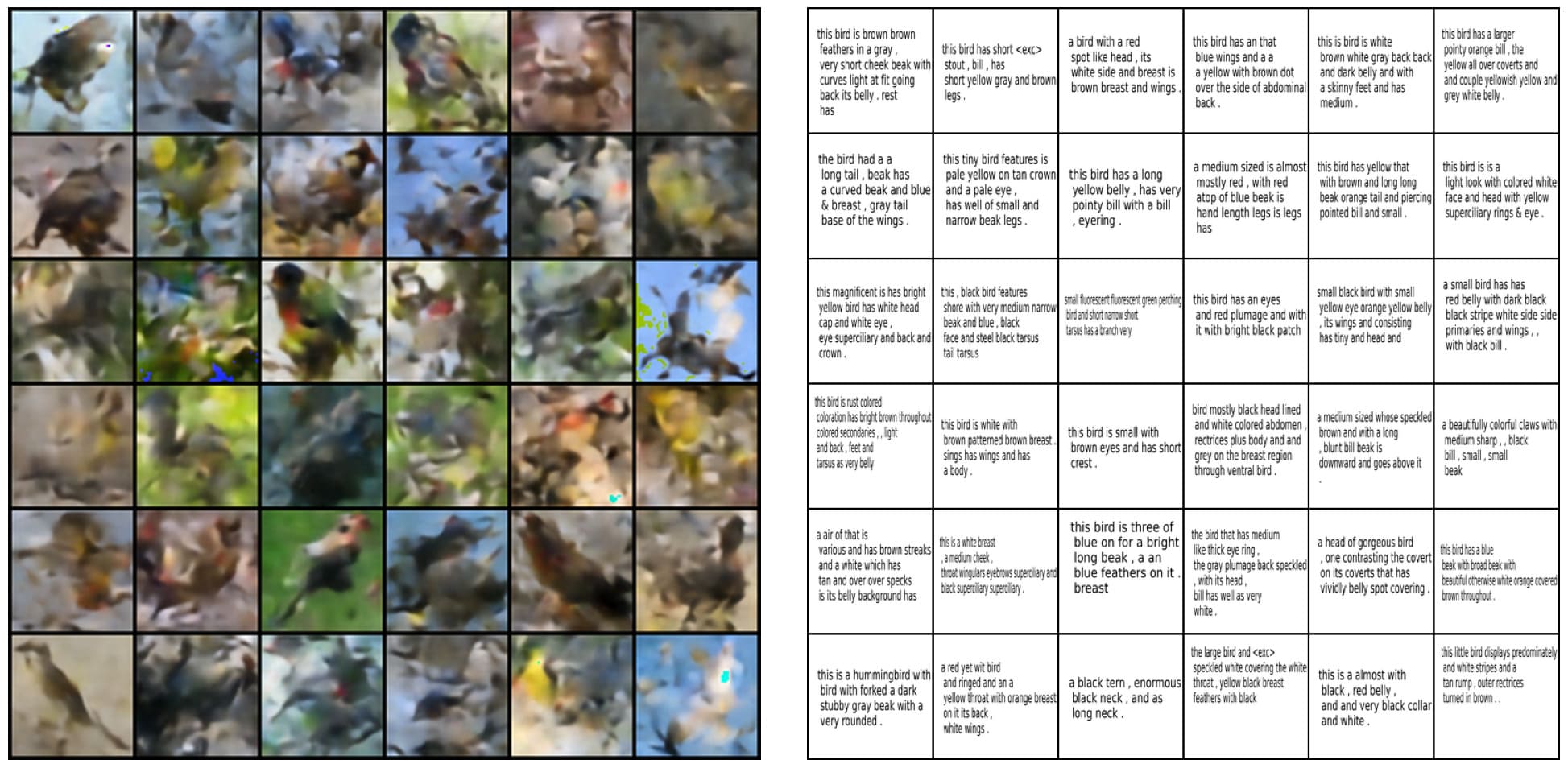}
         \caption*{\gls{MVTCAE}}
     \end{subfigure}
 \begin{subfigure}{0.45\textwidth}
         \centering
        \includegraphics[page=1,width=\linewidth]{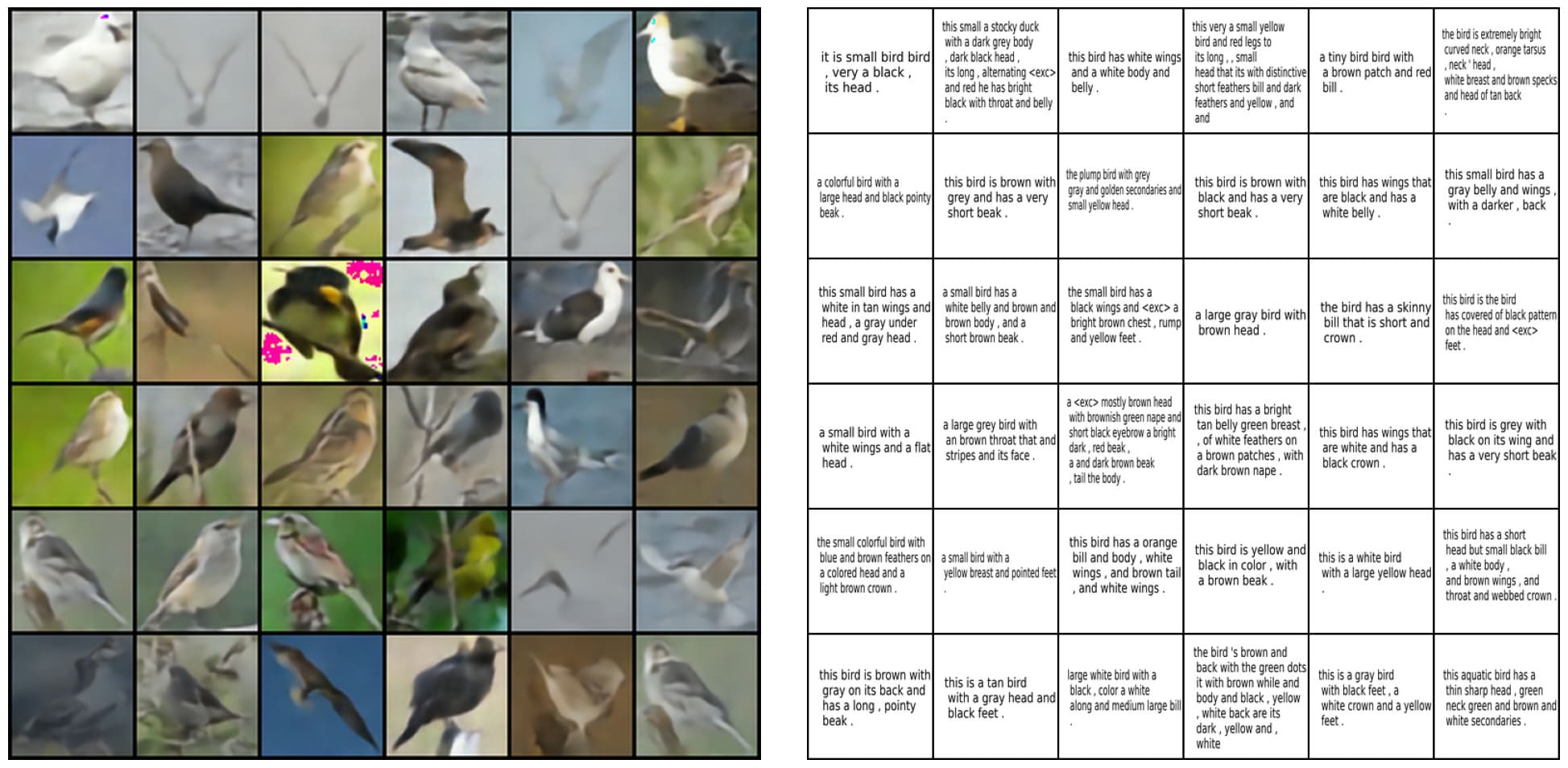}
         \caption*{\gls{MLD Inpaint}  }
\end{subfigure}

 \begin{subfigure}{0.45\textwidth}
         \centering
         \includegraphics[page=1,width=\linewidth]{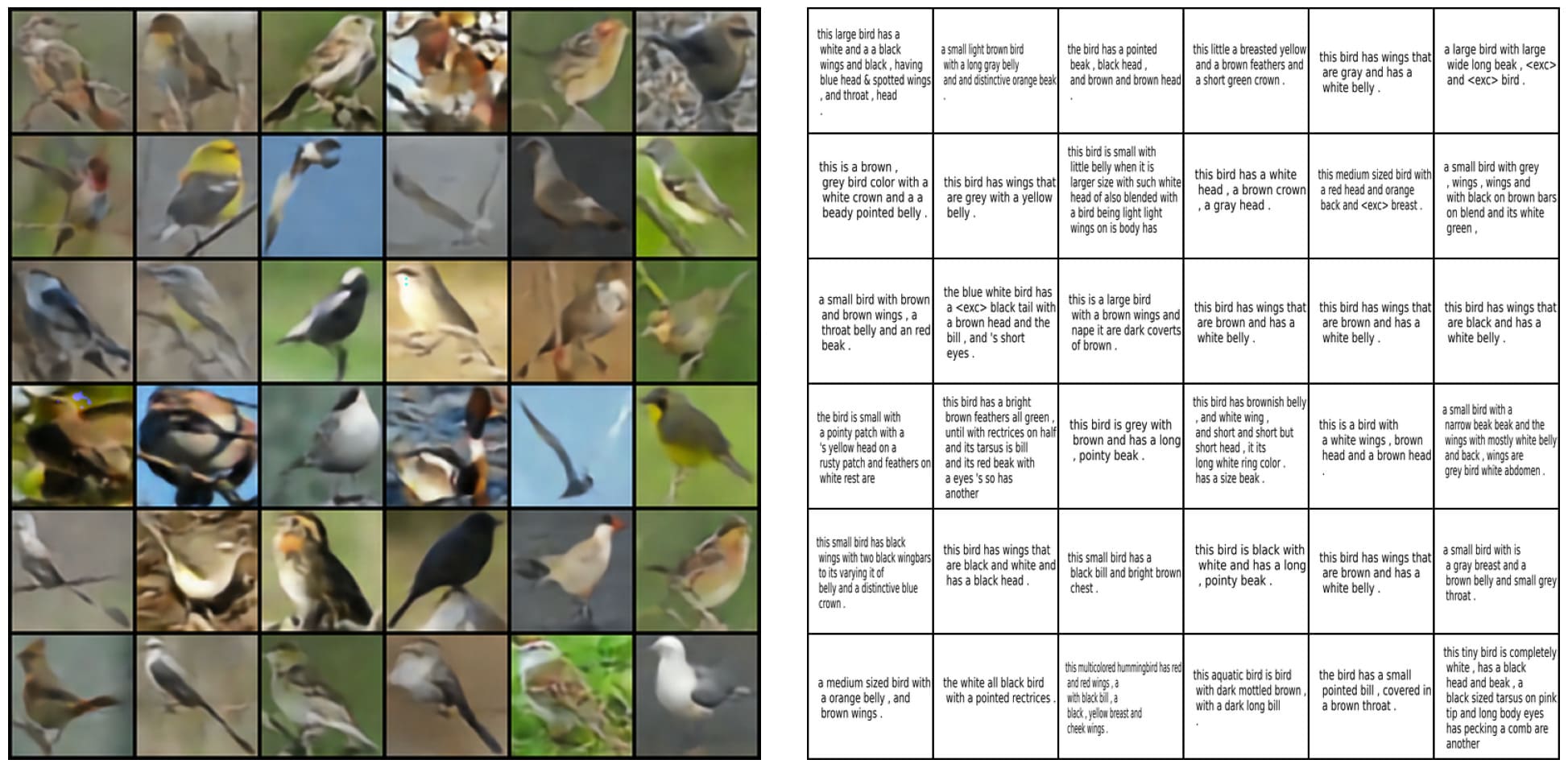}
         \caption*{\gls{MLD Uni}  }
\end{subfigure}
\begin{subfigure}{0.45\textwidth}
    \centering
    \includegraphics[page=1,width=\linewidth]{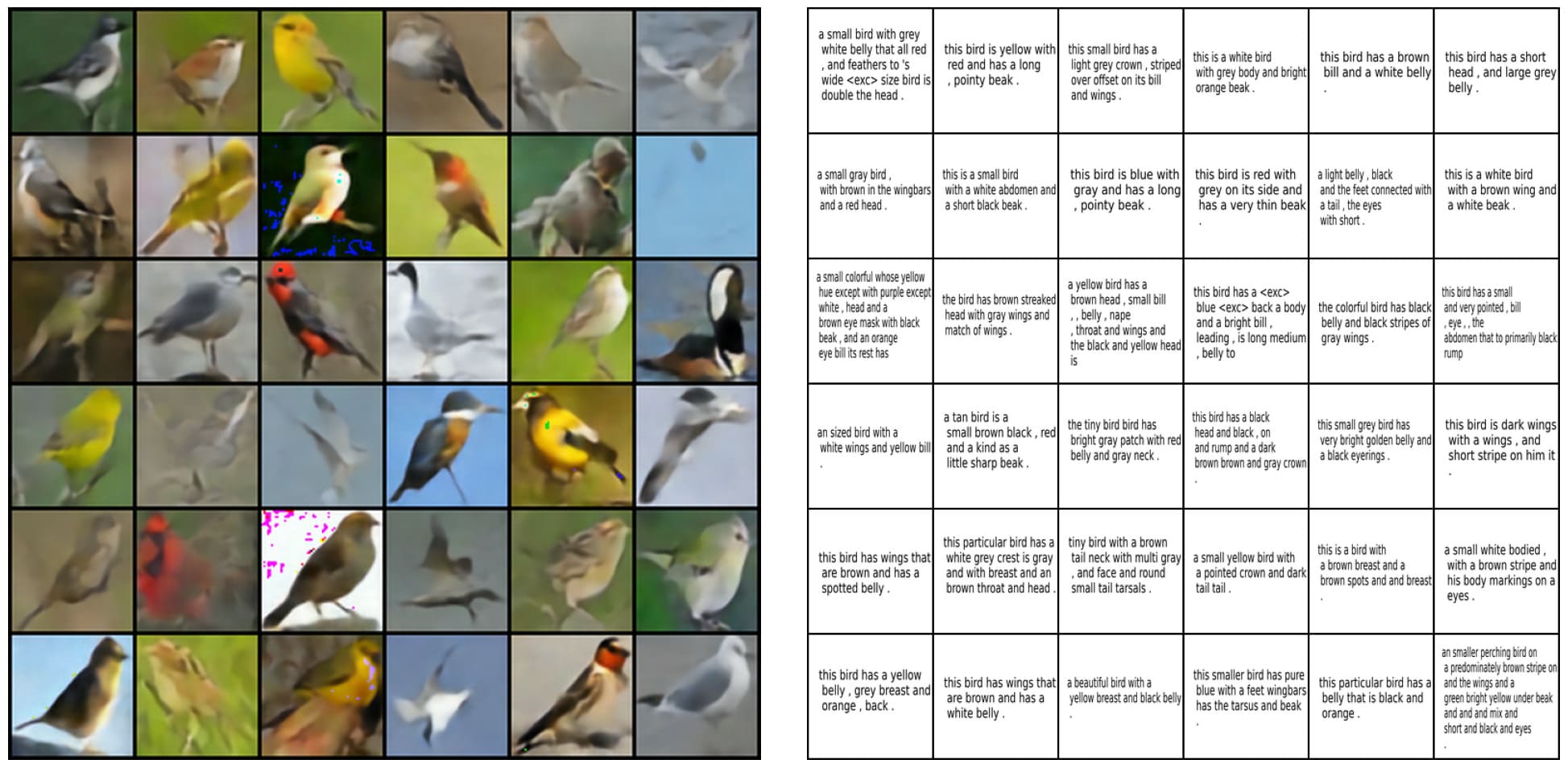}
    \caption*{\gls{MLD} (ours) }
\end{subfigure}
\caption{Qualitative results for joint generation on \textbf{\cub}.}
\label{cap_joint_detailed}
        
\end{figure}

\begin{figure}[H]
     \centering
     
     \begin{subfigure}{0.25\textwidth}
         \centering
         \includegraphics[width=\linewidth]{figures/datasets/cub/new_64/MLD_cond_text_image_mld64_new.jpg}
         \caption{ Conditional generation.}
    \end{subfigure}
  \begin{subfigure}{0.52\textwidth}
         \centering
         \includegraphics[width=\linewidth]{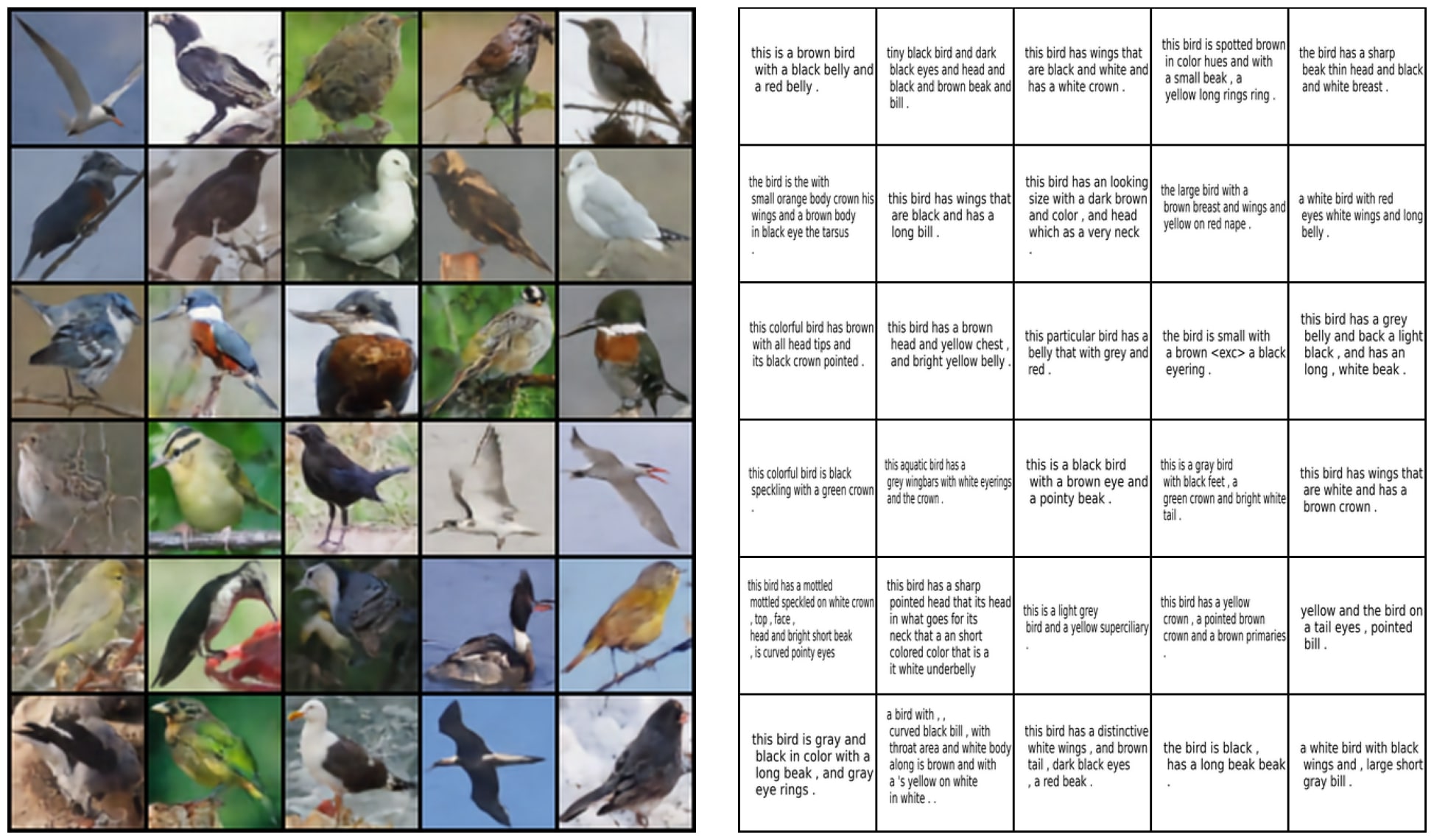}
            \caption{Joint generation.}
         
\end{subfigure}
        \caption{Qualitative results of \textbf{\gls{MLD}*} on \textbf{CUB} data-set with powerful image autoencoder.}
    
\end{figure}

\subsection{CelebAMask-HQ}

In this section, we present additional experiments on the CelebAMask-HQ dataset~\citep{Lee2019MaskGANTD}, which consists of face images, each having a segmentation mask, and text attributes, so 3 modalities. We follow the same experimentation protocol as in \citep{wesego2023scorebased} including the autoencoder base architecture. Note that for \gls{MLD} we use deterministic autoencoders instead of variational autoencoders~\citep{Lee2019MaskGANTD}.
Similarly, the CelebAMask-HQ dataset is restricted to take into account 18 out of 40 attributes from the original dataset and the images are resized to 128$\times$128 resolution, as done in~\citep{mvaeWu2018,wesego2023scorebased}. Please refer to \Cref{apdx:implementation}, for additional implementation details of \gls{MLD}.

The Image generation quality is evaluated in terms of \gls{FID} score. The attributes and the mask having binary values, are evaluated in terms of $F1$ Score against the ground truth. The competitors performance results are reported from~\cite{wesego2023scorebased}. 

The quantitative results in \Cref{coh_qua:celebA} show that \gls{MLD} outperforms the competitors on the generation quality. It achieves the best F1 score in the attributes generation given Image and Mask modalities. The mask generation best performance is achieved by \gls{MOPOE}. \gls{MLD} achieves the second best performance on the mask generation conditioning on both image and attributes modalities. Overall, \gls{MLD} stands out with the best image quality generation while being on-par with competitors in the mask and attribute generation coherence.

\begin{figure}[H]
     \centering
     
     \begin{subfigure}{0.49\textwidth}
         \centering
         \includegraphics[width=\linewidth]{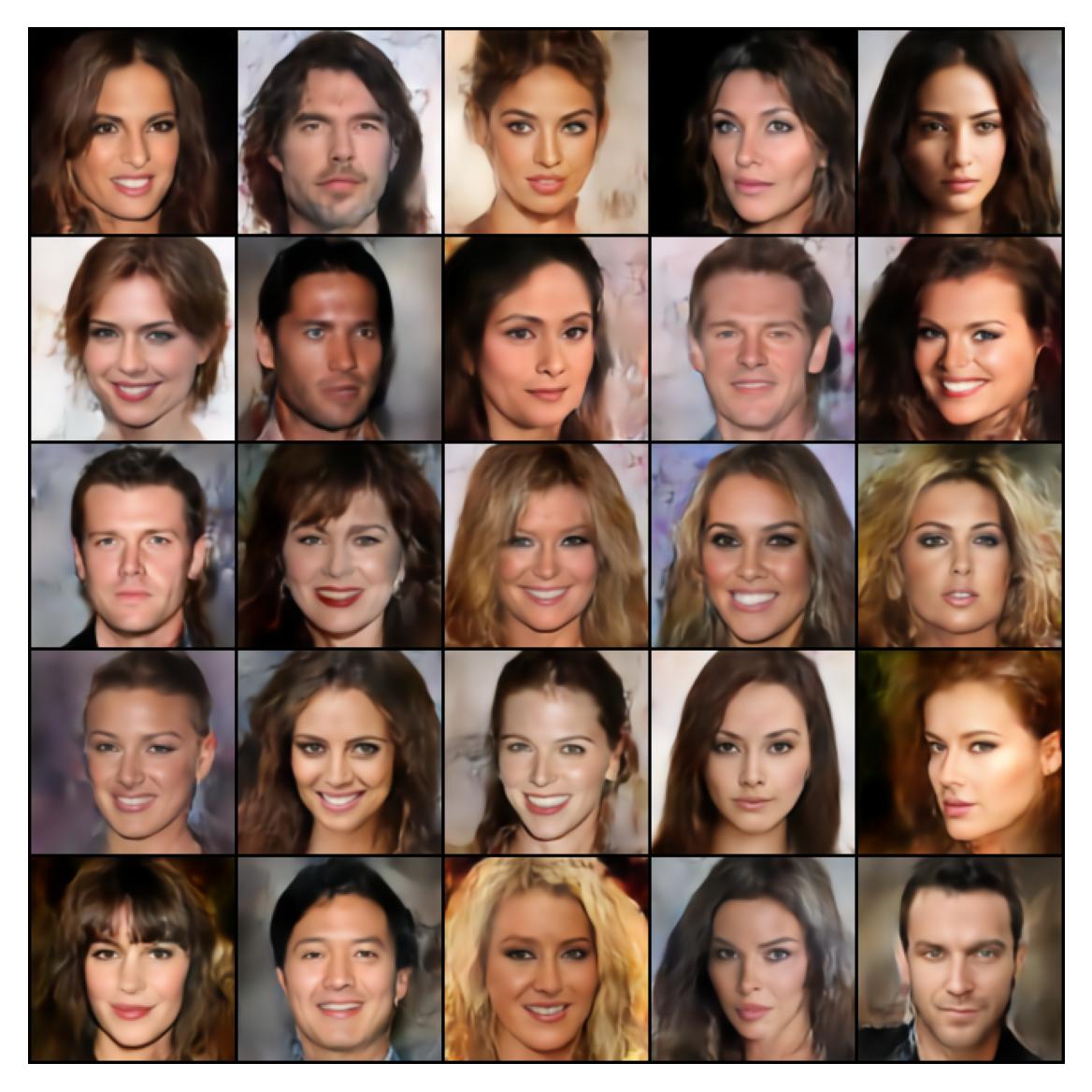}
         \caption{Image}
    \end{subfigure}
  \begin{subfigure}{0.49\textwidth}
         \centering
         \includegraphics[width=\linewidth]{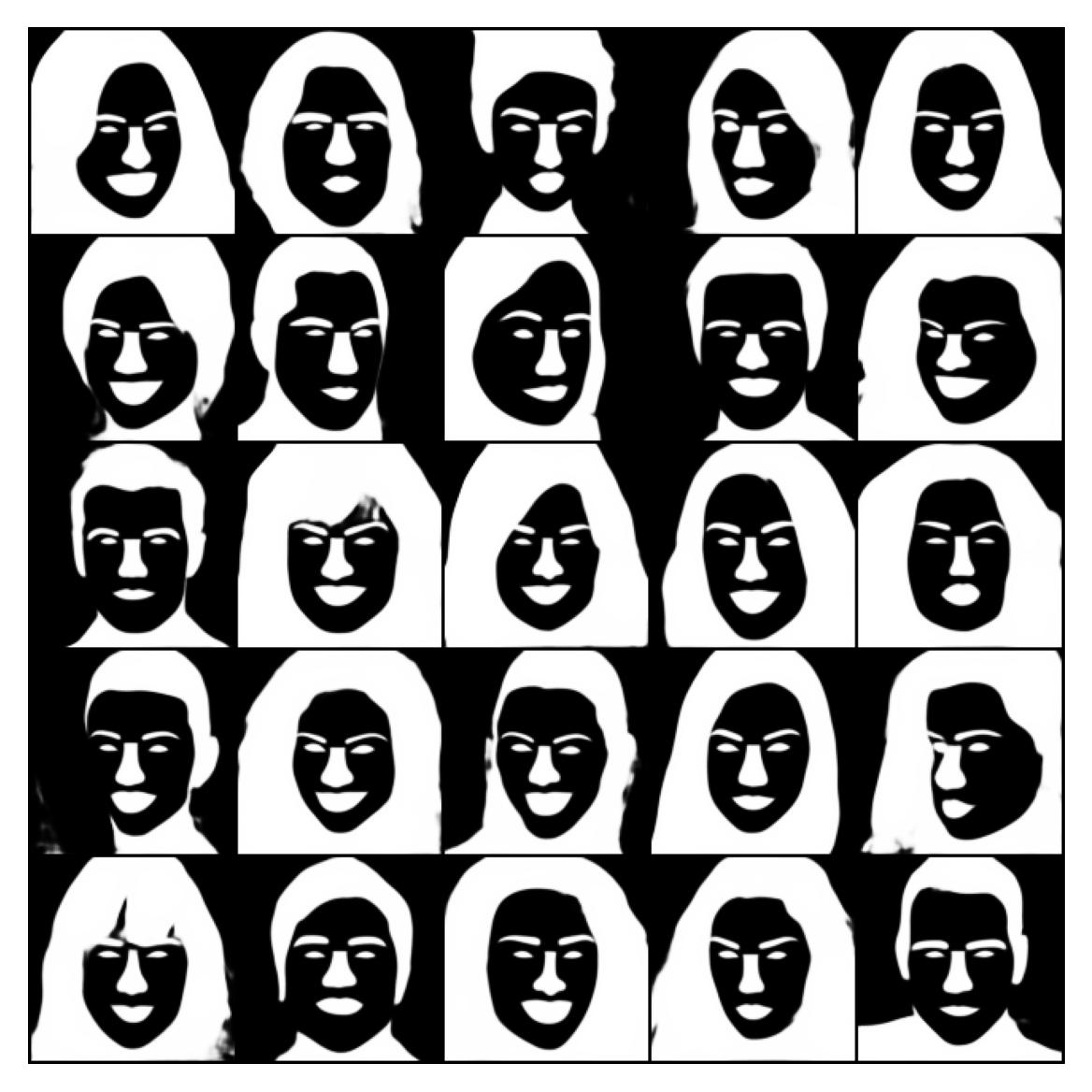}
            \caption{Mask}
\end{subfigure}
  \begin{subfigure}{0.49\textwidth}
         \centering
         \includegraphics[width=\linewidth]{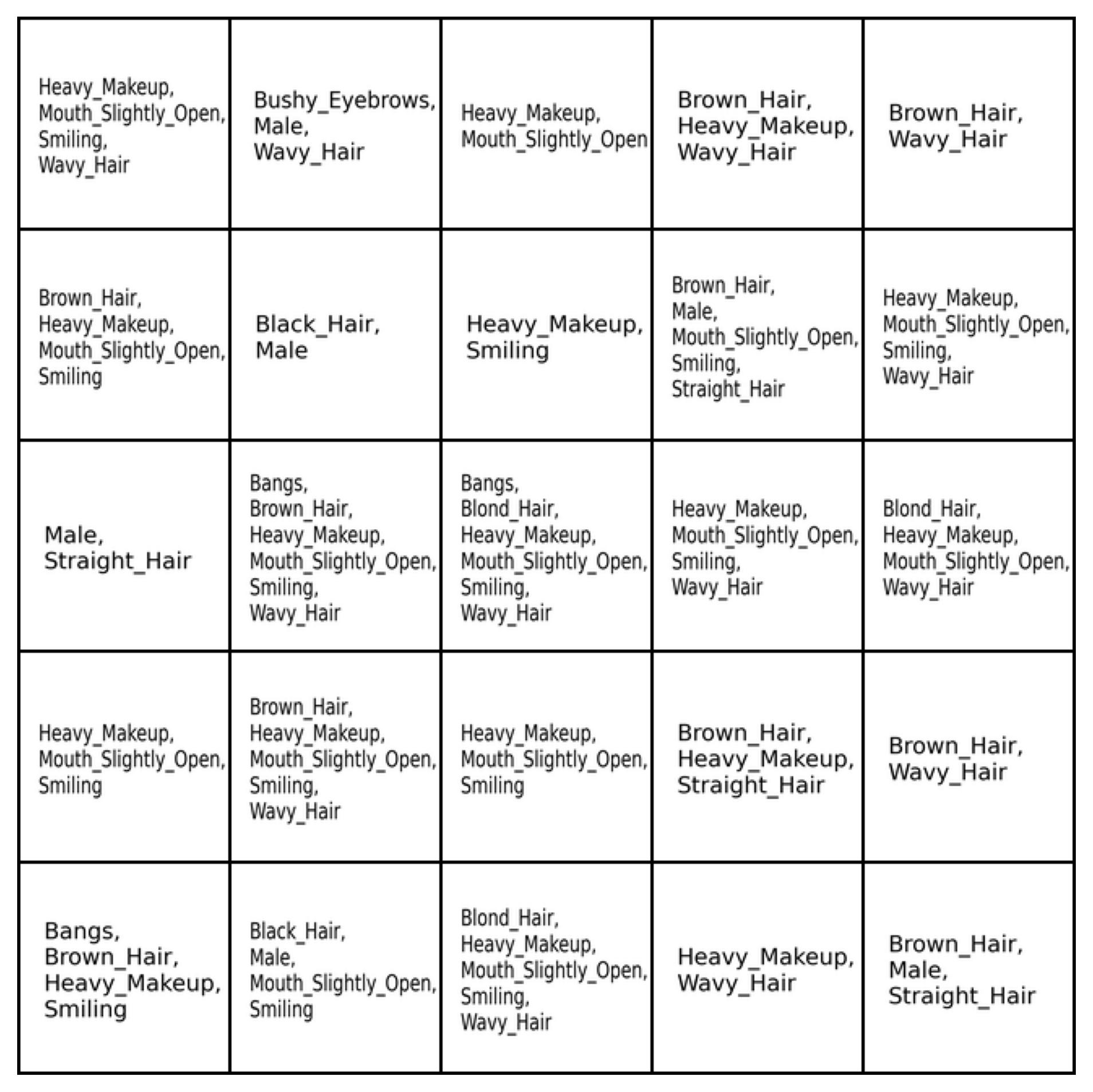}
            \caption{Attributes}
\end{subfigure}
        \caption{Joint (Unconditional) generation  qualitative results of \textbf{\gls{MLD}} on CelebAMask-HQ.}

\end{figure}

\begin{table}[H]
\tiny
\centering
\begin{tabular}{c|cc|cccc|cc}
\toprule
\multirow{3}{*}{ Models }  & \multicolumn{2}{c}{ Attributes }  &  \multicolumn{4}{c}{Image }  &  \multicolumn{2}{c}{Mask } \\
    \cmidrule{2-9}
    &  Img + Mask & Img &  Att + Mask & Mask  & Att & Joint & Img+Att & Img \\
   
    &  F1 & F1 &  FID  & FID  & FID & FID & F1  & F1 \\
     \midrule
     \\
     \cite{wesego2023scorebased} 
     \\
     SBM-RAE  & 0.62 & 0.6 & 84.9& 86.4 & 85.6& 84.2 & 0.83 & 0.82
    \\
     SBM-RAE-C & 0.66 & 0.64& 83.6& 82.8& 83.1& 84.2 & 0.83 & 0.82
     \\
     SBM-VAE& 0.62 & 0.58 & 81.6& 81.9 & 78.7&79.1& 0.83 & 0.83
     \\
     SBM-VAE-C& 0.69 & 0.66 & 82.4& 81.7 & 76.3& 79.1 & 0.84 & 0.84

     \\
          \cmidrule{2-9}
     \gls{MOPOE} & 0.68 & \textbf{0.71} &114.9 & 101.1 &186.8 & 164.8 & 0.85 & \textbf{0.92}
     \\
      \gls{MVTCAE} & 0.71 & 0.69 & 94 & 84.2 &87.2 & 162.2 &\textbf{0.89} & 0.89
      \\
    \gls{MMVAEplus} & 0.64 & 0.61 &133 & 97.3 & 153 &103.7 & 0.82 & 0.89 \\
     \midrule
   Supervised classifier &   &
   0.79  & &  &   &  & 0.94 \\
  \midrule
\textbf{ \gls{MLD} (ours) } & 
 $ \textbf{0.72}$&
 $0.69$ &

 $\textbf{52.75}$ & 
    $\textbf{51.73}$ & 
    $\textbf{53.09}$ &
 $\textbf{54.27}$ 
 & 0.87 & 0.87 \\
    \bottomrule

\end{tabular}
\caption{Quantitative results on CelebAMask-HQ dataset. The performance is measured in terms of \gls{FID} ($ \downarrow $) and F1 score ($ \uparrow $). The first row is the generated modality while the second row is the modalities used as condition.
Supervised classifier designates a classifier performance to predict the attributes or the mask from an image. }
\label{coh_qua:celebA}

 \end{table}

\begin{figure}[H]
     \centering
     
     \begin{subfigure}{0.14\textwidth}
                \begin{subfigure}{1\textwidth}
    
         \includegraphics[width=\linewidth]{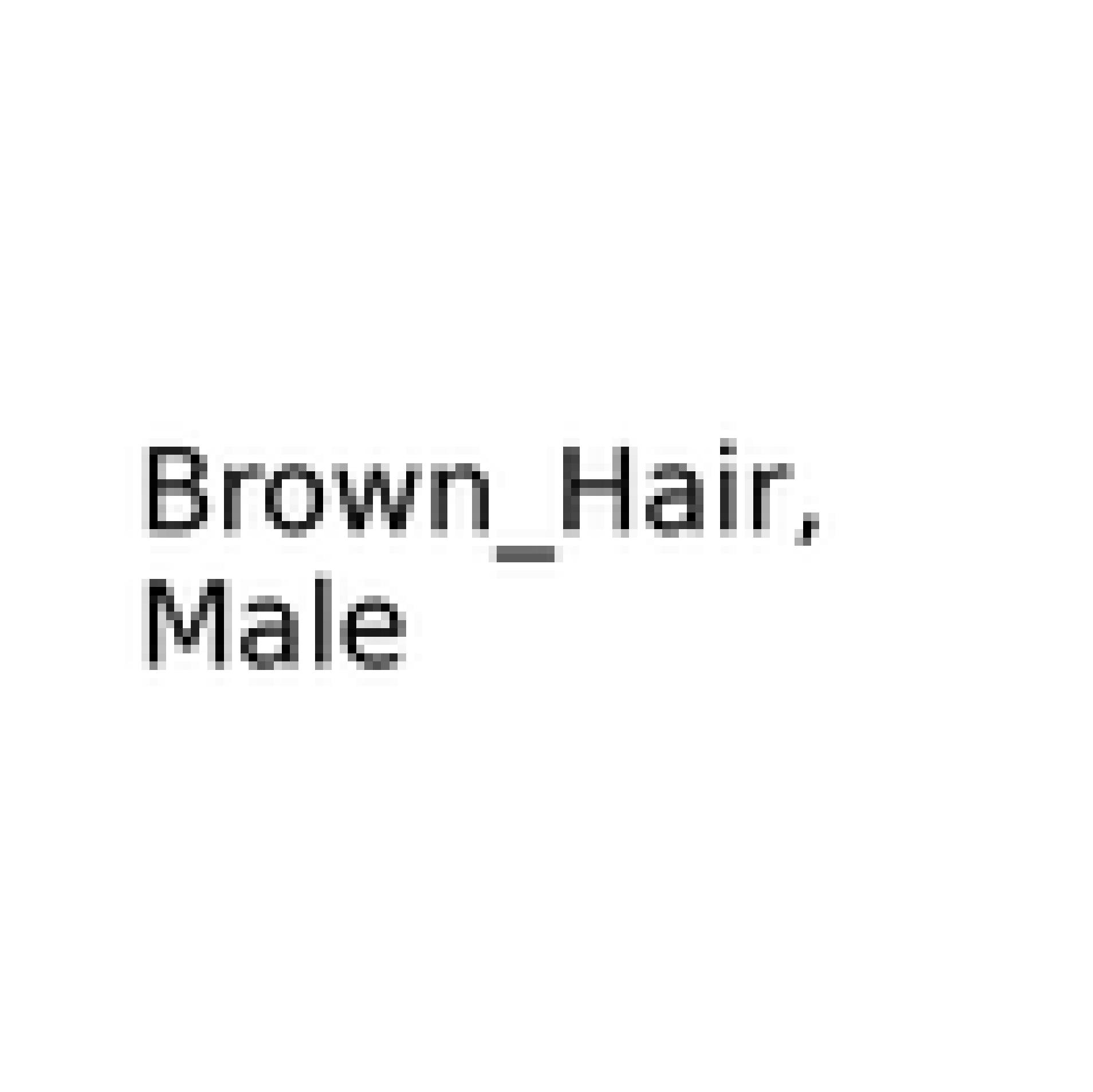}
         \caption*{ }
          \end{subfigure}

                    \begin{subfigure}{1\textwidth}
    
         \includegraphics[width=\linewidth]{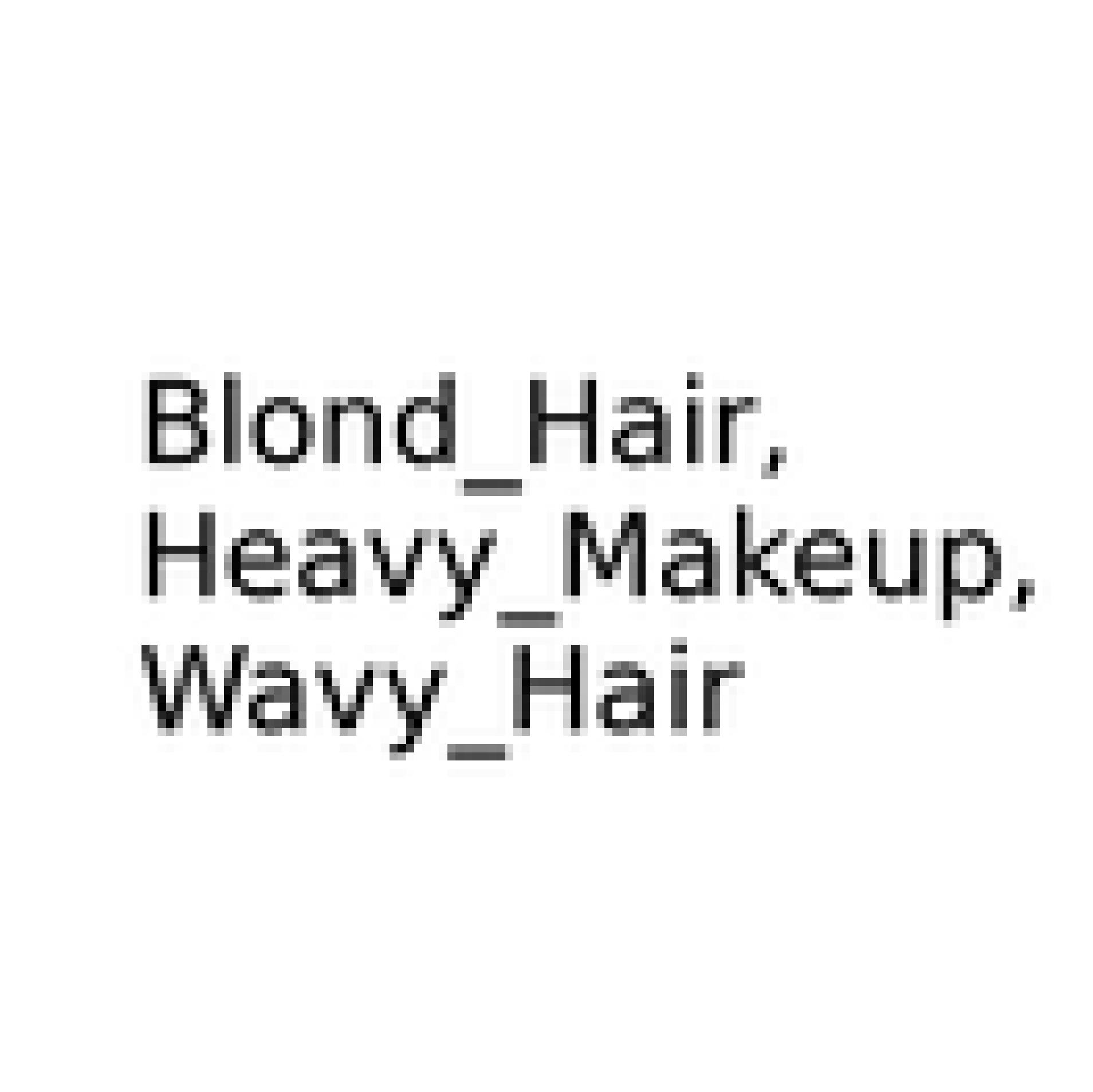}
         \caption*{ }
          \end{subfigure}

              \begin{subfigure}{1\textwidth}
    
         \includegraphics[width=\linewidth]{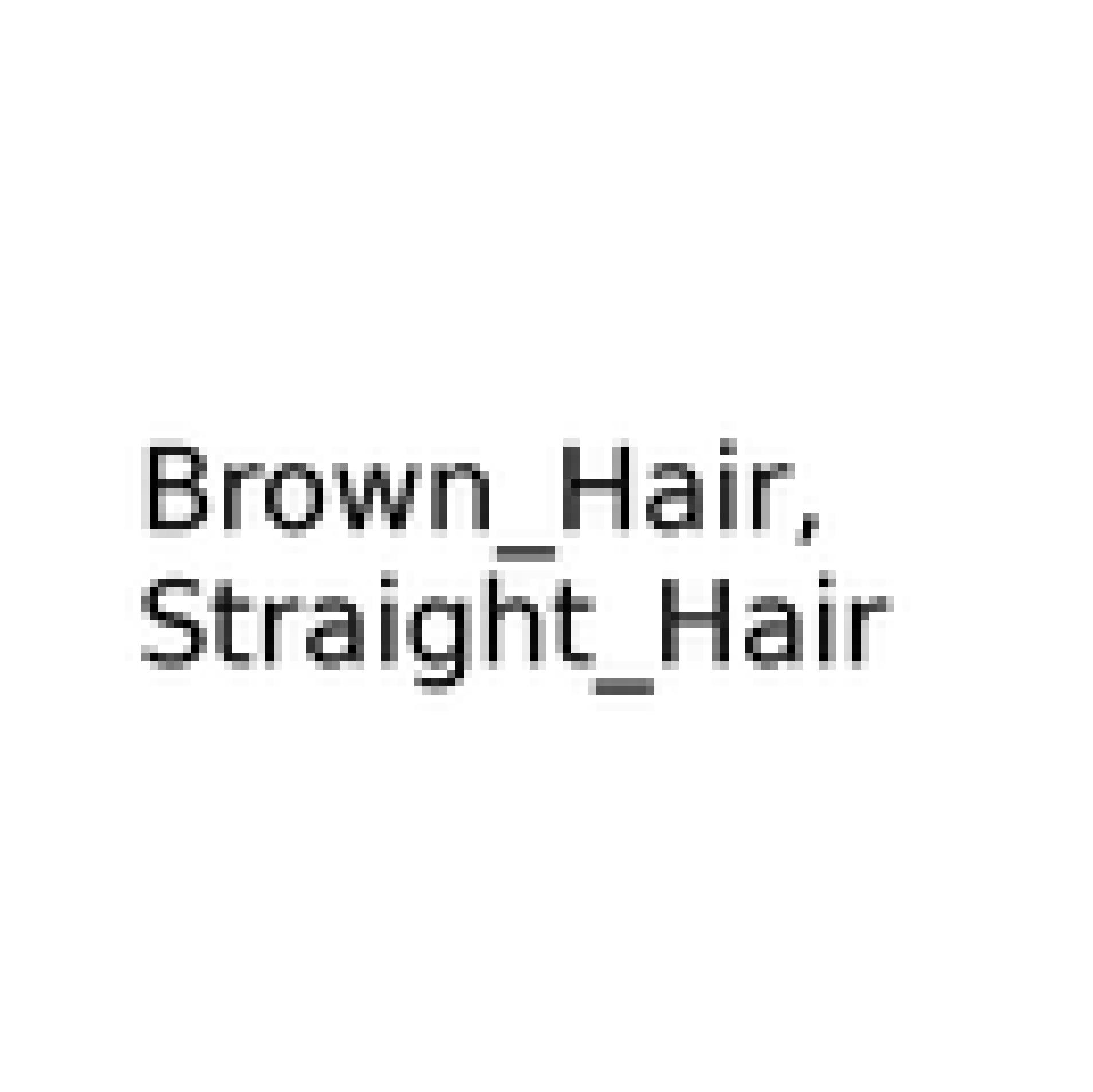}
         \caption*{ }
          \end{subfigure}

    \end{subfigure}
  \begin{subfigure}{0.14\textwidth}

\end{subfigure}
  \begin{subfigure}{0.7\textwidth}
  
     \begin{subfigure}{1\textwidth}
      
         \includegraphics[width=\linewidth]{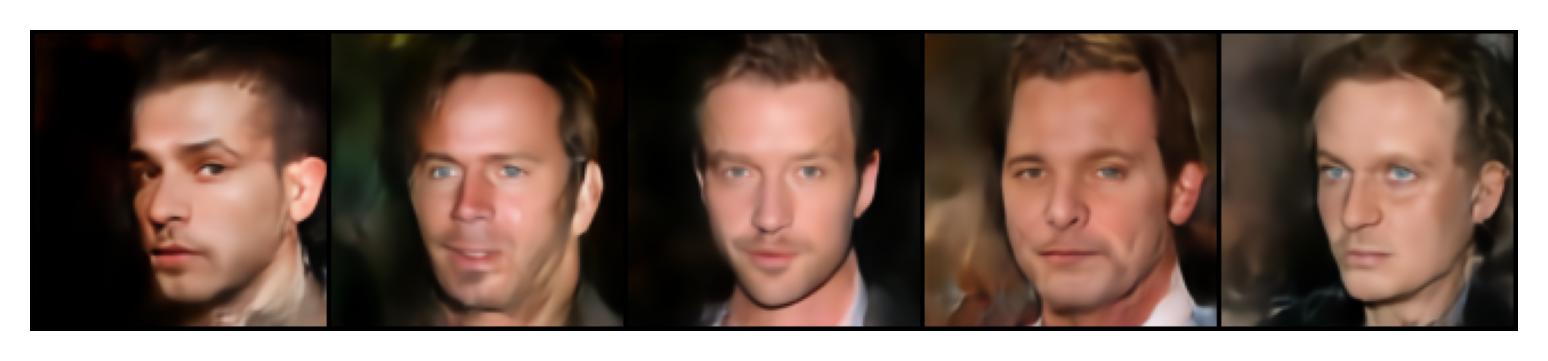}
    \end{subfigure}

        \begin{subfigure}{1\textwidth}
      
         \includegraphics[width=\linewidth]{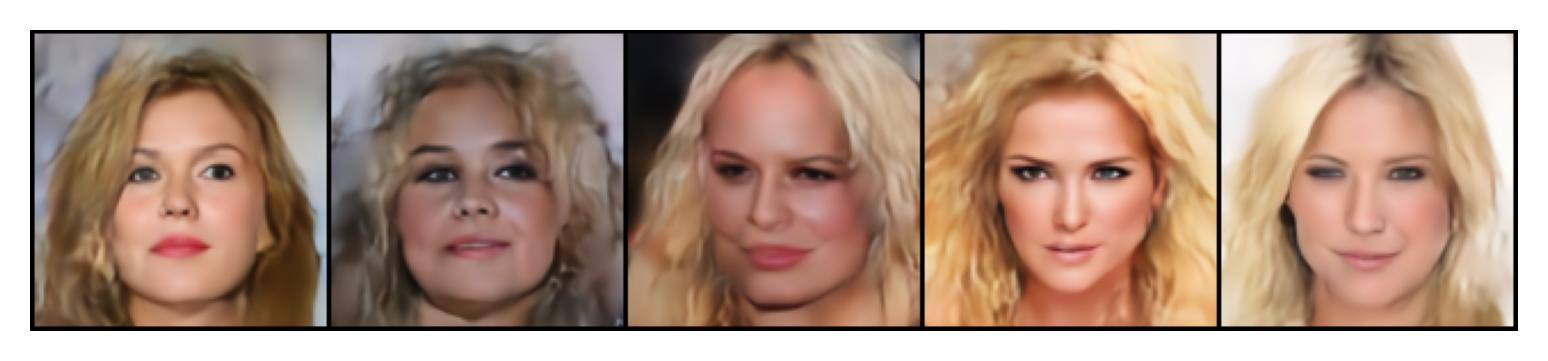}
    \end{subfigure}
    
      \begin{subfigure}{1\textwidth}
         \includegraphics[width=\linewidth]{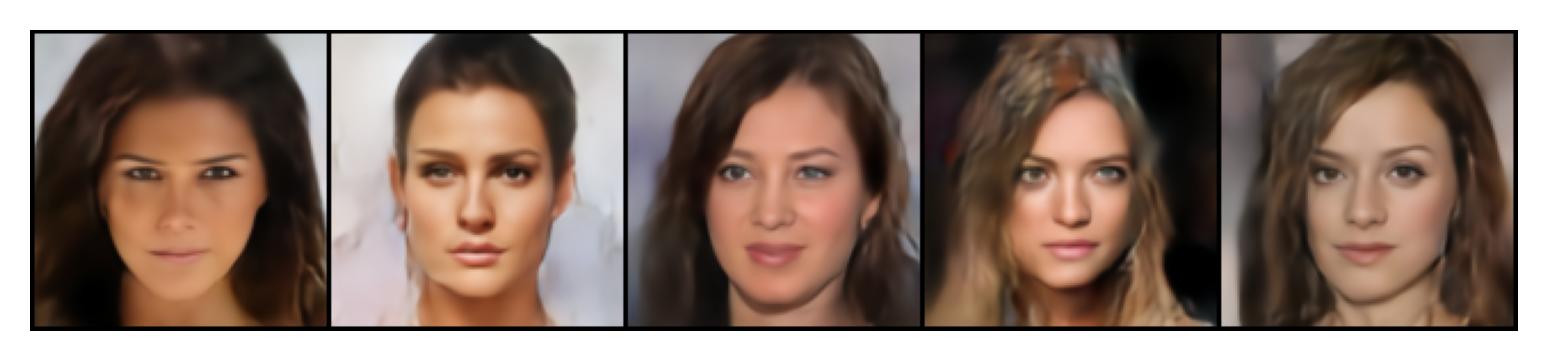}
    \end{subfigure}

            \caption{Generated Images}
\end{subfigure}
        \caption{ (Attributes $\rightarrow$ Image ) Conditional generation of \textbf{\gls{MLD}} on CelebAMask-HQ.}

\end{figure}

\begin{figure}[H]
     \centering
     
     \begin{subfigure}{0.13\textwidth}

      \begin{subfigure}{1\textwidth}
     
         \includegraphics[width=\linewidth]{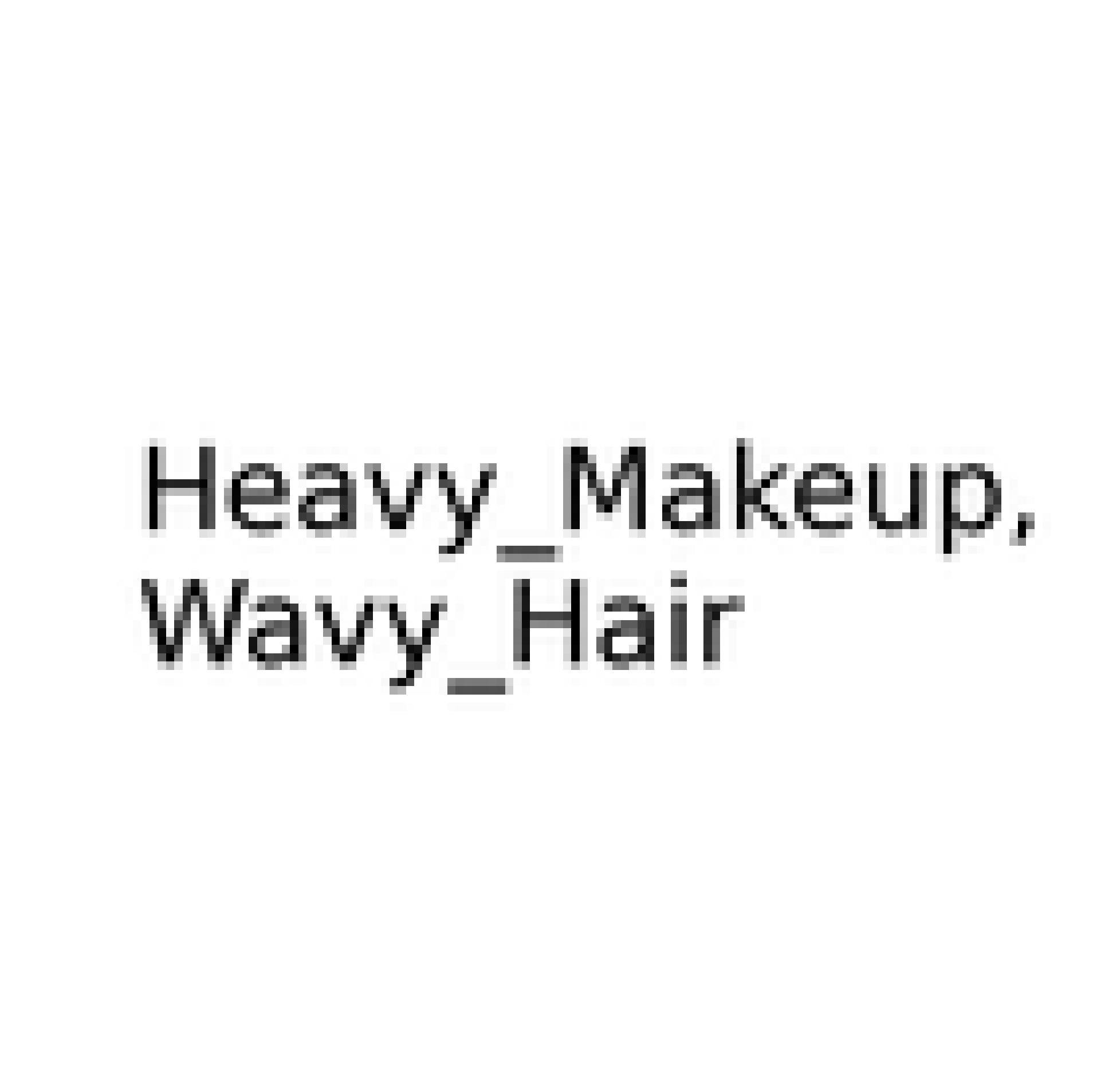}
     \caption*{ }
          \end{subfigure}
          
                \begin{subfigure}{1\textwidth}
    
         \includegraphics[width=\linewidth]{figures/datasets/celebA/examp2/2_attributes_c.jpg}
  \caption*{ }
          \end{subfigure}
              \begin{subfigure}{1\textwidth}
     
         \includegraphics[width=\linewidth]{figures/datasets/celebA/examp4/4_attributes_c.jpg}
 \caption*{ }
          \end{subfigure}

                \begin{subfigure}{1\textwidth}
     
         \includegraphics[width=\linewidth]{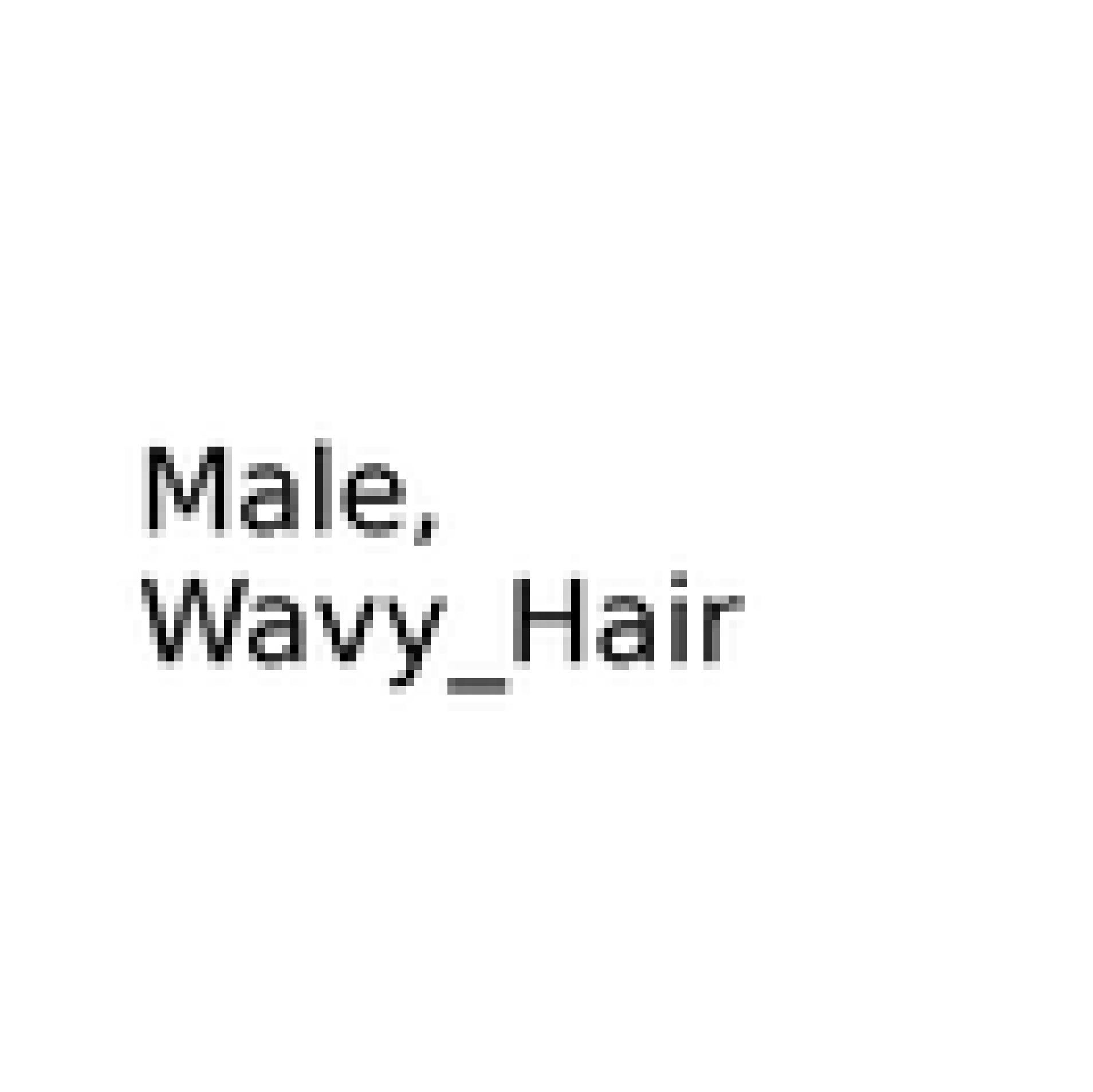}
 \caption*{ }
          \end{subfigure}
          
    \end{subfigure}
  \begin{subfigure}{0.14\textwidth}
  
   \begin{subfigure}{1\textwidth}

         \includegraphics[width=\linewidth]{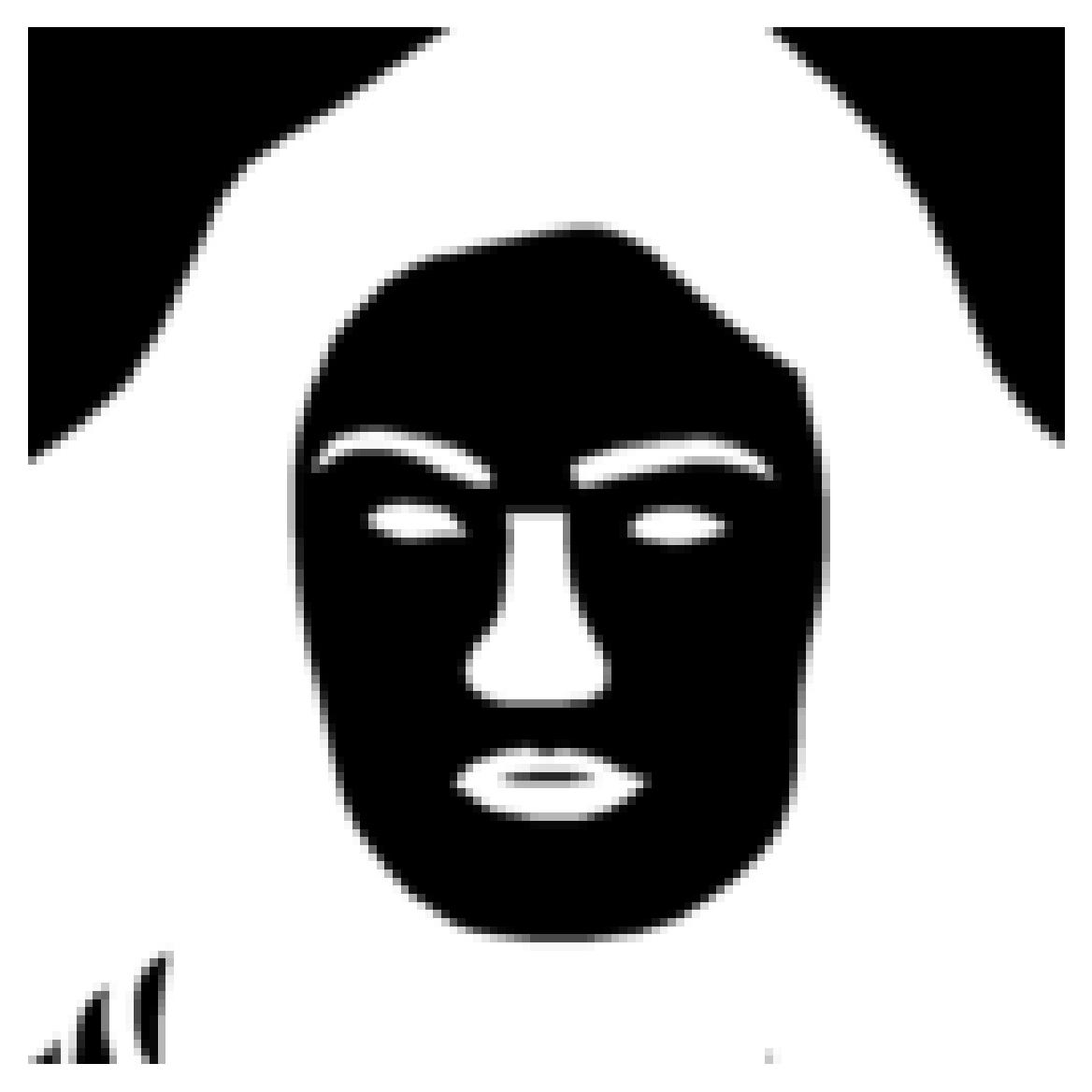}
        
    \end{subfigure}

      \begin{subfigure}{1\textwidth}
         \includegraphics[width=\linewidth]{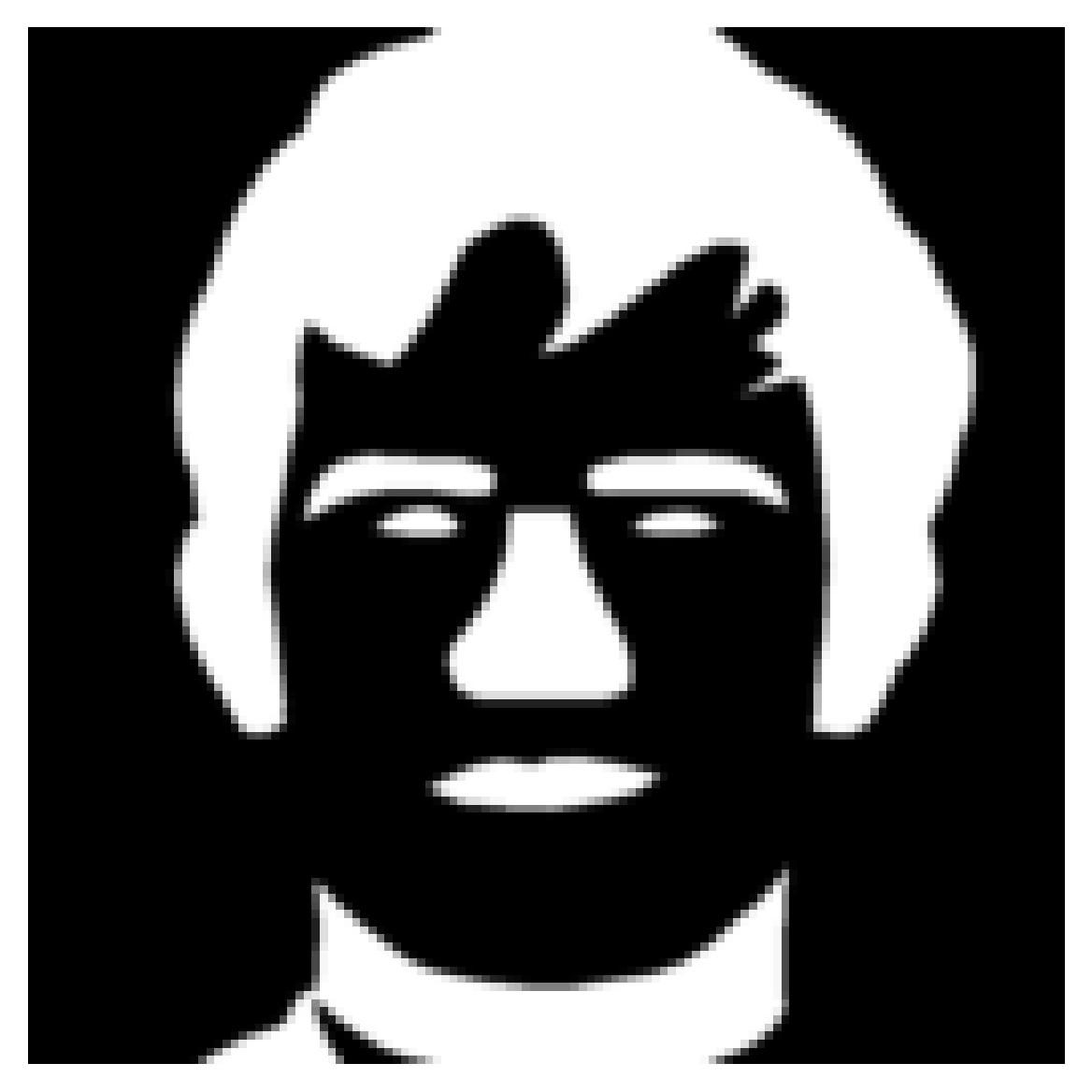}
    \end{subfigure}
    
     

               \begin{subfigure}{1\textwidth}
     
         \includegraphics[width=\linewidth]{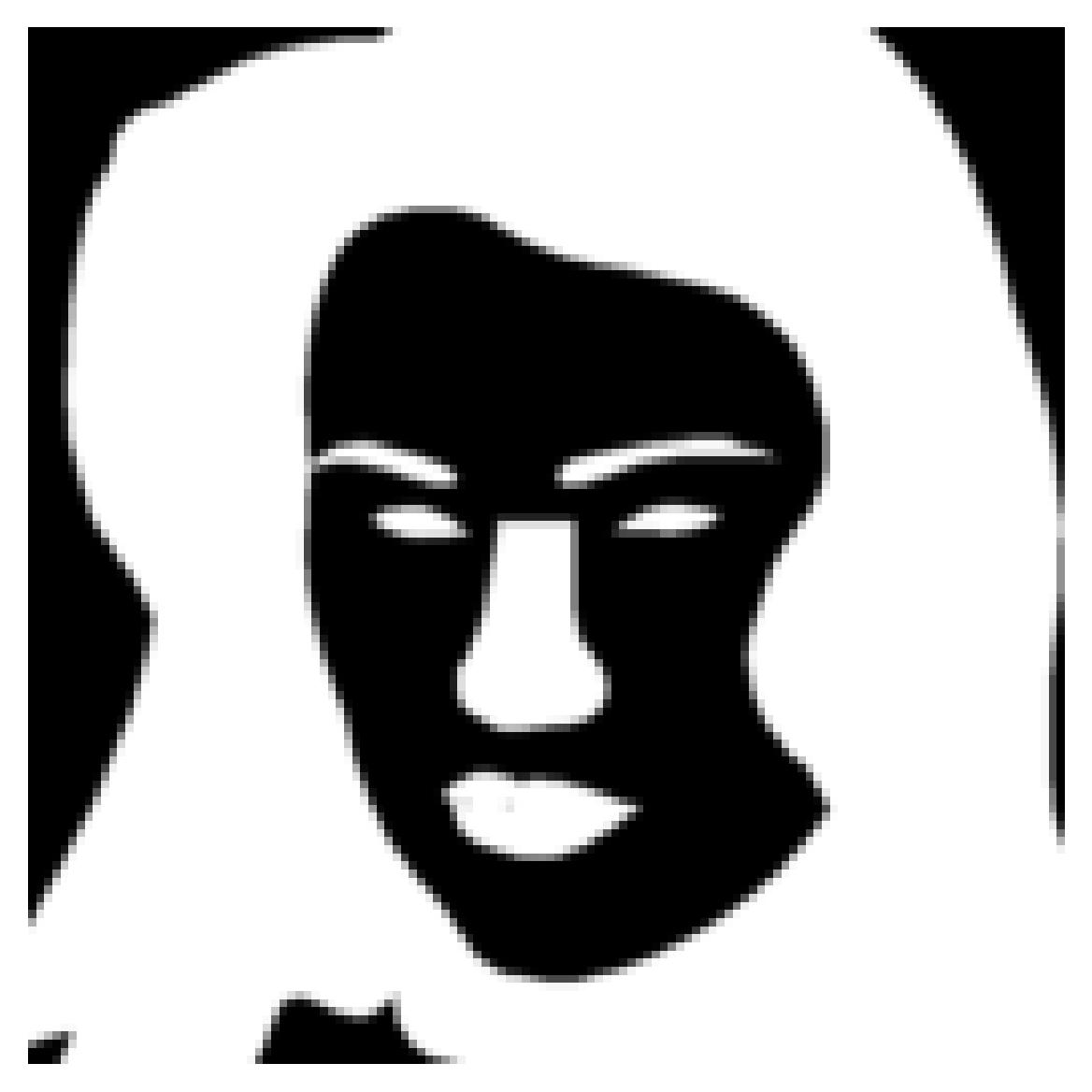}
         \caption*{ }
          \end{subfigure}

             \begin{subfigure}{1\textwidth}
     
         \includegraphics[width=\linewidth]{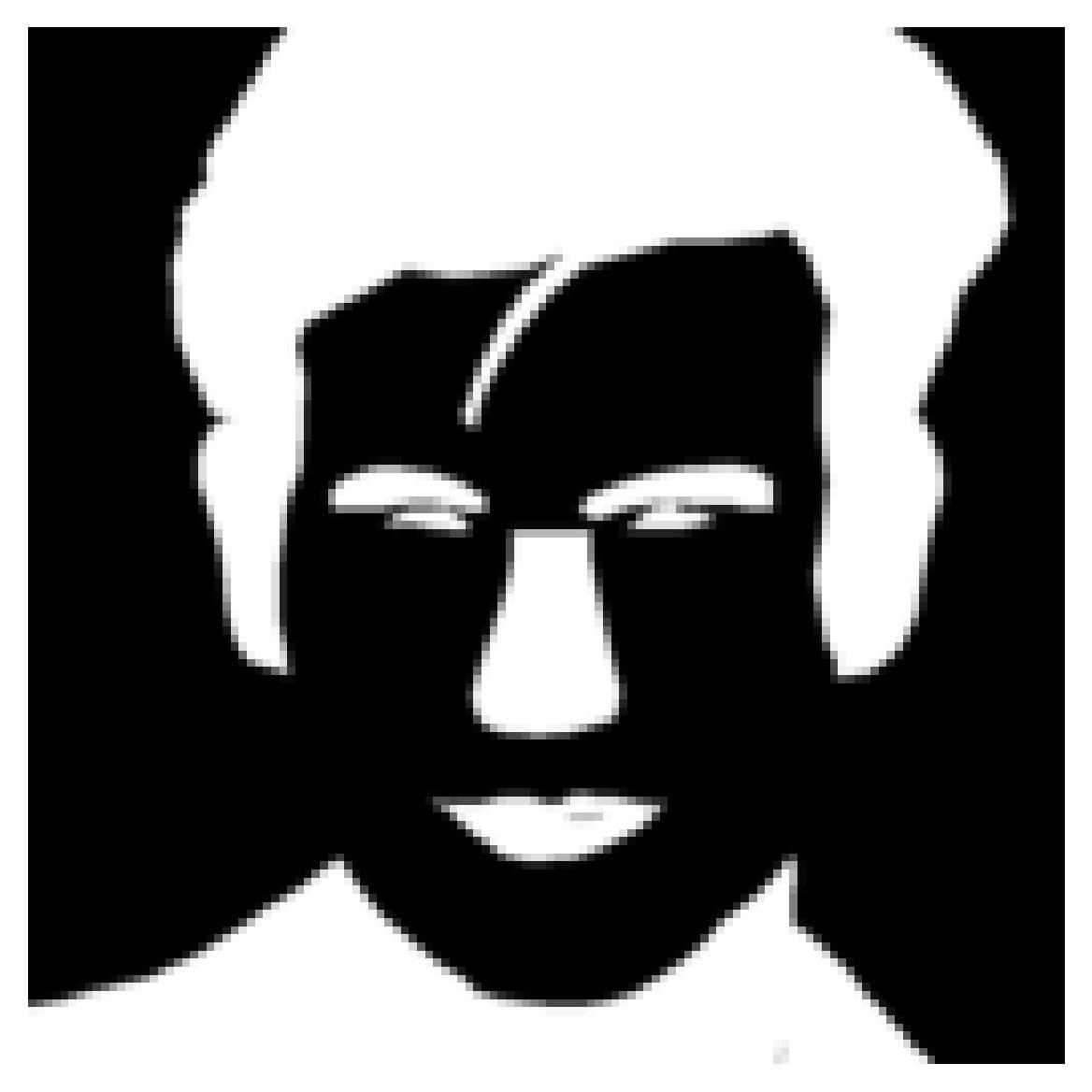}
         \caption*{ }
          \end{subfigure}
          
\end{subfigure}
  \begin{subfigure}{0.65\textwidth}
    \begin{subfigure}{1\textwidth}
 
         \includegraphics[width=\linewidth]{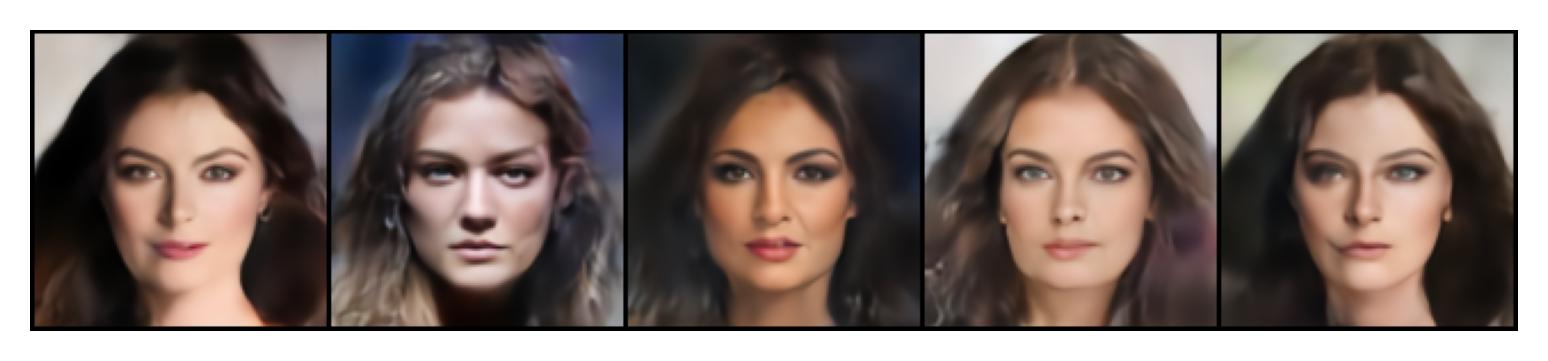}
    \end{subfigure}
    
     \begin{subfigure}{1\textwidth}
      
         \includegraphics[width=\linewidth]{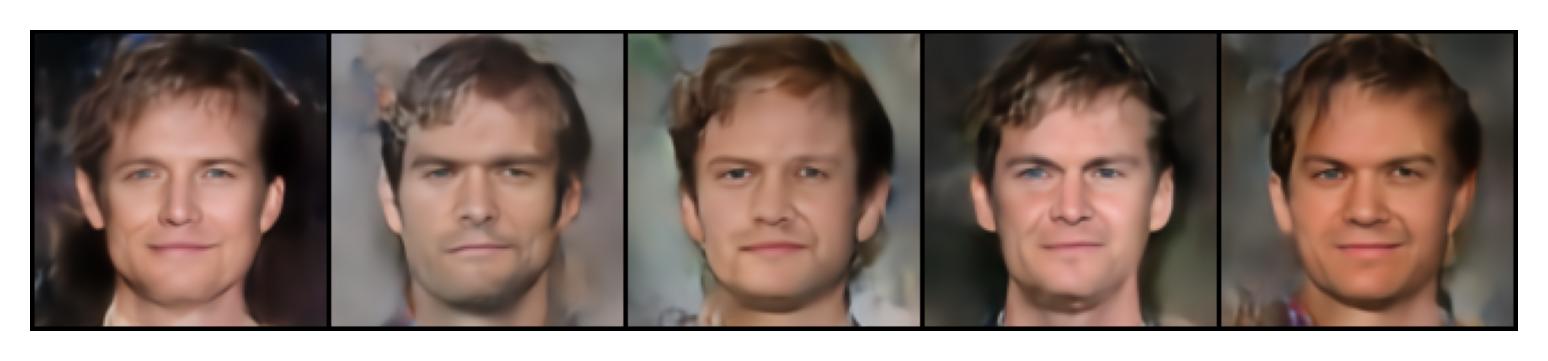}
    \end{subfigure}

      

     \begin{subfigure}{1\textwidth}
      
         \includegraphics[width=\linewidth]{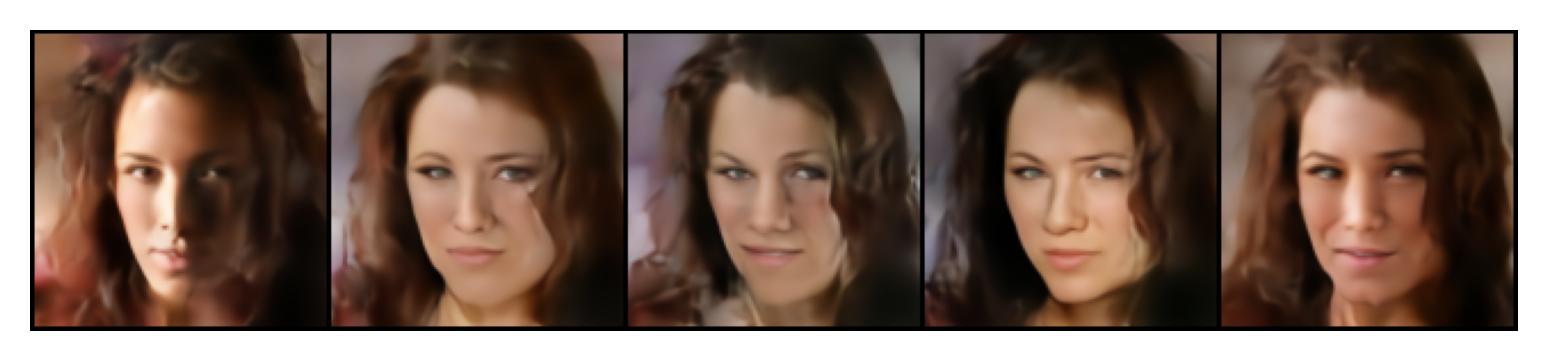}
    \end{subfigure}

       \begin{subfigure}{1\textwidth}
      
         \includegraphics[width=\linewidth]{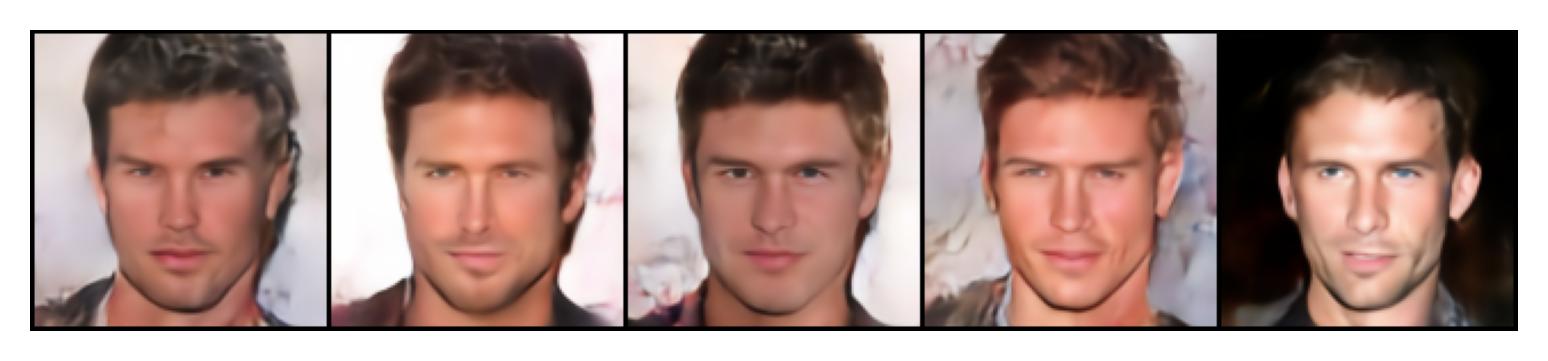}
    \end{subfigure}
            \caption{Generated Images}
\end{subfigure}
        \caption{ (Attributes,Mask $\rightarrow$ Image ) Conditional generation of \textbf{\gls{MLD}} on CelebAMask-HQ.}

\end{figure}

\begin{figure}[H]
     \centering
     
     \begin{subfigure}{0.14\textwidth}
                \begin{subfigure}{1\textwidth}
         \includegraphics[width=\linewidth]{figures/datasets/celebA/examp2/2_mask_c.jpg}
          \end{subfigure}

               \begin{subfigure}{1\textwidth} 
         \includegraphics[width=\linewidth]{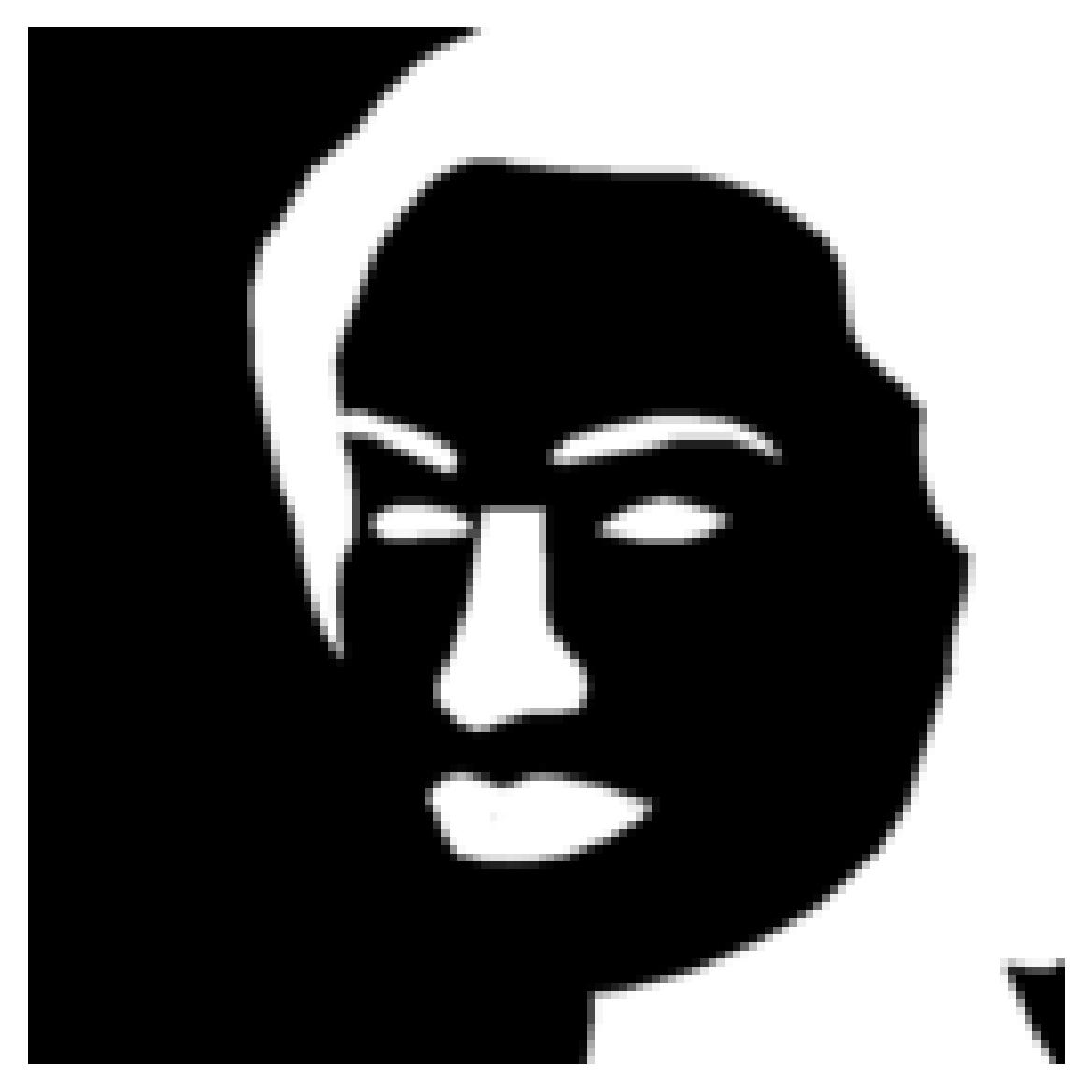}
          \end{subfigure}

             \begin{subfigure}{1\textwidth}
         \includegraphics[width=\linewidth]{figures/datasets/celebA/examp4/4_mask_c.jpg}
          \end{subfigure}

                 \begin{subfigure}{1\textwidth}
         \includegraphics[width=\linewidth]{figures/datasets/celebA/examp5/5_mask_c.jpg}
         \caption*{ }
          \end{subfigure}
    \end{subfigure}
      \begin{subfigure}{0.14\textwidth}
      \end{subfigure}
  \begin{subfigure}{0.7\textwidth}
  
     \begin{subfigure}{1\textwidth}
      
         \includegraphics[width=\linewidth]{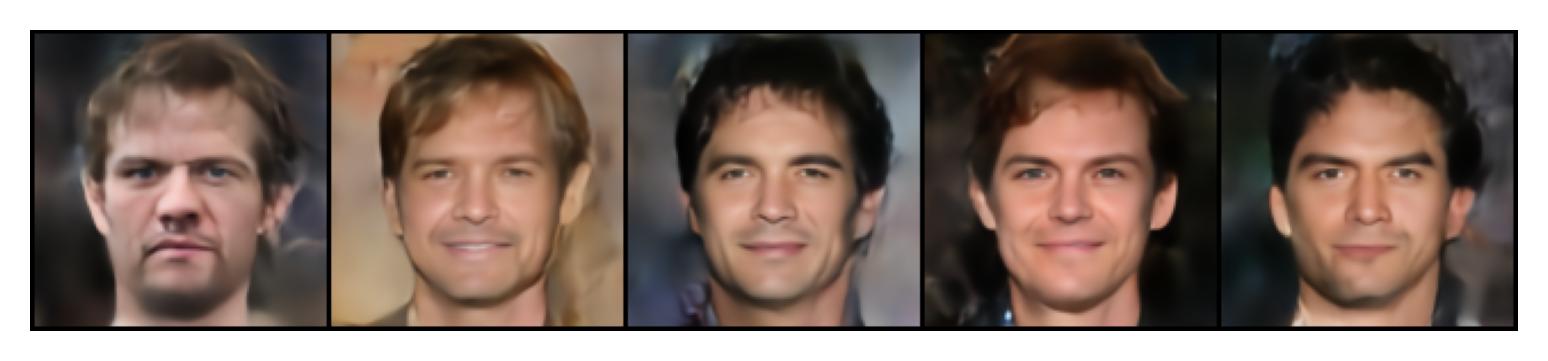}
    \end{subfigure}

     \begin{subfigure}{1\textwidth}
      
         \includegraphics[width=\linewidth]{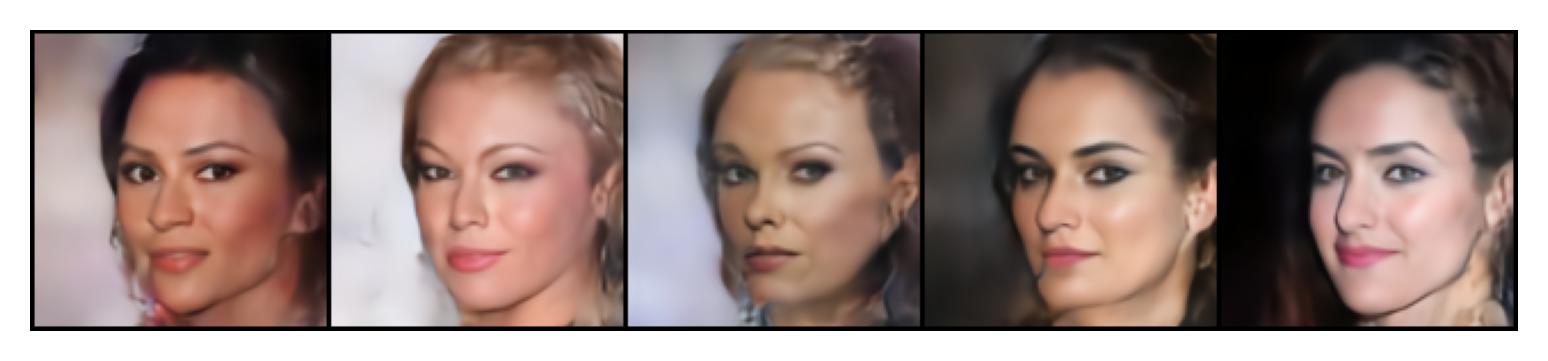}
    \end{subfigure}

      \begin{subfigure}{1\textwidth}
      
         \includegraphics[width=\linewidth]{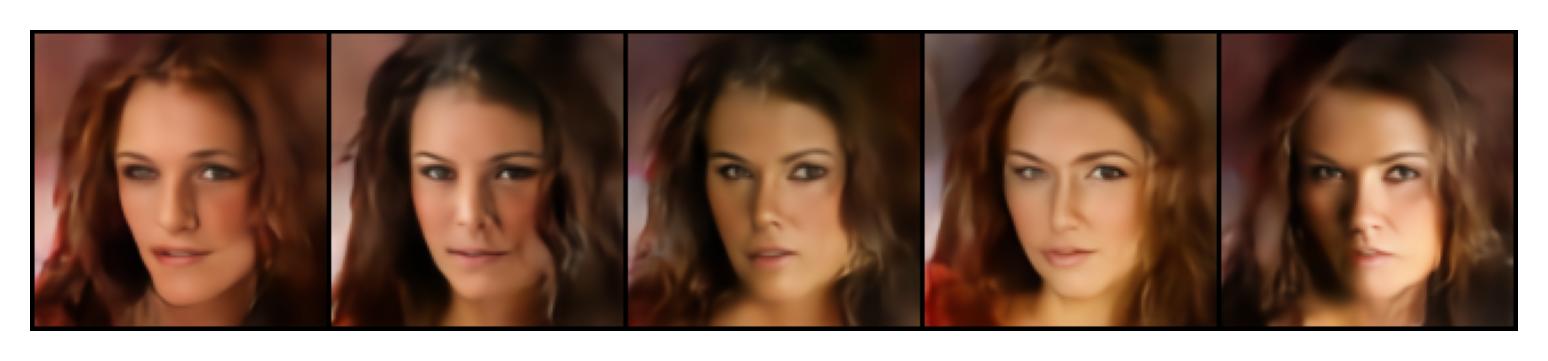}
    \end{subfigure}

         \begin{subfigure}{1\textwidth}
    
         \includegraphics[width=\linewidth]{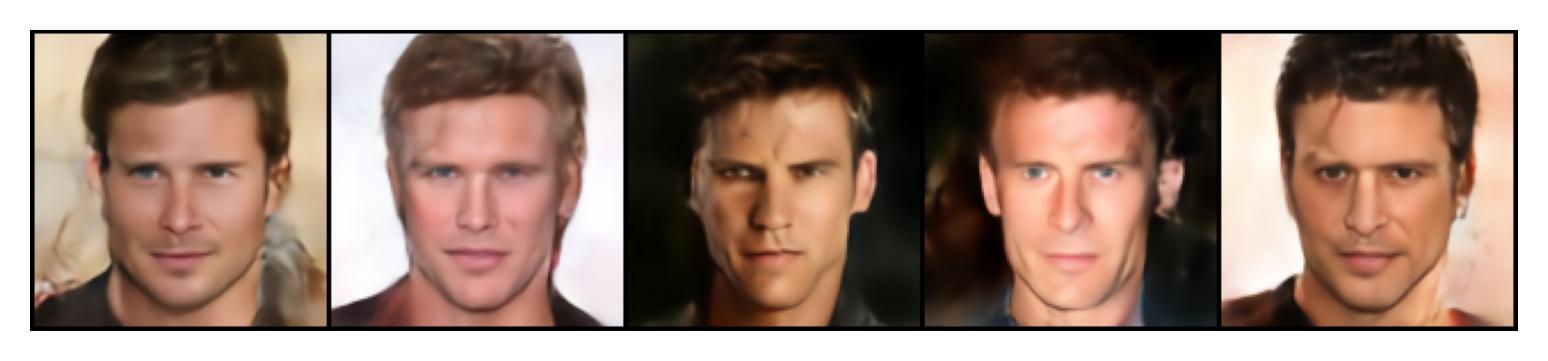}
      
          \end{subfigure}
    
            \caption{Generated Images}
\end{subfigure}
        \caption{ (Mask $\rightarrow$ Image ) Conditional generation of \textbf{\gls{MLD}} on CelebAMask-HQ.}

\end{figure}

\begin{figure}[H]
     \centering
     
     \begin{subfigure}{0.14\textwidth}
       \centering
                \begin{subfigure}{1\textwidth}
    
         \includegraphics[width=\linewidth]{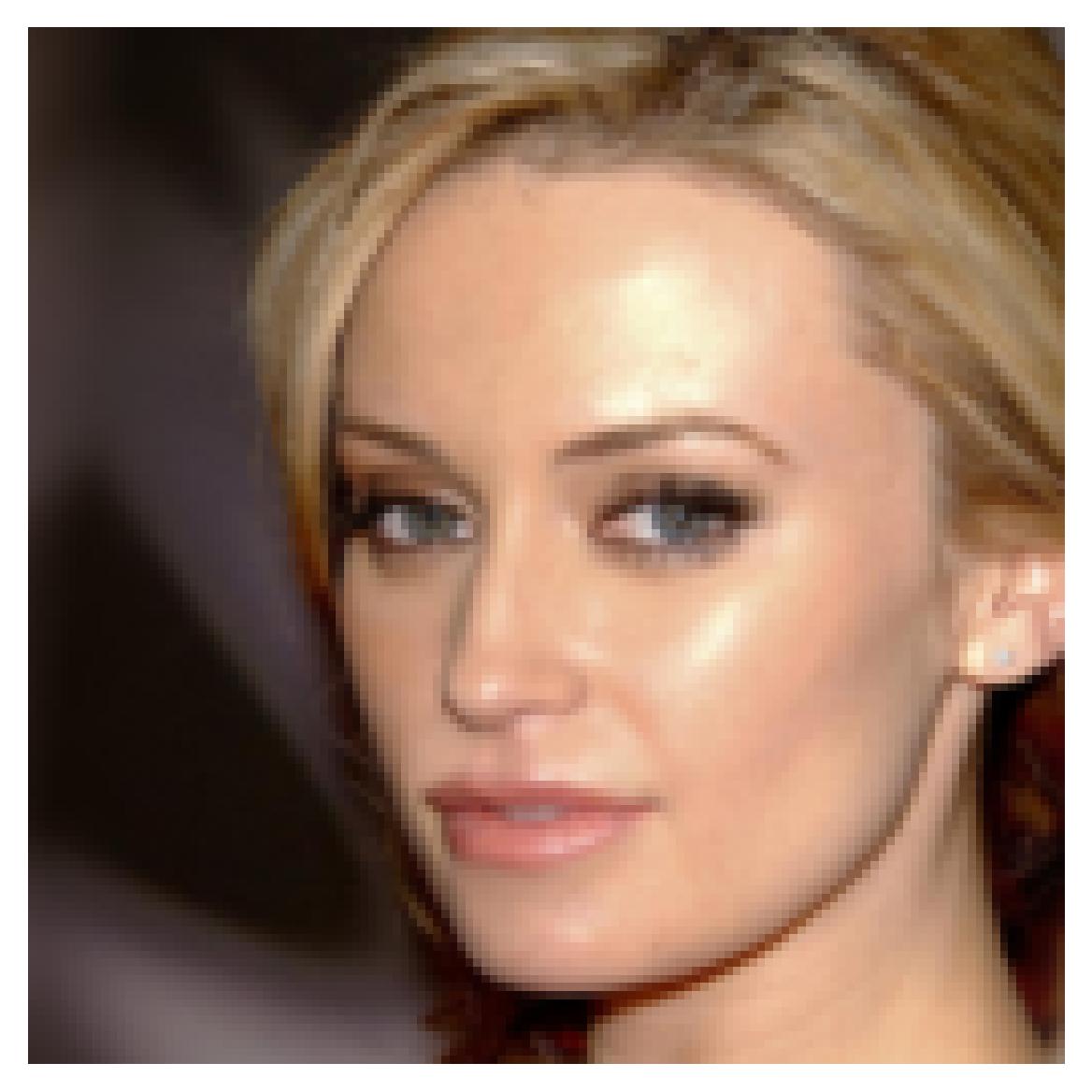}
         \caption*{ }\end{subfigure}

\vspace{3pc}

         \begin{subfigure}{1\textwidth}
    
         \includegraphics[width=\linewidth]{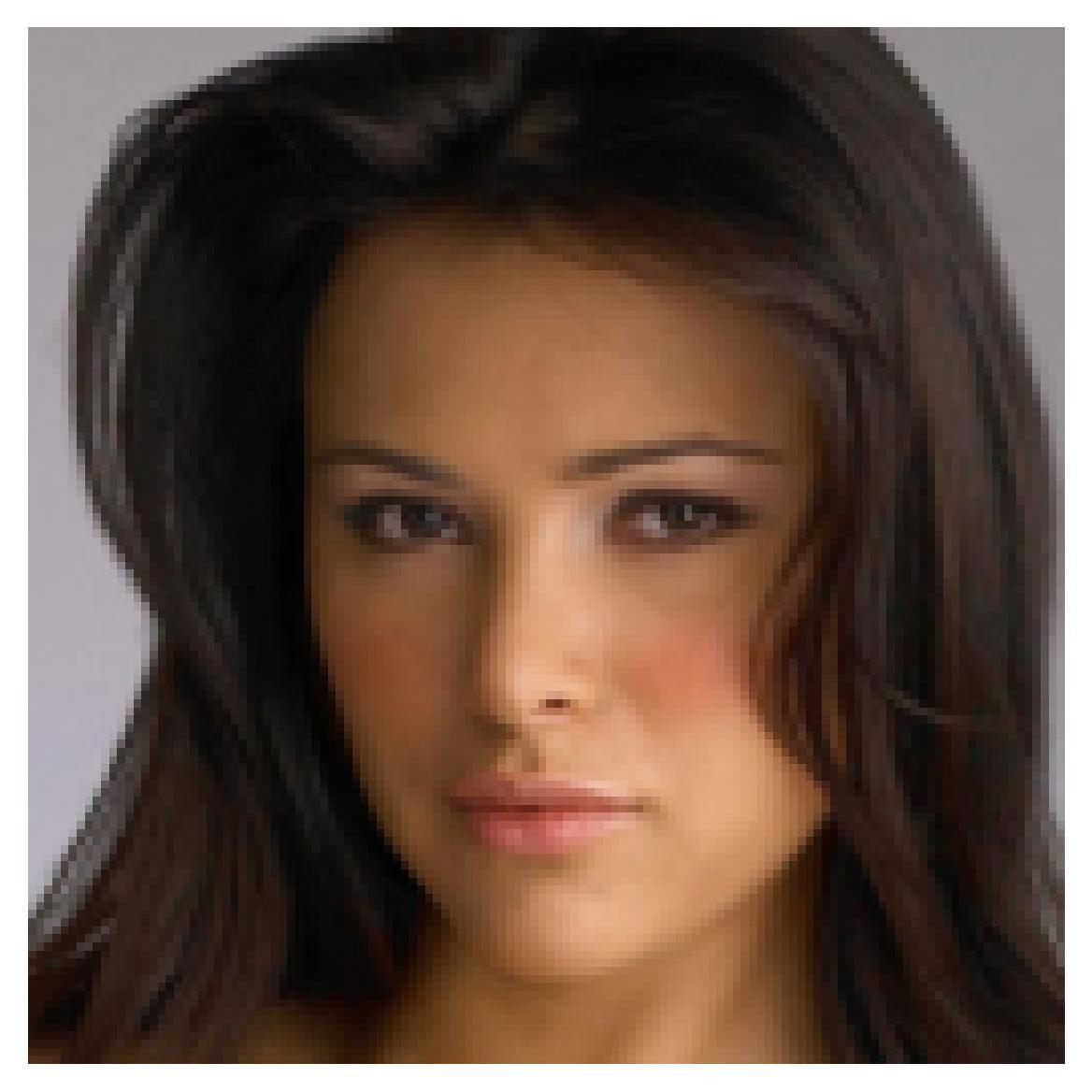}
         \caption*{ }\end{subfigure}
         
\vspace{3pc}

              \begin{subfigure}{1\textwidth}
    
         \includegraphics[width=\linewidth]{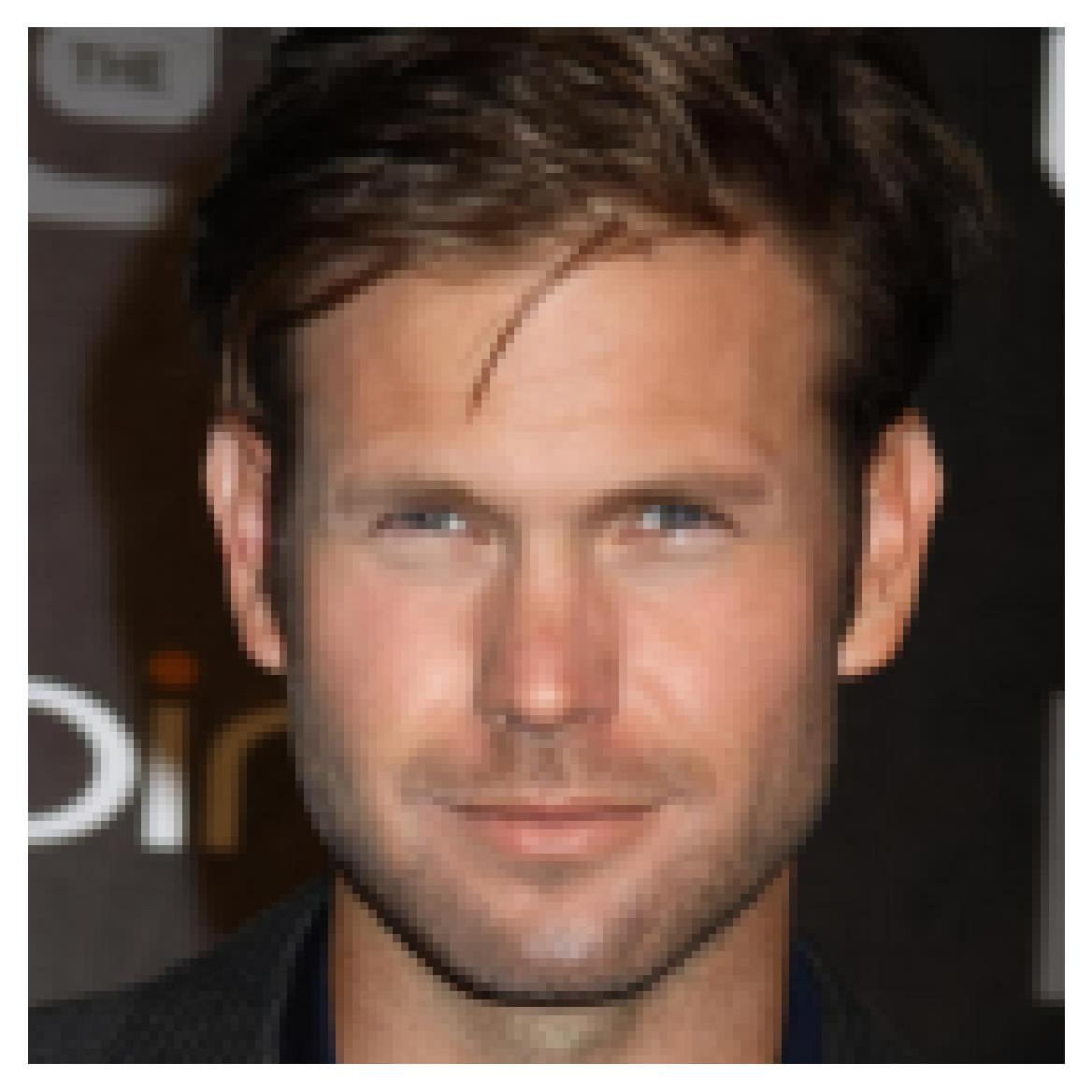}
         \caption*{ }
          \end{subfigure}
         
          \end{subfigure}
\begin{subfigure}{0.5\textwidth}
    \begin{subfigure}{1\textwidth}
         \includegraphics[width=\linewidth]{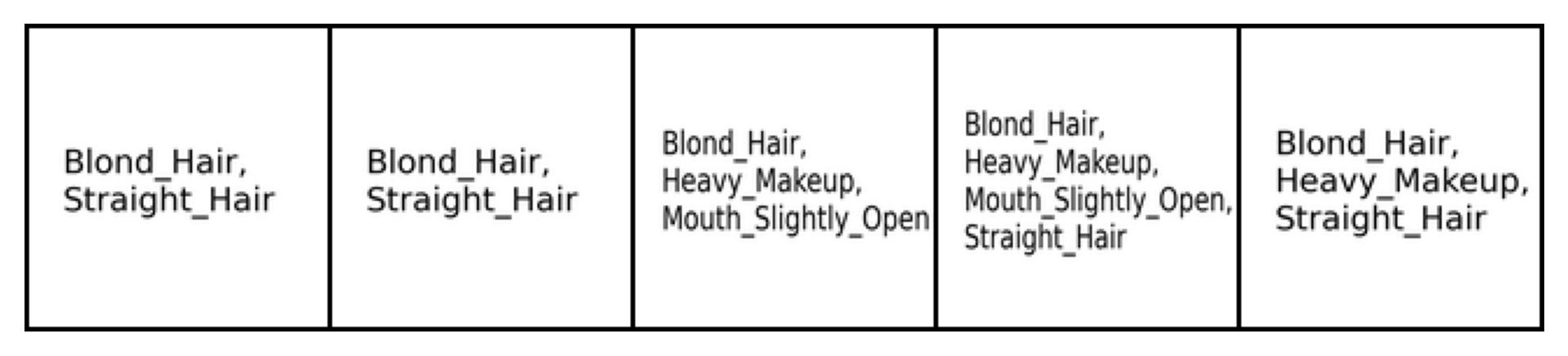}
         \caption*{ }
          \end{subfigure}

   \begin{subfigure}{1\textwidth}
    
         \includegraphics[width=\linewidth]{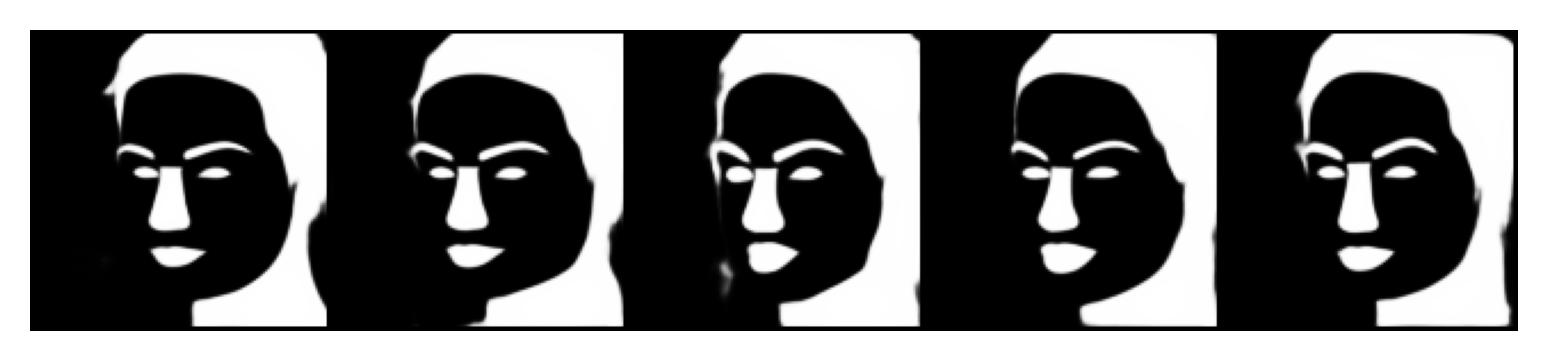}

     \begin{subfigure}{1\textwidth}
         \includegraphics[width=\linewidth]{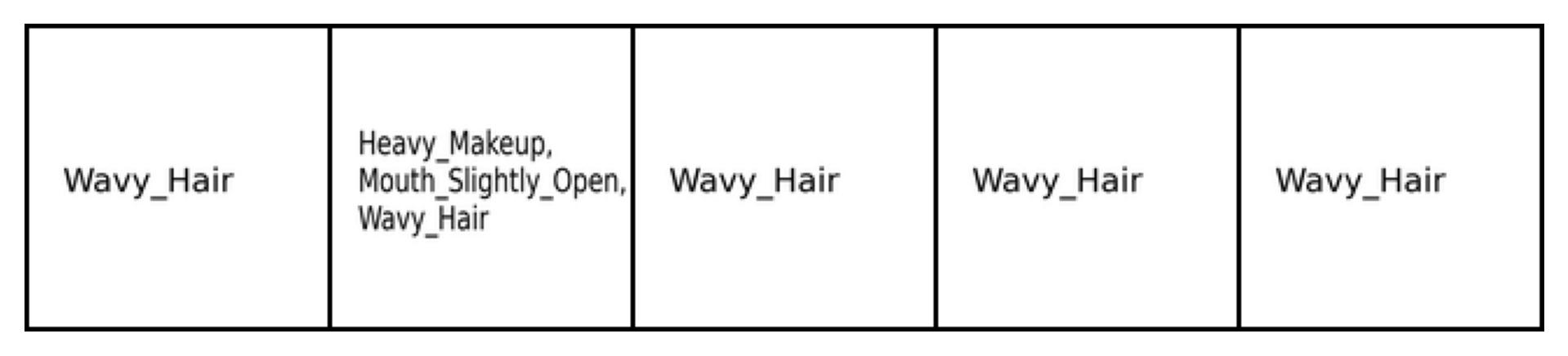}
      
          \end{subfigure}

           \begin{subfigure}{1\textwidth}
         \includegraphics[width=\linewidth]{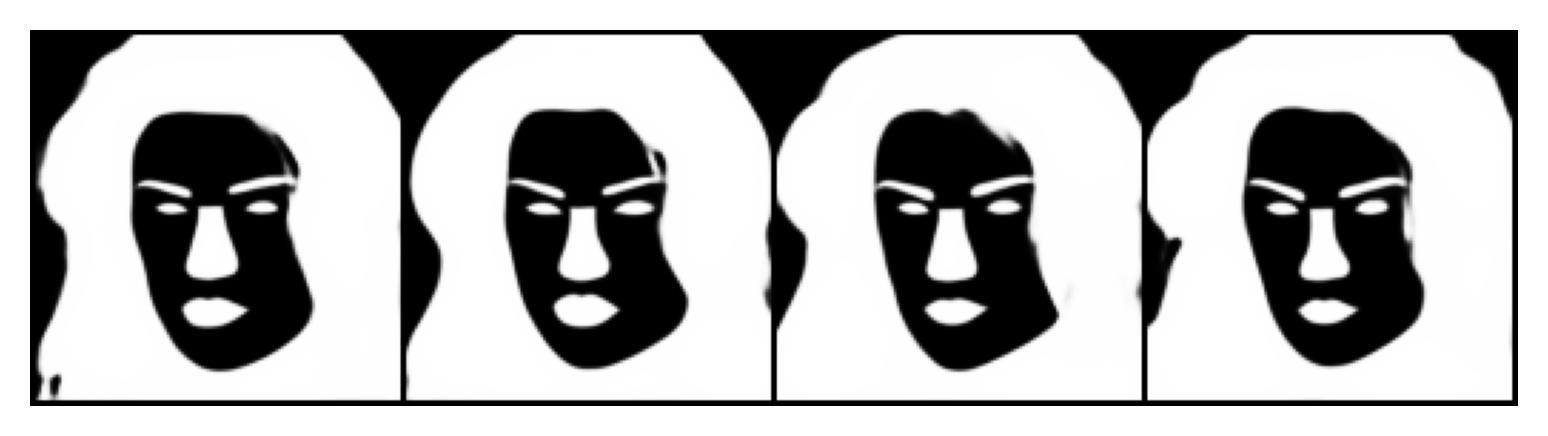}
       
          \end{subfigure}

            \begin{subfigure}{1\textwidth}
         \includegraphics[width=\linewidth]{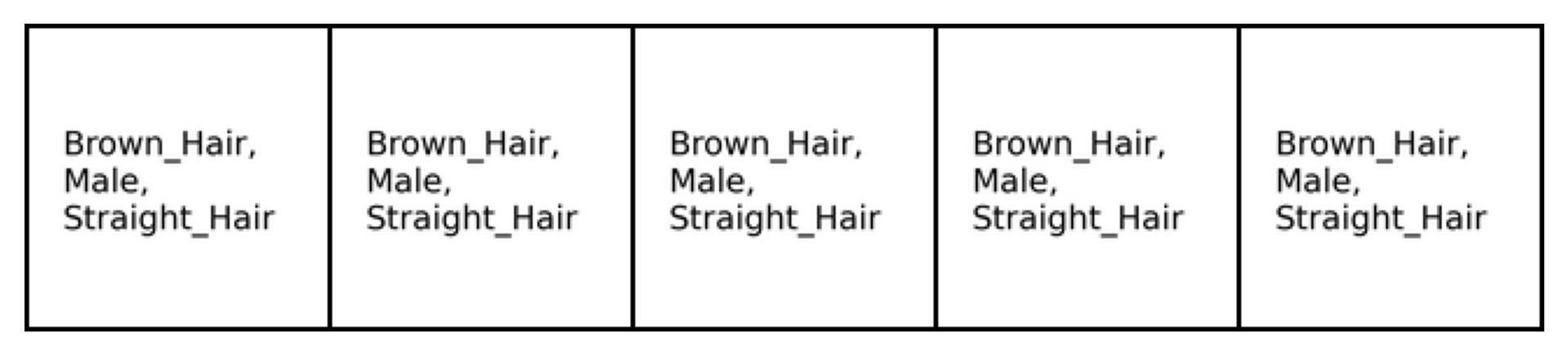}
      
          \end{subfigure}

           \begin{subfigure}{1\textwidth}
         \includegraphics[width=\linewidth]{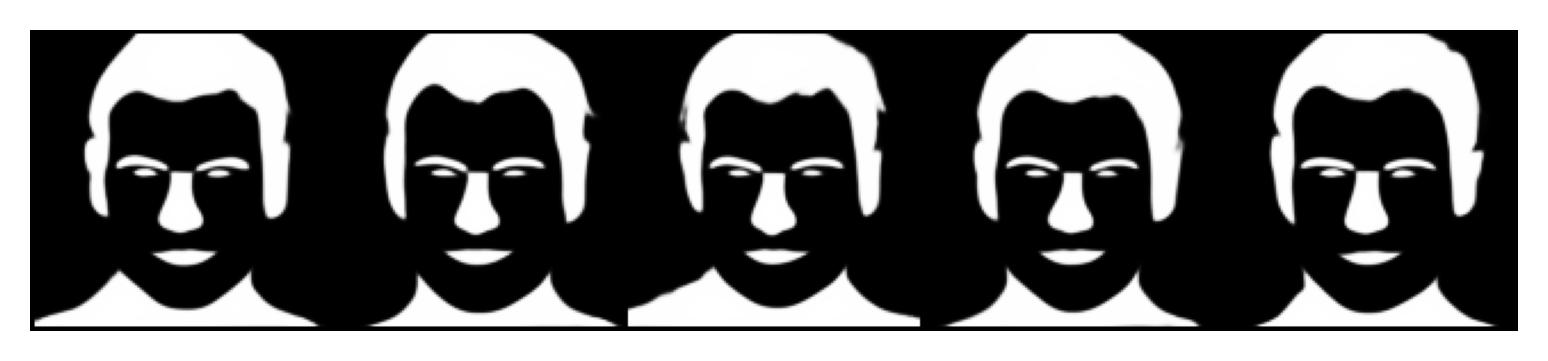}
         \caption*{ }
          \end{subfigure}

    \end{subfigure}
\end{subfigure}

        \caption{ (Image $\rightarrow$ Attribute,Mask ) Conditional generation of \textbf{\gls{MLD}} on CelebAMask-HQ.}

\end{figure}

\end{document}